\documentclass[acmtog,nonacm]{acmart}

\setcopyright{none}
\usepackage{subcaption}
\usepackage{multirow}
\usepackage{mathtools}
\usepackage{bbm}
\usepackage{tikz}
\usepackage{comment}
\usepackage{scalerel}
\usepackage{tcolorbox}
\usepackage{wrapfig}

\usepackage{appendix}

\makeatletter
\let\@authorsaddresses\@empty
\makeatother

\renewcommand\footnotetextcopyrightpermission[1]{}

\begin{document}

\title{Be Decisive: Noise-Induced Layouts for Multi-Subject Generation}

\author{Omer Dahary}
\email{omer11a@gmail.com}
\orcid{0000-0003-0448-9301}
\affiliation{%
    \institution{Tel Aviv University}
    \country{Israel}
}
\affiliation{%
    \institution{Snap Research}
    \country{Israel}
}
\author{Yehonathan Cohen}
\email{cohen19966@gmail.com}
\orcid{0009-0007-7919-4301}
\affiliation{%
    \institution{Tel Aviv University}
    \country{Israel}
}
\author{Or Patashnik}
\email{orpatashnik@gmail.com}
\orcid{0000-0001-7757-6137}
\affiliation{%
    \institution{Tel Aviv University}
    \country{Israel}
}
\affiliation{%
    \institution{Snap Research}
    \country{Israel}
}
\author{Kfir Aberman}
\email{kfiraberman@gmail.com}
\orcid{0000-0002-4958-601X}
\affiliation{%
    \institution{Snap Research}
    \country{United States of America}
}
\author{Daniel Cohen-Or}
\email{cohenor@gmail.com}
\orcid{0000-0001-6777-7445}
\affiliation{%
    \institution{Tel Aviv University}
    \country{Israel}
}
\affiliation{%
    \institution{Snap Research}
    \country{Israel}
}

\begin{abstract}
    Generating multiple distinct subjects remains a challenge for existing text-to-image diffusion models.
Complex prompts often lead to subject leakage, causing inaccuracies in quantities, attributes, and visual features.
Preventing leakage among subjects necessitates knowledge of each subject’s spatial location.
Recent methods provide these spatial locations via an external layout control.
However, enforcing such a prescribed layout often conflicts with the innate layout dictated by the sampled initial noise, leading to misalignment with the model's prior.
In this work, we introduce a new approach that predicts a spatial layout aligned with the prompt, derived from the initial noise, and refines it throughout the denoising process. By relying on this noise-induced layout, we avoid conflicts with externally imposed layouts and better preserve the model’s prior.
Our method employs a small neural network to predict and refine the evolving noise-induced layout at each denoising step, ensuring clear boundaries between subjects while maintaining consistency. Experimental results show that this noise-aligned strategy achieves improved text-image alignment and more stable multi-subject generation compared to existing layout-guided techniques, while preserving the rich diversity of the model’s original distribution.

\end{abstract}

\begin{teaserfigure}
    \centering
    \includegraphics[width=0.95\linewidth]{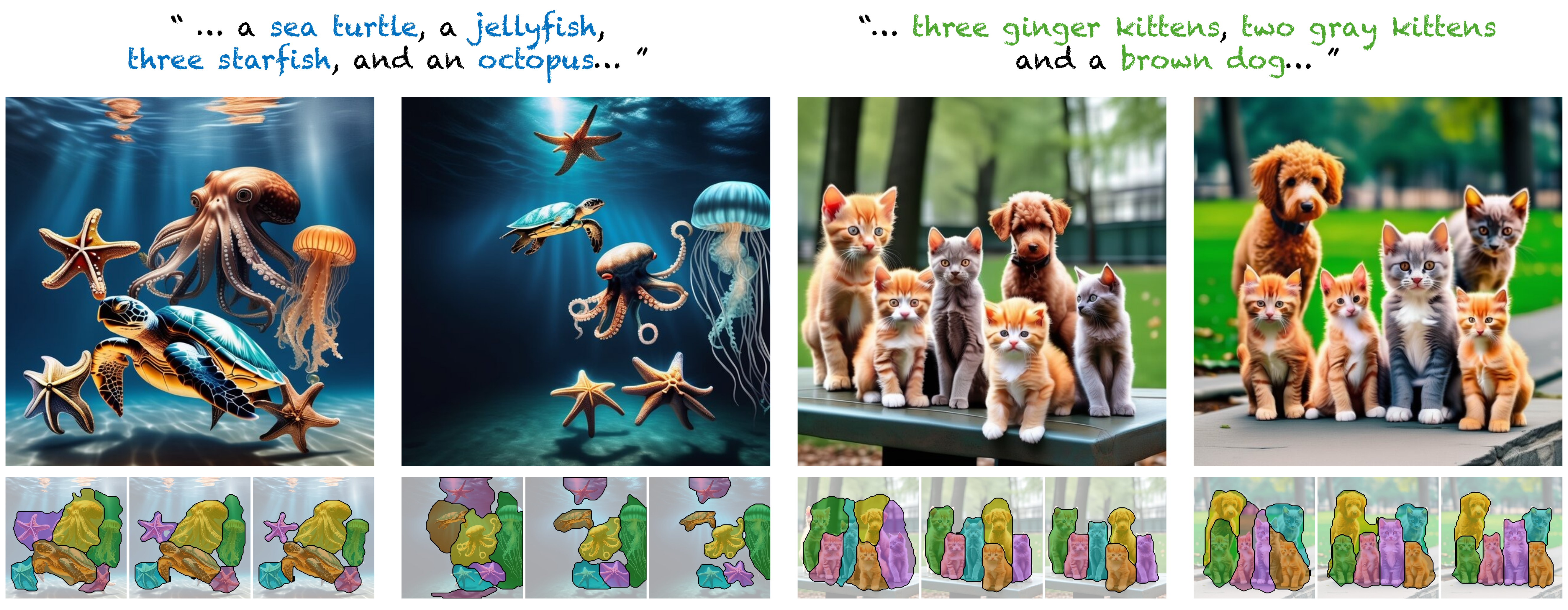}
    \caption{Our method generates multi-subject images by leveraging the layout encoded in the initial noise. Having a layout control allows to accurately generate each subject. We predict the layout based on the initial noise, and refine it throughout the denoising process, aligning it with the prompt and making it more fine-grained. Using the layout encoded in the initial noise we preserve the prior of the original model and generate diverse compositions. Below each of the generated images, we show the layout predicted at three timesteps along the generation process.}
    \label{fig:teaser2}
\end{teaserfigure}

\maketitle

\section{Introduction}
\label{sec:intro}

Diffusion models have revolutionized the field of image synthesis, enabling the creation of high-quality and diverse images from intuitive conditions such as textual prompts. However, despite their significant success, these models still struggle to accurately align to complex prompts~\cite{chefer2023attend}. Specifically, generating multiple subjects remains surprisingly challenging, often resulting in inaccurate quantities, attributes, and visual features~\cite{rassin2023linguistic,yang2024mastering,binyamin2024make}.

Recent works have identified harmful leakage between subjects as a primary source to text-image misalignment. To address this issue, previous methods manipulate the denoising process by limiting inter-attention among distinct subjects~\cite{dahary2025yourself}. This approach requires knowing each subject’s spatial location, which is not explicitly represented within the model, and hence it relies on a prescribed layout control. 

However, an externally imposed layout~\cite{zheng2023layoutdiffusion,qu2023layoutllm,yang2024mastering,feng2024layoutgpt,feng2024ranni} can conflict with the layout implied by the sampled initial noise, creating tension with the model’s prior and potentially leading to inferior results or deviations from the model's prior.

Specifically, as the image's low frequencies are defined early in the denoising process, the initial noise plays a fundamental role in shaping the final layout of the generated image~\cite{patashnik2023localizing,guo2024initno,ban2024crystal}. 
Therefore, steering the denoising trajectory toward a specific layout requires actively countering the model's intrinsic prior, which naturally encodes a layout intent within the initial noise. This often pushes the generated image away from the image manifold, resulting in semantic misalignment and degradation of image quality.

In this work, we introduce a method that derives a prompt-aligned spatial layout from the initial noise and iteratively refines it throughout the denoising process, as illustrated in Figure~\ref{fig:teaser2}.
By anchoring the layout around the initial noise, this approach stays consistent with the model’s prior, avoiding the conflicts introduced by externally imposed layouts.
We argue that this approach promotes more natural and diverse compositions by minimizing resistance to the input noise, and hence succeeds in generating images that better adhere to the prompt.

To produce the layout, we train a small neural network that predicts the layout induced by the latent noisy image using features extracted from the denoising model.
This network is applied throughout the denoising process, gradually refining the layout at each timestep to guide the generation toward layouts that remain both prompt-aligned and consistent across timesteps.

Our work embraces the motto ``Be Decisive''. At each denoising step, we guide the process toward a well-defined layout, ensuring clear boundaries between subjects. In this approach, each subject is assigned to a distinct image region, preventing leakage and enhancing text-image alignment. Meanwhile, only minimal adjustments are made to the layout between steps, maintaining consistency with the noise-induced layout throughout the process.

\begin{figure}[t]
    \setlength{\tabcolsep}{1pt}
    \centering
    \begin{tabular}{c c c c}
        \multicolumn{4}{c}{``... science fiction movie poster with \textbf{two astronauts},} \\
        \multicolumn{4}{c}{a \textbf{robot}, and a \textbf{spaceship}''} \\
        \includegraphics[width=0.24\linewidth]{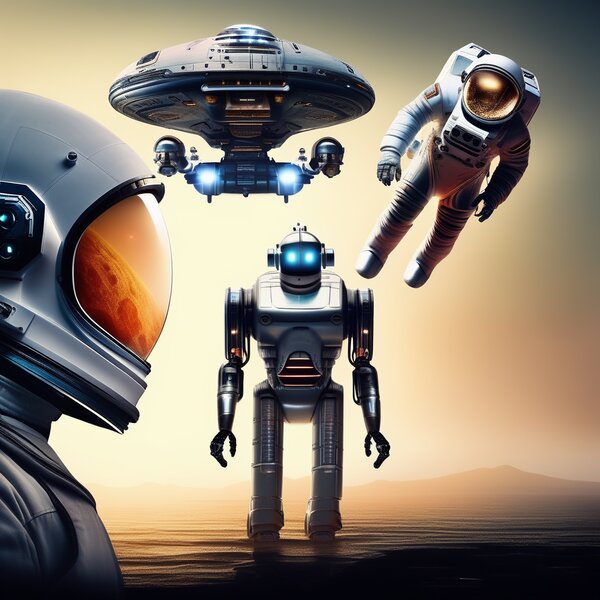} &
        \includegraphics[width=0.24\linewidth]{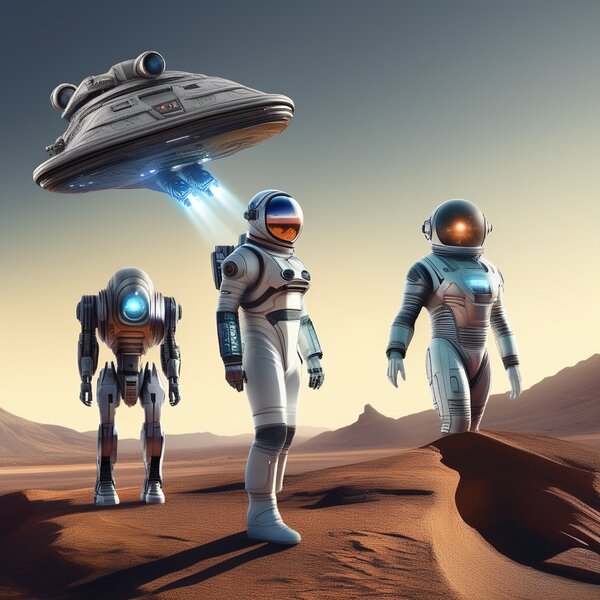} &
        \includegraphics[width=0.24\linewidth]{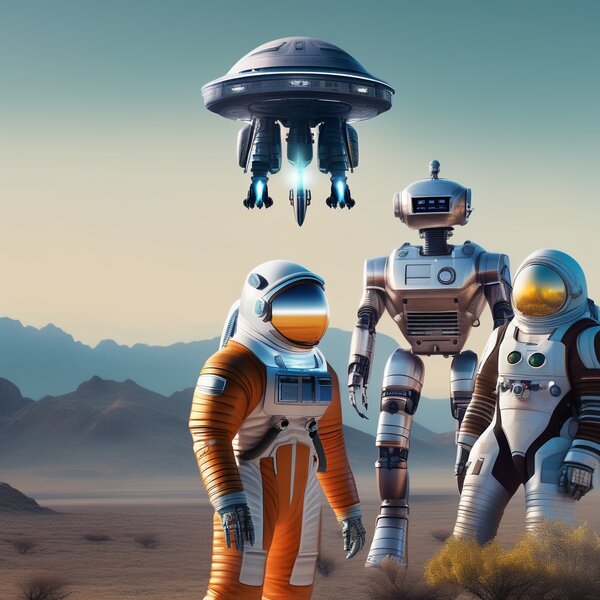} &
        \includegraphics[width=0.24\linewidth]{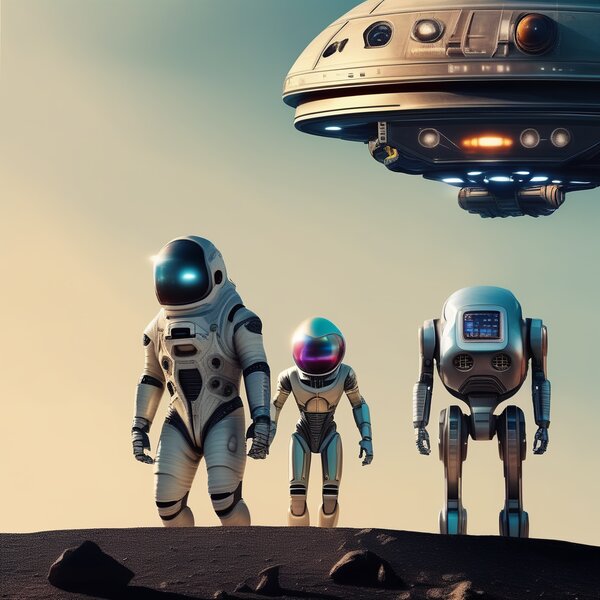} \\
        \includegraphics[width=0.24\linewidth]{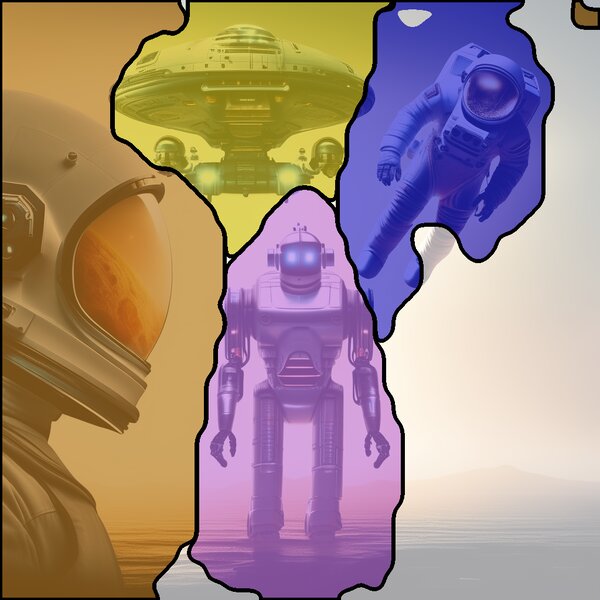} &
        \includegraphics[width=0.24\linewidth]{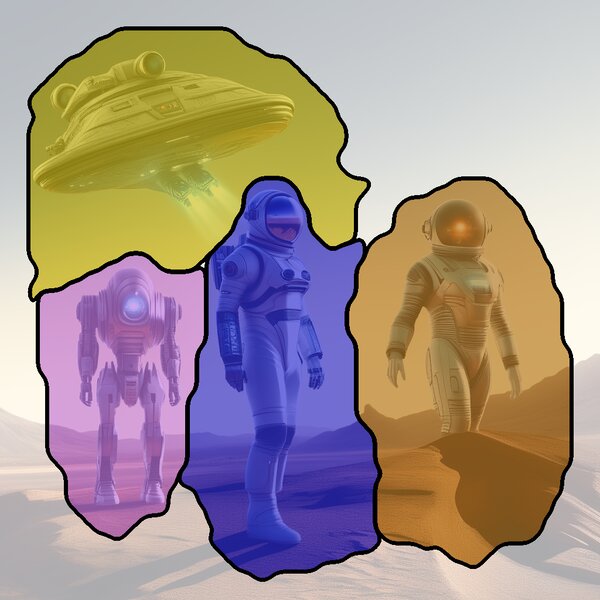} &
        \includegraphics[width=0.24\linewidth]{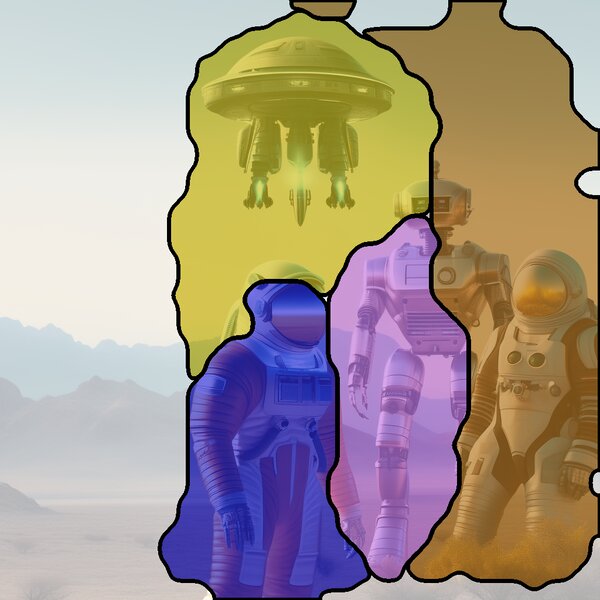} &
        \includegraphics[width=0.24\linewidth]{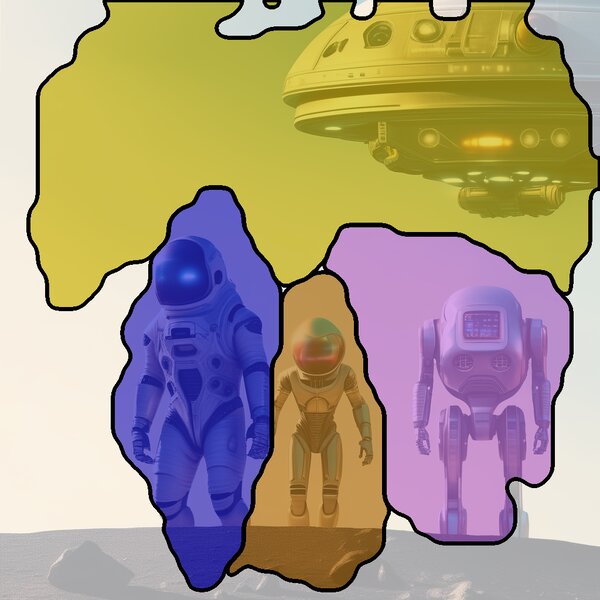} \\
    \end{tabular}
    \caption{Our method generates images with multiple subjects without requiring external layout inputs. By following the innate noise-induced layout encoded in the sampled initial noise, we preserve the model's prior and achieve diverse compositions. The second row show the initial noise-induced layout of the corresponding output images above. As can be seen, the initial layouts reflect the final composition of the generated images.
    }
    \label{fig:initial-clusters}
\end{figure}

Through extensive experiments, we demonstrate our method's power in adhering to complex multi-subject prompts, and compare it with previous methods. 
Specifically, we demonstrate that our method generates combinations of classes, adjectives, and quantities while maintaining diverse layouts that are natural, as they remain consistent with the model's prior layouts.
Figure~\ref{fig:initial-clusters} highlights this diversity, showcasing compositions obtained by sampling different initial noises.

\section{Related Work}
\label{sec:related_work}

Diffusion models~\cite{dhariwal2021diffusion,rombach2022high,ramesh2022hierarchical,saharia2022photorealistic,podell2023sdxl} have achieved remarkable success in modeling the complex distribution of natural images. However, despite their advantages, these models still face limitations in adhering to detailed prompts, particularly those involving multiple subjects. 
Previous works have addressed challenges in multi-subject generation through two distinct approaches: conditioning the generation on a spatial layout or applying heuristics to attention maps to enforce the generation of each subject mentioned in the prompt.

\paragraph{Layout-Based Multi-Subject Generation.}

Layout-based methods have demonstrated greater consistency in multi-subject generation compared to text-to-image models. Early efforts incorporated layout information through techniques such as multiple diffusion compositions~\cite{bar2023multidiffusion,ge2023expressive}, guidance from model features~\cite{kim2022dag, luo2024readout, voynov2023sketch}, specifically attention features~\cite{chen2023training,xie2023boxdiff,couairon2023zero,phung2024grounded,kim2023dense,liu2023customizable}, or fine-tuning~\cite{zhang2023adding,li2023gligen,yang2023reco,avrahami2023spatext,nie2024compositional}.

Recent studies highlight the architectural tendency of attention layers to leak visual features between subjects -- a phenomenon that complicates multi-subject generation~\cite{dahary2025yourself}. 
To address this, prior methods~\cite{wang2024instancediffusion,zhou2024migc,dahary2025yourself,wang2024ms} introduce techniques that mitigate such leakage by modifying the operation of attention layers within the model. However, these approaches rely on a predefined spatial layout to identify the subjects among which leakage should be prevented.
In our work, we propose a method to dynamically define the spatial locations of subjects during image generation by extracting the layout throughout the process. This extracted layout is then used to prevent leakage, enabling the generation of accurate multi-subjects images.

\begin{figure*}[th!]
    \centering
    \includegraphics[width=1\linewidth]{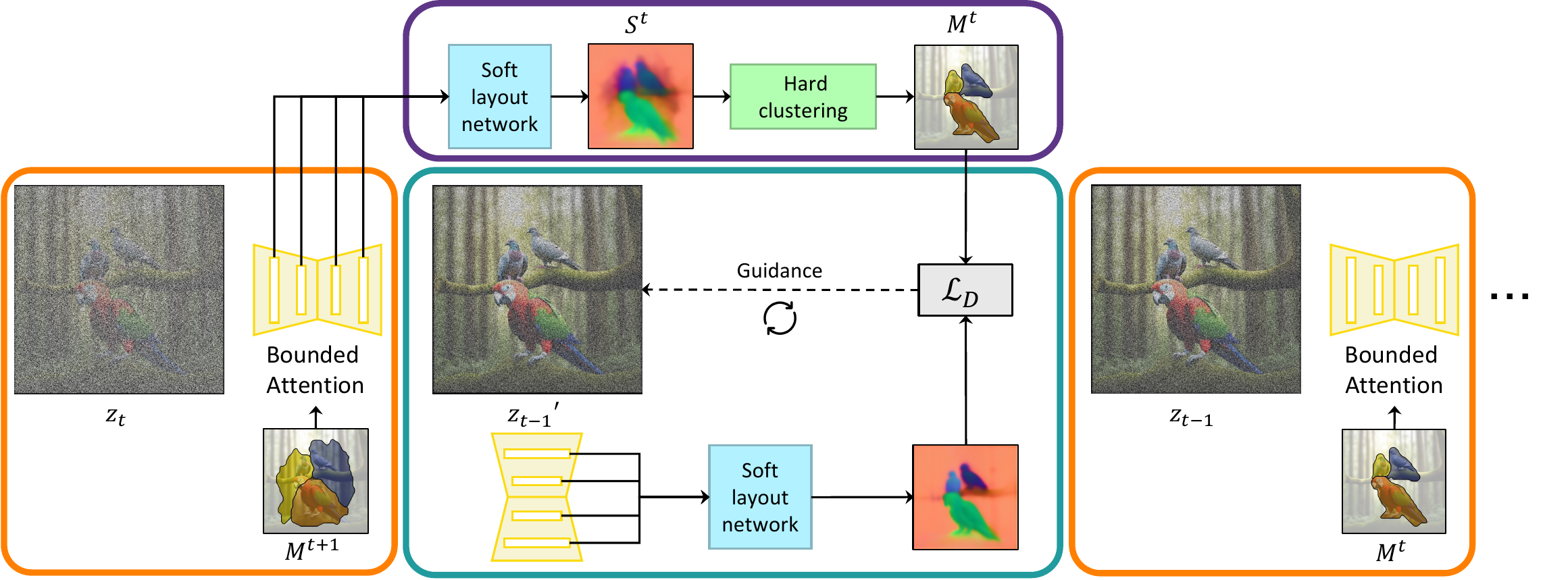}
    \caption{Our method steers the denoising process by applying iterative guidance (turquoise box) after each denoising step (orange regions). At denoising step $t$ (left orange box), we predict a soft-layout $S^t$ based on the diffusion model's features, and cluster it to form a hard-layout $M^t$ (purple box). This hard-layout is then used to control the layout of the next denoising step (right orange box). In the guidance stage, we optimize the latent image, with the objective to align its associated updated soft-layout with the hard-layout $M^t$.}
    \label{fig:method}
\end{figure*}

To simplify the image generation process for users, a common practice is to automatically generate a layout prior to image generation. Several works leverage large language models (LLMs) for this task, employing in-context learning or chain-of-thought reasoning~\cite{lian2023llm,qu2023layoutllm,chen2023reason,feng2024layoutgpt,yang2024mastering}. While these methods excel in producing plausible layouts, the separation between the prompt-to-layout and layout-to-image models often leads to inaccuracies or unnatural results in multi-subject images. Notably, Ranni~\cite{feng2024ranni} proposes overcoming this limitation by jointly fine-tuning the LLM and diffusion model on a shared dataset. However, their approach demands significant resources, with results obtained using a large proprietary model trained on millions of examples.

\paragraph{Layout-Free Multi-Subject Generation.}

Numerous approaches have sought to address specific aspects of subject misalignment during inference without relying on a predefined spatial layout. Some manipulate text embeddings~\cite{feng2022training,tunanyan2023multi}, while others guide the model to disentangle the attention distributions of distinct subjects and attributes~\cite{chefer2023attend,rassin2023linguistic,agarwal2023star,li2023divide,meral2024conform}. While these methods show some success, their effectiveness often hinges on the initial noise, resulting in unstable outcomes. To enhance robustness, other techniques~\cite{wang2023tokencompose,bao2024separate} employ fine-tuning based on similar heuristics in a self-supervised manner. Nevertheless, due to the model's limitations in interpreting quantities and distinguishing between numerous subjects, these approaches often struggle to generate more than two or three distinct subjects and fail to support multiple instances of the same class effectively.

Other works specifically tailor solutions for accurate subject quantities~\cite{zhang2023zero,kang2023counting,binyamin2024make}. While these methods perform well for single-class scenarios, they lack the generality needed for complex compositions involving multi-class subjects and attributes. In contrast, our approach provides comprehensive control over multi-class subjects, quantities, and attributes, addressing the limitations of existing layout-free methods.

\section{Preliminary: Bounded Attention}
\label{sec:preliminary}

Text-to-image diffusion models struggle to generate accurate multi-subject images due to visual leakage between subjects. Prior work~\cite{dahary2025yourself} identified the model’s attention layers as the primary source of this leakage — where features of semantically similar subjects are indiscriminately blended — and proposed Bounded Attention as a training-free solution to mitigate it.

Given an input layout, Bounded Attention modifies the attention layers during the denoising process by masking the attention between queries and keys of different subjects. In cross-attention layers, it constrains each subject’s attention to its corresponding textual tokens. In self-attention layers, it restricts attention to pixels within the subject’s own region and the background, explicitly excluding other subjects. This masking scheme reduces the influence of irrelevant visual and textual tokens on each pixel, maintaining the distinct visual features of each subject.

During generation, Bounded Attention alternates between denoising steps and guidance steps, both of which adopt the masking scheme. In guidance mode, the latent representation is optimized to adhere to the input layout: $ z_t^\text{opt} = z_t - \beta \nabla_{z_t} \left( \mathcal{L}_{\textit{cross}} + \mathcal{L}_{\textit{self}} \right) $, where $\mathcal{L}_{\textit{cross}}$ and $\mathcal{L}_{\textit{self}}$ are loss terms that encourage the respective cross- and self-attention maps to focus within each subject's designated mask. By isolating attention for each subject, the masking scheme avoids guidance artifacts caused by forcing similar queries to diverge, maintaining a trajectory that is better aligned with the data manifold.

In our work, we adopt Bounded Attention’s masking scheme to reduce leakage, but instead of relying on a prescribed layout, we extract the noise-induced layout and refine it between denoising steps. We further modify the guidance procedure to promote decisiveness — that is, enforcing strict subject boundaries throughout the layout refinement process.

\section{Method}
\label{sec:method}

Our method aims to facilitate the generation of multiple distinct subjects using an existing text-to-image model~\cite{podell2023sdxl}. 
We steer the denoising process to adhere to a layout that allows preventing unwanted leakage among the subjects. Our key idea is to progressively define a prompt-aligned spatial layout based on features extracted from the noisy latent images along the denoising process. We then encourage the denoising process to follow these layouts, upholding this initial ``decision''.

Figure~\ref{fig:method} illustrates the overall structure of our inference pipeline.
Our method is built on a denoising process to which we apply Bounded Attention~\cite{dahary2025yourself} (marked in orange boxes) controlled by layout masks $M^t$. 
We add two components to the denoising process. First, a component that predicts a prompt-aligned layout $M^t$ from a noisy latent image $z_t$ based on features extracted from the diffusion model (purple box). Second, a guidance mechanism that optimizes a noisy latent image so that its induced layout aligns with the previous layout (turquoise box). This mechanism encourages a ``decisive'' generation process, where each subject mentioned in the prompt is consistently assigned to its own distinct image region across timesteps.

Both components rely on a \emph{soft-layout} $S^t$. The soft-layout is a timestep-dependent feature map that reflects the likelihood that two pixels will be associated with a common subject. In the following, we elaborate on the soft-layout and its use.

\subsection{Soft-Layout}

We begin by explaining the motivation behind our soft-layouts. Extracting fine-grained layouts directly from the initial noise is inherently challenging since the image is formed in a gradual manner. Moreover, predicted layouts might not perfectly correspond to the subjects specified in the prompt. To address these challenges, we introduce the notion of soft-layout, a feature map that represents each pixel as a descriptor encapsulating its potential to associate with other pixels in composing a single subject. In the first timesteps, due to high uncertainty, the soft-layout encodes a coarse layout. At later timesteps, the soft-layout is more granular and precise. 
Our use of the soft-layout is two-fold. First, it is used to predict the masks $M^t$, termed as \textit{hard-layout}, which bounds the attention in the denoising steps. Second, we optimize the noisy latent image to produce a soft-layout that agrees with $M^t$.

\begin{figure}
    \setlength{\tabcolsep}{0.002\textwidth}
    \scriptsize
    \centering

    \begin{tabular}{c c c c c c}
        \multicolumn{6}{c}{``A \textbf{parrot} and \textbf{two doves} sitting on a branch in a lush forest at daylight''} \\
         & $t=50$ & $t=45$ & $t=40$ & $t=35$ & \\[2pt]

        \raisebox{8pt}{\rotatebox{90}{Soft-layout}} & 
        \includegraphics[width=0.18\linewidth]{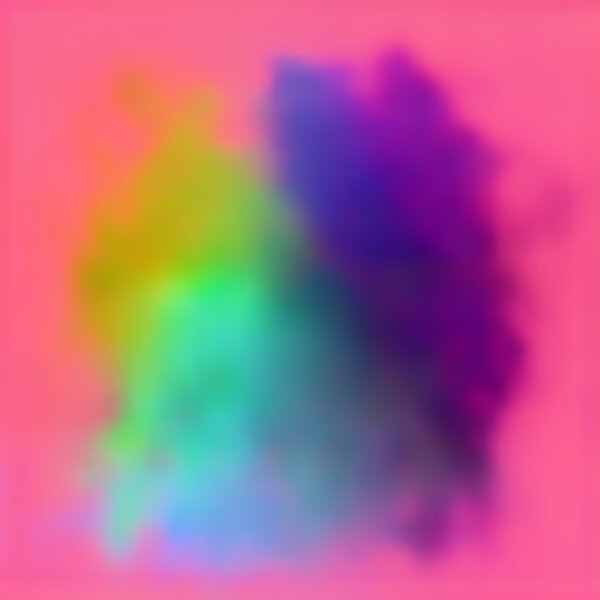} & 
        \includegraphics[width=0.18\linewidth]{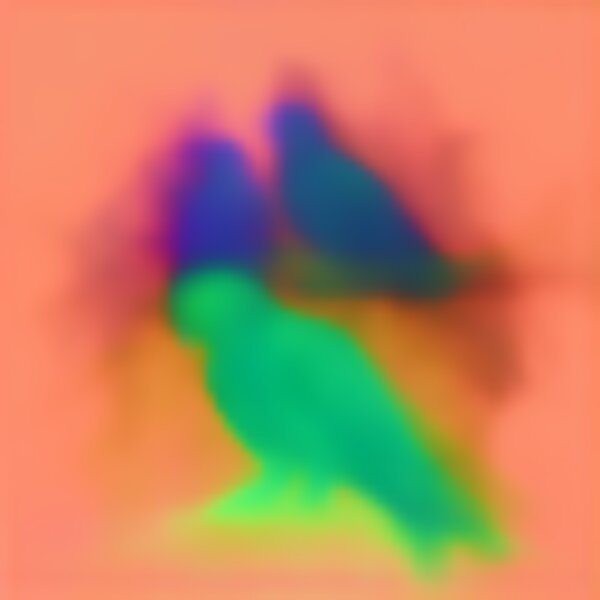} & 
        \includegraphics[width=0.18\linewidth]{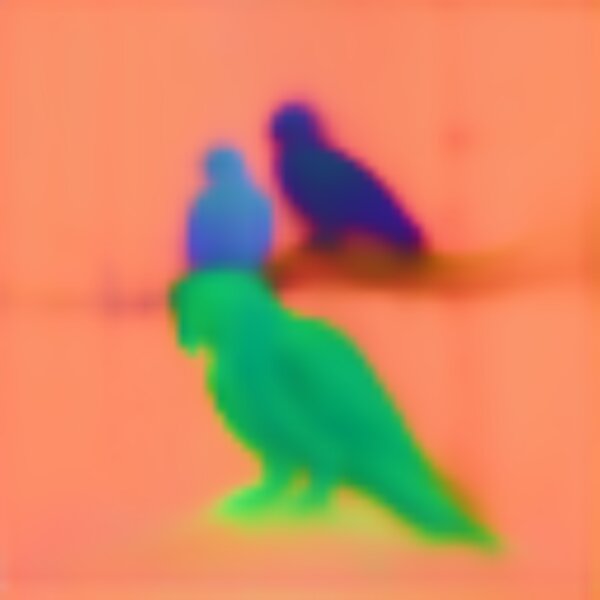} & 
        \includegraphics[width=0.18\linewidth]{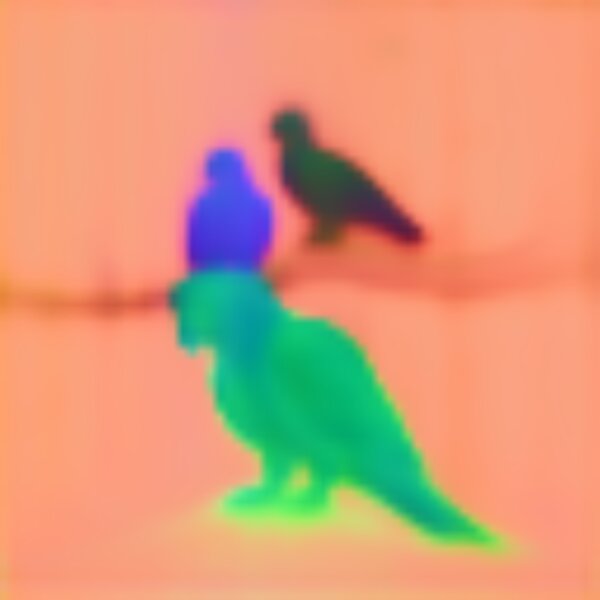}
        &
        \includegraphics[width=0.18\linewidth]{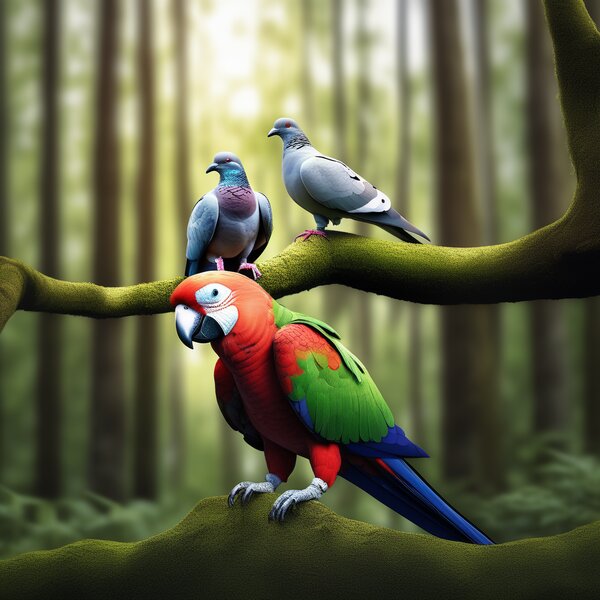}
        \\
        \raisebox{8pt}{\rotatebox{90}{Hard-layout}} & 
        \includegraphics[width=0.18\linewidth]{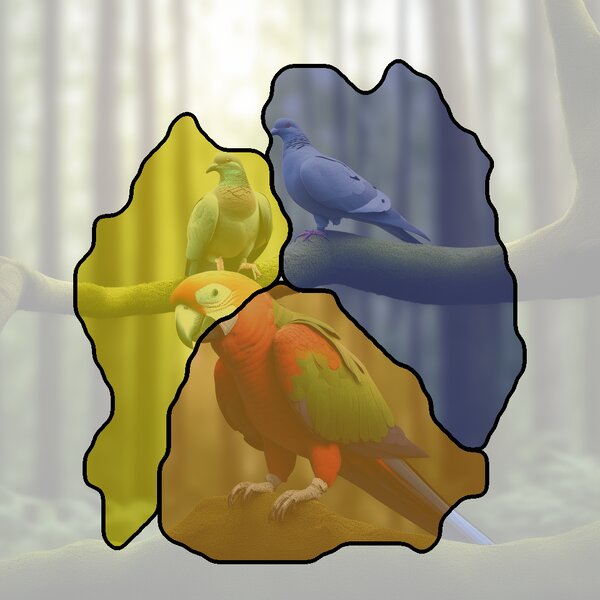} & 
        \includegraphics[width=0.18\linewidth]{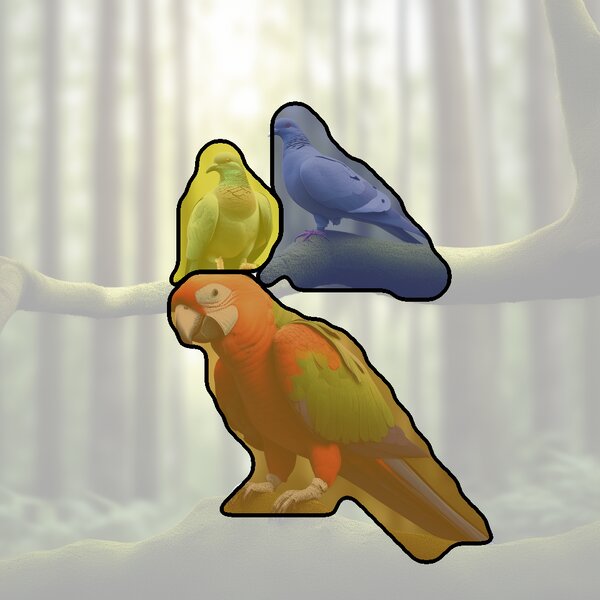} & 
        \includegraphics[width=0.18\linewidth]{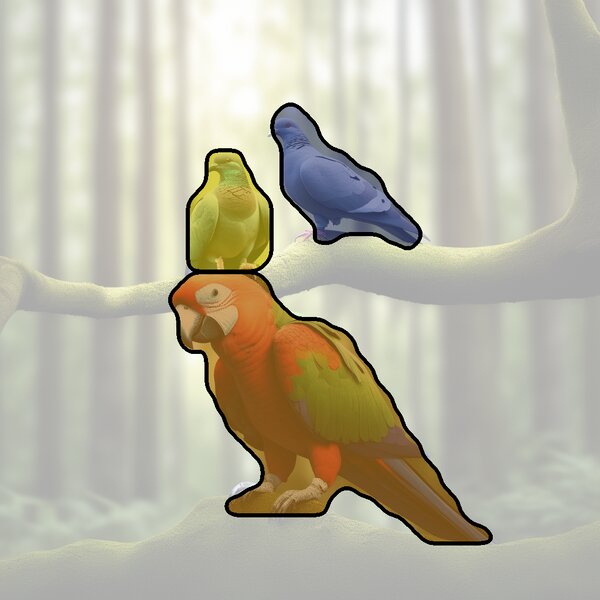} & 
        \includegraphics[width=0.18\linewidth]{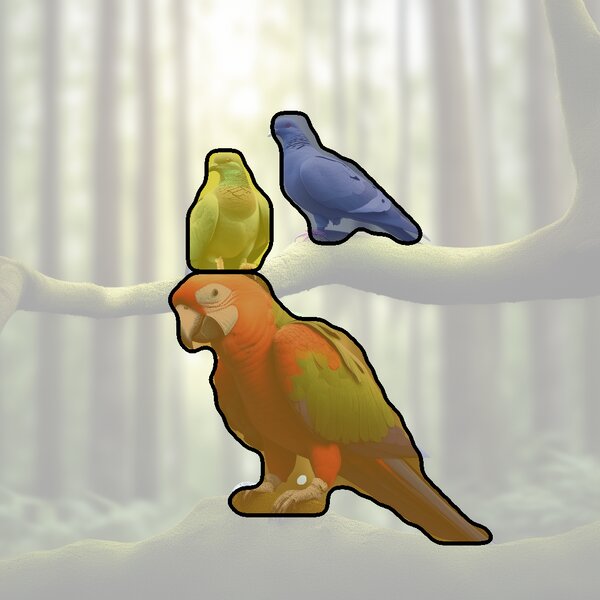} &
        \includegraphics[width=0.18\linewidth]{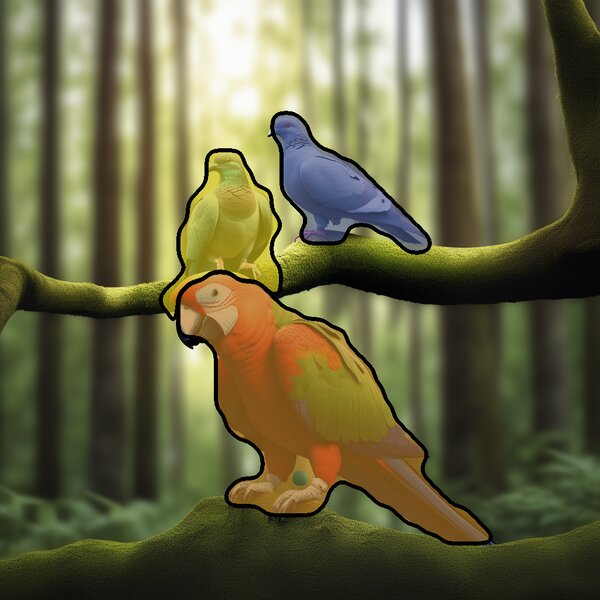} 
        \\
        \multicolumn{6}{c}{Our full method} \\[2pt]
        
        \raisebox{8pt}{\rotatebox{90}{Soft-layout}} & 
        \includegraphics[width=0.18\linewidth]{images/layout/saved_layout_0.jpg} & 
        \includegraphics[width=0.18\linewidth]{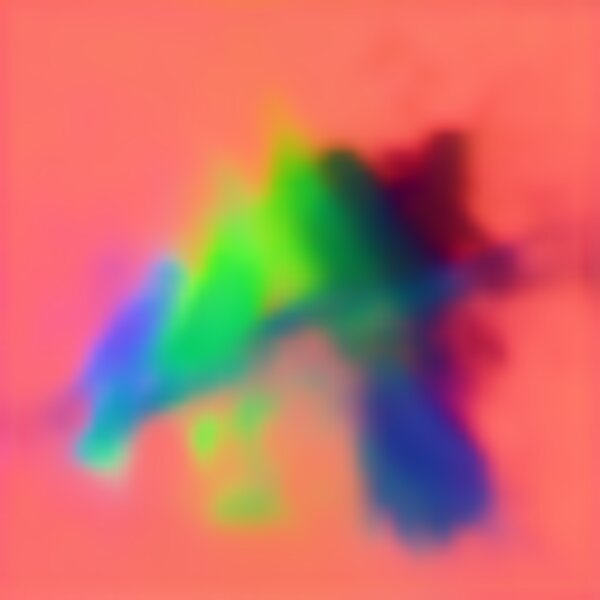} & 
        \includegraphics[width=0.18\linewidth]{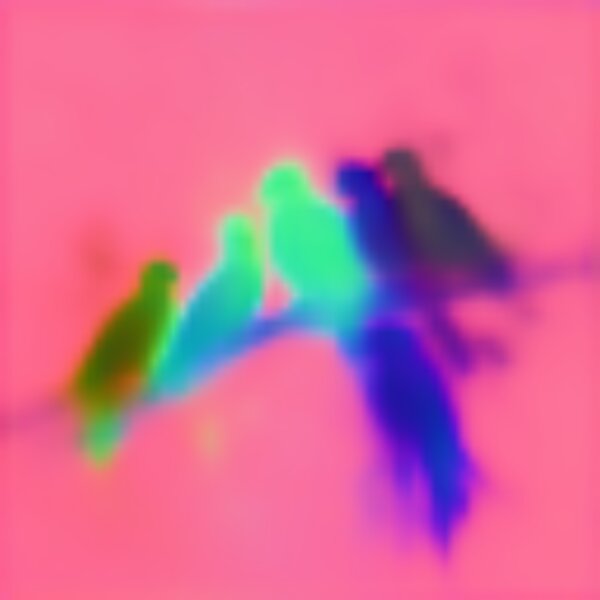} & 
        \includegraphics[width=0.18\linewidth]{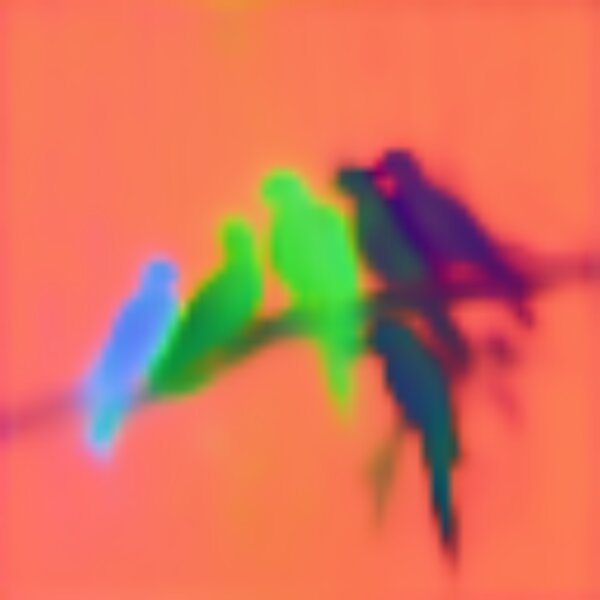}
        &
        \includegraphics[width=0.18\linewidth]{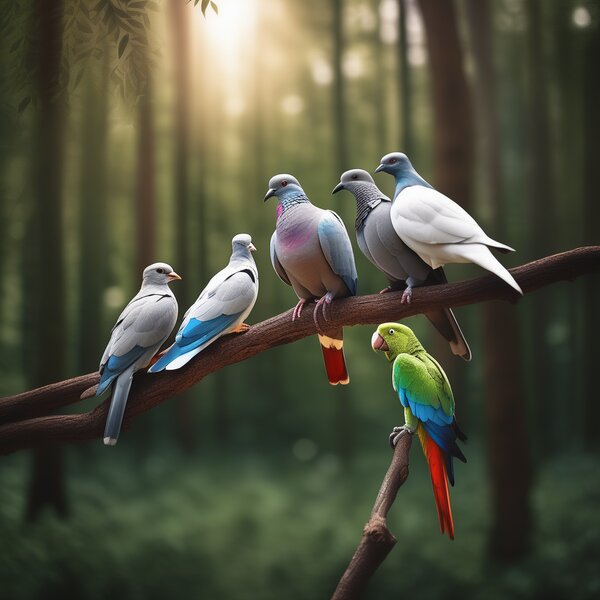}
        \\
        \raisebox{8pt}{\rotatebox{90}{Hard-layout}} & 
        \includegraphics[width=0.18\linewidth]{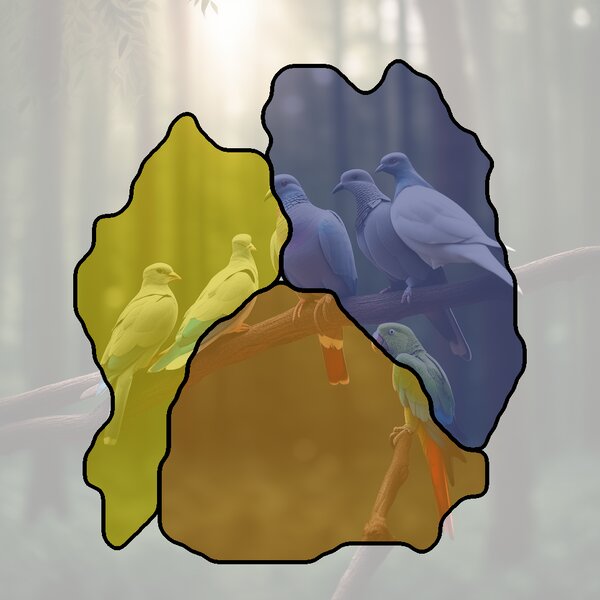} & 
        \includegraphics[width=0.18\linewidth]{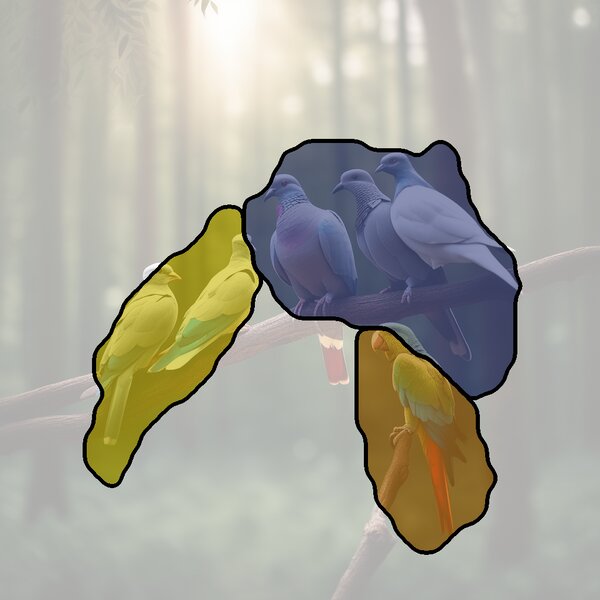} & 
        \includegraphics[width=0.18\linewidth]{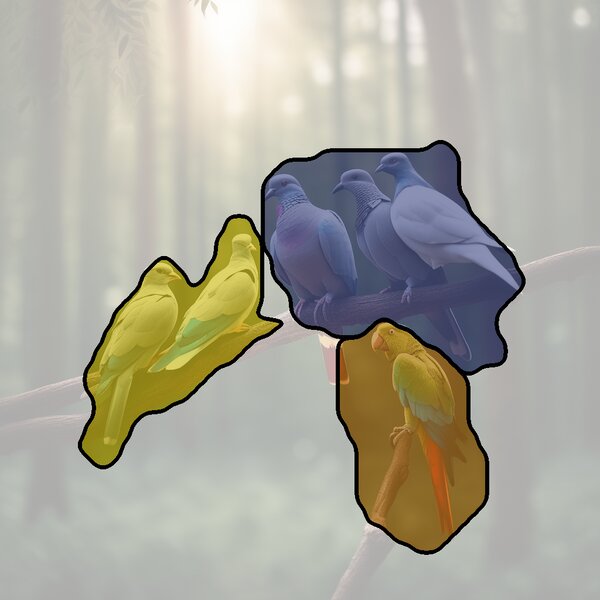} & 
        \includegraphics[width=0.18\linewidth]{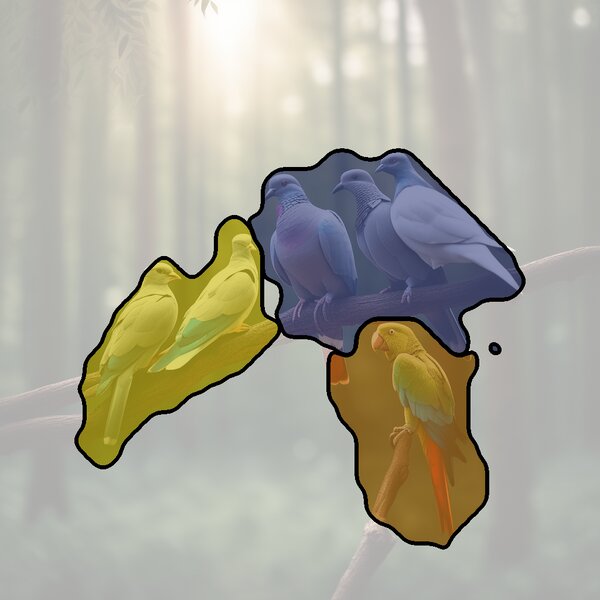} &
        \includegraphics[width=0.18\linewidth]{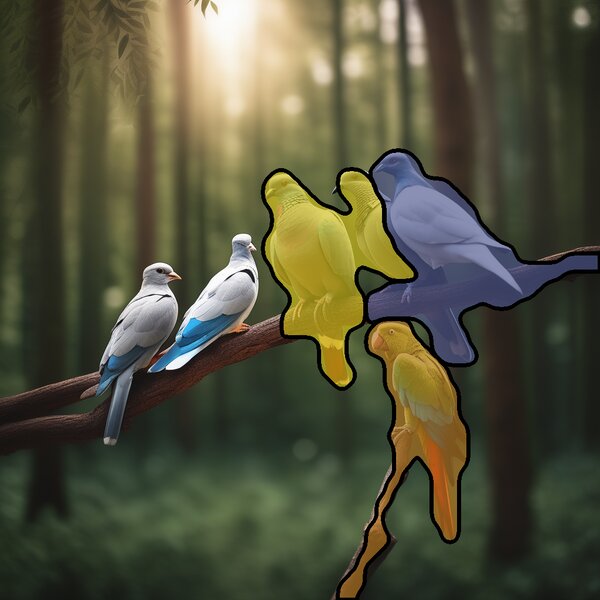}  \\
        \multicolumn{6}{c}{Our method w/o guidance} \\[4pt]
        
        \raisebox{8pt}{\rotatebox{90}{Soft-layout}} & 
        \includegraphics[width=0.18\linewidth]{images/layout/saved_layout_0.jpg} & 
        \includegraphics[width=0.18\linewidth]{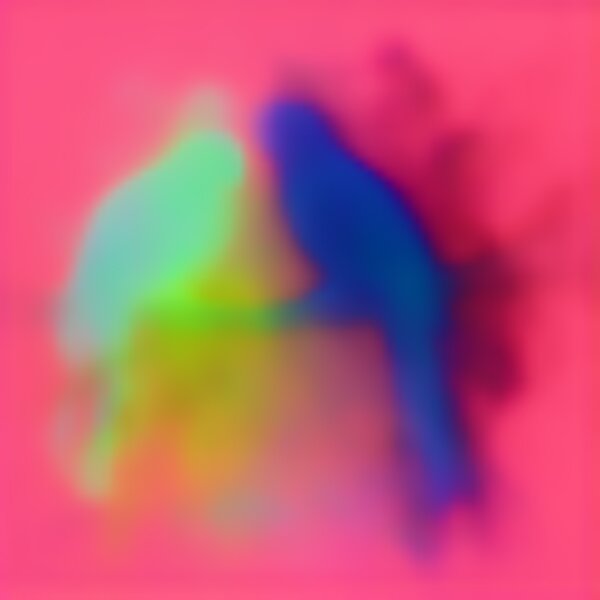} & 
        \includegraphics[width=0.18\linewidth]{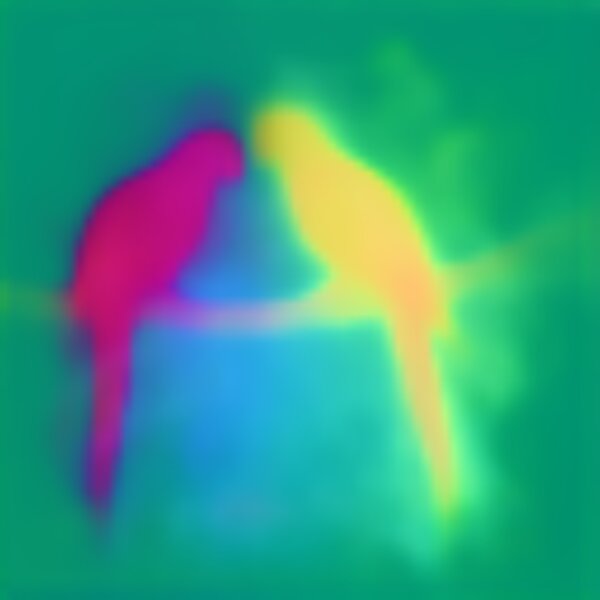} & 
        \includegraphics[width=0.18\linewidth]{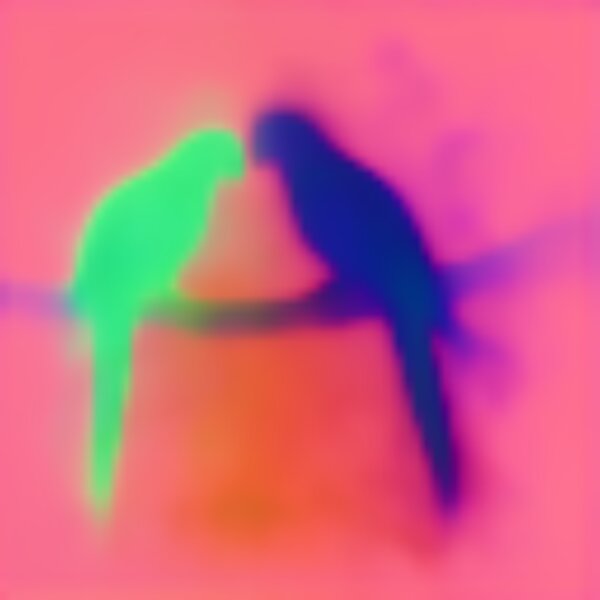} &
        \includegraphics[width=0.18\linewidth]{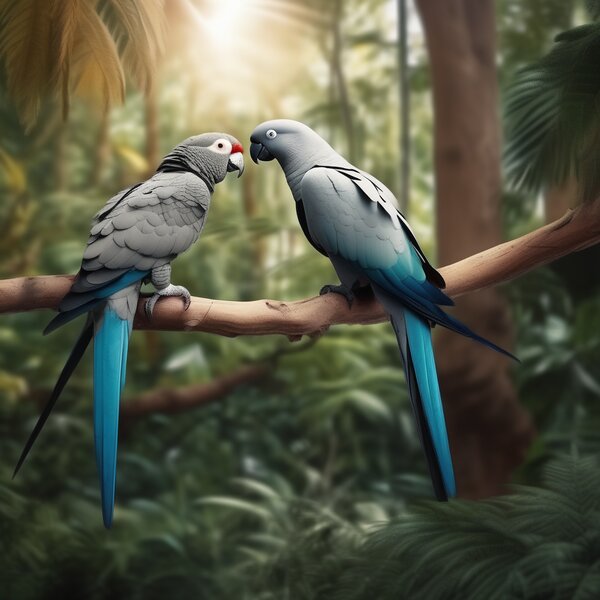} \\
        \multicolumn{6}{c}{Vanilla SDXL}
    \end{tabular}
    \caption{The figure illustrates the progression of the soft- and hard-layouts in three cases. The top row shows results from our full method. The middle row presents our method without guidance. The bottom row shows vanilla SDXL, where only the soft-layout extracted from the noisy latents is displayed. Below each image, we show the hard-layout obtained at the final timestep.
    }
    \label{fig:layout}
\end{figure}

At the top of Figure~\ref{fig:layout}, we display the progressive layouts produced by our full pipeline. In the middle, we show the corresponding layouts without guidance. As illustrated, guidance is crucial for maintaining consistent hard-layouts across timesteps, thereby facilitating convergence to a prompt-aligned layout by the end of the denoising process.

We now turn to formally define the soft-layout and elaborate on the network we train to predict it.
A soft-layout $S^t \in \mathbb{R}^{n \times d}$ is a feature map, encoding $n$ pixels as $d$-dimensional vectors, where the similarity of two feature vectors $S^t\left[x_1\right],S^t\left[x_2\right]$ indicates correspondence to the same subject in the generated image. To produce the soft-layout, we train a network that takes as input a set of features extracted from various layers of the diffusion model.

\paragraph{Dataset}

To train our network, we automatically construct a small dataset of $\sim 1500$ images synthesized by the diffusion model, along with their segmentation maps.
First, we randomly generate a set of prompts specifying multiple subject classes and their quantities (see full details in the supplemental). Then, we synthesize images based on these prompts, and segment them by feeding the corresponding subject names to GroundedSAM~\cite{Ren2024GroundedSA}. We filter out ambiguous examples, where two segmentation masks share a large overlap, and select a single label for each segment based on the segmentation model's confidence score.

Notably, we do not apply any filtering based on prompt alignment. This allows the network to predict soft-layouts that match the diffusion model's intent, even if it does not adhere to the prompt. In turn, this enables our guidance mechanism to detect misalignments early in the denoising process and apply corrective updates to the latent.

\paragraph{Architecture}

Following Readout Guidance~\cite{luo2024readout}, we design our model as a collection of lightweight convolutional heads, each processing different features from the denoising model along with the current time embedding. The outputs of these heads are then averaged using learnable weights, and fed into a convolutional bottleneck head, which outputs a $64\times64\times10$ feature map, representing the soft-layout.

We attach our heads to the attention layers, which are known to be highly indicative of the image structure and subject boundaries~\cite{hertz2022prompt,tumanyan2023plug,patashnik2023localizing}. Specifically, we use the cross-attention queries and the self-attention keys at the decoder layers. See the supplemental for full architectural details.

\paragraph{Training}

We train the soft-layout network with a triplet loss \cite{schroff2015facenet}, encouraging feature similarity between pixels of the same segment, and dissimilarity between different segments.

Formally, given a random timestep $t$ and an image with $k$ subject segments $\left\{M_j\right\}_{j=1}^k$ and a background $M_0$, we sample triplets of pixel coordinates $x_{i_{\textit{a}}},x_{i_{\textit{p}}} \in M_{j_{\textit{p}}}$ and $x_{i_{\textit{n}}} \in M_{j_{\textit{n}}}$, where $M_{j_{\textit{p}}} \neq M_{j_{\textit{n}}}$. Then, we compute the following loss

\begin{equation}
\sum_{i} \left[ \text{sim}(S^t\left[x_{i_{\textit{a}}}\right], S^t\left[x_{i_{\textit{n}}}\right]) - \text{sim}\left(S^t\left[x_{i_{\textit{a}}}\right], S^t\left[x_{i_{\textit{p}}}\right] \right) + \alpha \right]_+,
\end{equation}

where $\text{sim}$ is the cosine-similarity, $\alpha$ is the similarity margin between positive and negative samples, and $\left[\cdot\right]_+$ is the ReLU operation.

\subsection{From Soft to Hard Layouts}

While the soft-layout represents the original model's future intent, to successfully generate multiple prompt-aligned subjects, it is necessary to uphold clear subject boundaries in accordance to the prompt. To achieve this, we derive a hard-layout from the soft-layout produced by our network.

More specifically, given $k$ subjects mentioned in the prompt, we apply K-Means to cluster the soft-layout into $k+1$ segments: $k$ for the subjects, and one for the background. We set the background $M_0$ as the cluster that has the biggest overlap with the image's border, and recursively cluster each of the other segments into two sub-clusters, continuing the process with the bigger sub-cluster, until the variance is smaller than $\sigma^2_{\textit{cluster}}$. Any sub-cluster dropped during this process is added to $M_0$.

Finally, we must tag each subject cluster with an appropriate label representing a specific subject instance. After the first denoising step, at $t=T$, we compute the average cross-attention map of each subject noun~\cite{hertz2022prompt,epstein2023diffusion,patashnik2023localizing} and use 
the Hungarian algorithm to assign instances to clusters such that the corresponding cross-attention response in each cluster is maximized.

To avoid leakage, our initial decision regarding each subject's location must be respected throughout the generation process. Thus, for $t < T$, we stack the soft-layout $S^t$ with the previous soft-layouts $S^{t+1},\dots,S^{\min \left( t+w , T \right) }$ from $w$ earlier timesteps, before performing hard-clustering. Since clusters may shift over time, we reassign their labels at each timestep using the Hungarian algorithm, matching each cluster in $M^t$ to a cluster in $M^{t+1}$ such that their intersection-over-union (IoU) is maximized.

\subsection{Decisive Guidance}

\begin{figure}
    \setlength{\tabcolsep}{0.002\textwidth}
    \scriptsize
    \centering

    \begin{subfigure}{0.5\linewidth}
    \centering
    \begin{tabular}{c c}
        Soft-layout & 
        Hard-layout \\
        \includegraphics[width=0.5\linewidth]{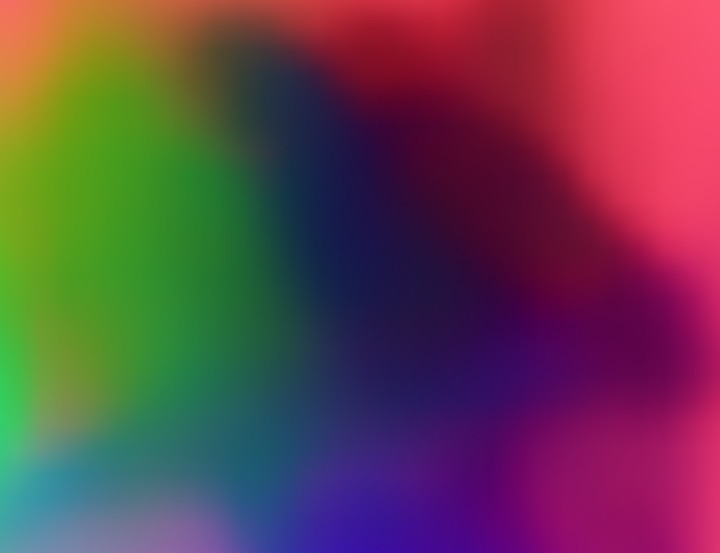} & 
        \includegraphics[width=0.5\linewidth]{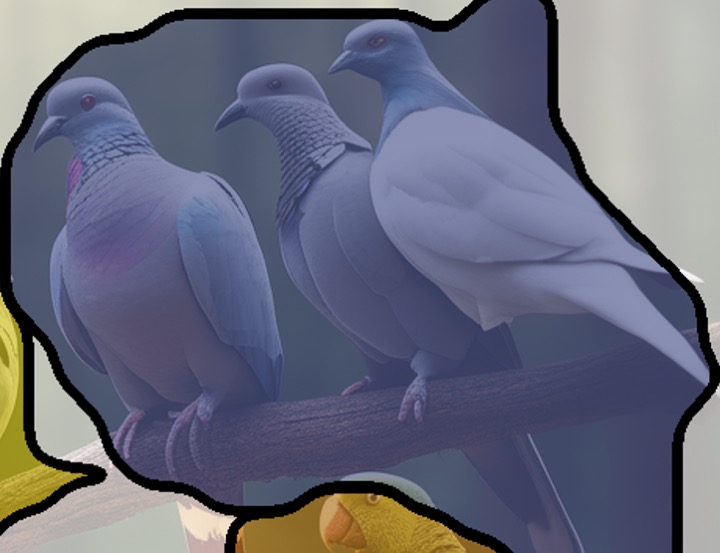} \\
    \end{tabular}
    \caption{intra-cluster over-generation}
    \label{fig:losses-intra}
    \end{subfigure}%
    \begin{subfigure}{0.5\linewidth}
    \centering
    \begin{tabular}{c c c}
        \multicolumn{3}{c}{Hard-layouts over successive timesteps} \\
        \includegraphics[height=0.385\linewidth]{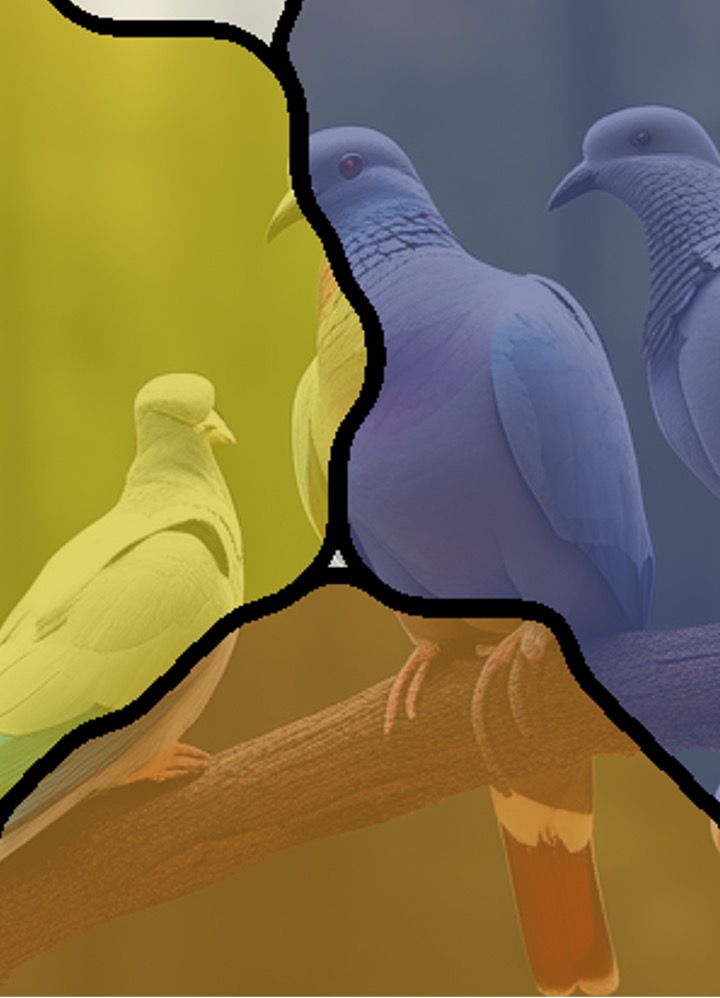} & 
        \includegraphics[height=0.385\linewidth]{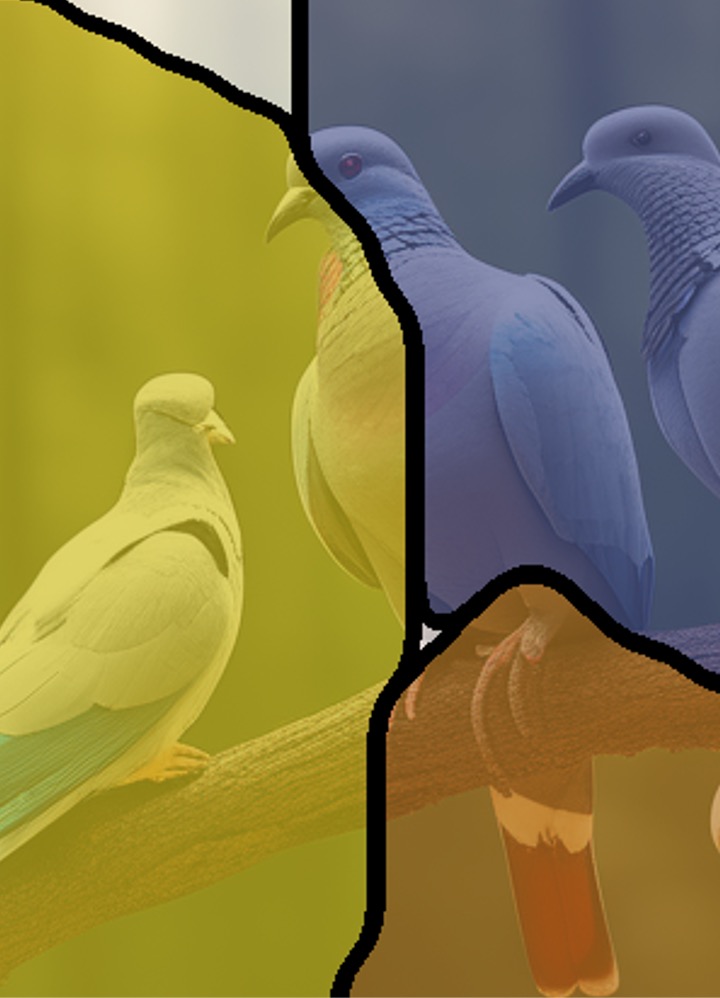} & 
        \includegraphics[height=0.385\linewidth]{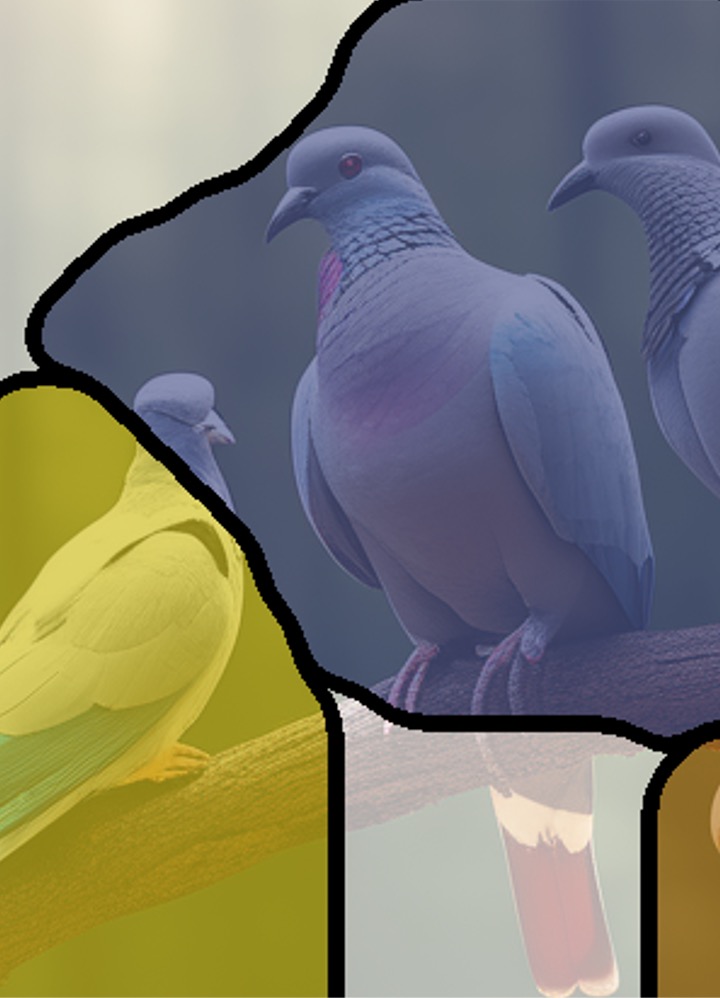} \\
    \end{tabular}
    \caption{inconsistent cluster borders}
    \label{fig:losses-inter}
    \end{subfigure}
    \caption{
    Without guidance, we observe two types of layout failures: (a) intra-cluster over-generation, where multiple subjects are assigned to a single cluster due to high variance in the soft-layout; and (b) inconsistent cluster borders across timesteps, leading to subject over-generation and leakage caused by oscillating boundaries.
    }
    \label{fig:losses}
\end{figure}

To encourage decisiveness — in the sense of maintaining consistent subject boundaries throughout generation — we perform guidance steps after each denoising step. These steps optimize the intermediate latent $z_{t-1}$ to align the predicted soft-layout $S^{t-1}$ with the previous hard-layout $M^t$ (see turquoise box in Figure~\ref{fig:method}).

First, to integrate each subject's semantics to its designated segment in $M^{t}$, we apply the cross-attention loss $\mathcal{L}_{\textit{cross}}$ from Bounded Attention (see Section \ref{sec:preliminary}).
Notably, Bounded Attention also introduces a self-attention loss $\mathcal{L}_{\textit{self}}$ to mitigate subject neglect in fixed, externally provided layouts by discouraging background attention. However, we omit this term, as it is unnecessary with our noise-induced layouts and its removal significantly reduces runtime.

Nonetheless, our evolving layouts give rise to two distinct failure modes. These are illustrated in Figure~\ref{fig:losses}, which presents zoomed-in views of the soft- and hard-layouts generated without guidance (originally shown in Figure~\ref{fig:layout}).
On the left (Figure~\ref{fig:losses-intra}), the soft-layout contains three spatially separated foreground regions (colored green, purple, and dark red) within a single hard-cluster, each corresponding to a distinct dove. On the right (Figure~\ref{fig:losses-inter}), the middle dove is generated at the intersection of three hard-clusters and does not maintain consistent membership in any single cluster during denoising. As a result, its lower body is initially assigned to the parrot cluster, leading to a hybrid generation in which the dove inherits a parrot-like tail.

To address the first issue (Figure~\ref{fig:losses-intra}), we introduce a variance loss $\mathcal{L}_{\textit{var}}$ that encourages low cluster variance in $S^{t-1}$ with respect to the previous hard-layout $M^t$:
\begin{equation}
    \mathcal{L}_{\textit{var}} = \frac{1}{k+1} \sum_{j=0}^{k} \frac{1}{\left| M^t_j \right|} \sum_{x_i \in M^t_j} \text{sim}^2 \left( S^{t-1}\left[x_i\right] , \mu^{t-1}_j \right),
\end{equation}
where $\mu^{t-1}_j$ is the mean soft-layout feature vector of cluster $j$: \begin{equation}
\mu^{t-1}_j = \frac{1}{\left| M^t_j \right|} \sum_{x_i \in M^t_j} S^{t-1}\left[x_i\right].
\end{equation}
This loss promotes intra-cluster similarity, encouraging each cluster to represent a coherent subject instance.

To avoid cluster boundaries from oscillating between timesteps (Figure~\ref{fig:losses-inter}), we compute the Dice segmentation loss $\mathcal{L}_{\text{dice}}$~\cite{milletari2016v} between the hard-layout $M^t$ and a probabilistic layout $P^{t-1} \in \mathbb{R}^{n \times (k+1)}$, where each element $P^{t-1}\left[x_i,j\right]$ represents the probability that pixel $x_i$ belongs to cluster $j$: \begin{equation}
P^{t-1}\left[x_i,\cdot\right] = \text{softmax}\left( \left\{ \text{sim}\left( S^{t-1}\left[x_i\right] , \mu^{t-1}_j \right) / \tau \right\}_{j=0}^k \right) \in \mathbb{R}^{k+1},
\end{equation}
where $\tau$ is a temperature hyperparameter. This term penalizes ambiguous pixel-cluster associations, promoting sharper and more consistent cluster boundaries.

Together, these three terms address complementary aspects of the layout refinement process: $\mathcal{L}_{\textit{cross}}$ promotes the proper semantic alignment in each cluster, $\mathcal{L}_{\textit{var}}$ reduces intra-cluster ambiguity, and $\mathcal{L}_{\text{dice}}$ encourages temporal consistency and boundary sharpness. The final decisiveness loss is defined as:
\begin{equation}
    \mathcal{L}_{\textit{decisive}} = \alpha_{\textit{cross}} \mathcal{L}_{\textit{cross}} + \alpha_{\textit{var}} \mathcal{L}_{\textit{var}} + \alpha_{\textit{dice}} \mathcal{L}_{\textit{dice}},
\end{equation}
where $\alpha_{\textit{cross}}, \alpha_{\textit{var}}, \alpha_{\textit{dice}}$ are the respective weighting coefficients.

Ablation studies evaluating the contribution of each component are provided in the supplementary material.

\section{Experiments}

\begin{figure}
    \setlength{\tabcolsep}{1pt}
    \centering
    \begin{tabular}{c c c c}
        \multicolumn{4}{c}{``... a \textbf{huge lion}, and \textbf{two huge crocodiles} ... streets of London''} \\
        \includegraphics[width=0.24\linewidth, height=0.24\linewidth]{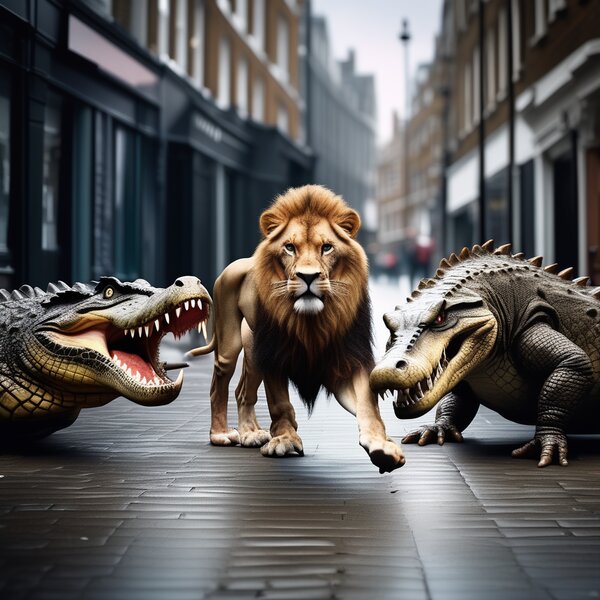} &
        \includegraphics[width=0.24\linewidth, height=0.24\linewidth]{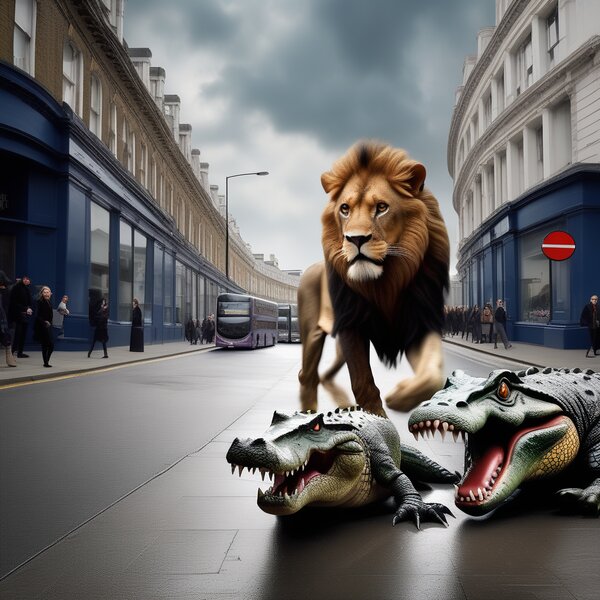} &
        \includegraphics[width=0.24\linewidth, height=0.24\linewidth]{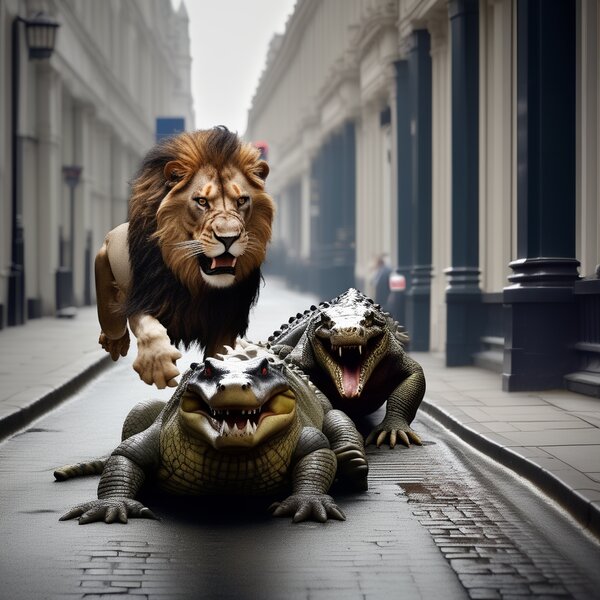} &
        \includegraphics[width=0.24\linewidth, height=0.24\linewidth]{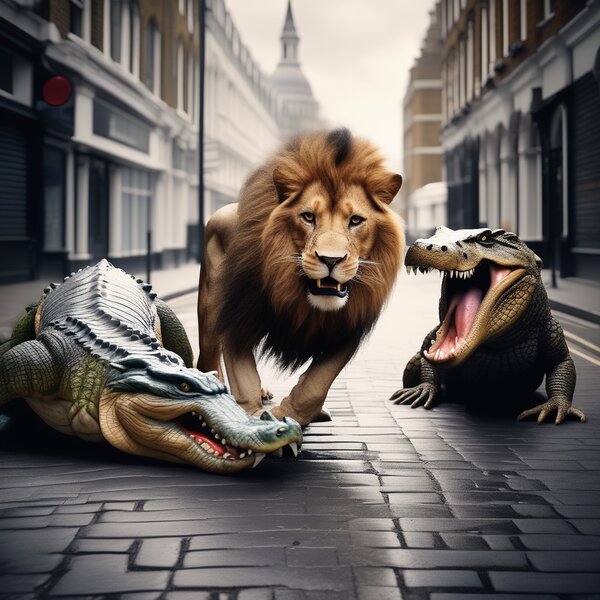} \\
        \multicolumn{4}{c}{``... a \textbf{pheonix}, a \textbf{dragon}, and a \textbf{unicorn} ... beautiful forest''} \\
        \includegraphics[width=0.24\linewidth, height=0.24\linewidth]{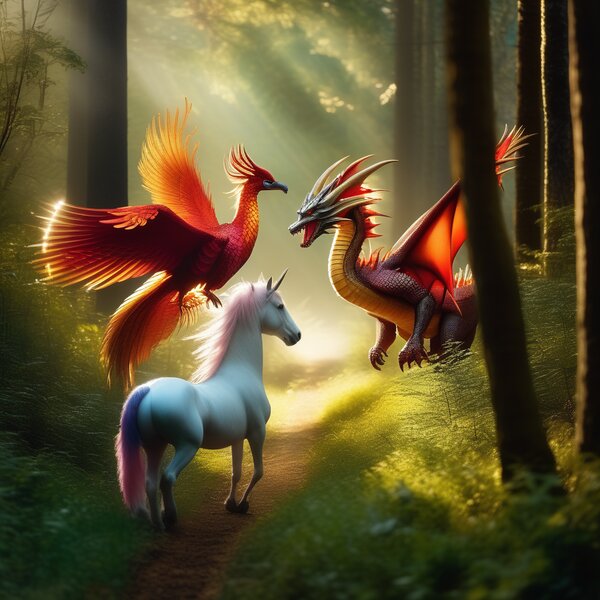} &
        \includegraphics[width=0.24\linewidth, height=0.24\linewidth]{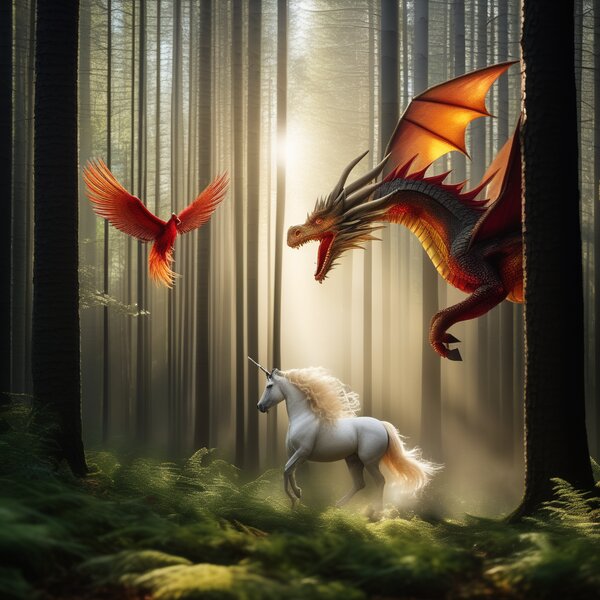} &
        \includegraphics[width=0.24\linewidth, height=0.24\linewidth]{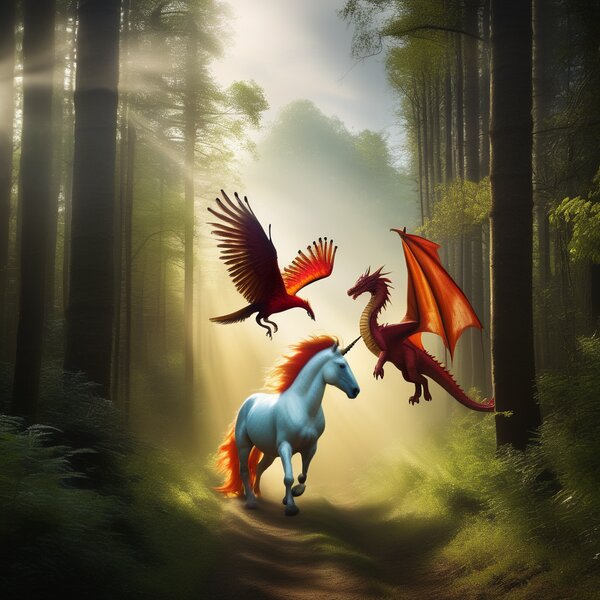} &
        \includegraphics[width=0.24\linewidth, height=0.24\linewidth]{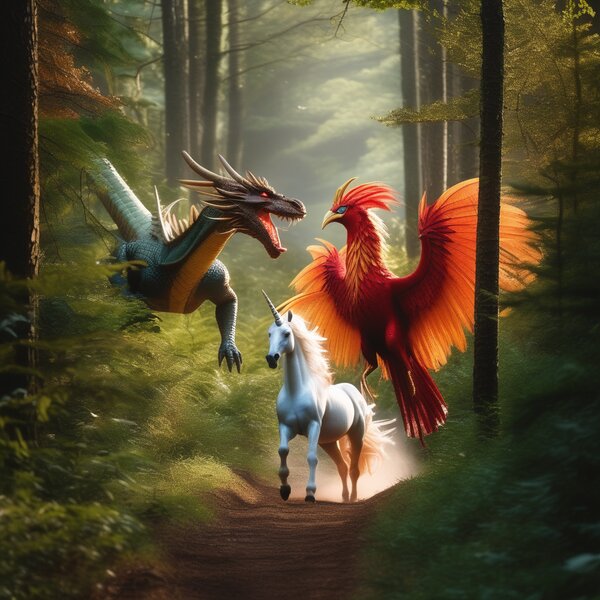} \\
    \end{tabular}
    \caption{Generated images across different seeds. Our method follows the noise-induce layouts to generate prompt-aligned images with diverse compositions.
    }
    \label{fig:diversity_small}
\end{figure}

In this section, we present both qualitative and quantitative experiments to evaluate the effectiveness of our method. We compare our approach against four training-free baseline methods: Make-It-Count (MIC)~\cite{binyamin2024make}, RPG~\cite{yang2024mastering}, Attend-and-Excite (A\&E)~\cite{chefer2023attend}, and Bounded Attention (BA) \cite{dahary2025yourself}. Since BA operates on layouts, we use an LLM to automatically provide it with layouts constructed from given prompts (denoted as LLM+BA). Furthermore, we include comparisons with LMD+\cite{lian2023llm} and Ranni~\cite{feng2024ranni}, which require training.

\subsection{Qualitative Results}

\paragraph{Layout diversity.}

We begin our experiments by showing the effectiveness of our method in generating diverse and natural layouts that adhere to the prompt. Each row of figures~\ref{fig:diversity_small},\ref{fig:diversity} depict images generated from a single prompt using different random seeds. As can be seen, our results exactly match subject descriptions, displaying proper combinations of classes, attributes and quantities, while still demonstrating unique and believable compositions.

\paragraph{Non-curated results.}

We conduct a non-curated comparison with our baseline in Figure~\ref{fig:non_curated} by sampling each method seven times, using a single prompt and the seeds $0$ to $6$. We also display the results obtained by Flux.

While LLM+BA is able to generate correct images four out of seven times, our method is able to correctly adhere to the prompt in each image without requiring an input layout. Notably, none of the other methods, including Flux, are able to generate even one sample that match the prompt, often depicting subject amalgamations due to severe leakage. Specifically, SDXL, LLM+BA and A\&E suffer from over-generation of subjects, while Flux, RPG and Ranni struggle due to under-generation. On the other hand, LMD+ is able to construct the correct quantities, but is prone to generating unnatural compositions, where subjects appear disjointed from the background.

\begin{figure}[!t]
    \setlength{\tabcolsep}{0.002\textwidth}
    \scriptsize
    \centering
    \begin{tabular}{c c c c c}
        & ``... in the snow'' & ``... in the snow'' & ``... in a video game'' \\
        \includegraphics[width=0.135\linewidth]{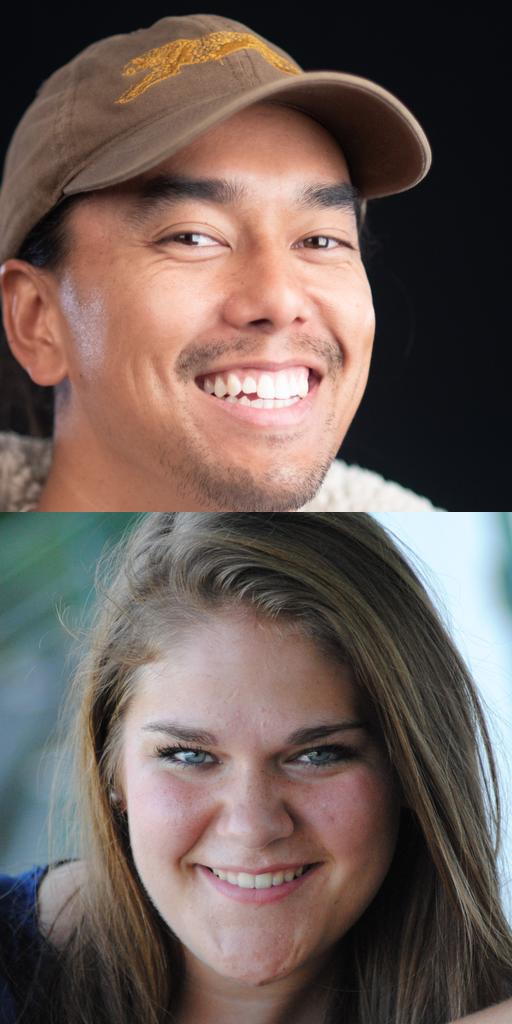} &
        \includegraphics[width=0.27\linewidth]{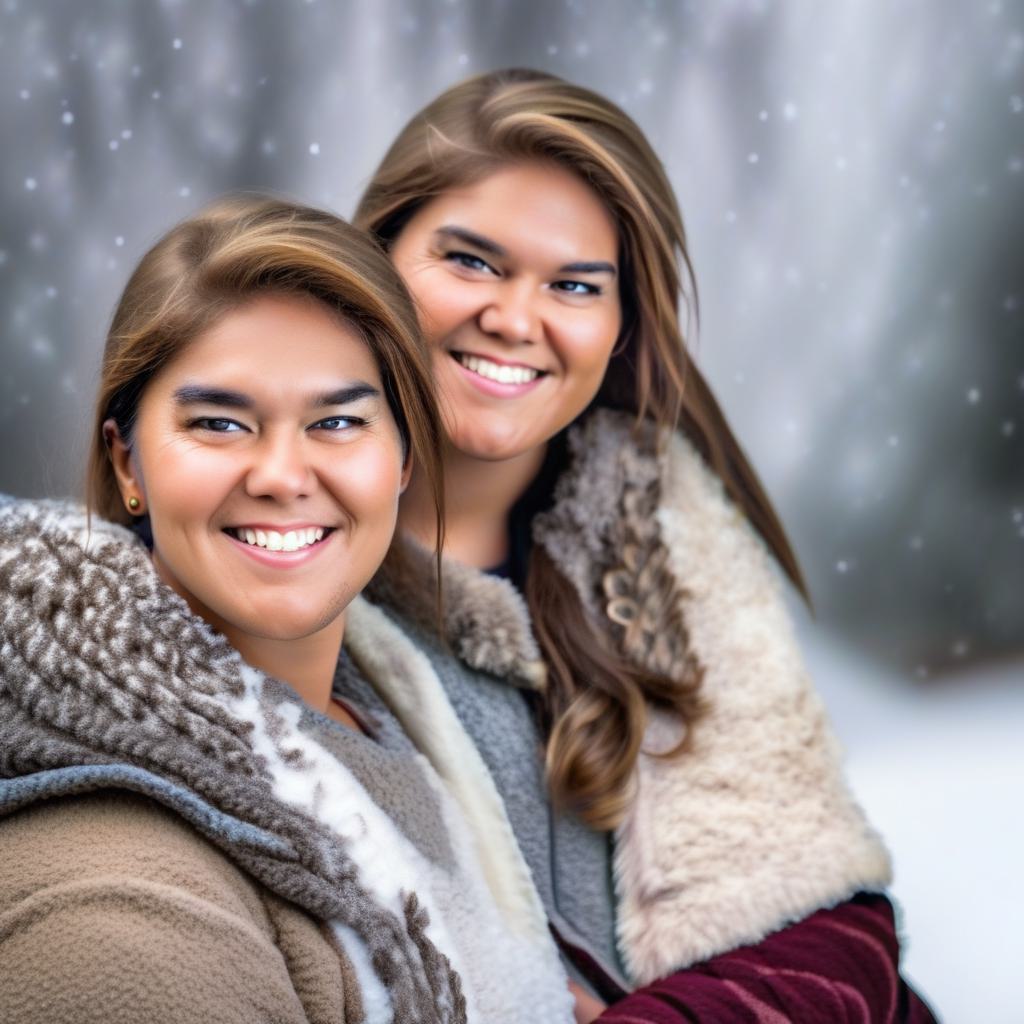} &
        \includegraphics[width=0.27\linewidth]{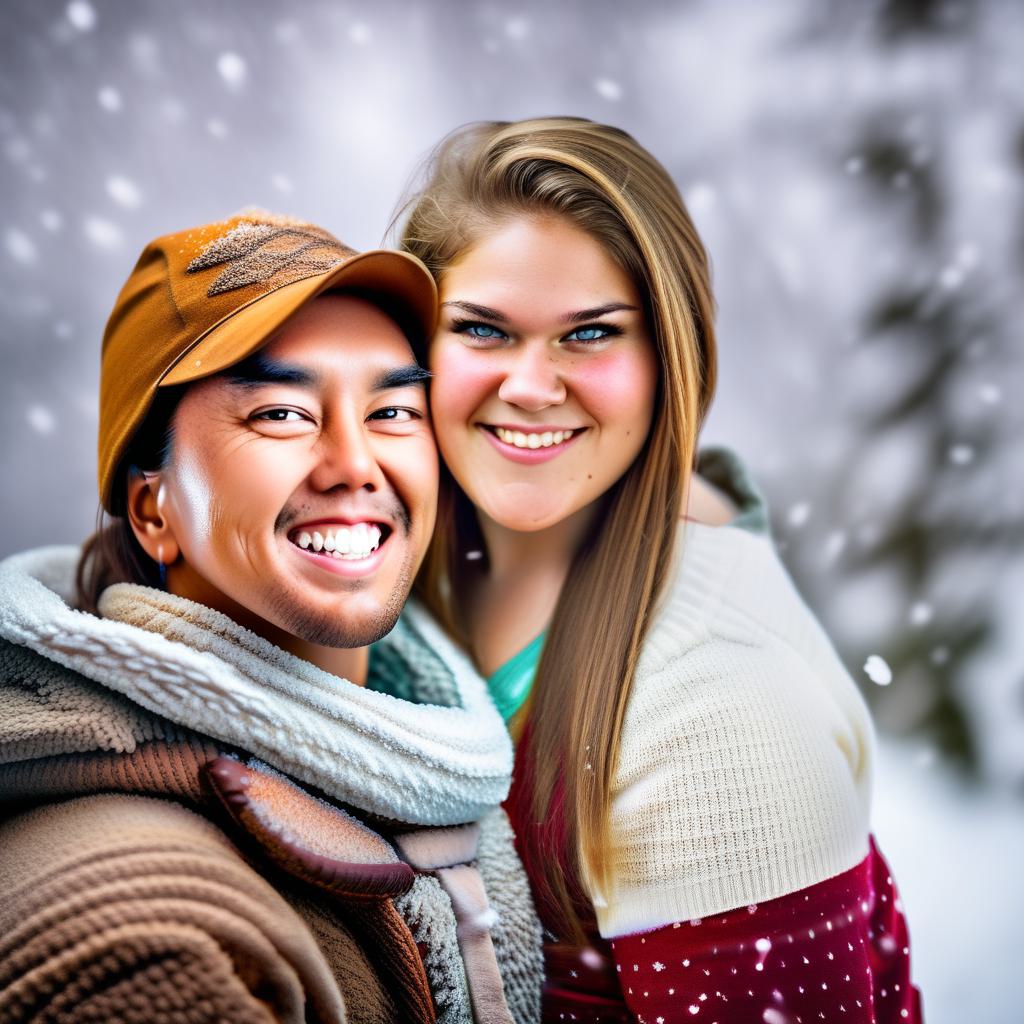} &
        \includegraphics[width=0.27\linewidth]{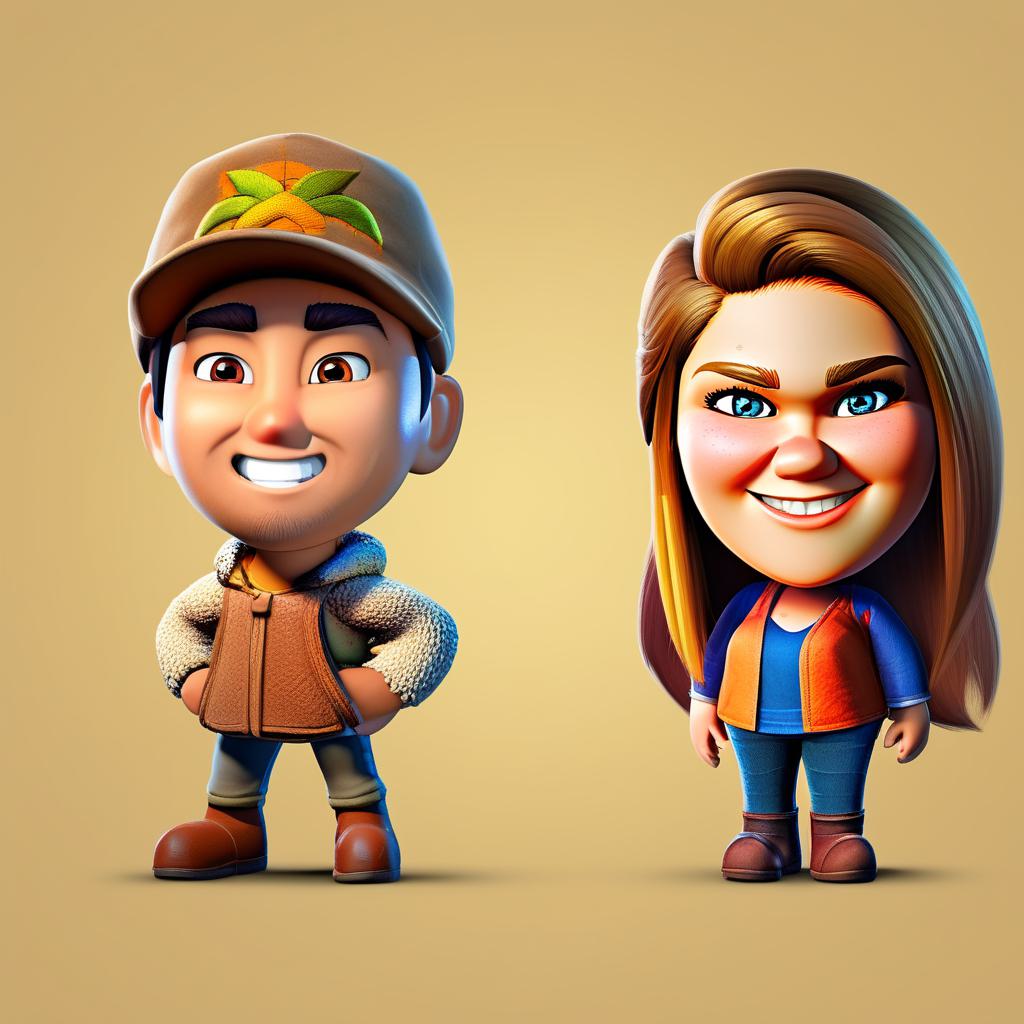} 
        \\
        & ``... as chefs in the kitchen'' & ``... as chefs in the kitchen'' & ``Anime painting ...'' \\
        \includegraphics[width=0.135\linewidth]{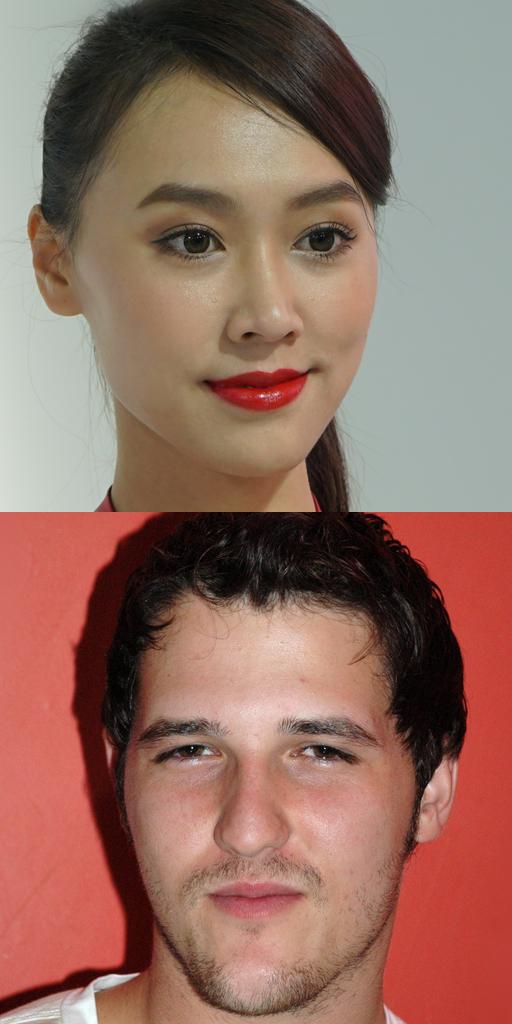} &
        \includegraphics[width=0.27\linewidth]{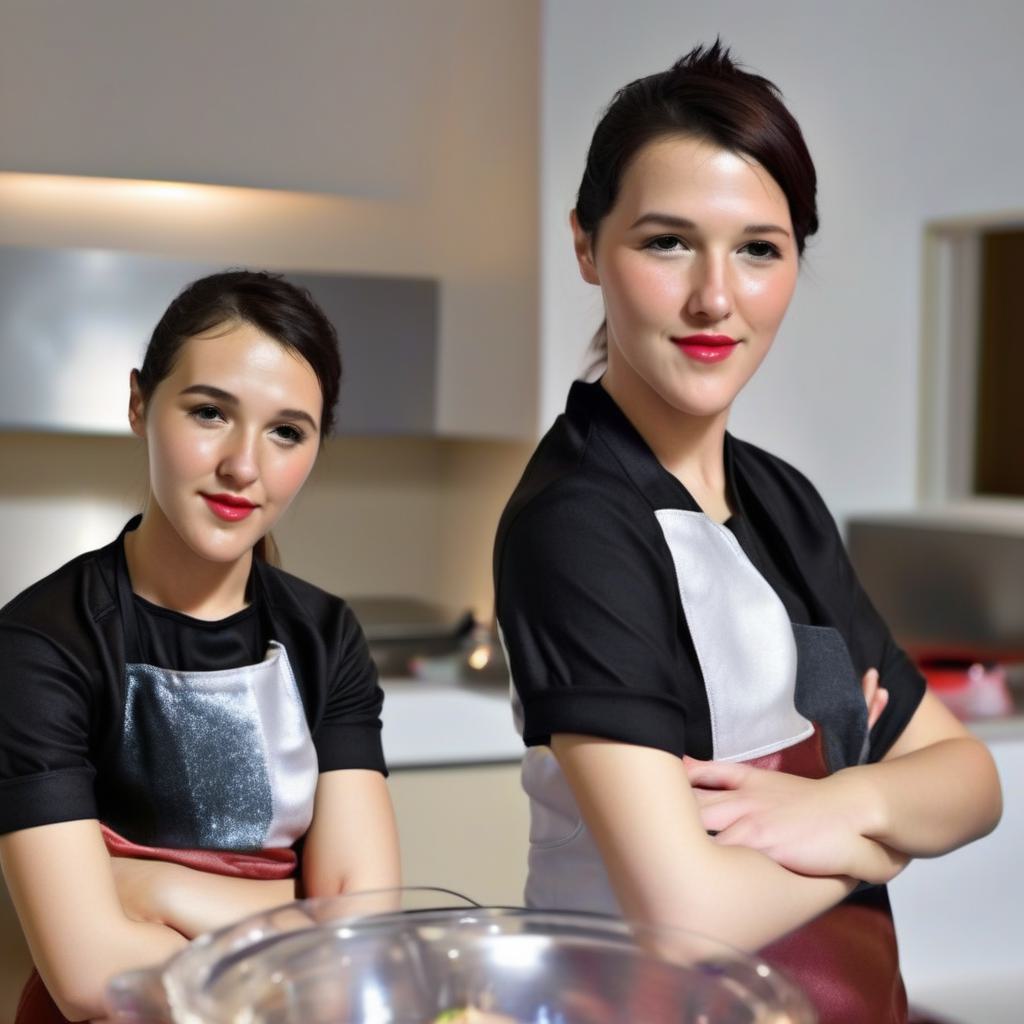} &
        \includegraphics[width=0.27\linewidth]{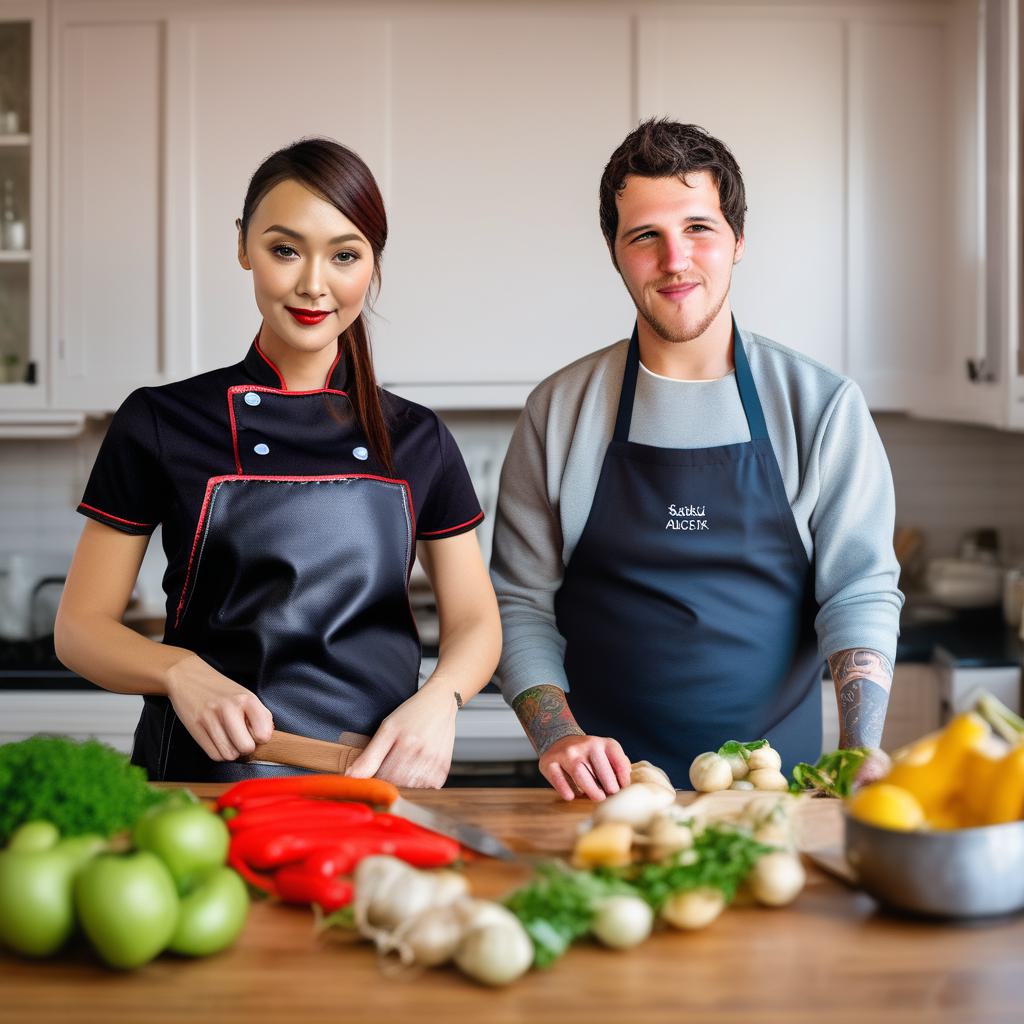} &
        \includegraphics[width=0.27\linewidth]{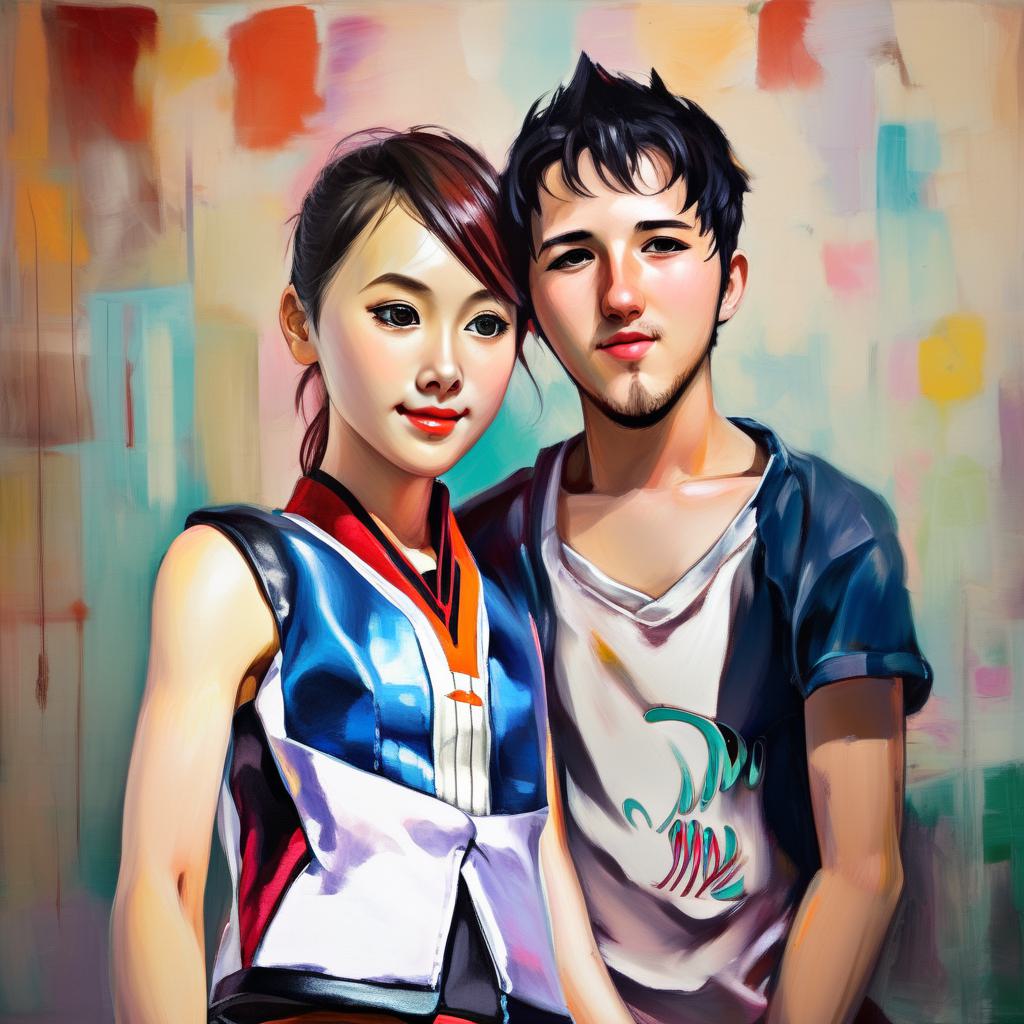} & 
        \\
         & ``... in a coffee shop'' & ``... in a coffee shop'' & ``... as pop figures'' \\
        \includegraphics[width=0.135\linewidth]{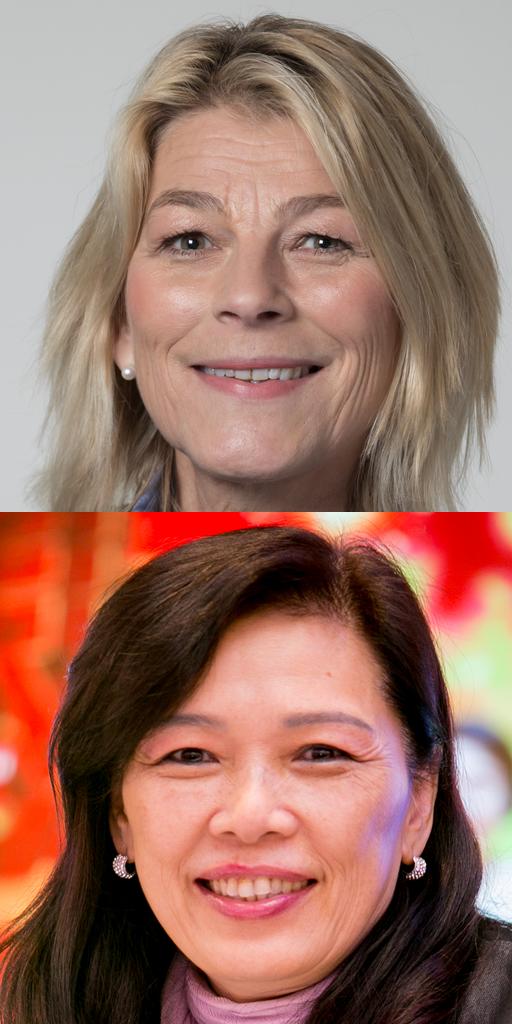} &
        \includegraphics[width=0.27\linewidth]{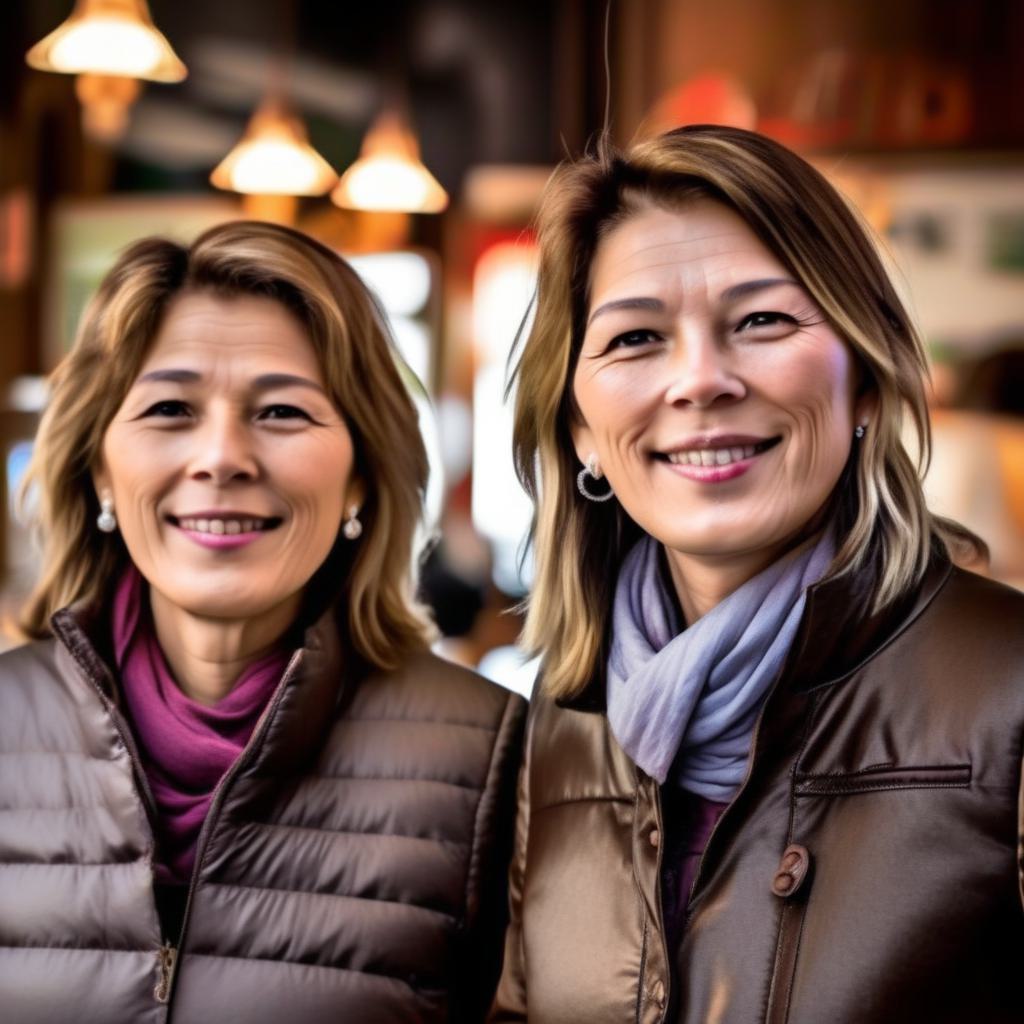} &
        \includegraphics[width=0.27\linewidth]{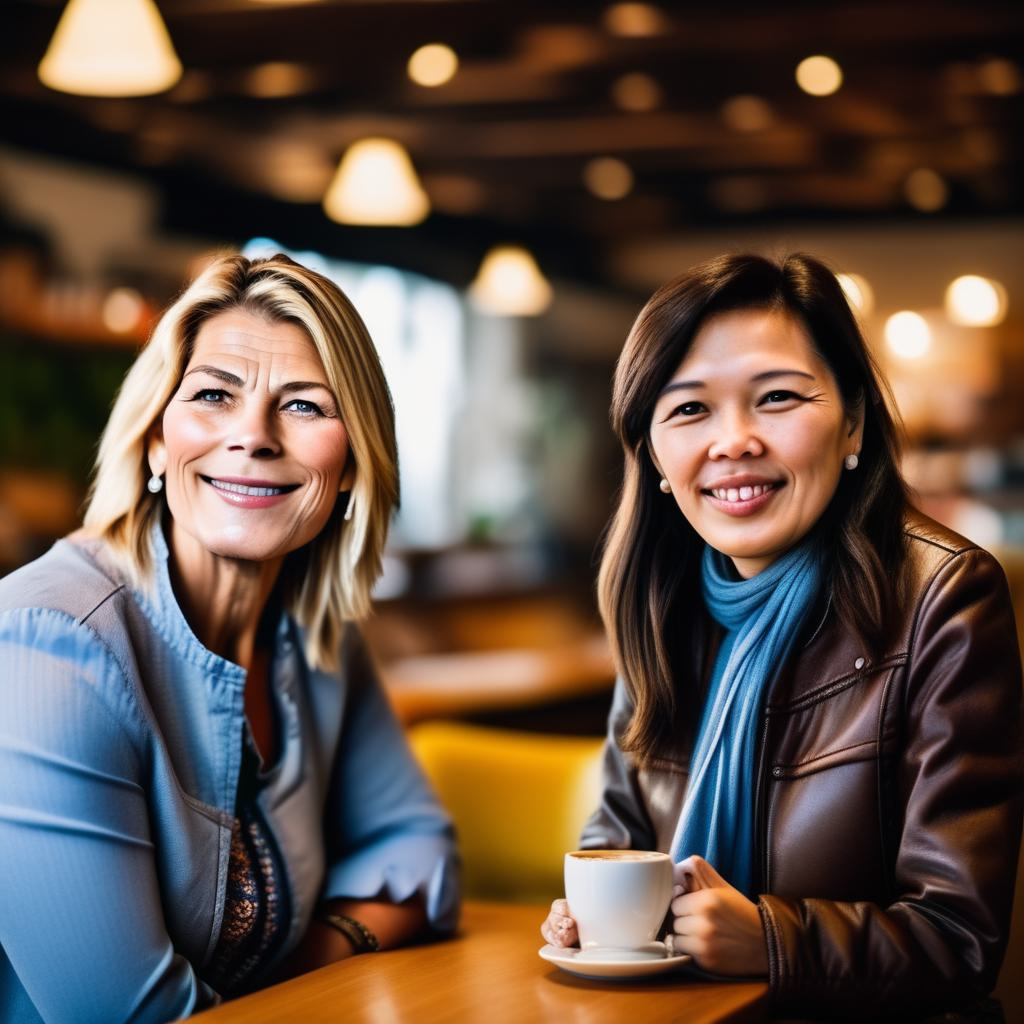} &
        \includegraphics[width=0.27\linewidth]{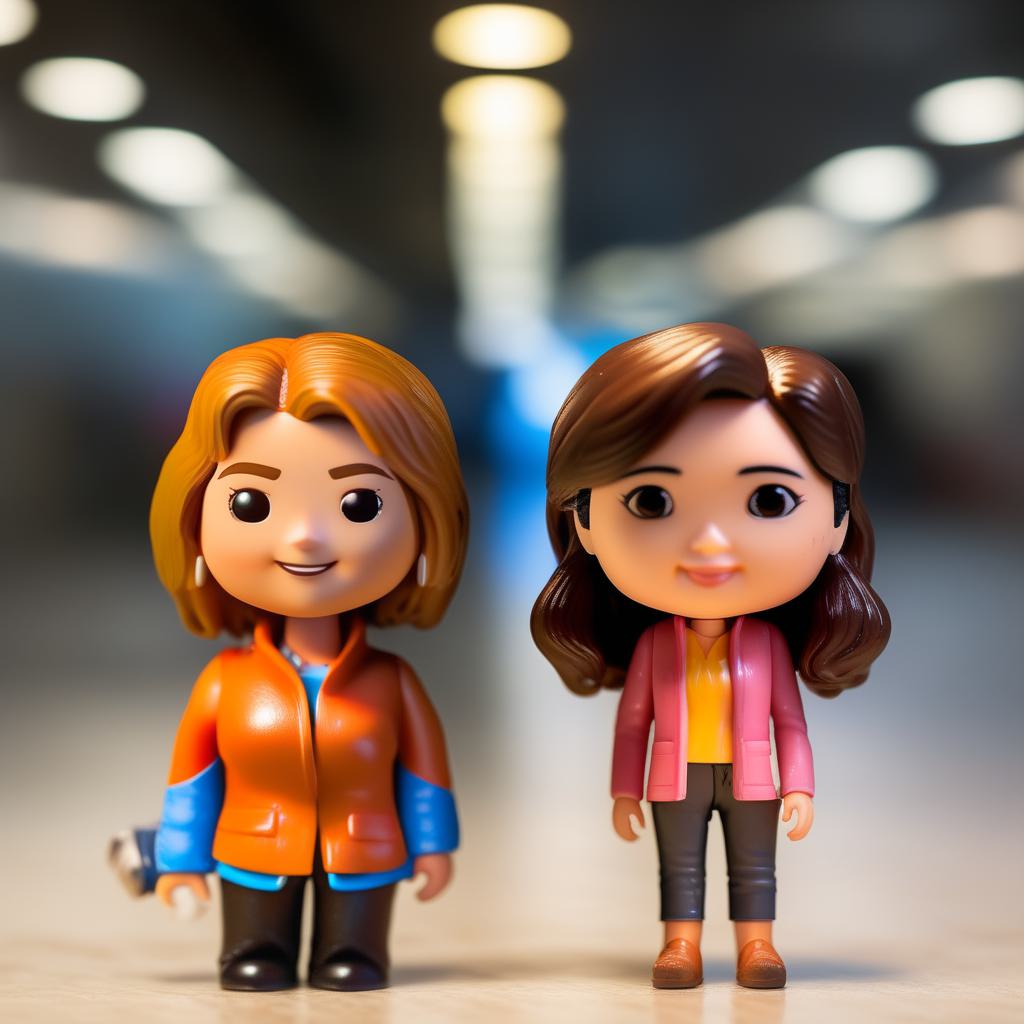} 
        \\
        Inputs & w/o Be Decisive & \multicolumn{2}{c}{w/ Be Decisive}
    \end{tabular}
    \caption{Results of integrating our method with an existing personalization method, enabling the generation of multiple personalized individuals within the same image.
    }
    \label{fig:personalization}
\end{figure}

\begin{figure*}[h!]
    \centering
    \setlength{\tabcolsep}{0.001\textwidth}
    {\small
    \begin{tabular}{c c c c c c c}
        \multicolumn{7}{c}{"A \textbf{polar bear}, a \textbf{grizzly bear}, a \textbf{panda bear}, and a \textbf{koala bear}"} \\
        \includegraphics[width=0.14\textwidth]{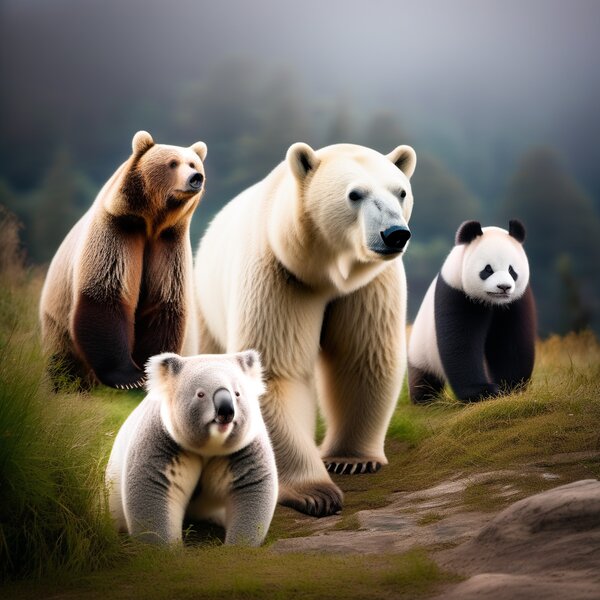} &
        \includegraphics[width=0.14\textwidth]{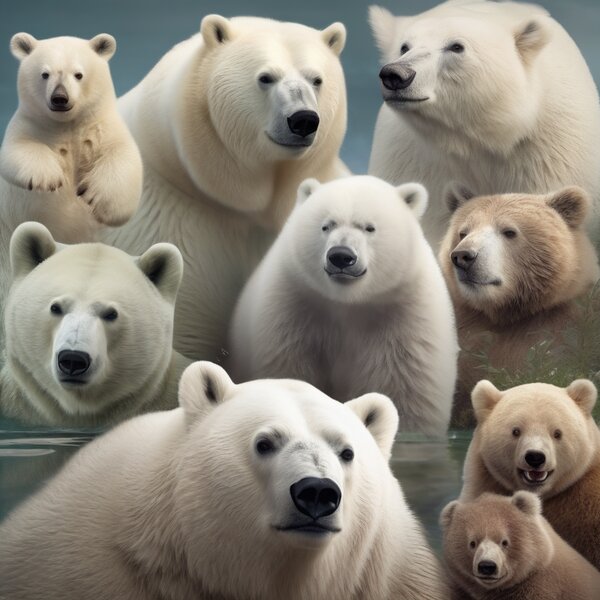} &
        \includegraphics[width=0.14\textwidth]{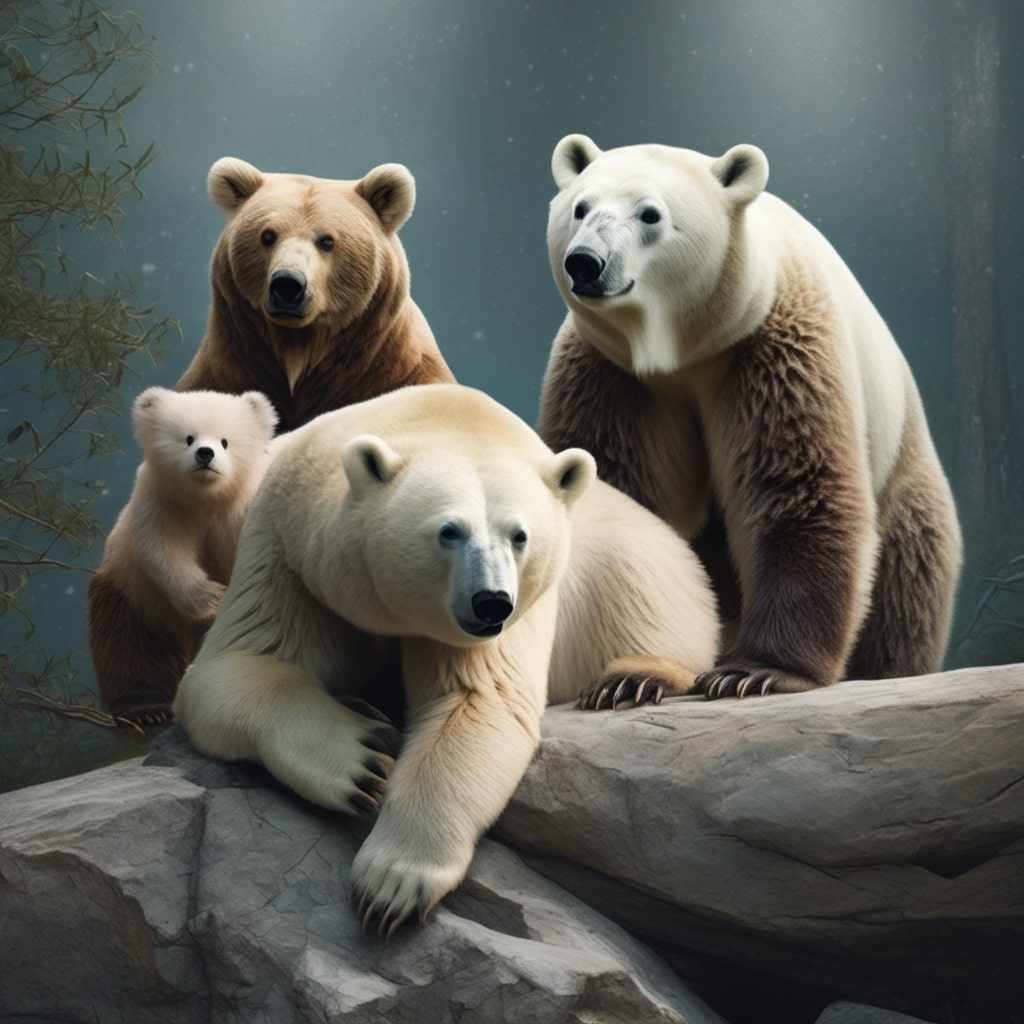} &
        \includegraphics[width=0.14\textwidth]{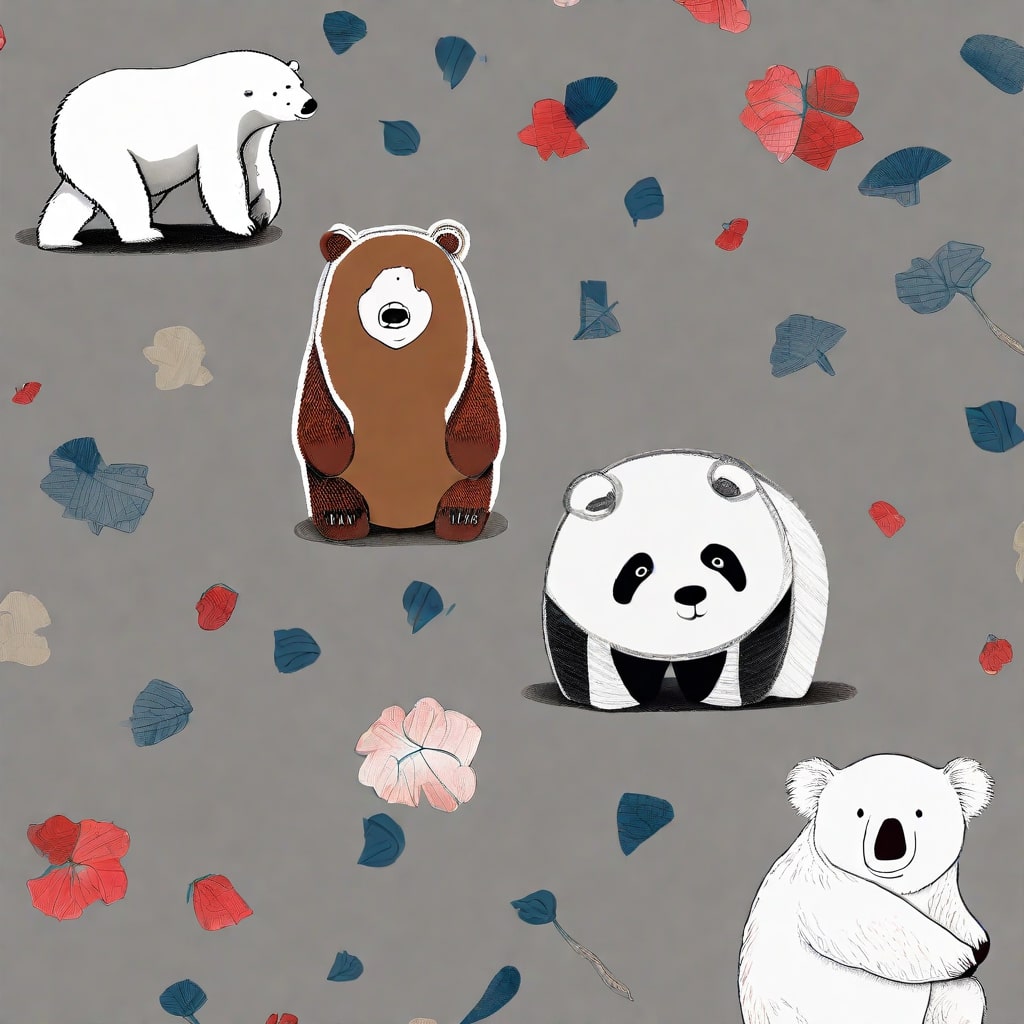} &
        \includegraphics[width=0.14\textwidth]{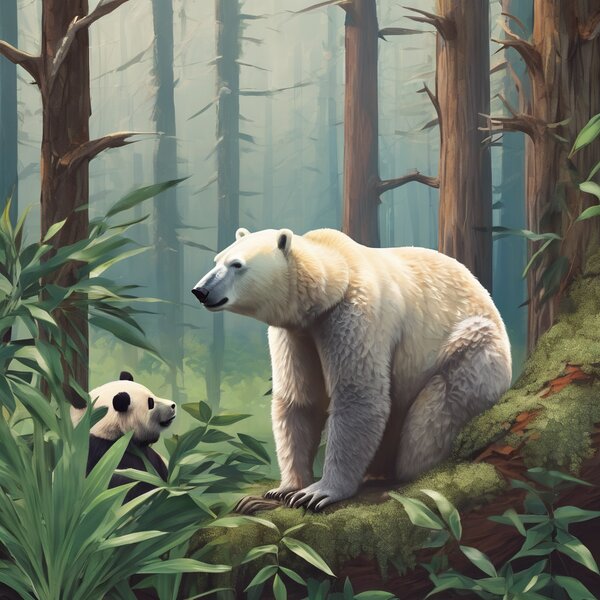} &
        \includegraphics[width=0.14\textwidth]{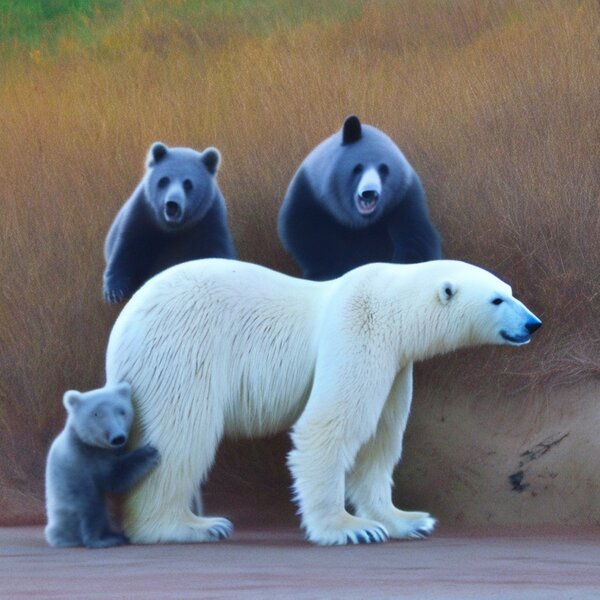} &
        \includegraphics[width=0.14\textwidth]{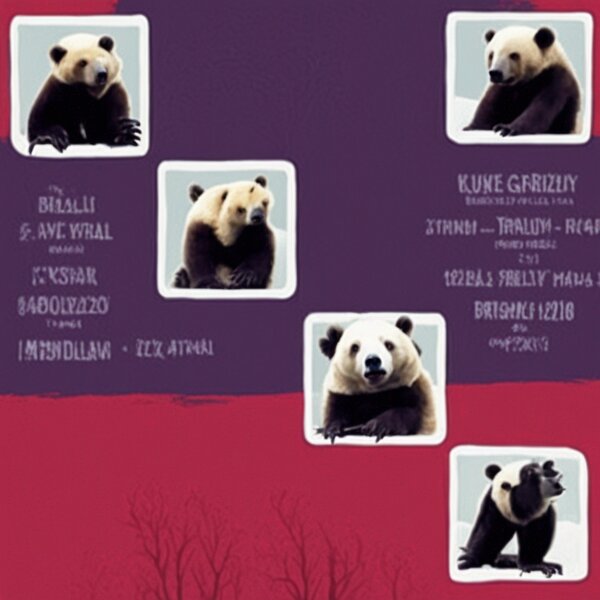} \\

        \multicolumn{7}{c}{"\textbf{Two teddy bears} and \textbf{four red toy cars} on a white carpet"} \\
        \includegraphics[width=0.14\textwidth]{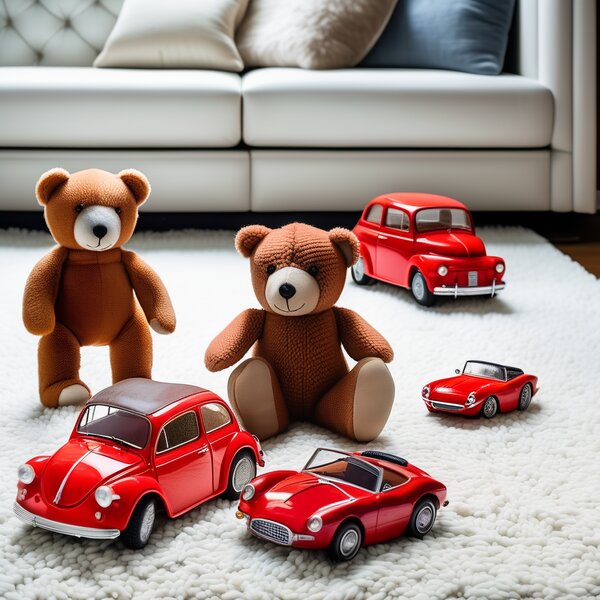} &
        \includegraphics[width=0.14\textwidth]{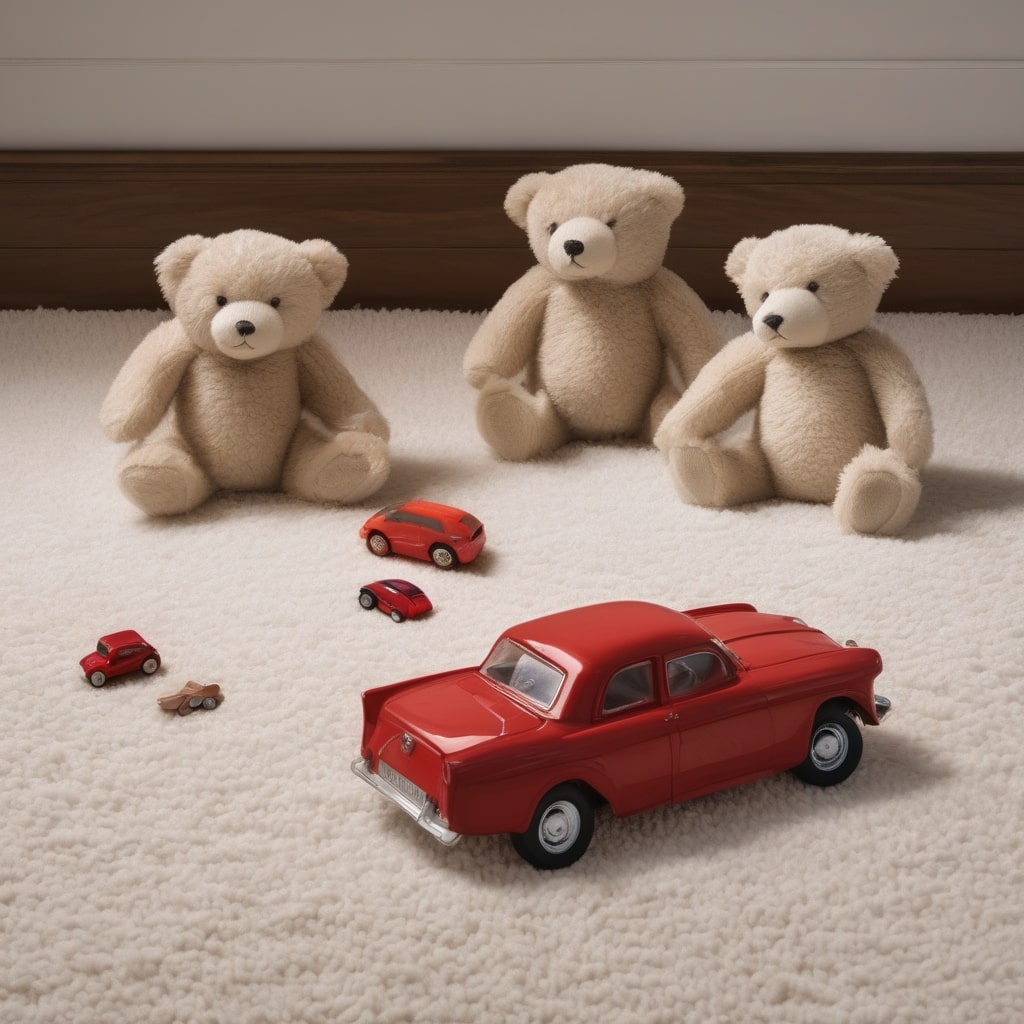} &
        \includegraphics[width=0.14\textwidth]{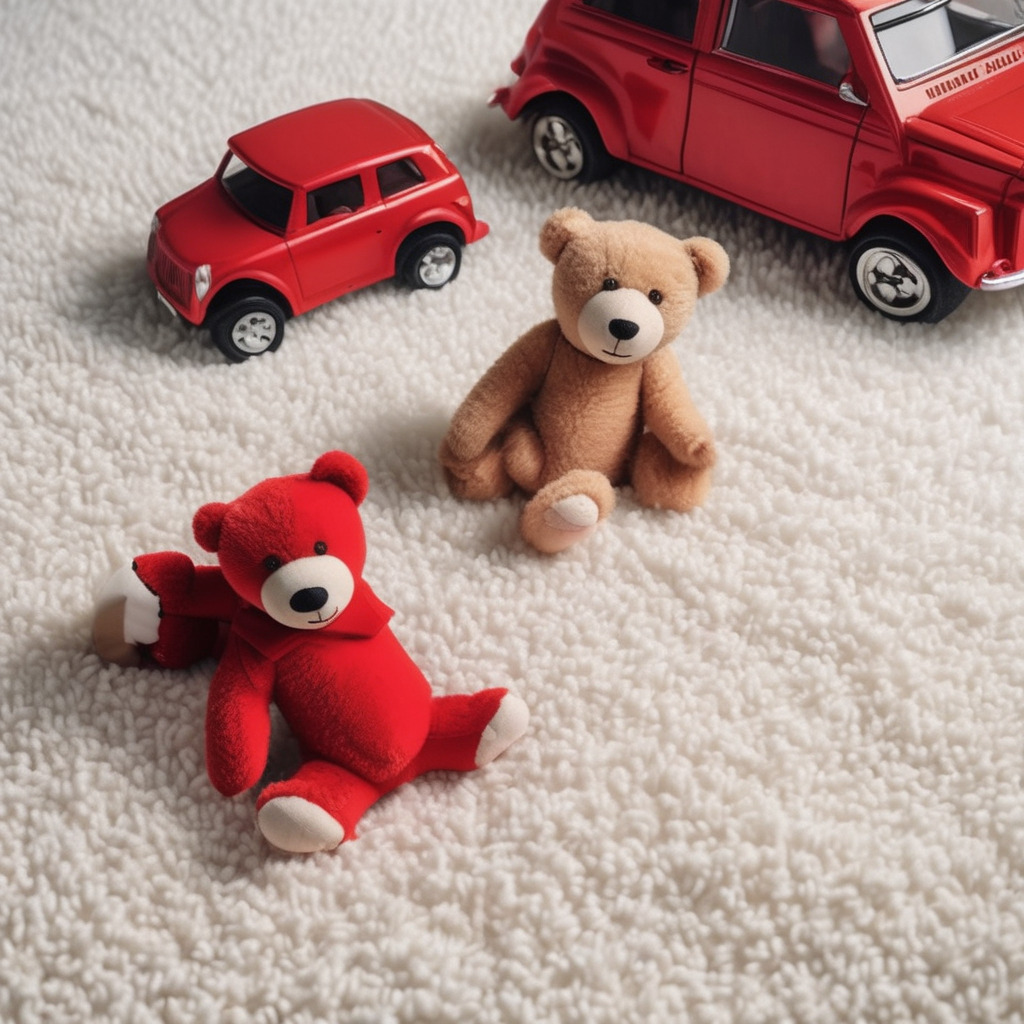} &
        \includegraphics[width=0.14\textwidth]{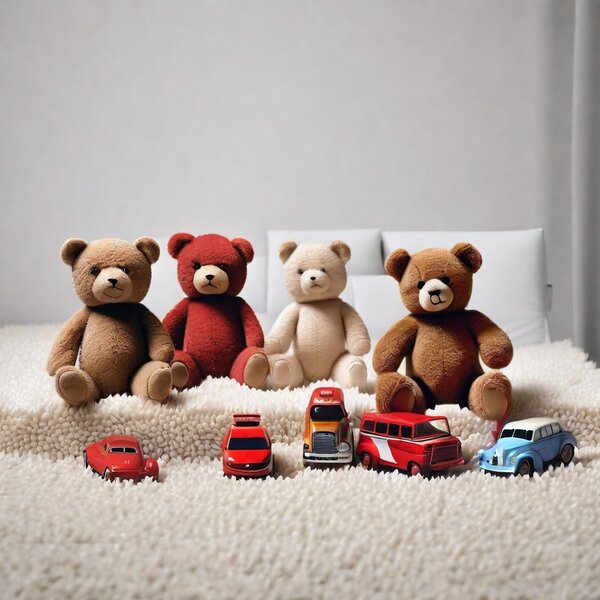} &
        \includegraphics[width=0.14\textwidth]{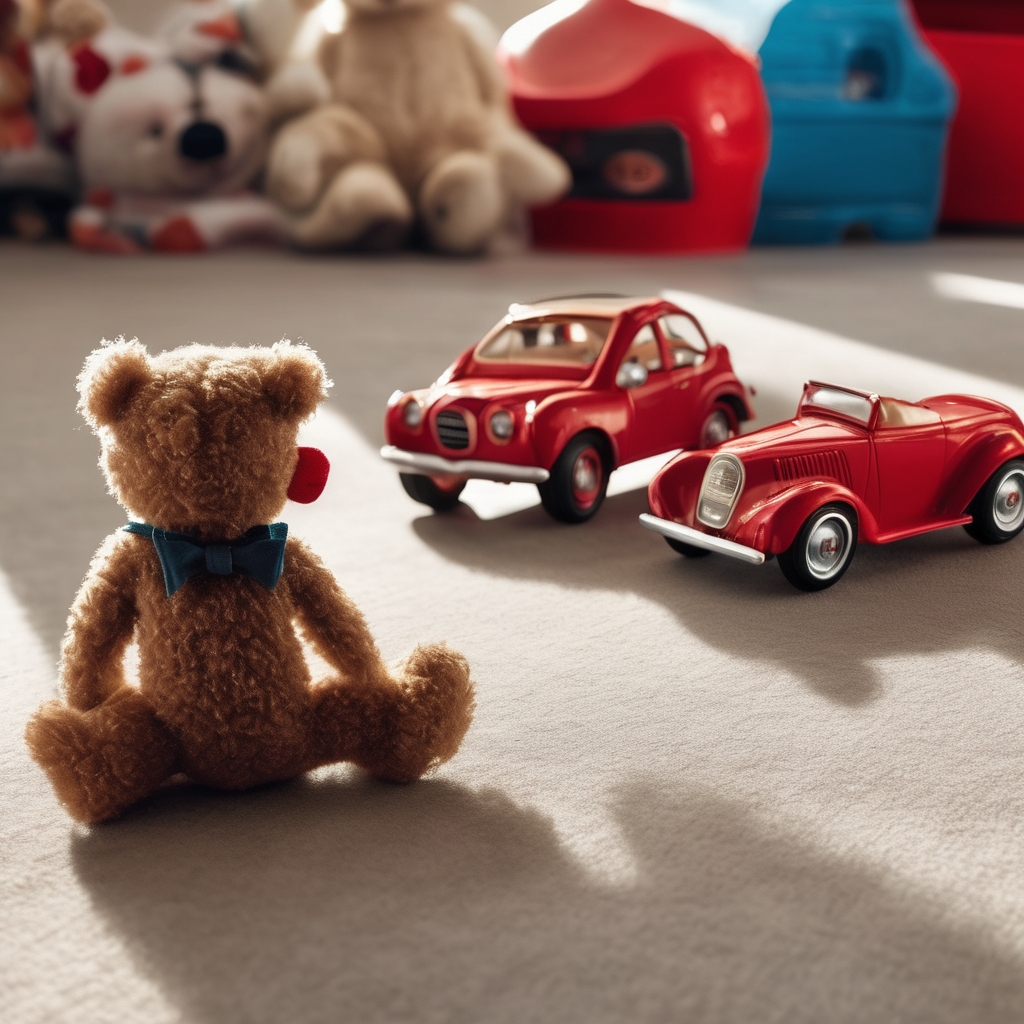} &
        \includegraphics[width=0.14\textwidth]{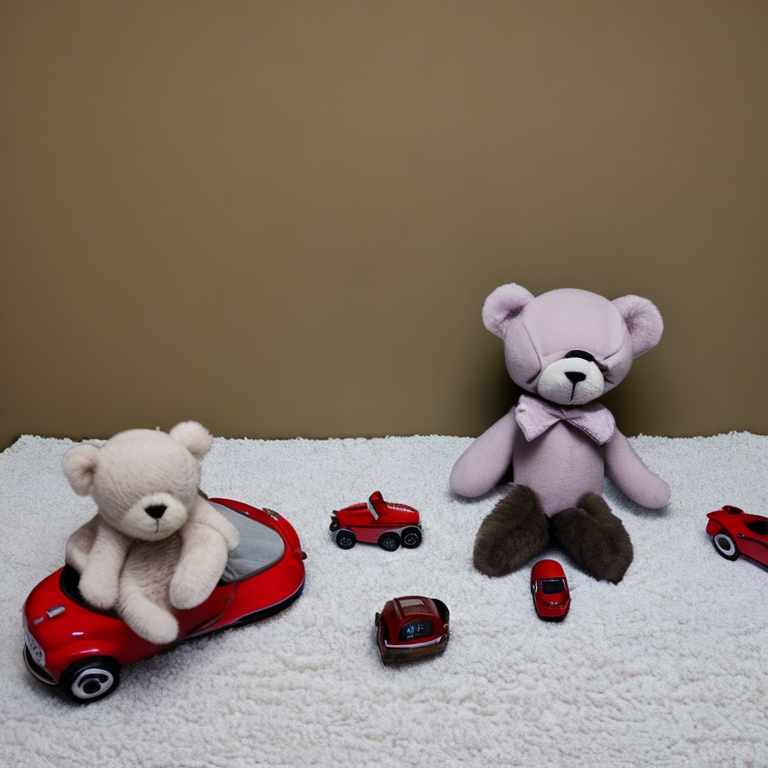} &
        \includegraphics[width=0.14\textwidth]{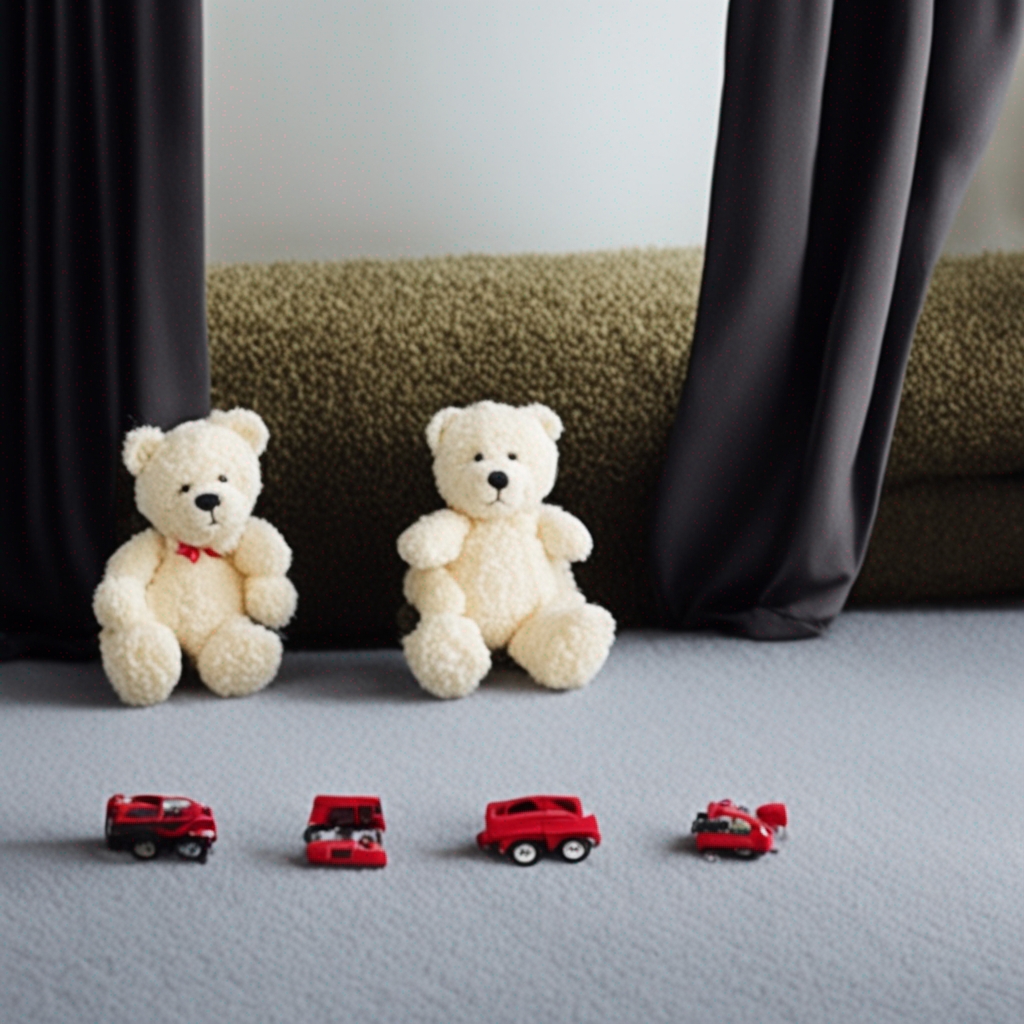} \\

        Ours & SDXL & A\&E & LLM+BA & RPG & Ranni & LMD+ \\
    \end{tabular}
    }
    \captionof{figure}{\textbf{Qualitative comparison} of our method with baseline methods. We provide more examples in the supplement.
    }
    \label{fig:comparisons}
\end{figure*}

\begin{figure}
    \setlength{\tabcolsep}{0.002\columnwidth}
    \centering
    \includegraphics[width=\columnwidth]{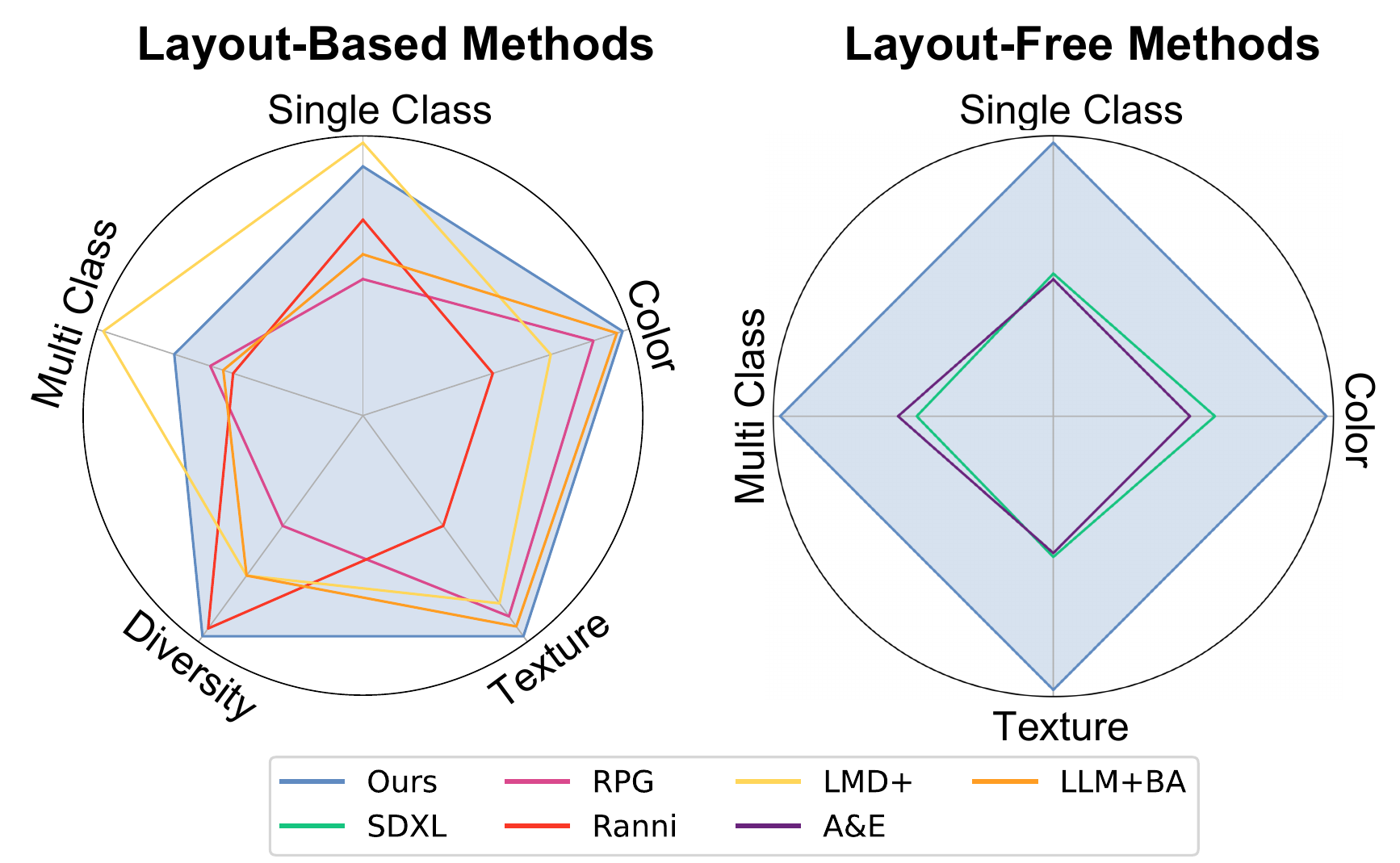}
    \caption{Quantitative comparison of our method against layout-based and layout-free methods. Results demonstrate that while other methods face trade-offs between metrics, our method consistently achieves high scores across all metrics.
    }
    \label{fig:radar_chart}
\end{figure}

\paragraph{Multiple Personalized Subjects.}
Leakage between subjects is particularly noticeable when generating personalized subjects. Here, we show that our method can be seamlessly integrated with an existing personalization method to facilitate the generation of multiple personalized subjects. Specifically, we utilize a method trained to generate specific individuals by injecting personalized features through the cross-attention layers of a text-to-image model~\cite{patashnik2025nested}. This method does not inherently support the generation of two individuals, as demonstrated in Figure~\ref{fig:personalization}. However, combining this method with ours, enables the accurate generation of diverse images with multiple individuals.

\paragraph{Comparisons with baselines.}

We present a qualitative comparison in Figure~\ref{fig:comparisons}. All other methods struggle to generate multi-subjects prompts due to leakage. In the first row, none of the competing methods are able to generate the distinct characteristics of each of the bears. In the second row, they either generate the wrong number of subjects, or leak the colors of the carpet or the cars into the teddy bears. Specifically, the current LLM-based methods exhibit either subpar control over subject quantities (LLM+BA, RPG, Ranni), or unnatural  grid-like subject arrangements (LLM+BA, LMD+). In comparison, our method successfully generates prompt-aligned images with natural-looking compositions.

\subsection{Quantitative Results}

\paragraph{Dataset evaluation.}

We perform quantitative evaluation on the T2I-CompBench dataset~\cite{huang2023t2i}, assessing our method's performance across the following key aspects: multi-class compositions, attribute binding, and numeracy. We further measure layout diversity, which quantifies the variability of generated compositions across different seeds. We summarize the results in Figure~\ref{fig:radar_chart}, and refer to the supplement for the full table.

To measure layout diversity, we sample 20 random prompts from CompBench's single-class dataset and generate five images per prompt using five random seeds. We align all five layouts by maximizing the $\text{IoU}$ between them using the Hungarian algorithm. Diversity is quantified as the average $1 - \text{IoU}$ between all layout pairs of the same prompt. As reported, our method achieves significantly higher diversity than the baseline, preserving the innate variability of the model's prior, in contrast to the limited diversity of LLM-based methods.

All other metrics were assessed on 200 prompts, sampled from the respective category in CompBench. Color and texture binding was evaluated using BLIP-VQA~\cite{huang2023t2i}, while single-class and multi-class compositions were evaluated using the F1 scores between ground-truth subject quantities and the quantities computed by GroundedSAM~\cite{ren2024grounded} on the generated images. While other methods are tailored towards enhancing specific metrics, our approach consistently achieves high performance across all measurements, surpassing competitors in most cases.

Lastly, since MIC is limited to single-class prompts, we only measure its performance on this specific metric. Our method achieves a score of 0.837, compared to MIC's score of 0.772.

\begin{table}
    \centering
    \setlength{\tabcolsep}{0.005\textwidth}
    \captionof{table}{
    User study results.
    }
    \begin{tabular}{l c c c c c c}
        \toprule
         & SDXL & LLM+BA & RPG & Ranni & LMD+ & MIC \\
        \midrule
        Our score vs. & 0.74 & 0.87 & 0.96 & 0.89 & 0.9 & 0.53 \\
    
        \bottomrule
    \end{tabular}
    \label{table:user_study}
\end{table}

\paragraph{User study.}

The automatic metrics in Figure~\ref{fig:radar_chart} fail to detect semantic leakage, as they rely on models trained on real images, where such issues do not arise. To address this limitation, we conduct a user study.
We utilize ChatGPT to generate 25 prompts enlisting three to four visually similar, but distinct, animals, with an appropriate background. For each prompt, participants were shown 10 images: five generated by our method, and five generated by a competing method. Users were than tasked with selecting images with realistic compositions that accurately reflect the prompt. For evaluation against MIC, we additionally generate five prompts with single-class quantities. We collected 192 responses from 32 participants. Table~\ref{table:user_study} reports the conditional probability of a selected image being generated by our method versus competitors. The results showcase our method's superiority in handling complex multi-subject prompts, with our scores substantially improving over layout-based methods. Notably, even though MIC is specifically designed to tackle single-class quantities, our method receives comparable scores, while still being versatile enough to support more complex prompts.

\section{Conclusions}

We have addressed the notorious difficulty of generating multiple distinct subjects from complex prompts in text-to-image diffusion models. Recognizing that inter-subject leakage is the primary issue, and that bounding mutual attention offers a viable solution, we designed a mechanism to define a layout for controlling inter-subject attention. Our key contribution lies in using the natural latent layout defined by the initial noise of the model, rather than imposing an external layout. By making only small adjustments on-the-fly, our approach remains rooted in the original distribution of the model, benefiting from denoising a signal already close to that distribution. Empirical evaluations confirm that this strategy provides a stronger balance between text-image alignment and visual diversity compared to layout-driven alternatives.

It is important to recognize that the multi-subject generation problem is intrinsically tied to the pretrained model’s prior. When the underlying network has not been sufficiently exposed to images featuring multiple distinct subjects, its learned distribution may be ill-equipped to handle complex multi-subject arrangements. As a result, any approach aiming to improve multi-subject generation, ours included, must contend with these fundamental distributional constraints. Although our method outperforms existing alternatives, there remains a ceiling imposed by the model training data, restricting how effectively multi-subject prompts can be addressed in practice.

The main limitation of our method lies in the computational cost of the iterative guidance and its tendency to push the optimized latent away from the prior distribution. In the future, we aim to explore regularization techniques to keep the latent closer to its original distribution or replace the optimization process with feature injection from a control map representing the target clusters.

\begin{figure*}[hp!]
    \setlength{\tabcolsep}{1pt}
    \centering
    \begin{tabular}{c c c c c c}
        \\
        \\
        \multicolumn{6}{c}{"... movie scene with a \textbf{meerkat}, a \textbf{bunny}, a \textbf{fox}, and a \textbf{frog} dancing hula in Hawaii"} \\
        \includegraphics[width=0.15\textwidth]{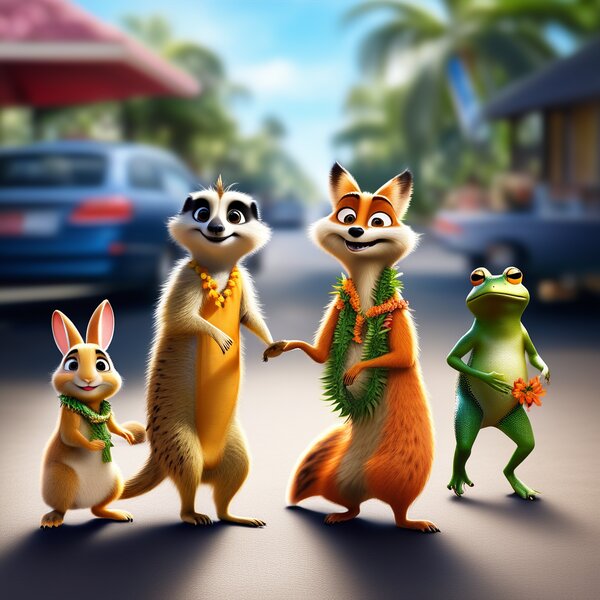} &
        \includegraphics[width=0.15\textwidth]{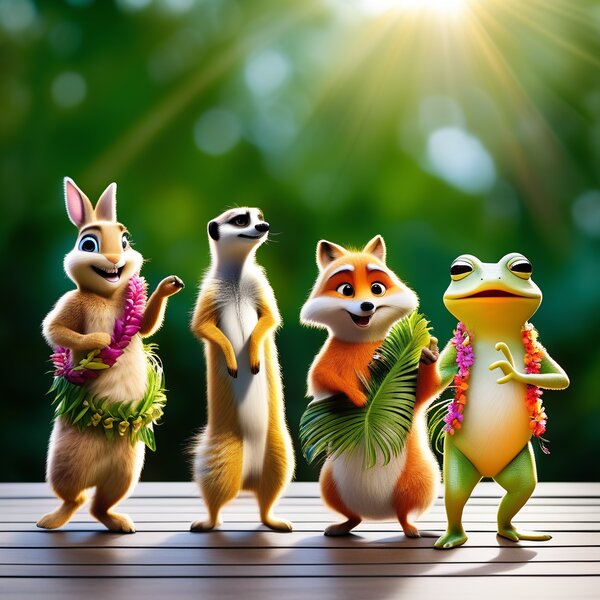} &
        \includegraphics[width=0.15\textwidth]{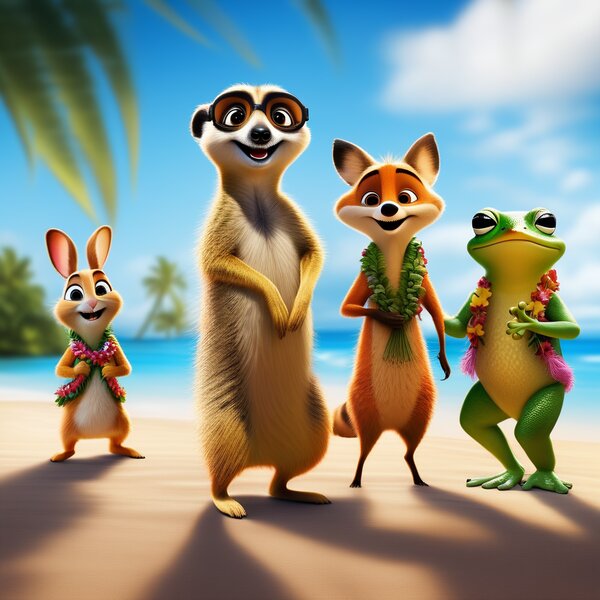} &
        \includegraphics[width=0.15\textwidth]{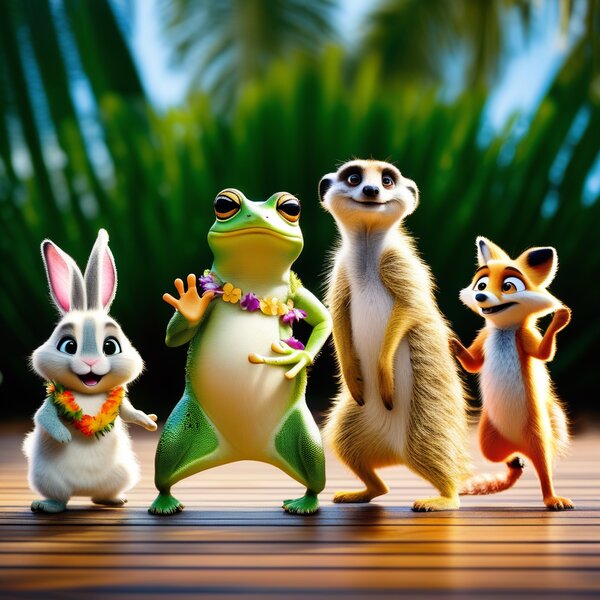} &
        \includegraphics[width=0.15\textwidth]{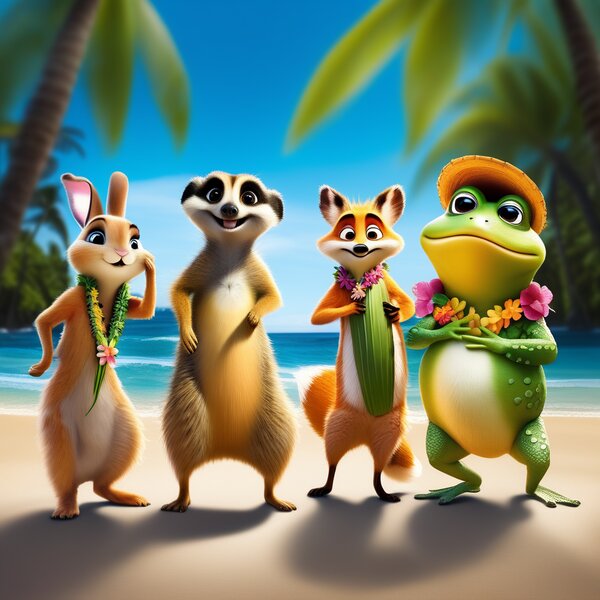} &
        \includegraphics[width=0.15\textwidth]{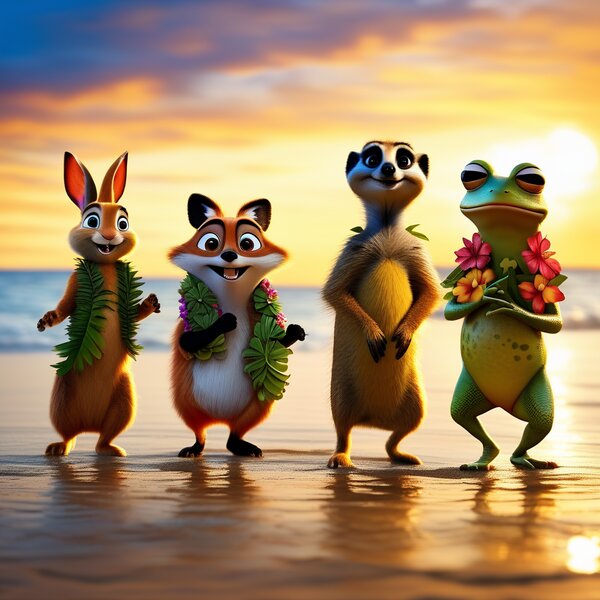} \\
        \multicolumn{6}{c}{"... serene and exotic scene with a \textbf{wooden cabin}, a \textbf{brick house}, and the \textbf{golden temple} on a hill"} \\
        \includegraphics[width=0.15\textwidth]{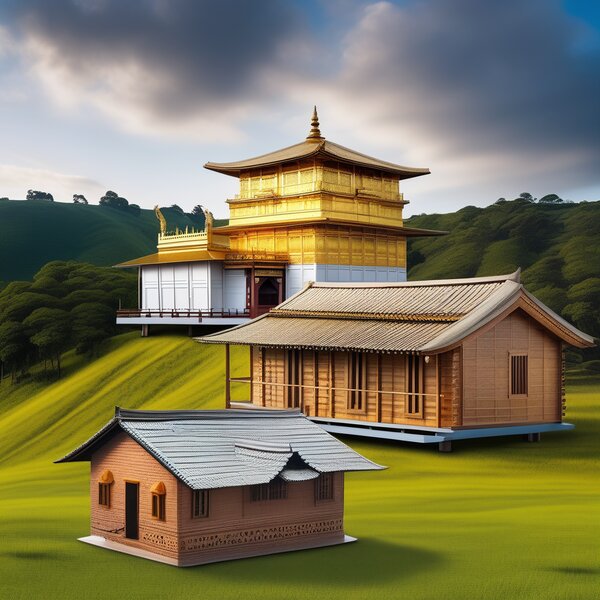} &
        \includegraphics[width=0.15\textwidth]{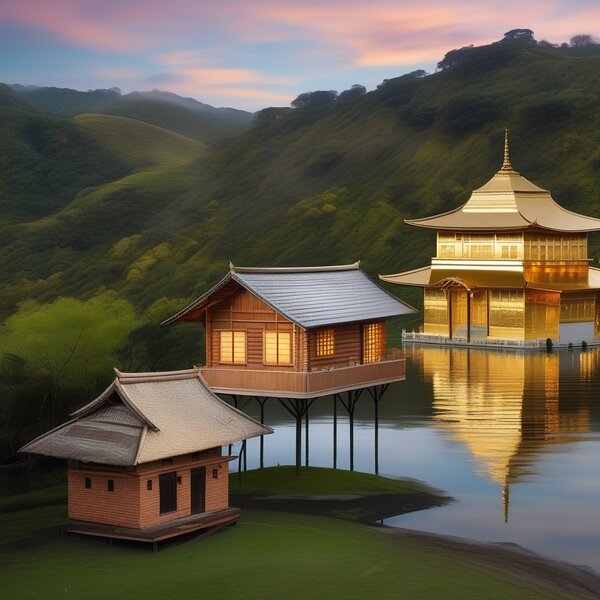} &
        \includegraphics[width=0.15\textwidth]{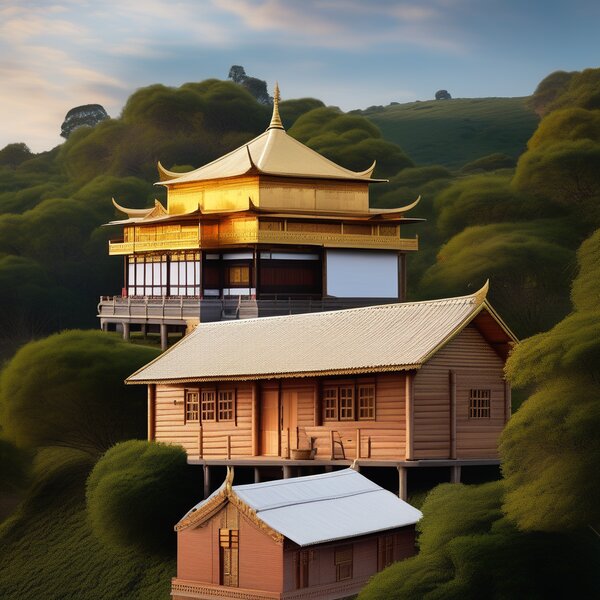} &
        \includegraphics[width=0.15\textwidth]{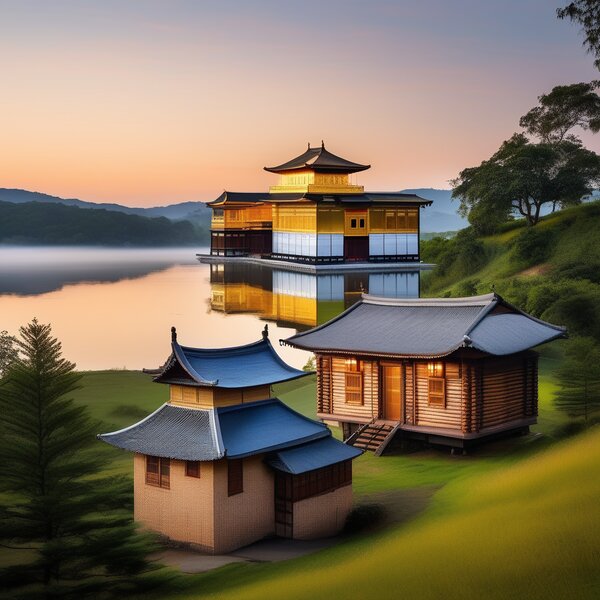} &
        \includegraphics[width=0.15\textwidth]{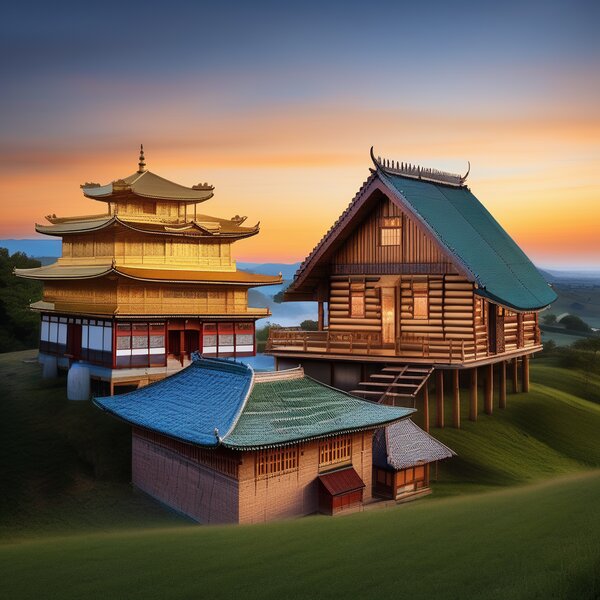} &
        \includegraphics[width=0.15\textwidth]{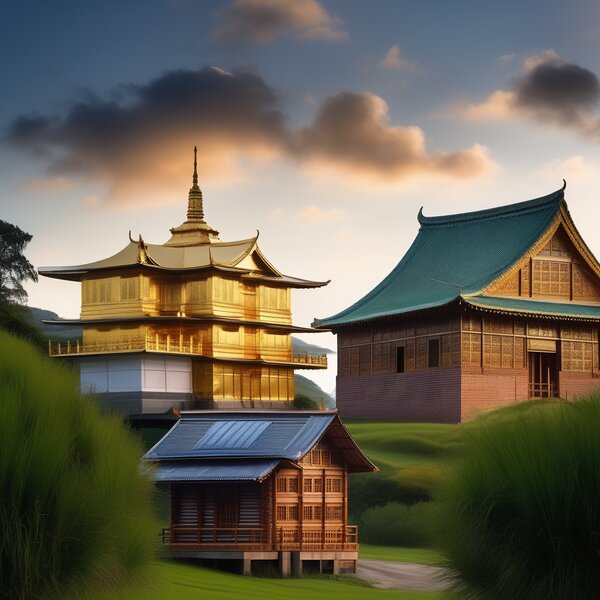} \\
        \multicolumn{6}{c}{"... a \textbf{sailboat}, a \textbf{motorboat}, and a \textbf{kayak} in a beautiful lake"} \\
        \includegraphics[width=0.15\textwidth]{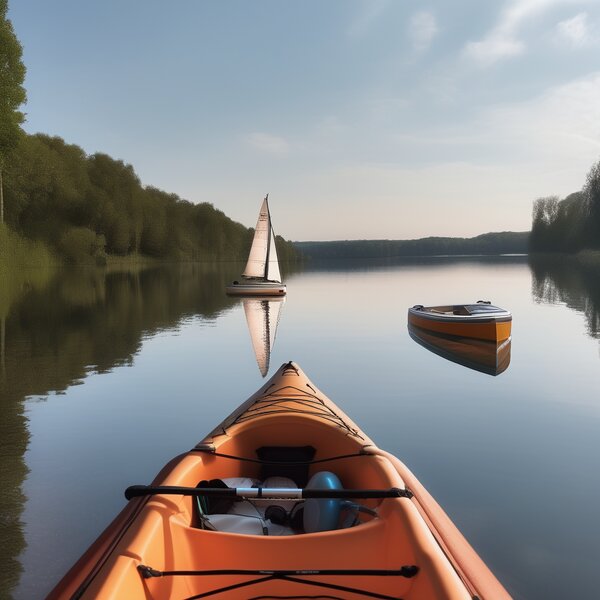} &
        \includegraphics[width=0.15\textwidth]{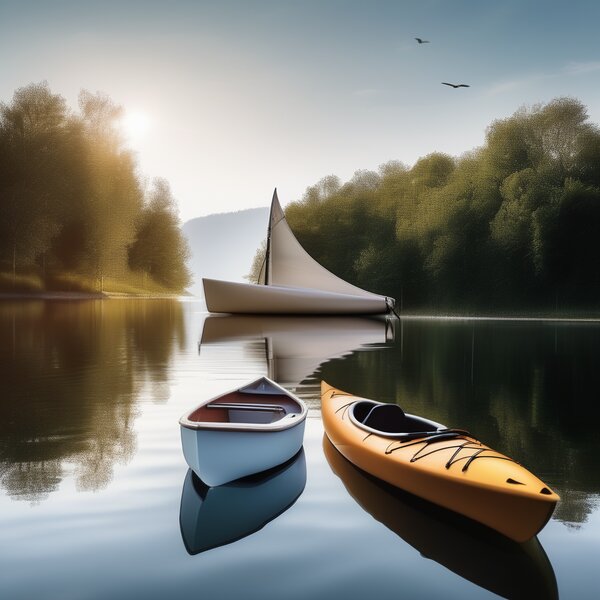} &
        \includegraphics[width=0.15\textwidth]{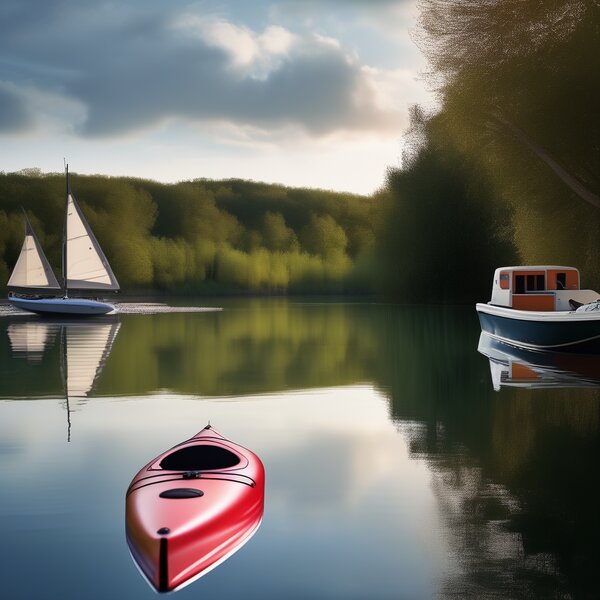} &
        \includegraphics[width=0.15\textwidth]{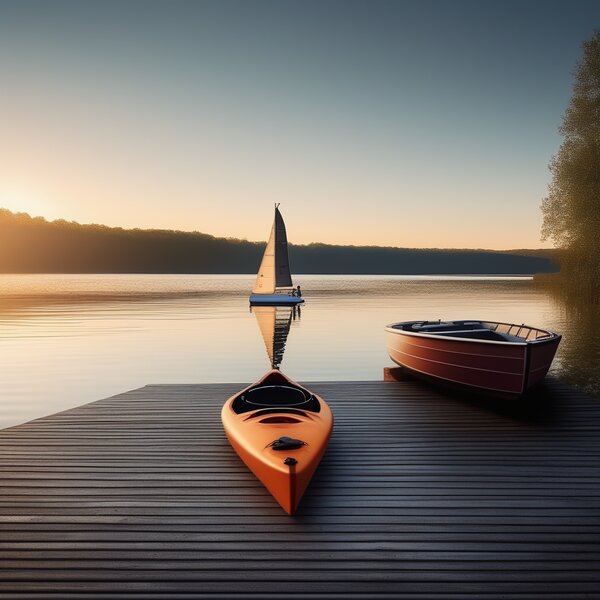} &
        \includegraphics[width=0.15\textwidth]{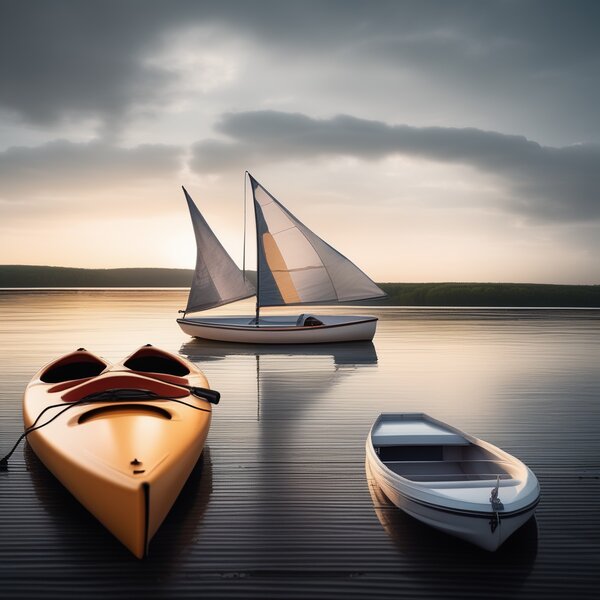} &
        \includegraphics[width=0.15\textwidth]{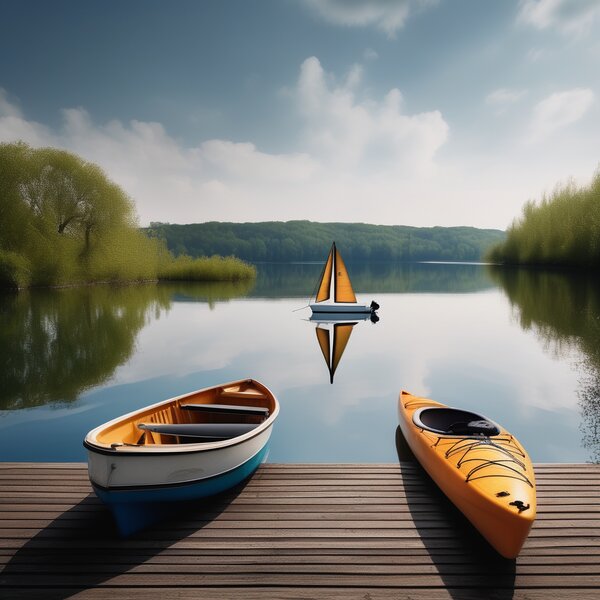} \\
        \multicolumn{6}{c}{"... \textbf{five horses} in a field"} \\
        \includegraphics[width=0.15\textwidth]{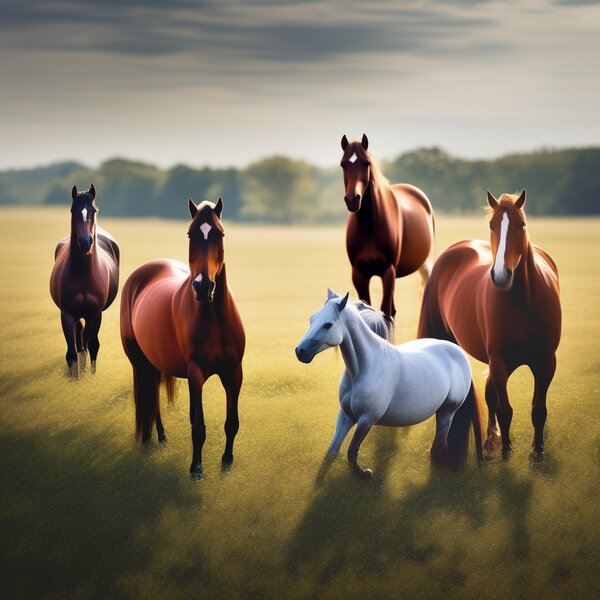} &
        \includegraphics[width=0.15\textwidth]{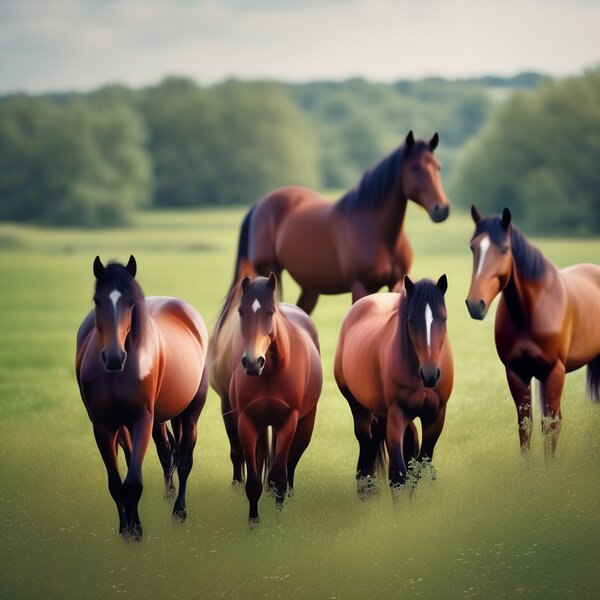} &
        \includegraphics[width=0.15\textwidth]{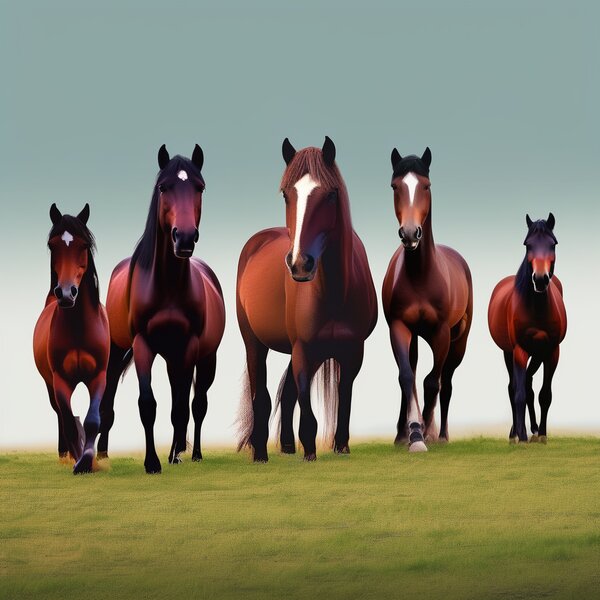} &
        \includegraphics[width=0.15\textwidth]{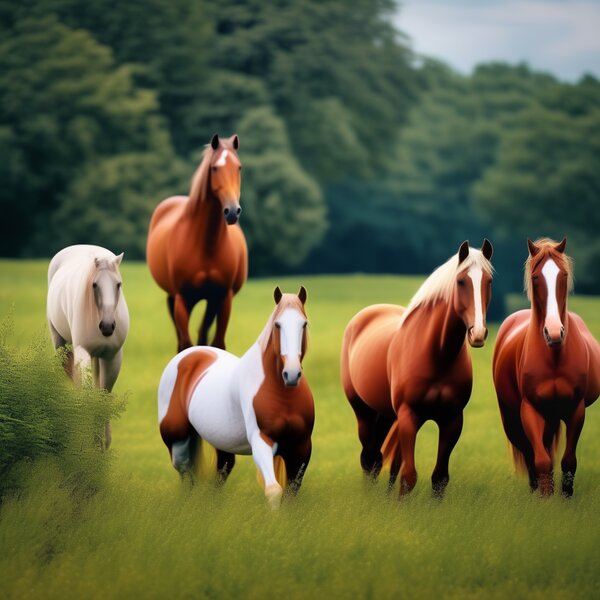} &
        \includegraphics[width=0.15\textwidth]{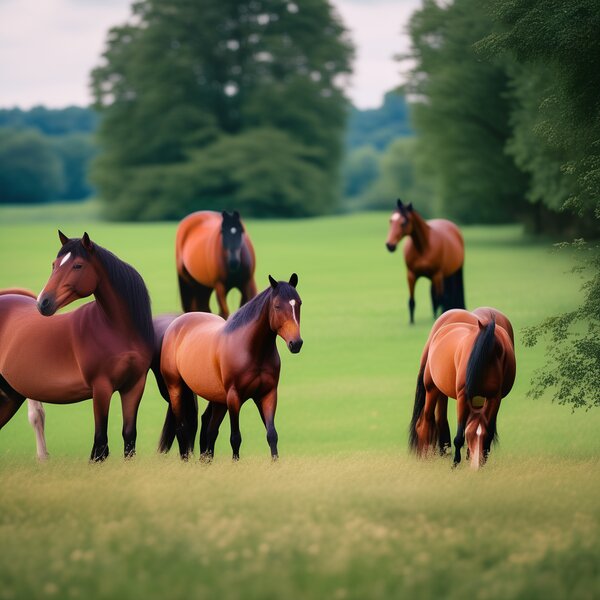} &
        \includegraphics[width=0.15\textwidth]{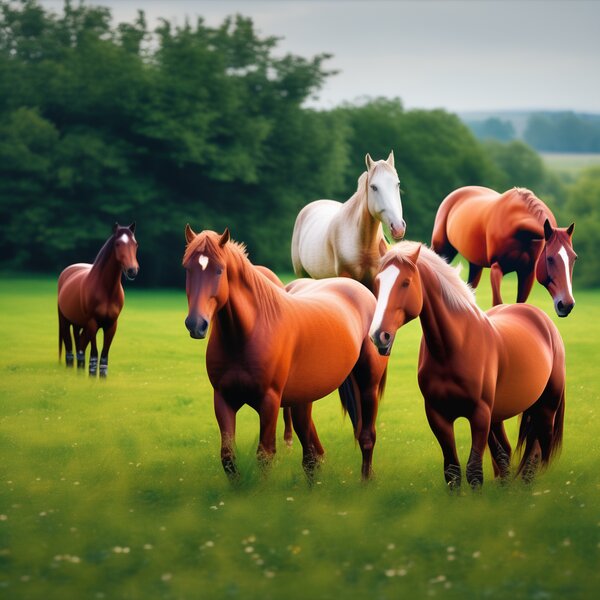} \\
        \multicolumn{6}{c}{"... \textbf{panda plush toy}, a \textbf{red panda plush toy}, and a \textbf{koala plush toy} on a shelf"} \\
        \includegraphics[width=0.15\textwidth]{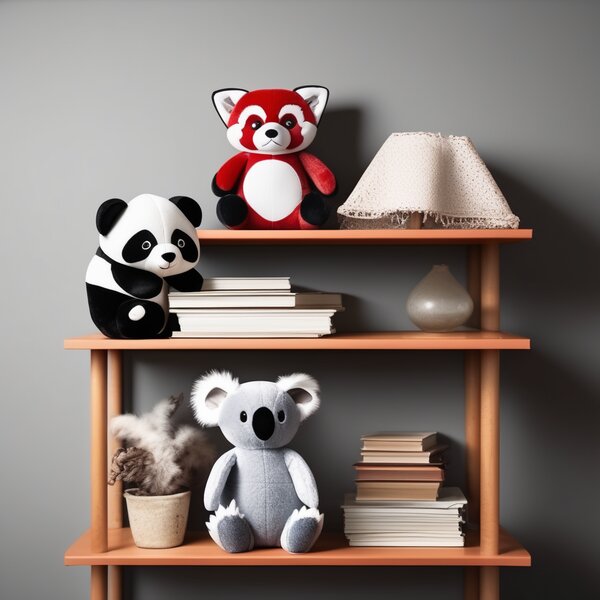} &
        \includegraphics[width=0.15\textwidth]{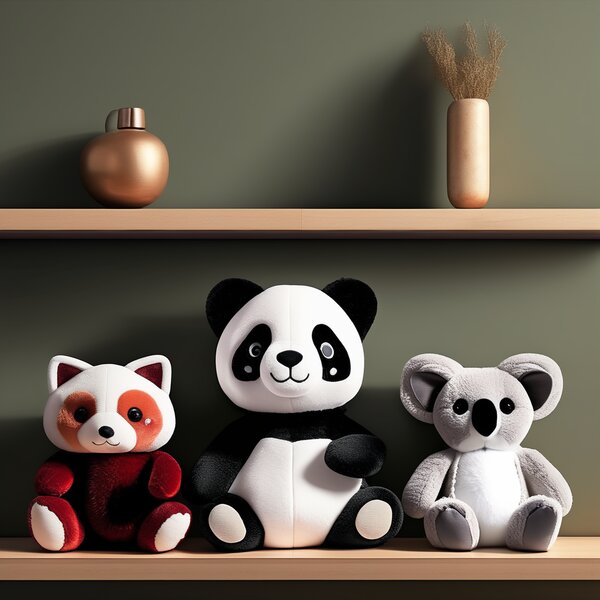} &
        \includegraphics[width=0.15\textwidth]{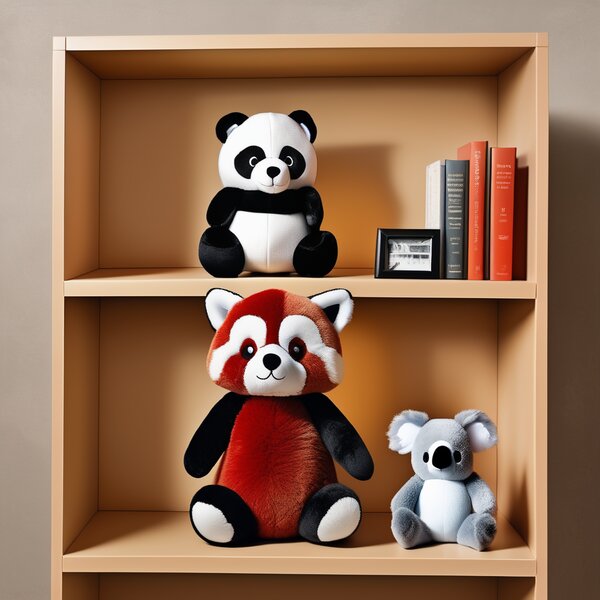} &
        \includegraphics[width=0.15\textwidth]{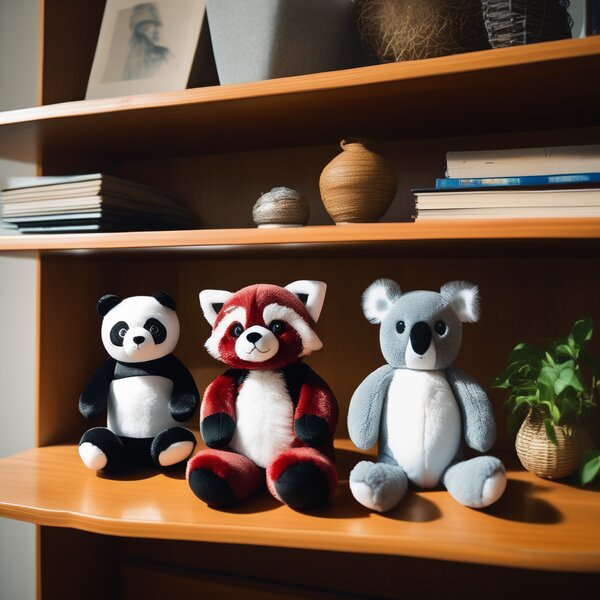} &
        \includegraphics[width=0.15\textwidth]{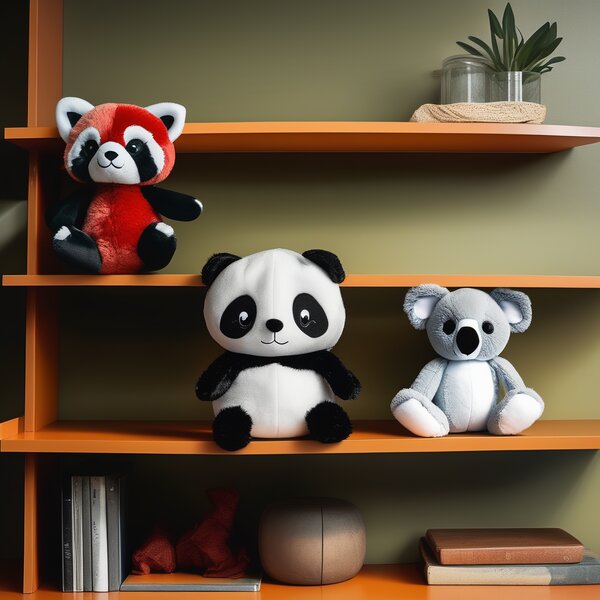} &
        \includegraphics[width=0.15\textwidth]{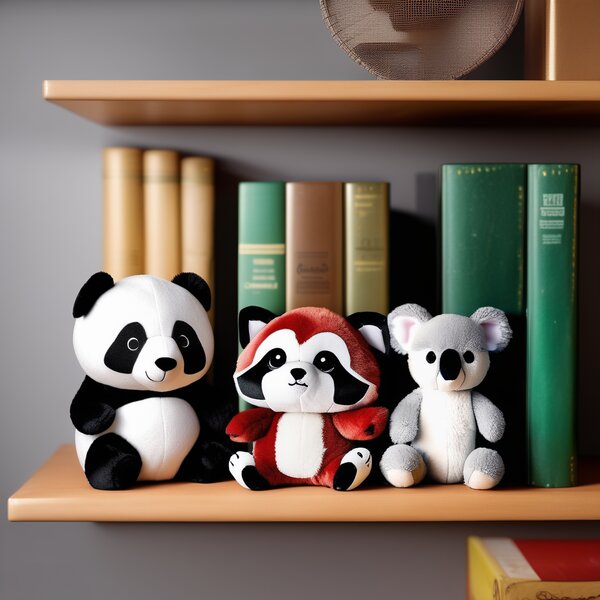} \\
        \multicolumn{6}{c}{"... a window dressing with three mannequins wearing a \textbf{blue velvet dress}, a \textbf{pink tulle gown}, and a \textbf{brown fur coat}"} \\
        \includegraphics[width=0.15\textwidth]{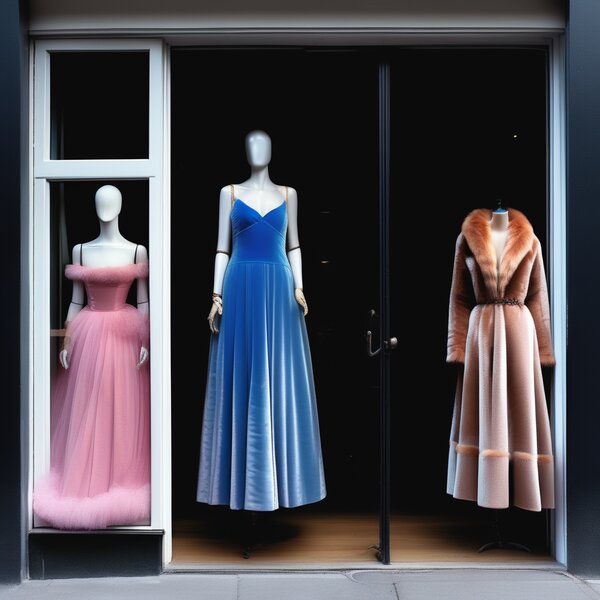} &
        \includegraphics[width=0.15\textwidth]{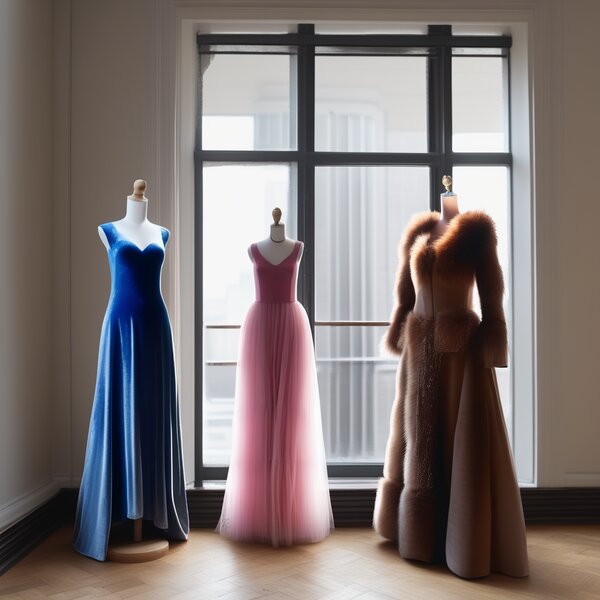} &
        \includegraphics[width=0.15\textwidth]{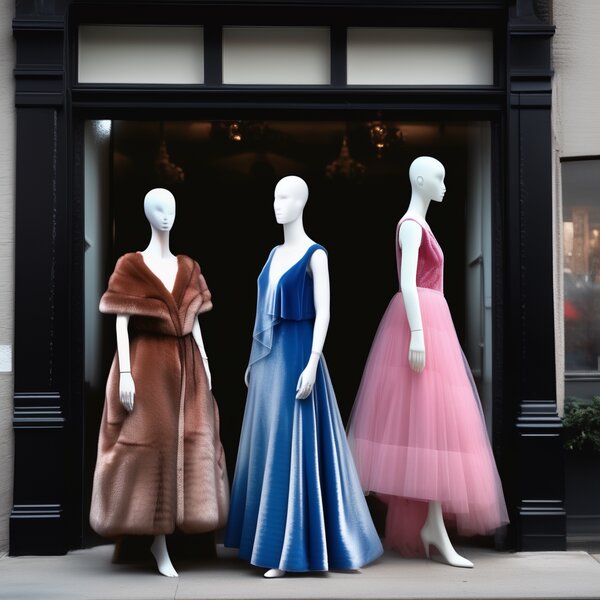} &
        \includegraphics[width=0.15\textwidth]{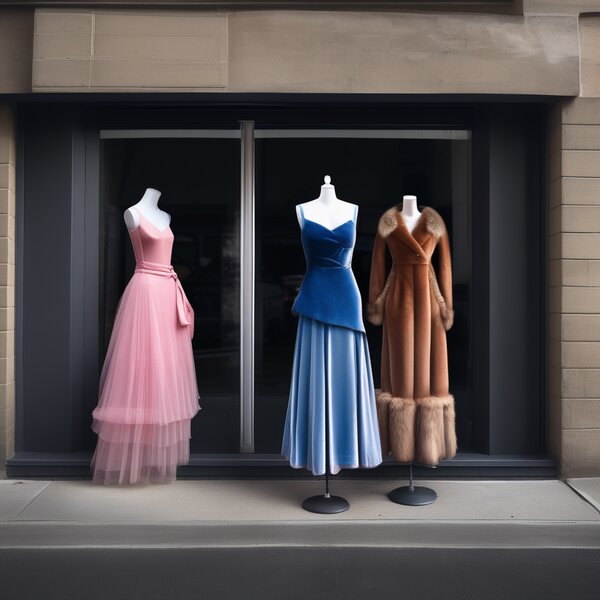} &
        \includegraphics[width=0.15\textwidth]{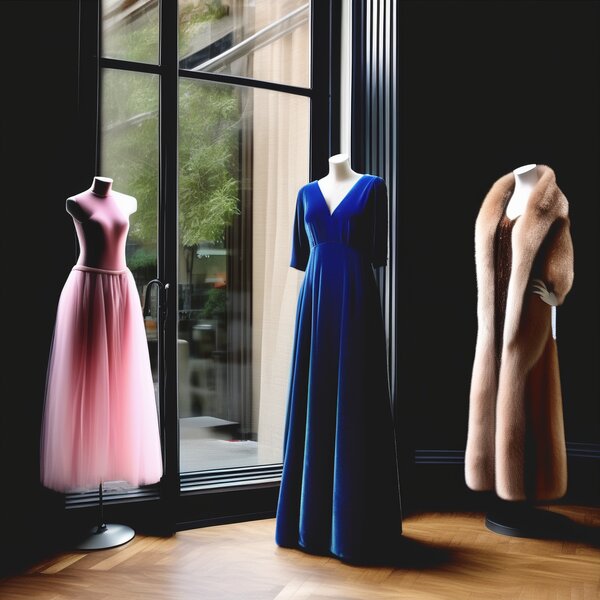} &
        \includegraphics[width=0.15\textwidth]{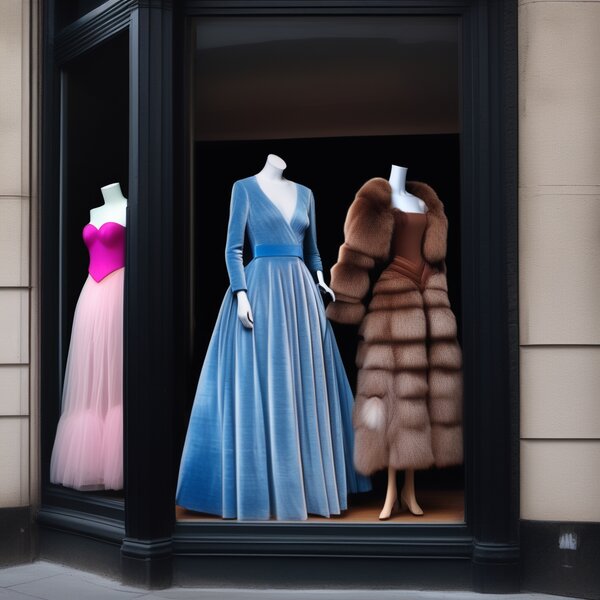} \\
    \end{tabular}
    \caption{Generated images across different seeds. Our method follows noise-induce layouts to generate diverse compositions, while still faithfully depicting subject characteristics such as class features, attributes and quantities. Note the rich layout diversity of the results.}
    \label{fig:diversity}
\end{figure*}

\begin{figure*}[hp!]
    \centering
    \setlength{\tabcolsep}{0.001\textwidth}
    {\small
    \begin{tabular}{c c c c c c c c}
        \multicolumn{8}{c}{\small{"A hyper-realistic photo of a \textbf{ferret}, a \textbf{squirrel}, and a \textbf{crow} in a beautiful garden"}} \\

         & Seed 0 & Seed 1 & Seed 2 & Seed 3 & Seed 4 & Seed 5 & Seed 6 \\
        
        \raisebox{27pt}{\rotatebox{90}{Ours}} &
        \includegraphics[width=0.14\textwidth]{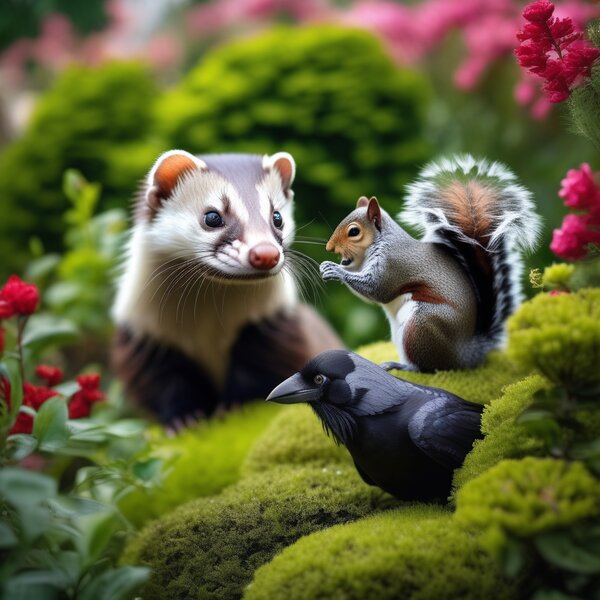} &
        \includegraphics[width=0.14\textwidth]{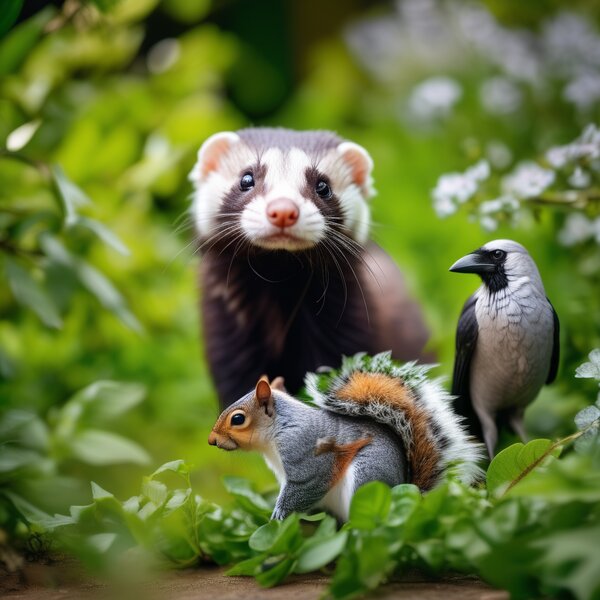} &
        \includegraphics[width=0.14\textwidth]{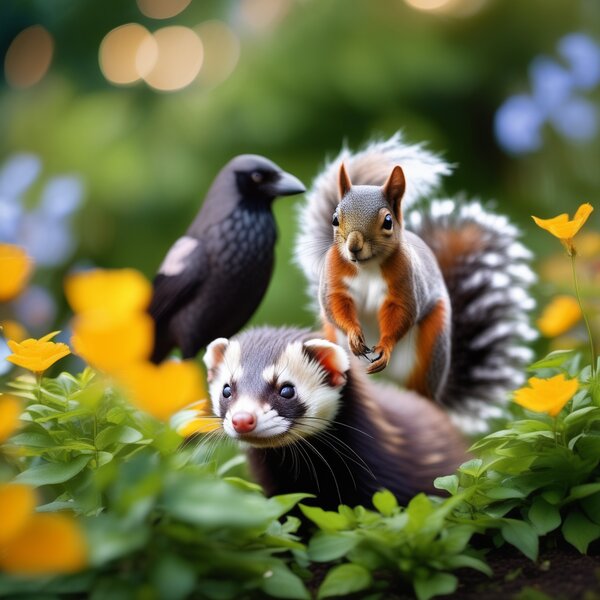} &
        \includegraphics[width=0.14\textwidth]{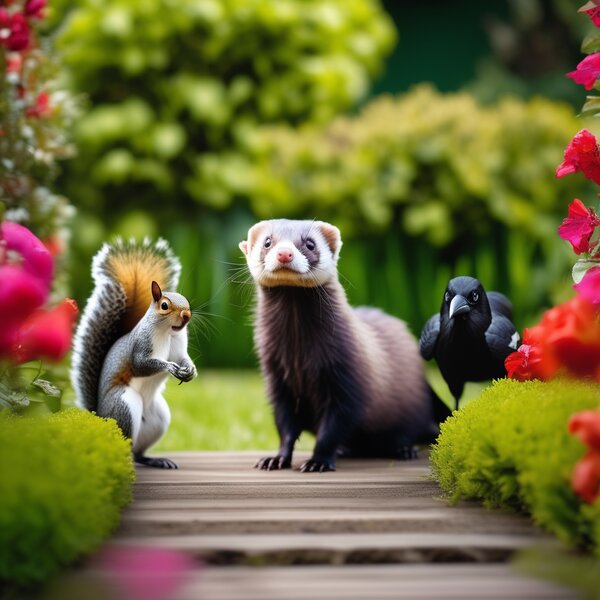} &
        \includegraphics[width=0.14\textwidth]{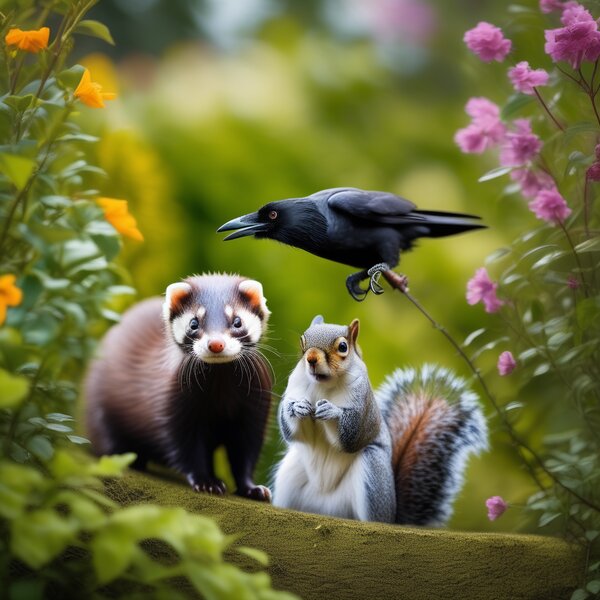} &
        \includegraphics[width=0.14\textwidth]{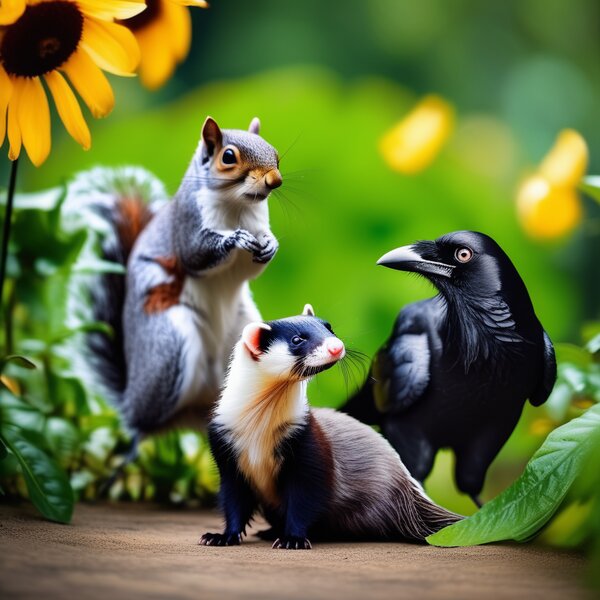} &
        \includegraphics[width=0.14\textwidth]{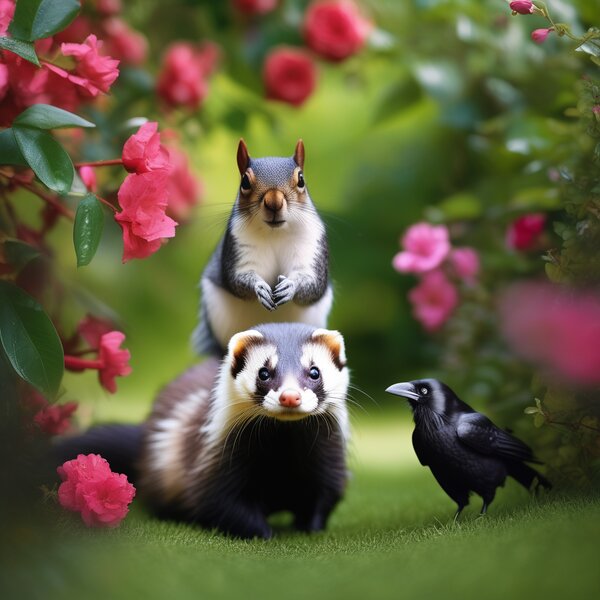} \\

        \raisebox{27pt}{\rotatebox{90}{SDXL}} &
        \includegraphics[width=0.14\textwidth]{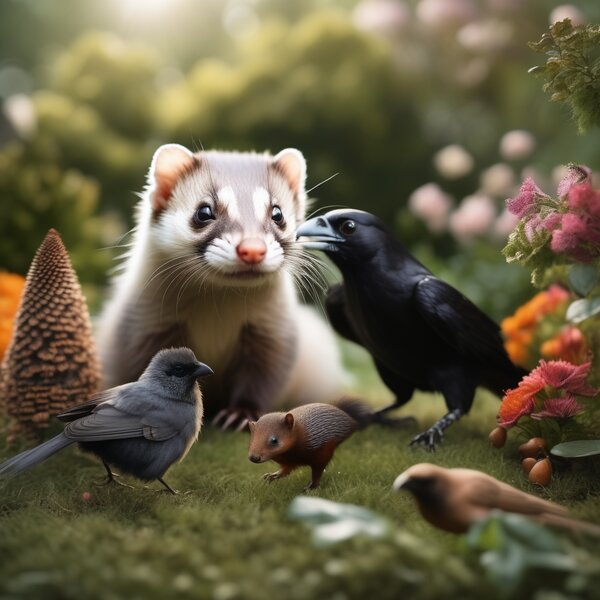} &
        \includegraphics[width=0.14\textwidth]{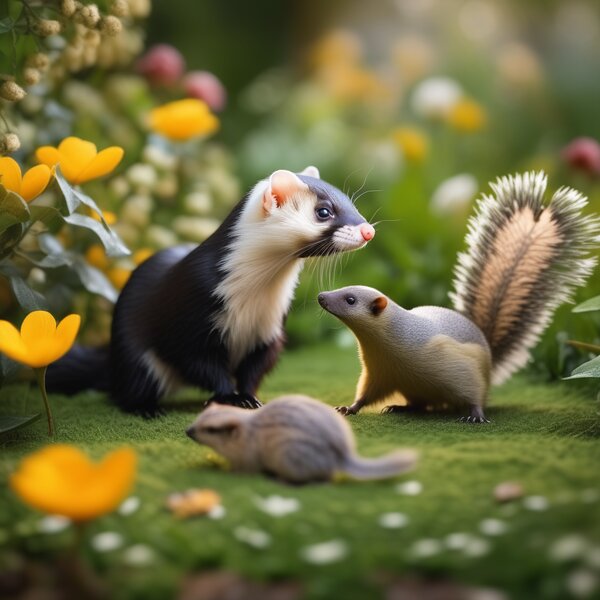} &
        \includegraphics[width=0.14\textwidth]{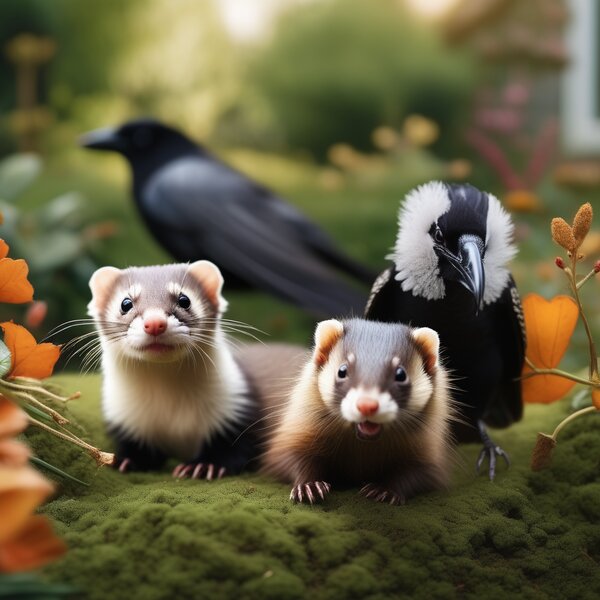} &
        \includegraphics[width=0.14\textwidth]{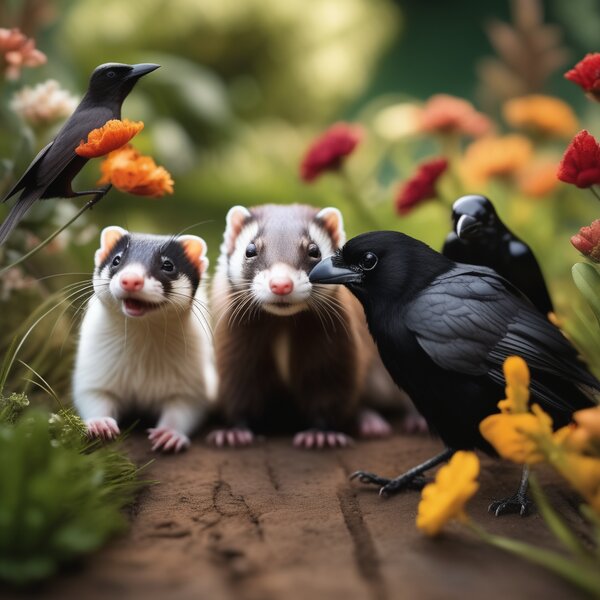} &
        \includegraphics[width=0.14\textwidth]{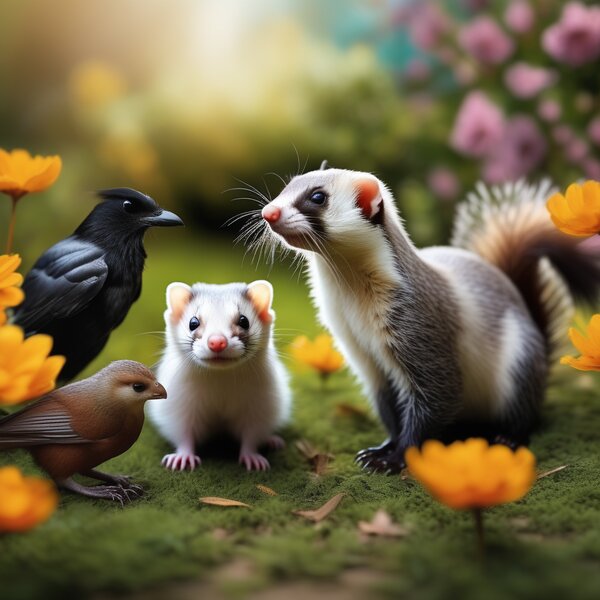} &
        \includegraphics[width=0.14\textwidth]{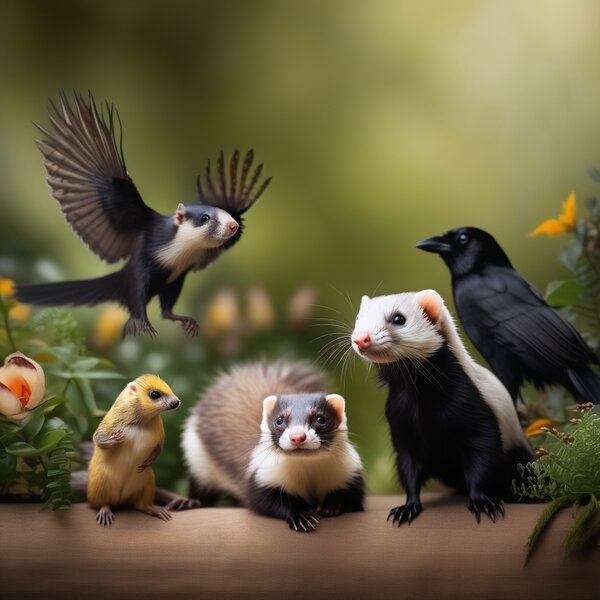} &
        \includegraphics[width=0.14\textwidth]{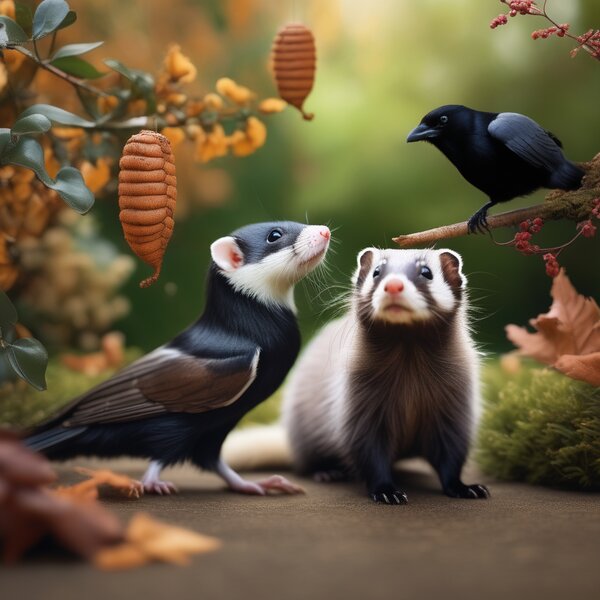} \\

        \raisebox{27pt}{\rotatebox{90}{Flux}} &
        \includegraphics[width=0.14\textwidth]{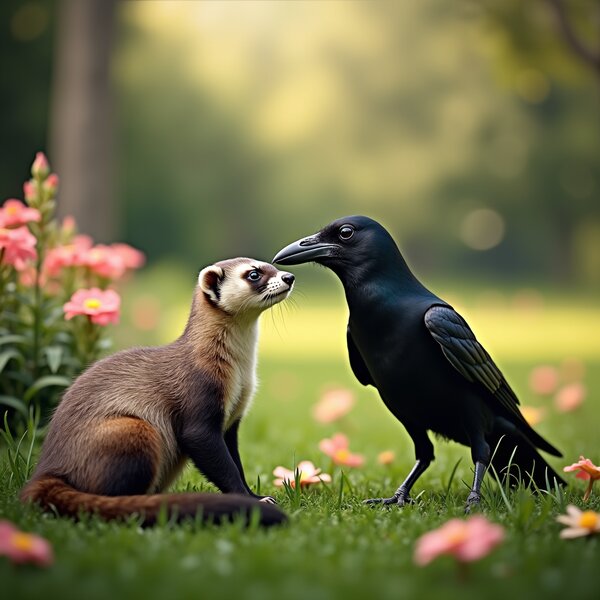} &
        \includegraphics[width=0.14\textwidth]{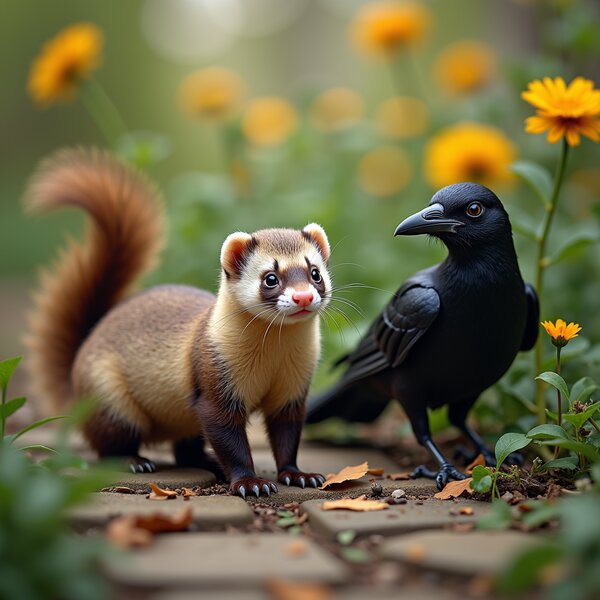} &
        \includegraphics[width=0.14\textwidth]{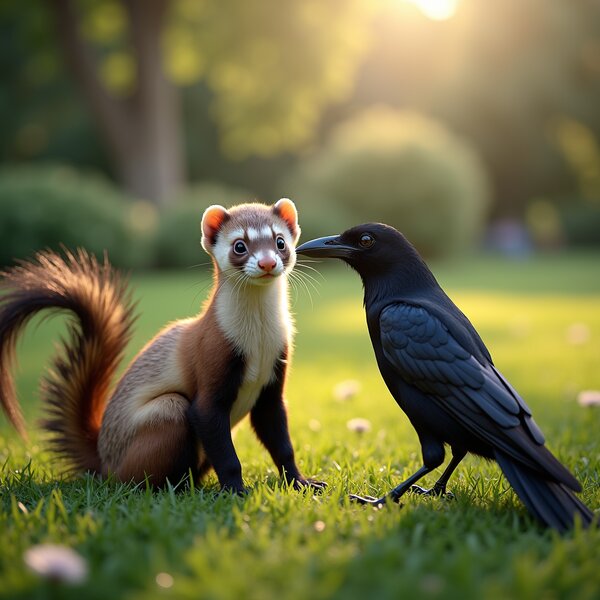} &
        \includegraphics[width=0.14\textwidth]{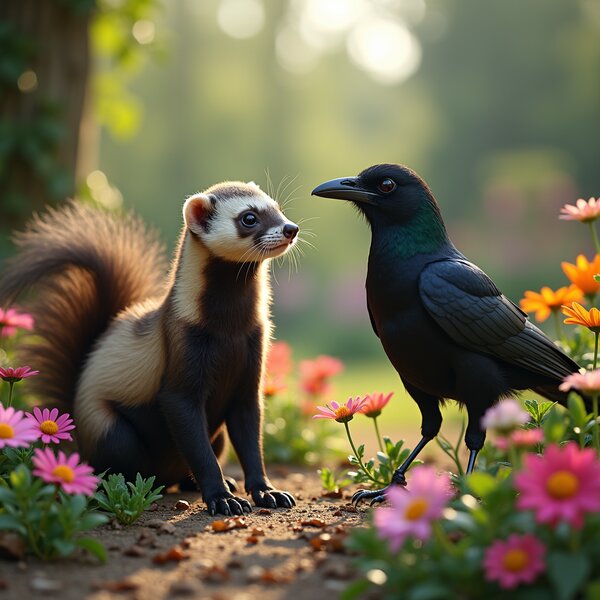} &
        \includegraphics[width=0.14\textwidth]{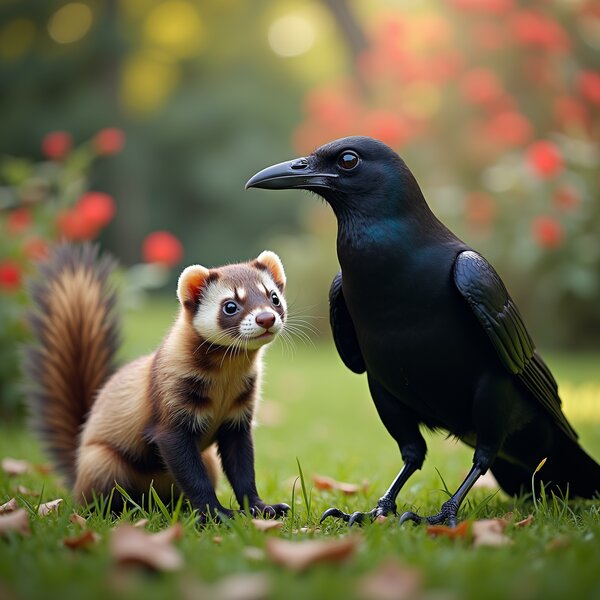} &
        \includegraphics[width=0.14\textwidth]{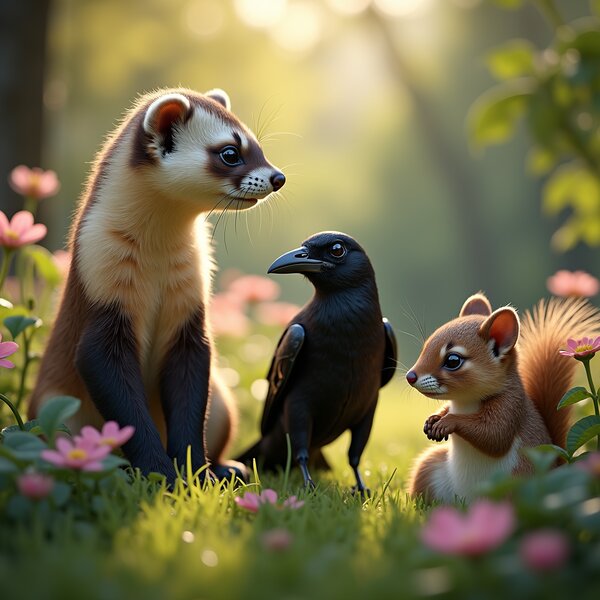} &
        \includegraphics[width=0.14\textwidth]{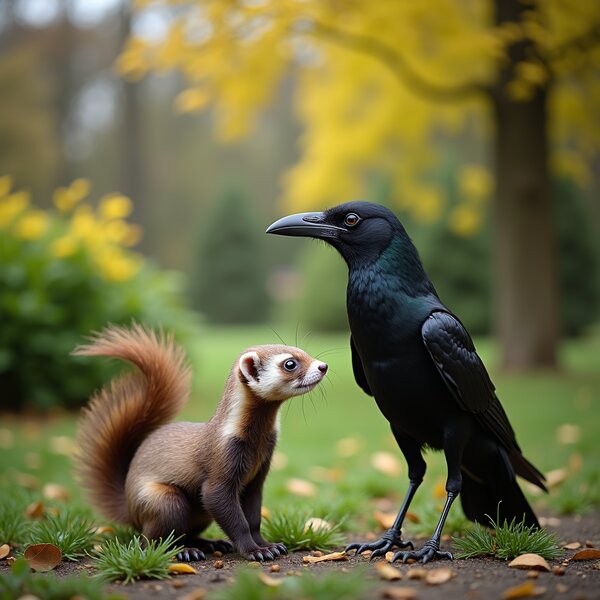} \\

        \raisebox{27pt}{\rotatebox{90}{A\&E}} &
        \includegraphics[width=0.14\textwidth]{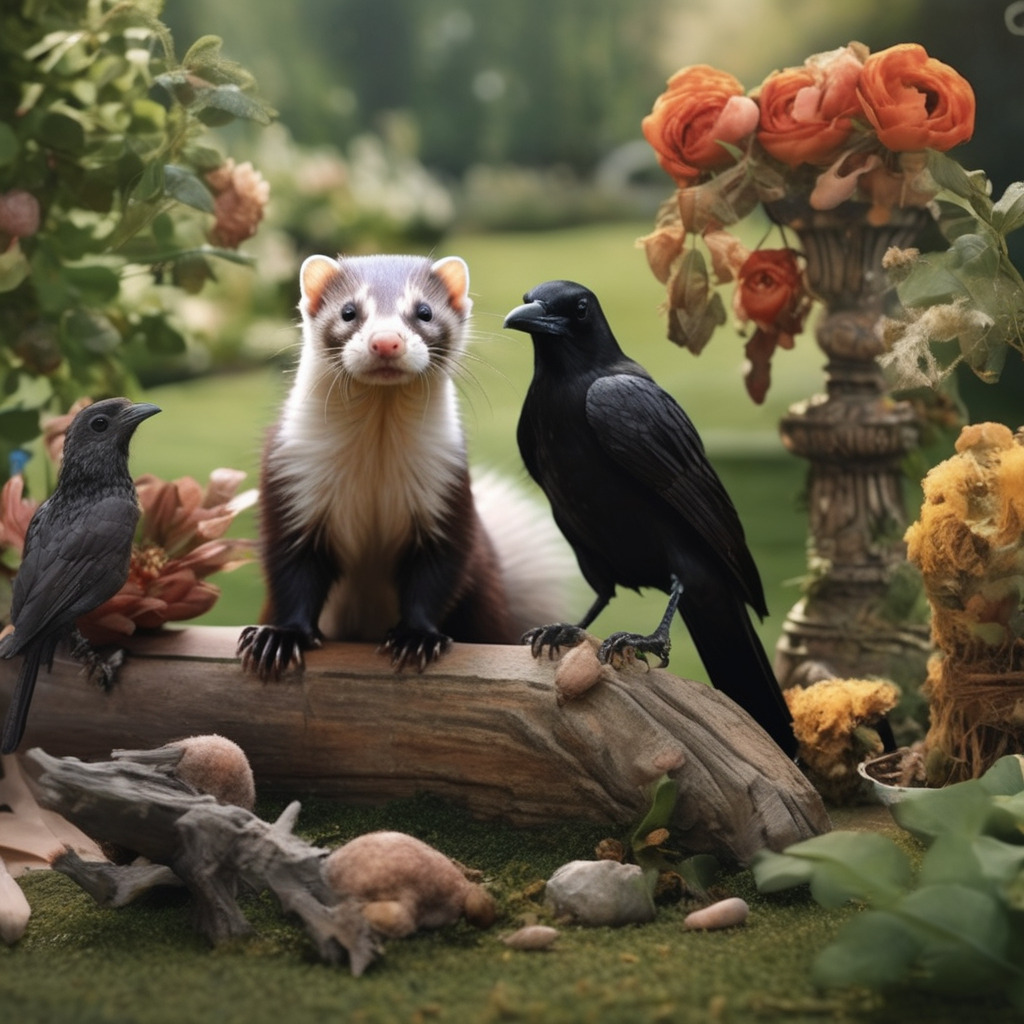} &
        \includegraphics[width=0.14\textwidth]{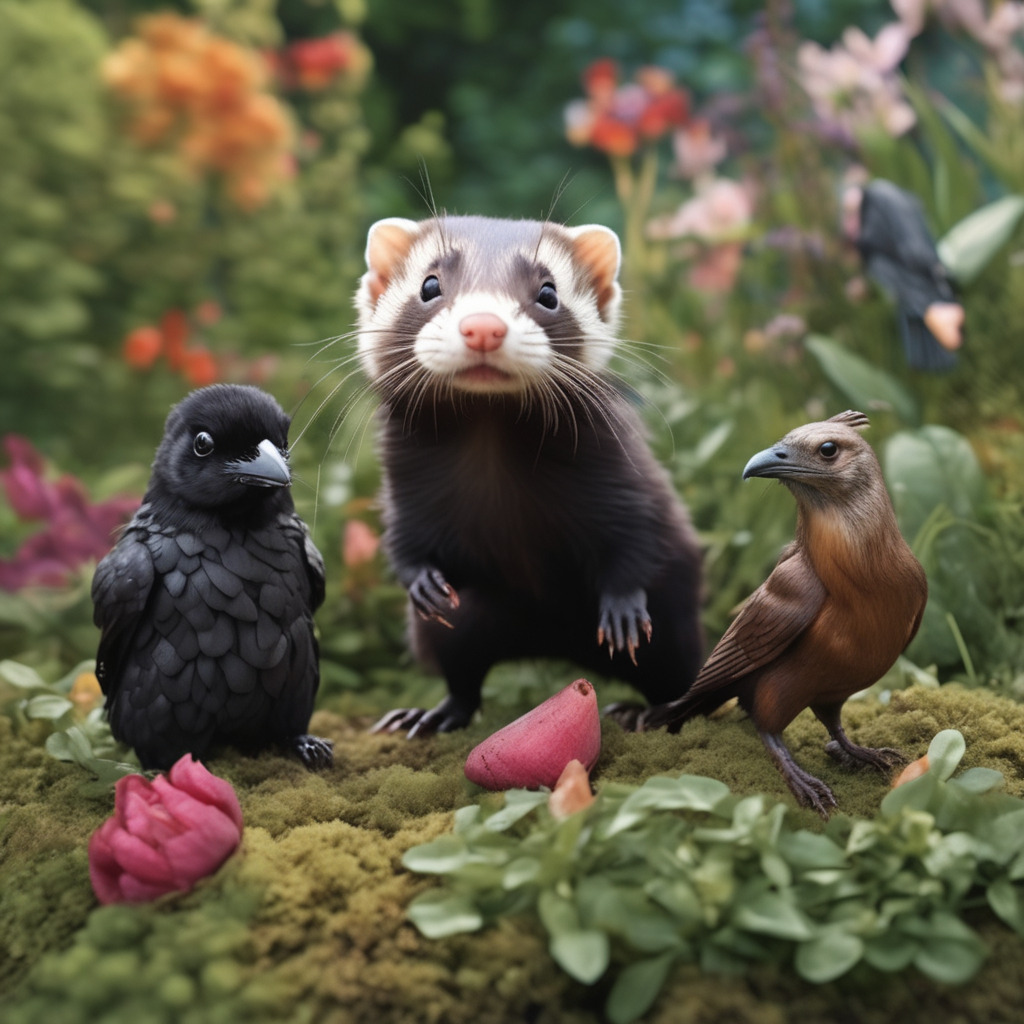} &
        \includegraphics[width=0.14\textwidth]{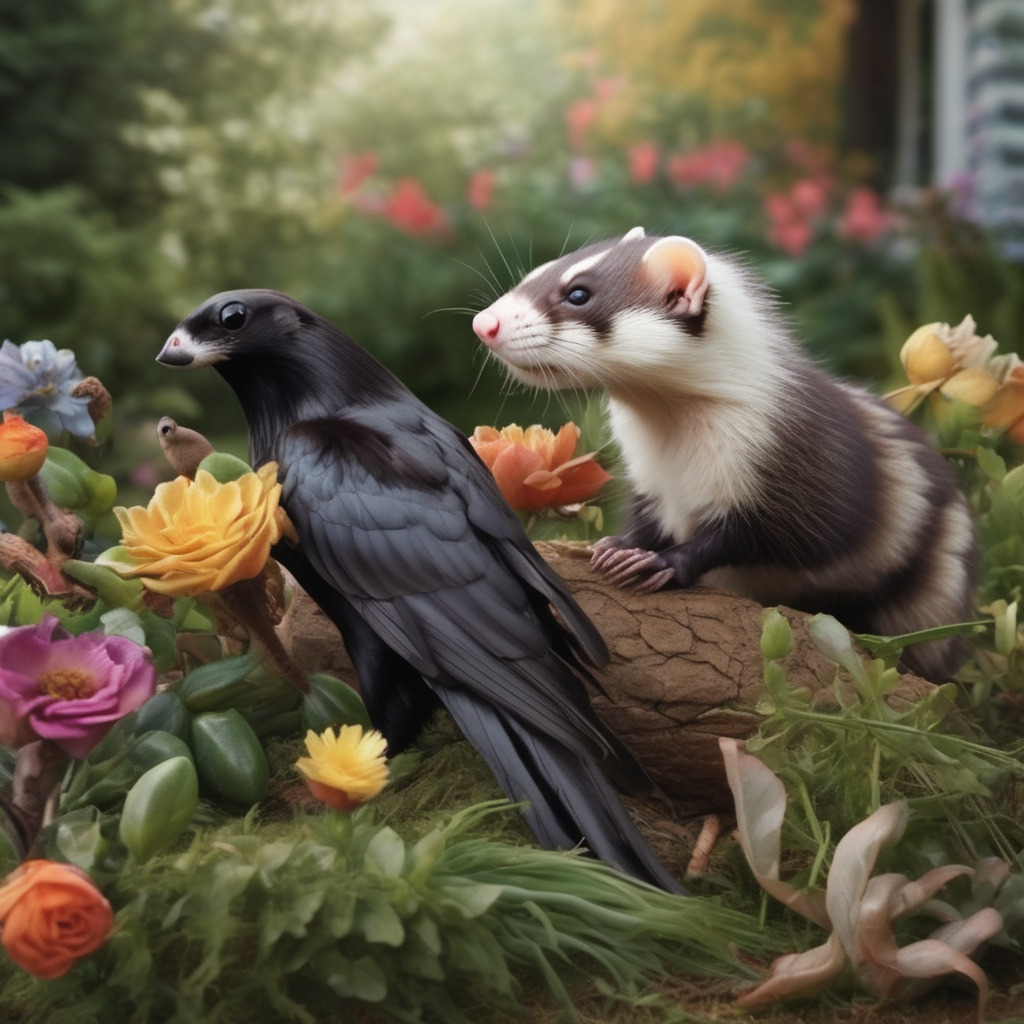} &
        \includegraphics[width=0.14\textwidth]{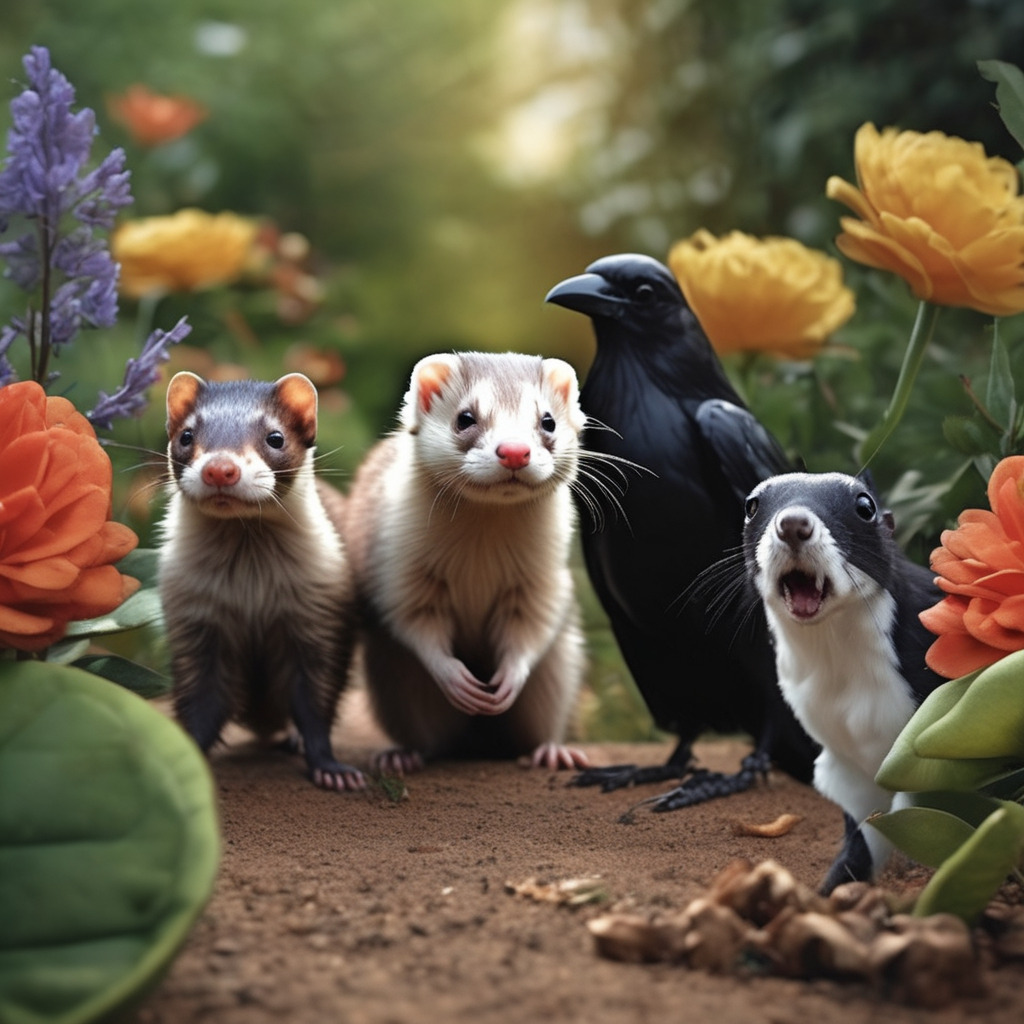} &
        \includegraphics[width=0.14\textwidth]{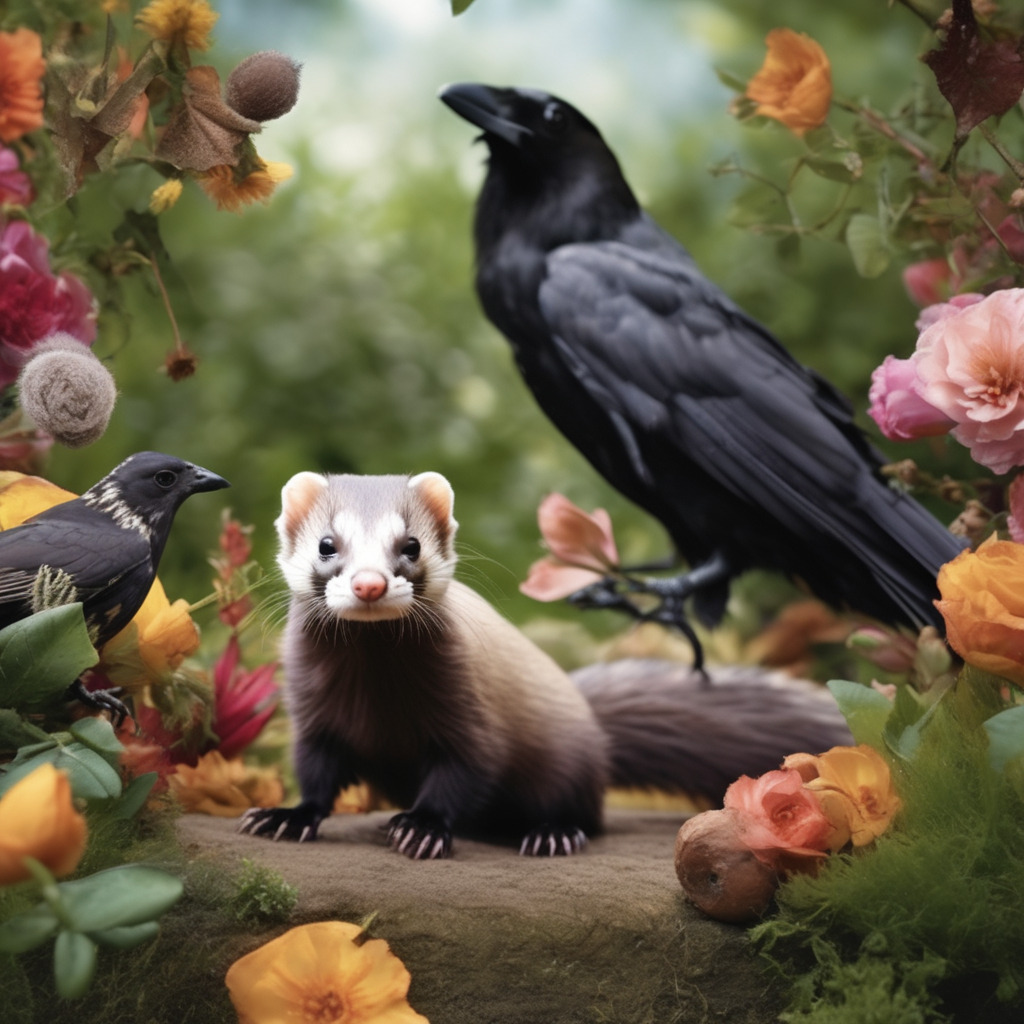} &
        \includegraphics[width=0.14\textwidth]{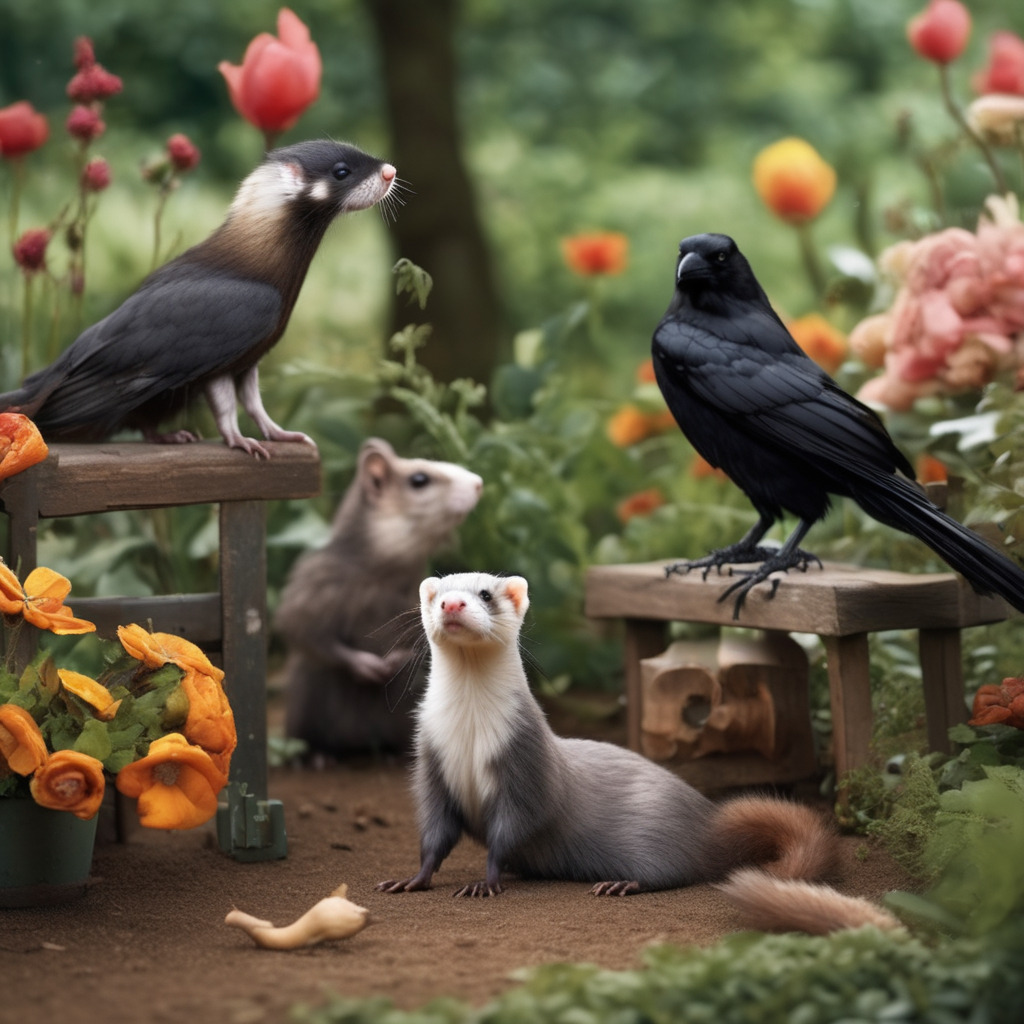} &
        \includegraphics[width=0.14\textwidth]{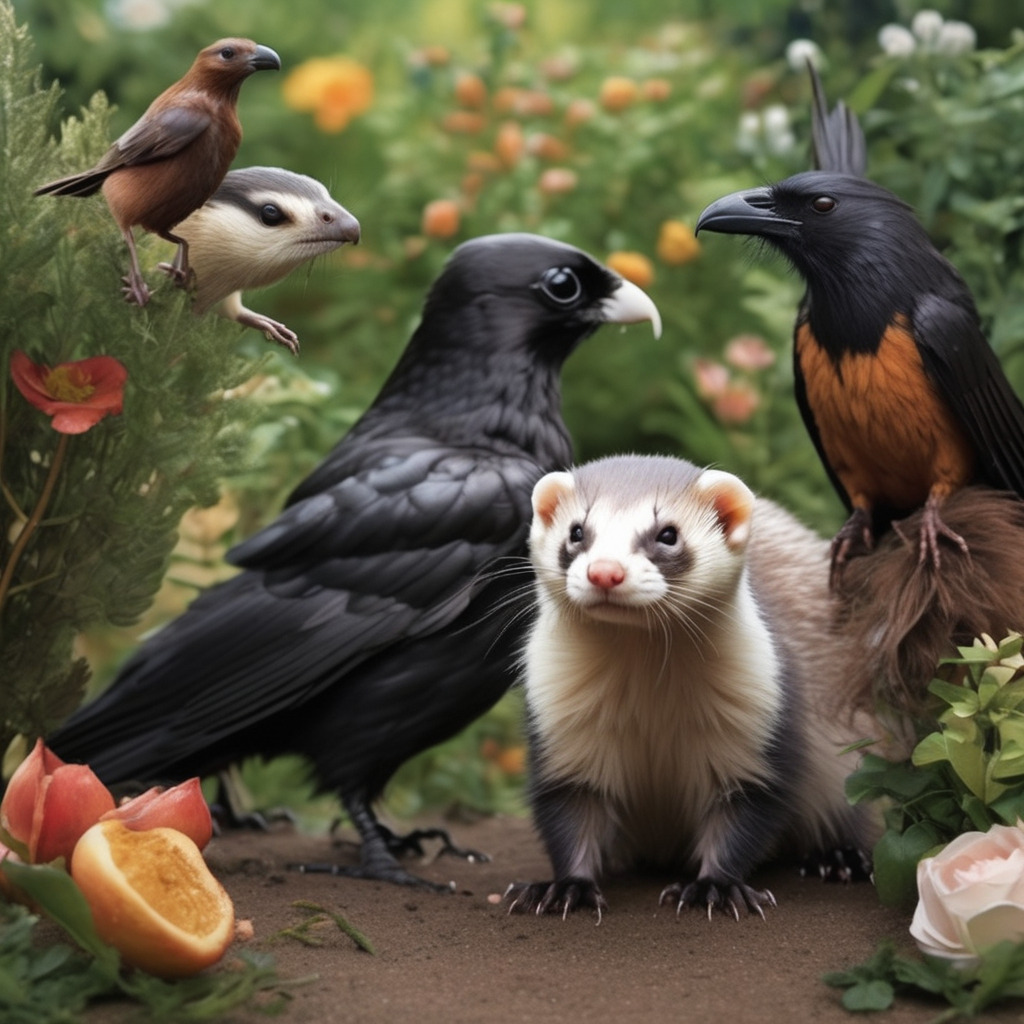} \\
        
        \raisebox{21pt}{\rotatebox{90}{LLM+BA}} &
        \includegraphics[width=0.14\textwidth]{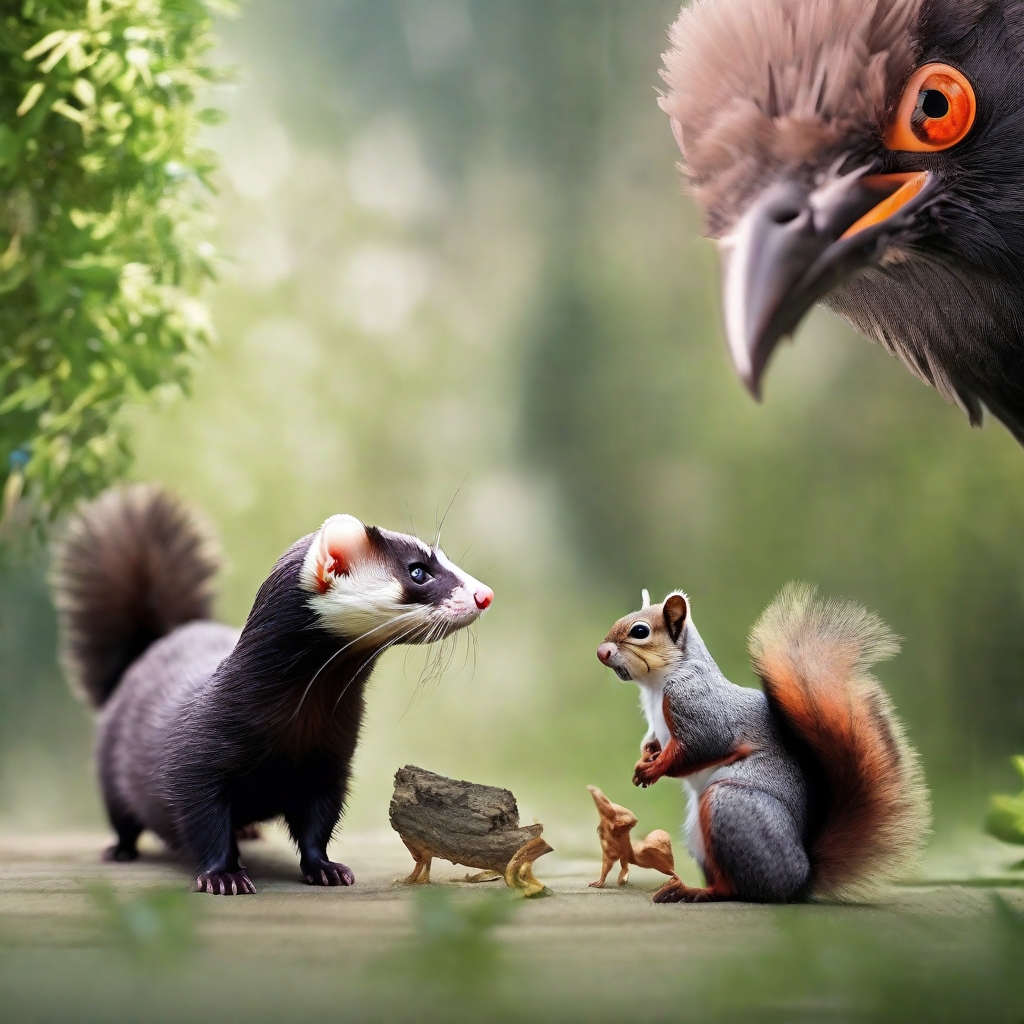} &
        \includegraphics[width=0.14\textwidth]{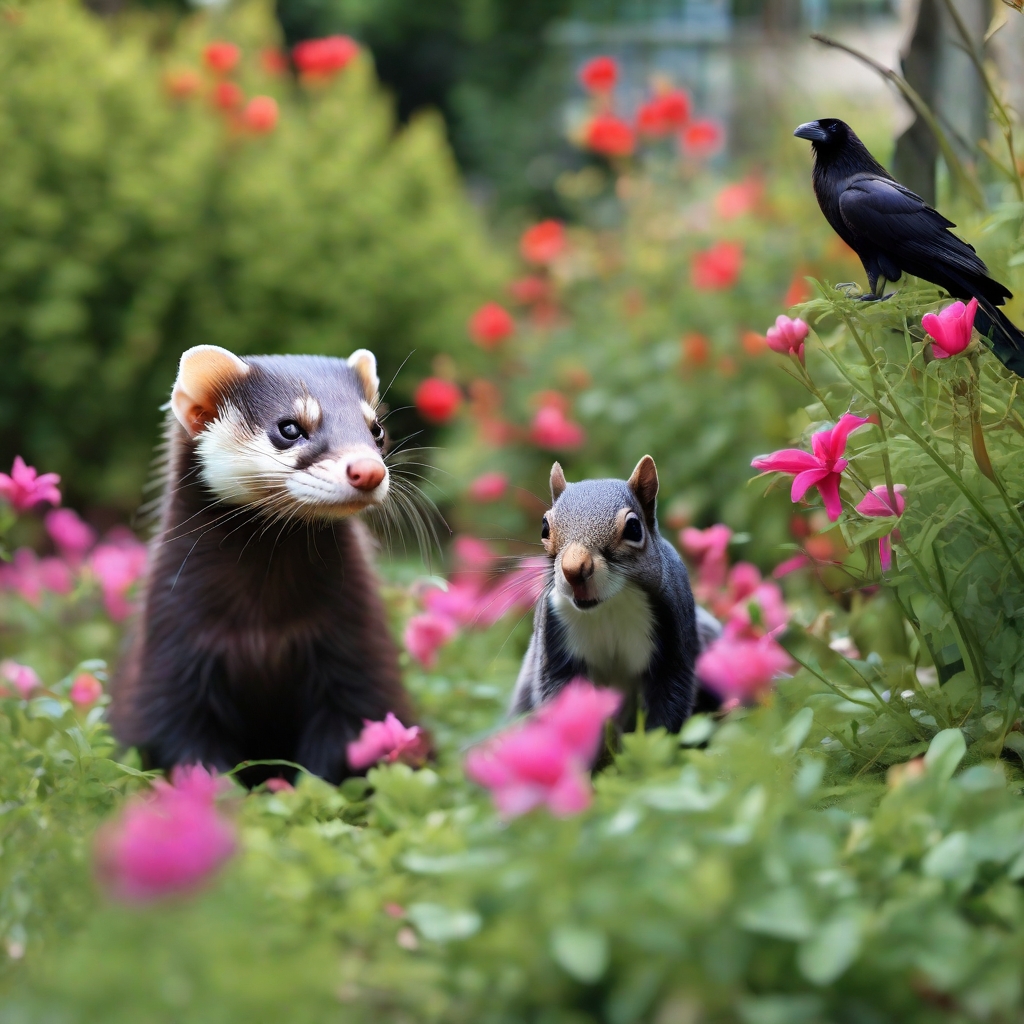} &
        \includegraphics[width=0.14\textwidth]{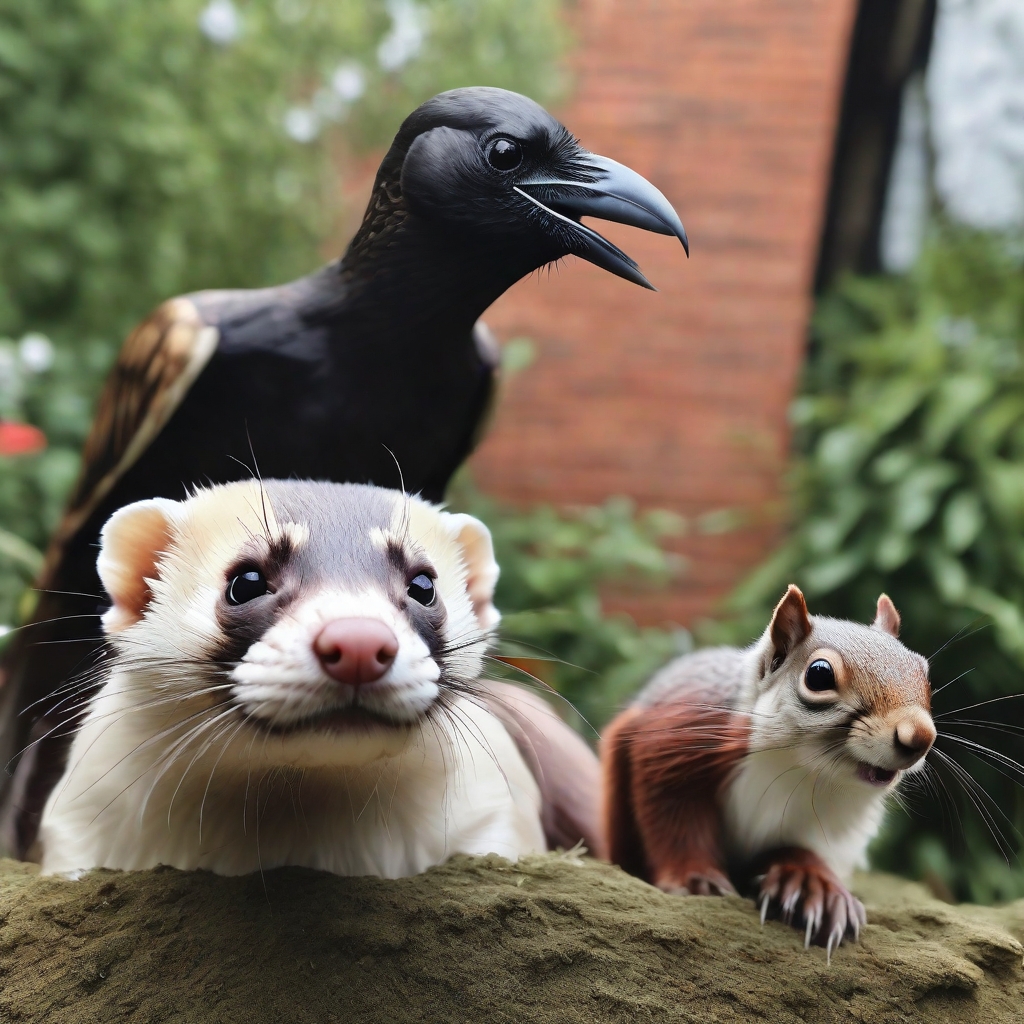} &
        \includegraphics[width=0.14\textwidth]{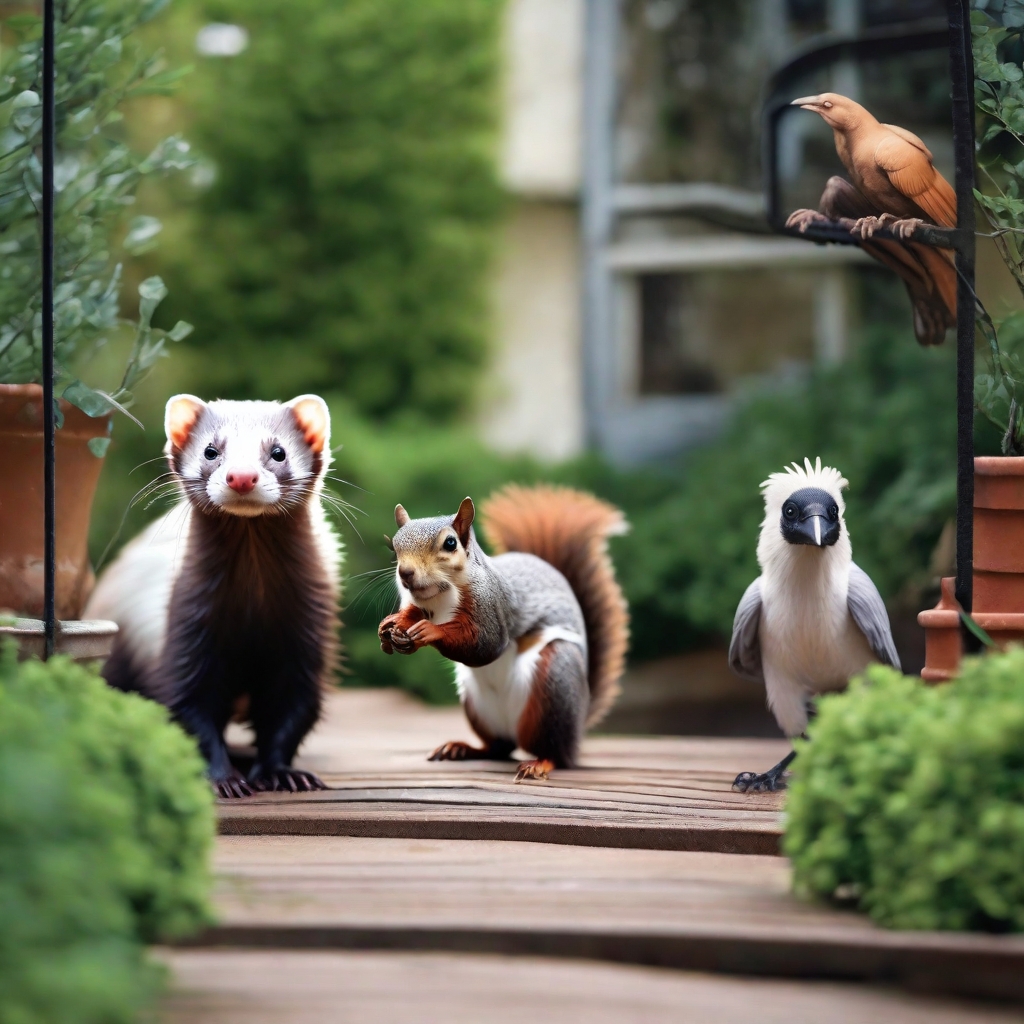} &
        \includegraphics[width=0.14\textwidth]{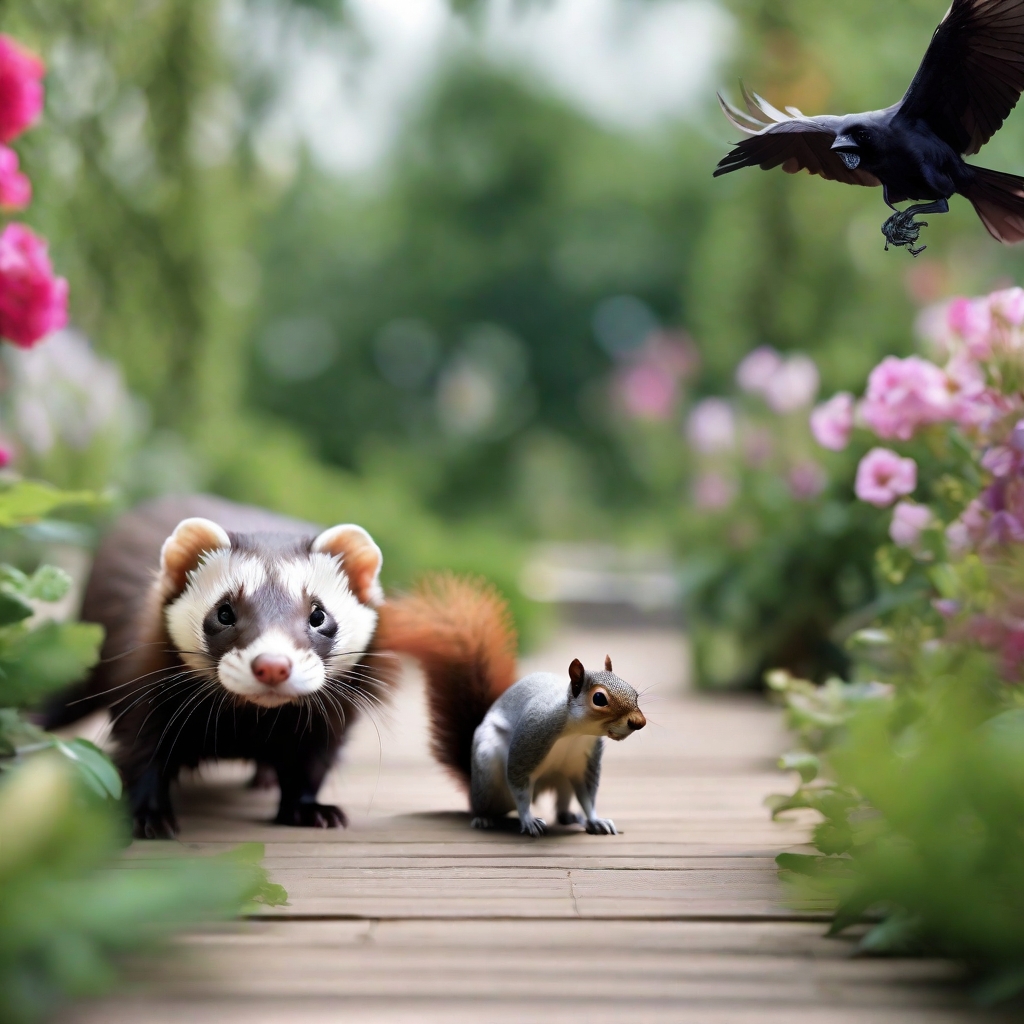} &
        \includegraphics[width=0.14\textwidth]{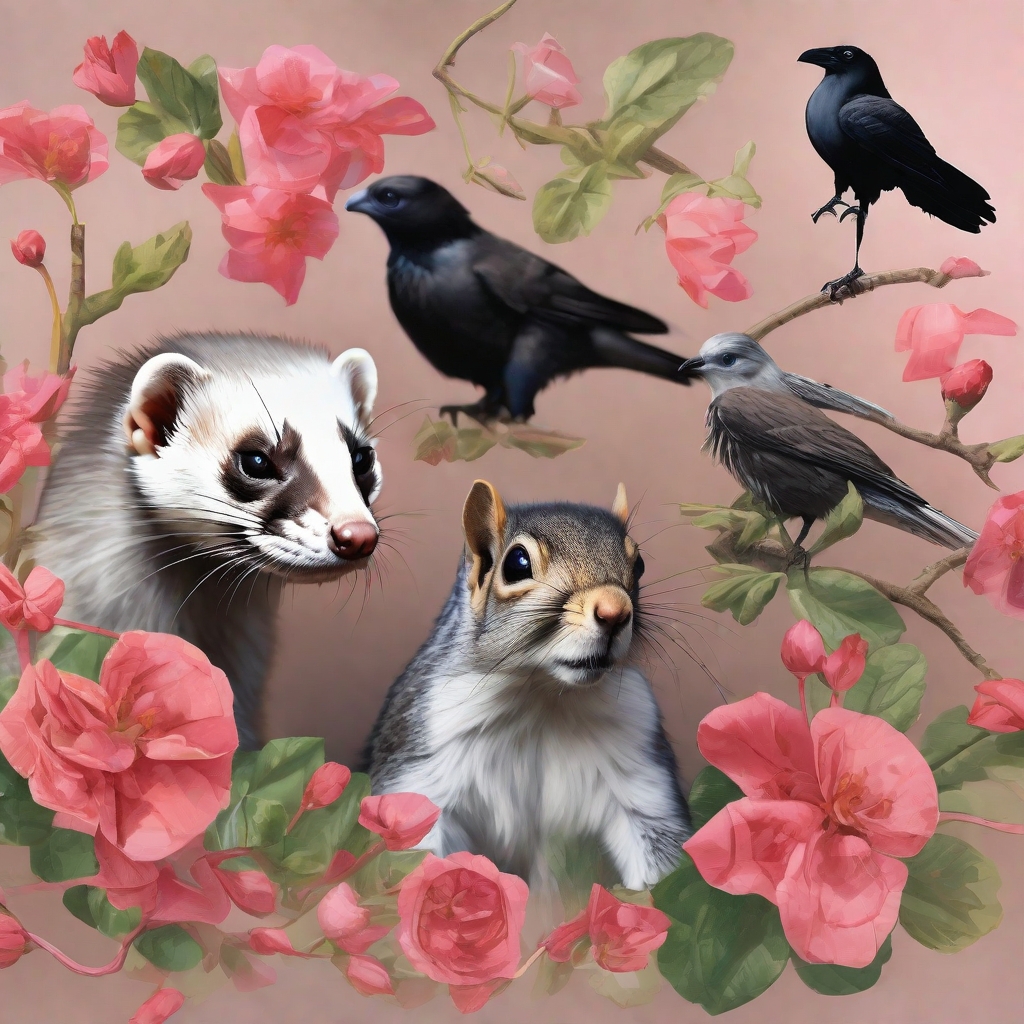} &
        \includegraphics[width=0.14\textwidth]{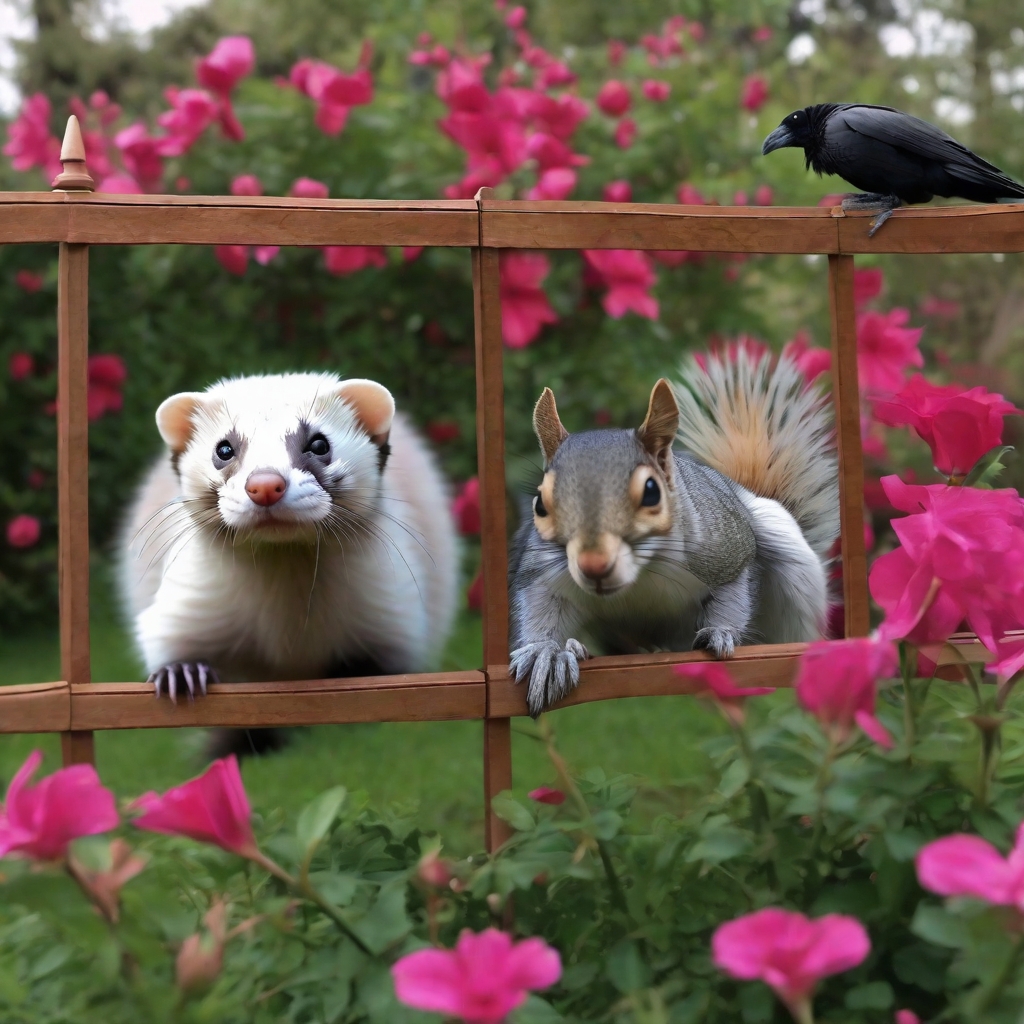} \\

        \raisebox{30pt}{\rotatebox{90}{RPG}} &
        \includegraphics[width=0.14\textwidth]{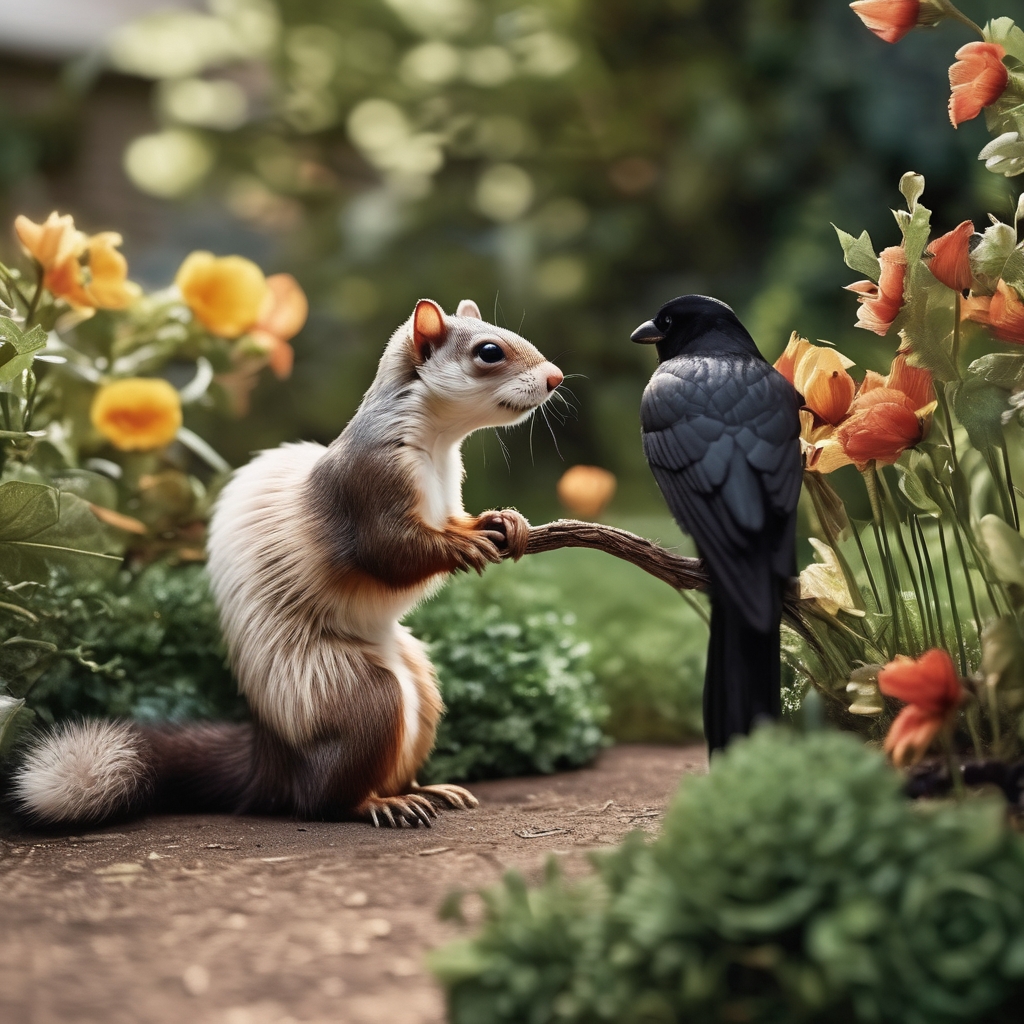} &
        \includegraphics[width=0.14\textwidth]{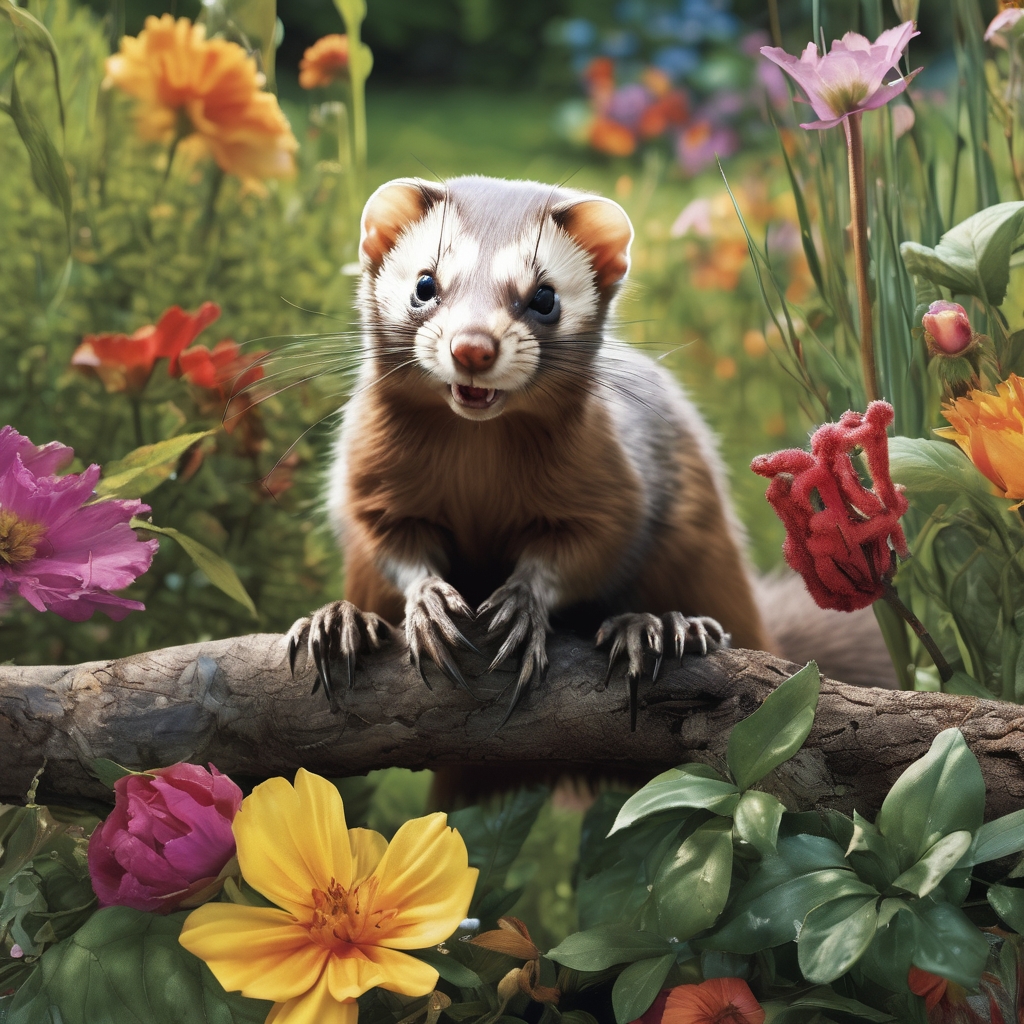} &
        \includegraphics[width=0.14\textwidth]{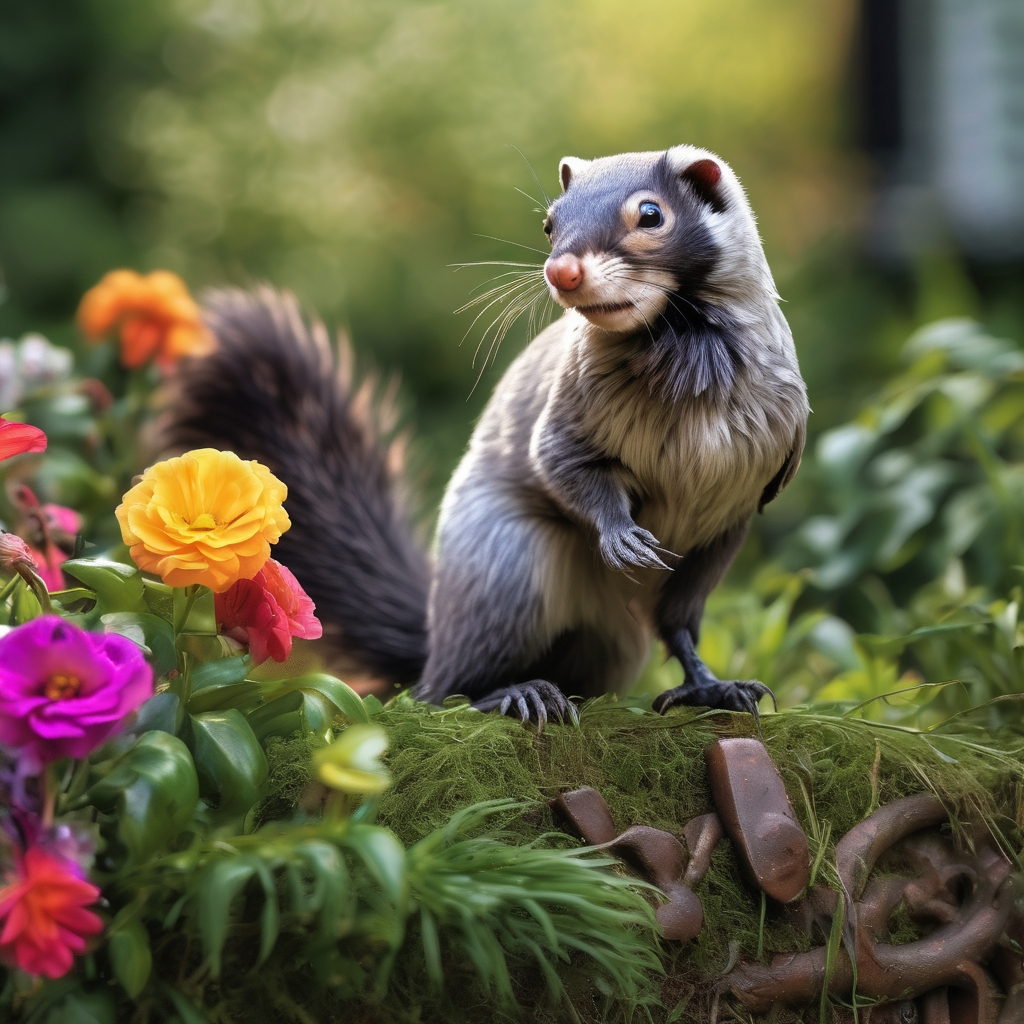} &
        \includegraphics[width=0.14\textwidth]{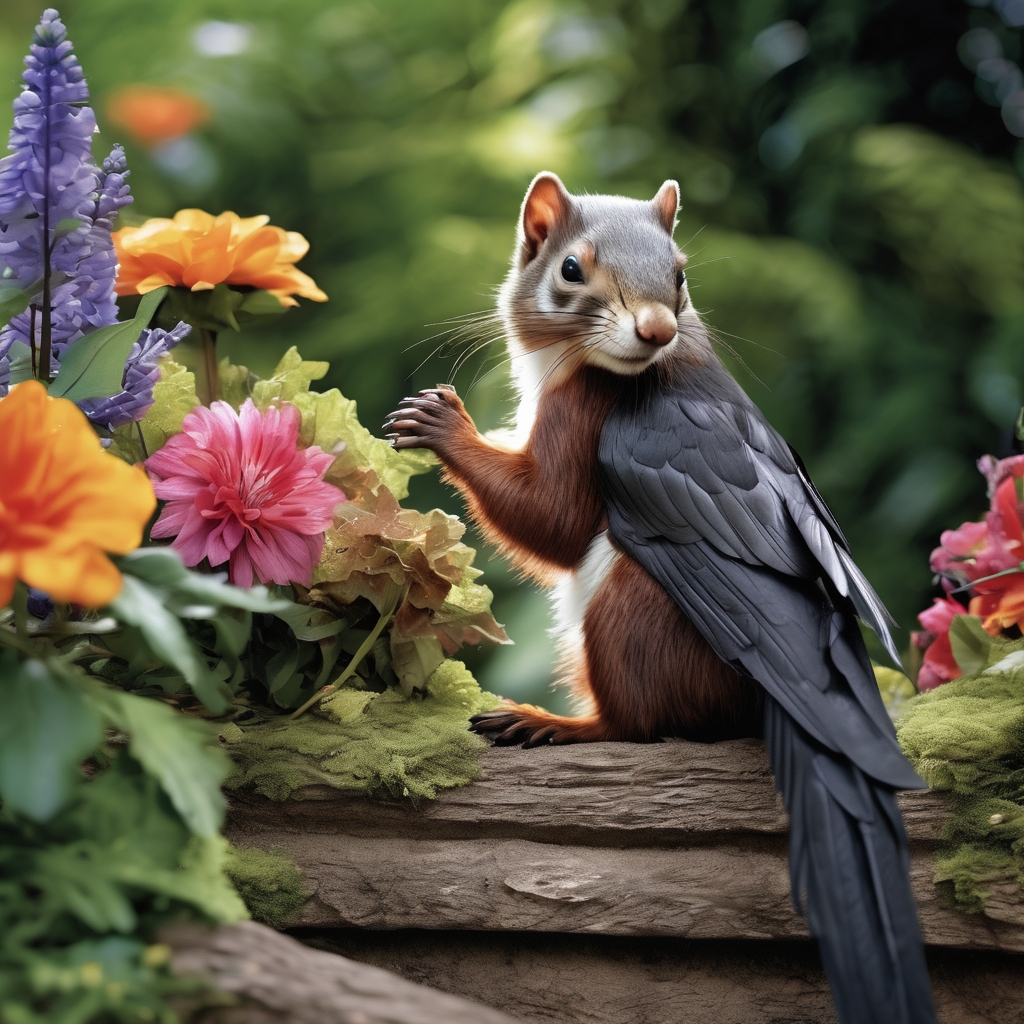} &
        \includegraphics[width=0.14\textwidth]{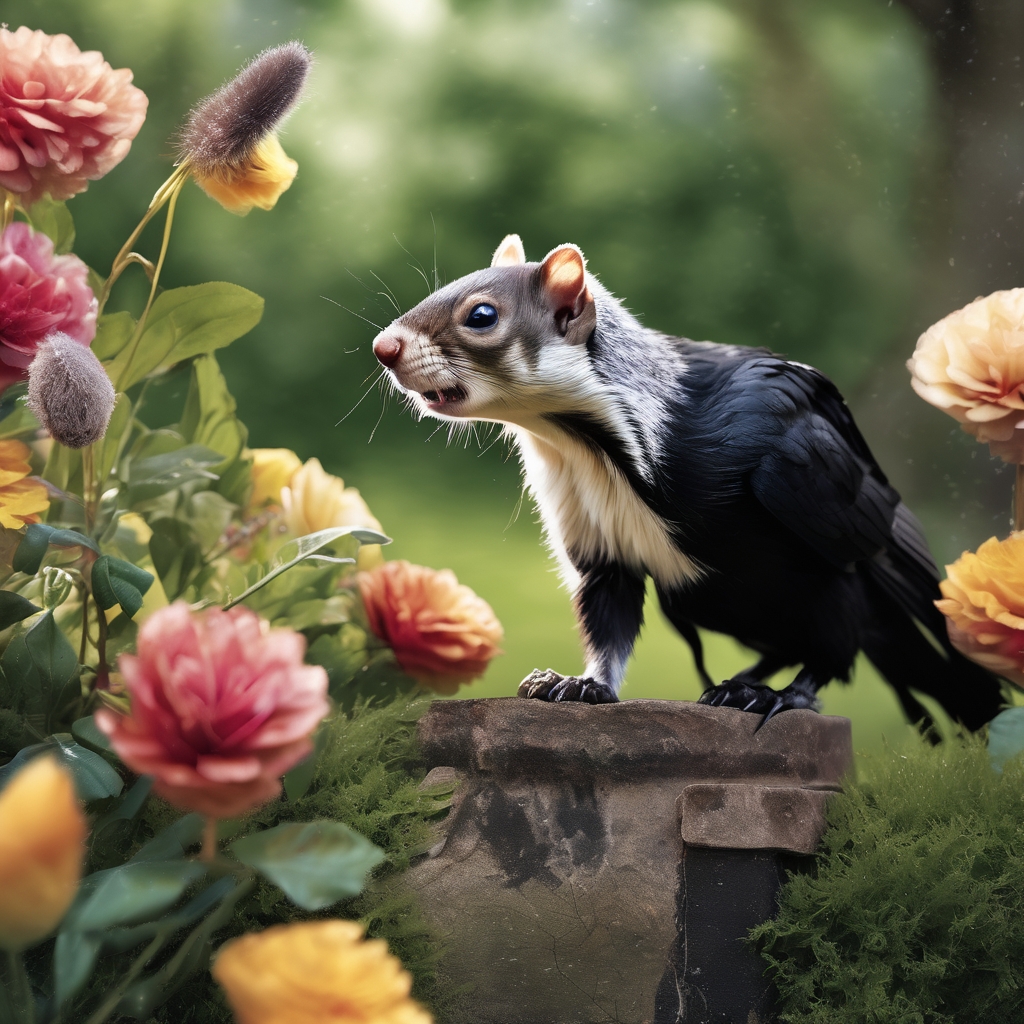} &
        \includegraphics[width=0.14\textwidth]{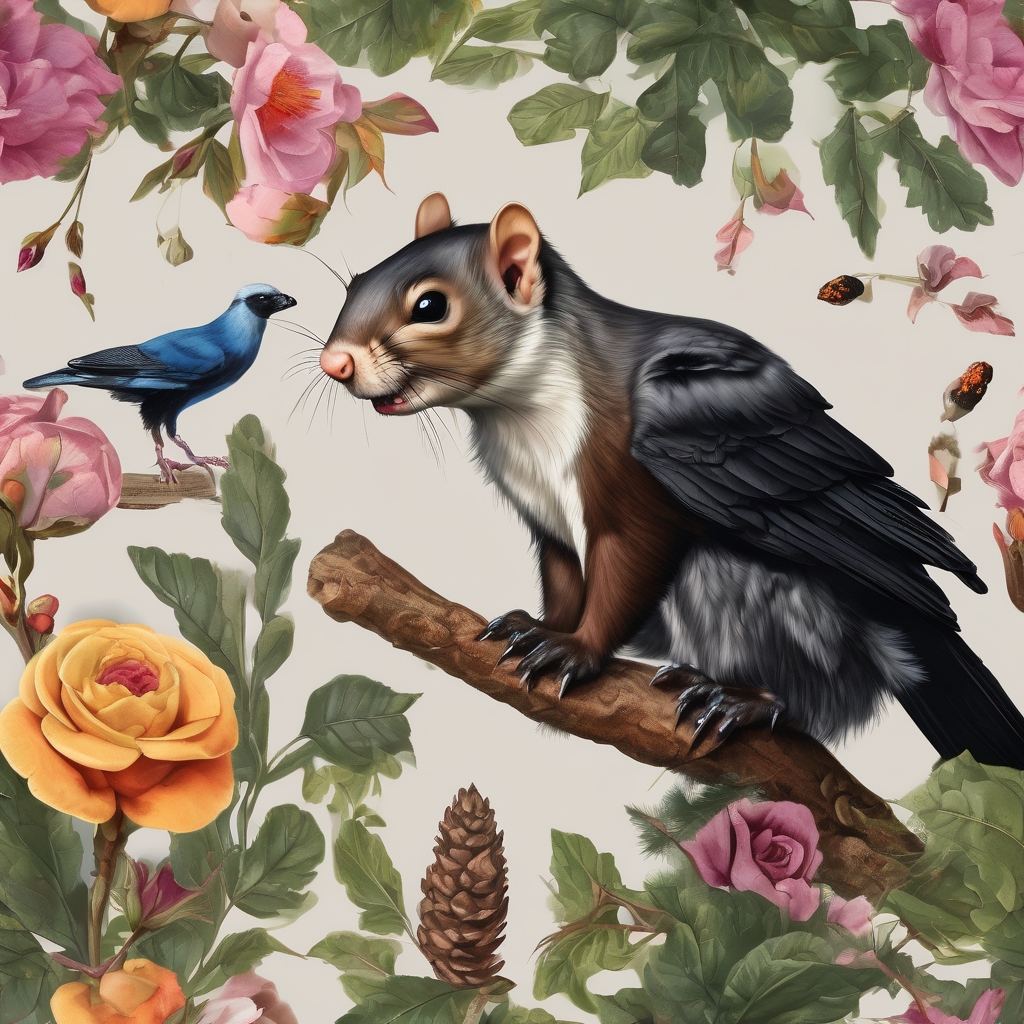} &
        \includegraphics[width=0.14\textwidth]{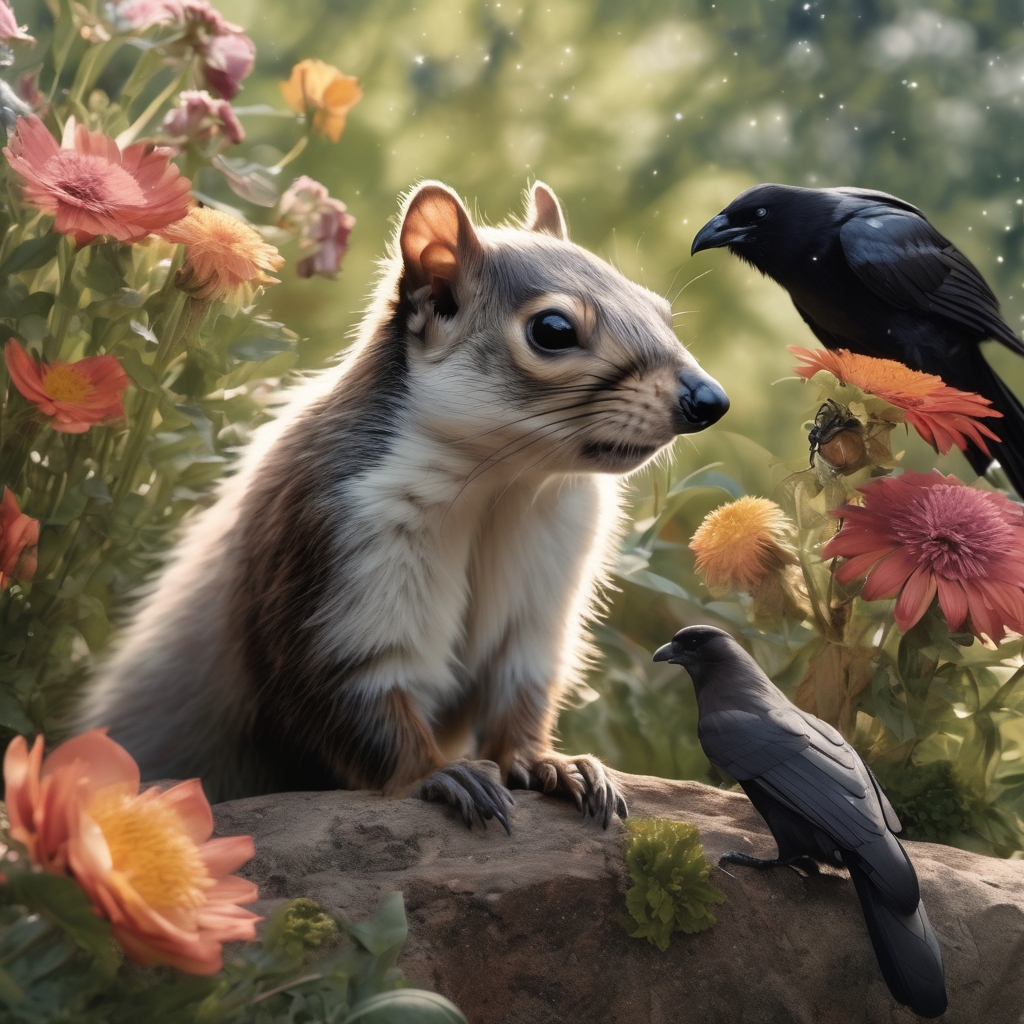} \\

        \raisebox{30pt}{\rotatebox{90}{Ranni}} &
        \includegraphics[width=0.14\textwidth]{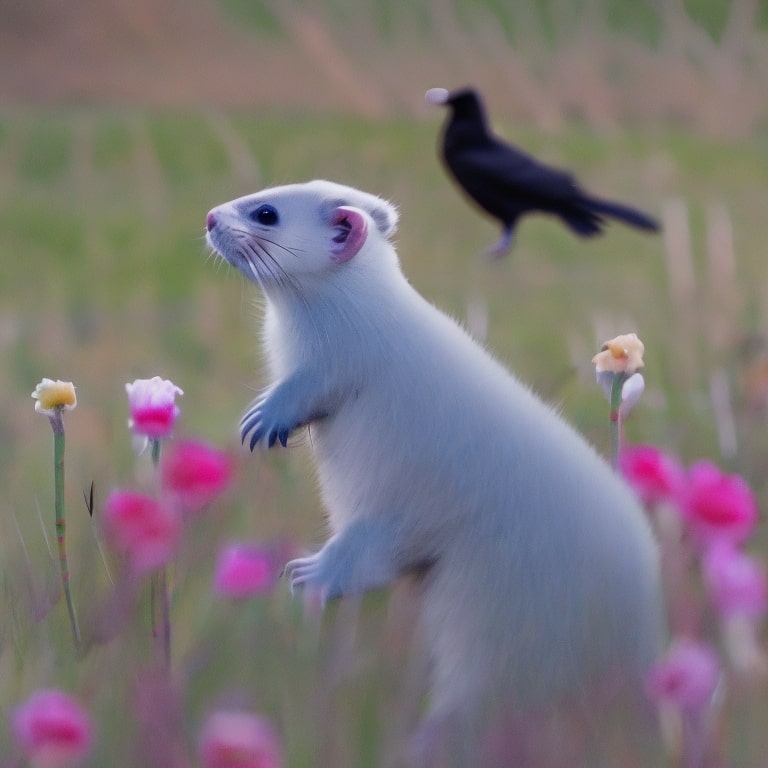} &
        \includegraphics[width=0.14\textwidth]{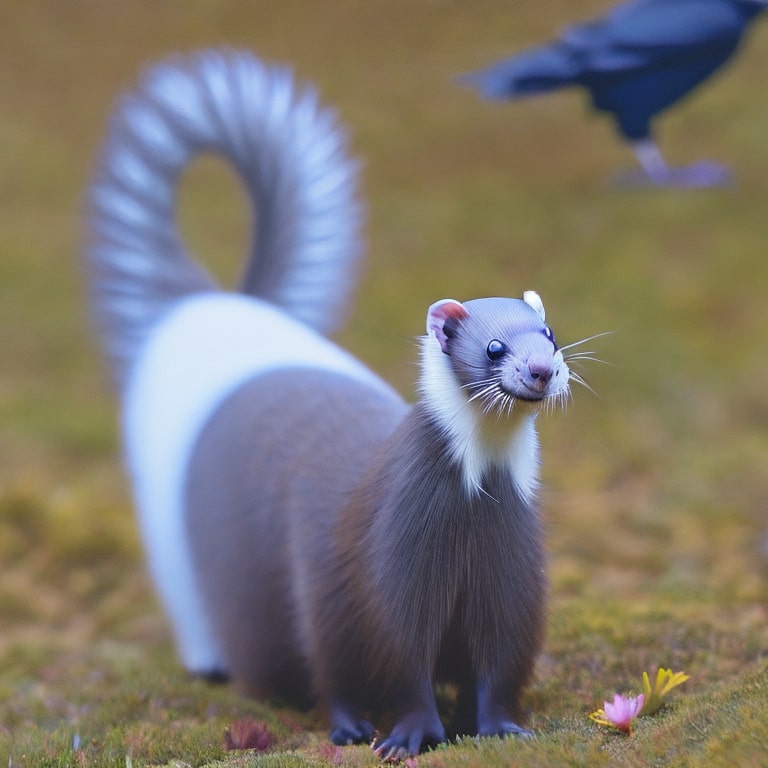} &
        \includegraphics[width=0.14\textwidth]{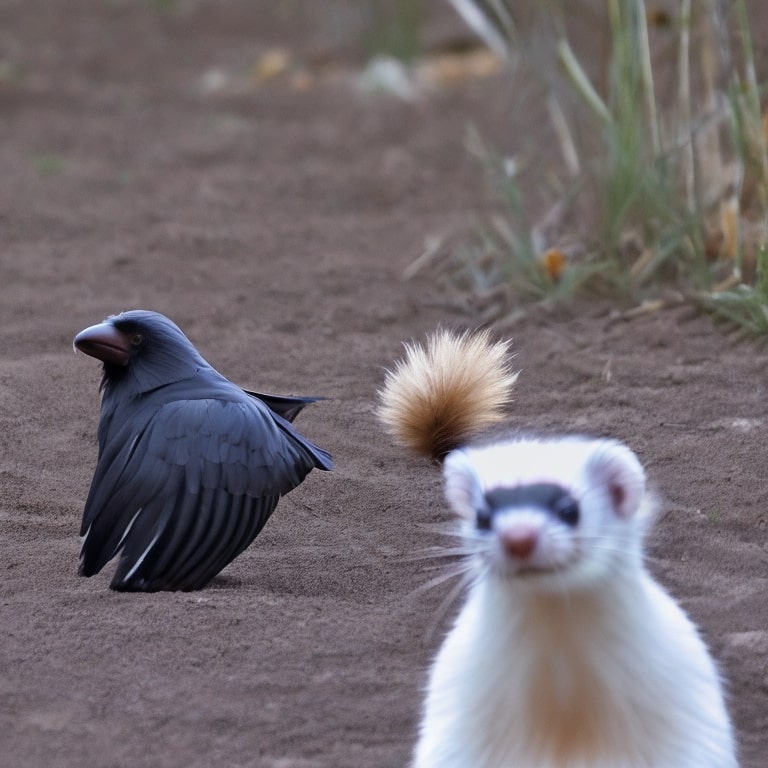} &
        \includegraphics[width=0.14\textwidth]{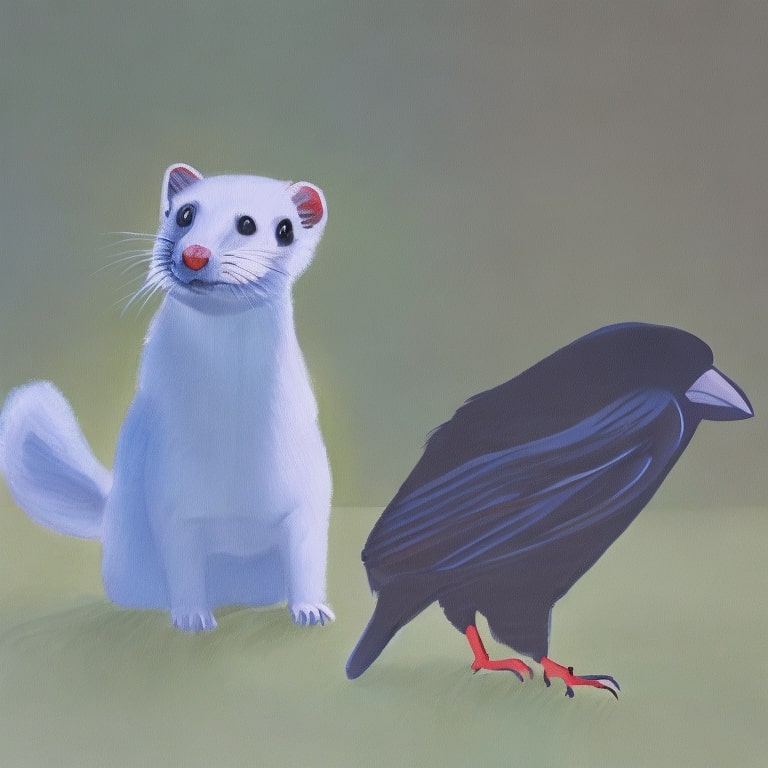} &
        \includegraphics[width=0.14\textwidth]{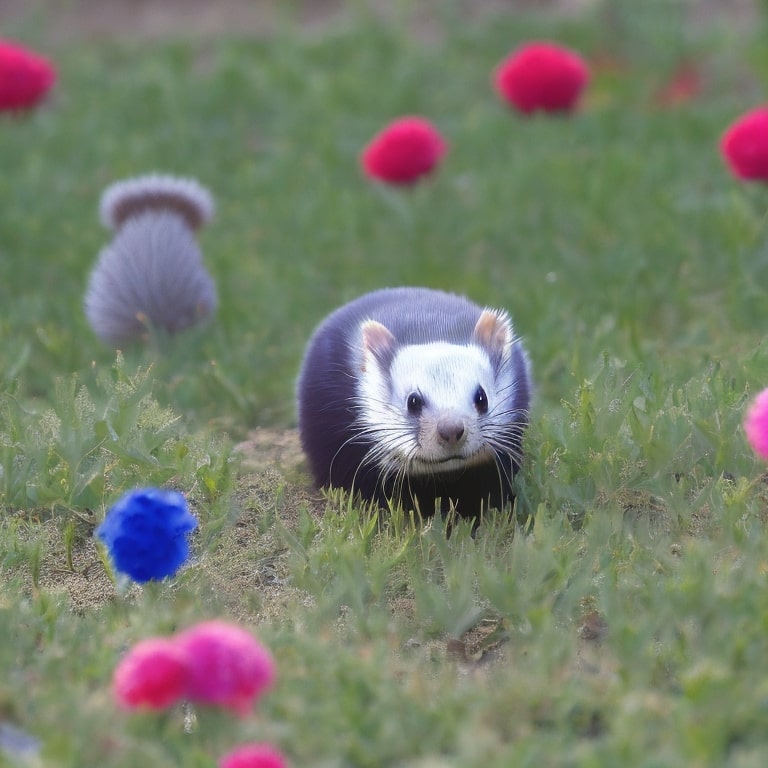} &
        \includegraphics[width=0.14\textwidth]{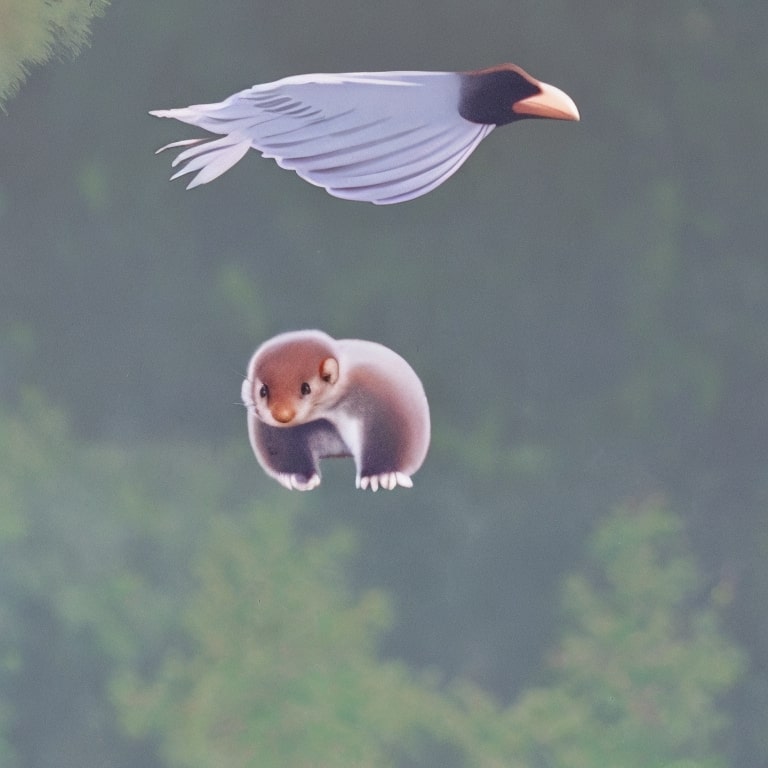} &
        \includegraphics[width=0.14\textwidth]{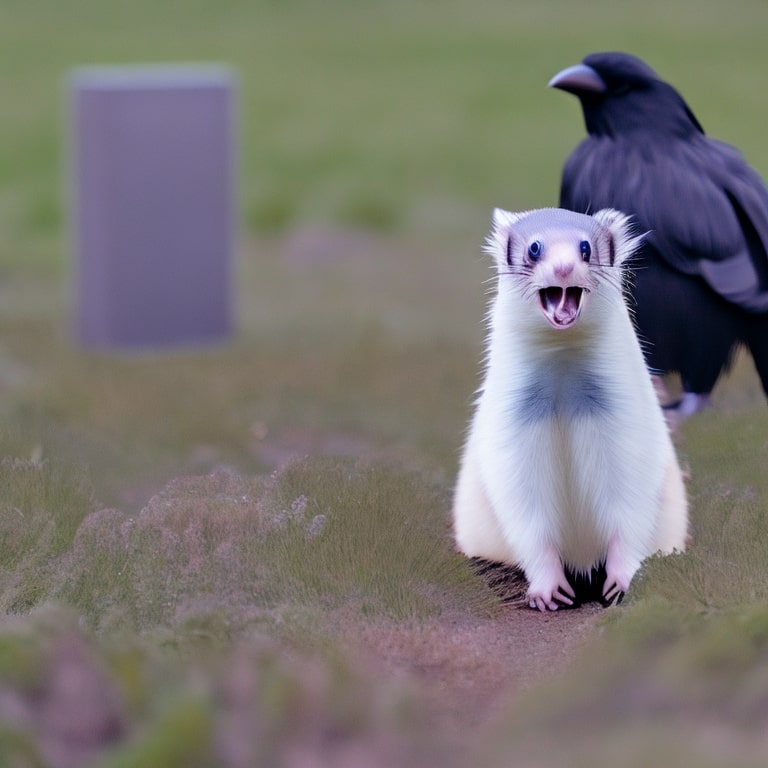} \\

        \raisebox{27pt}{\rotatebox{90}{LMD+}} &
        \includegraphics[width=0.14\textwidth]{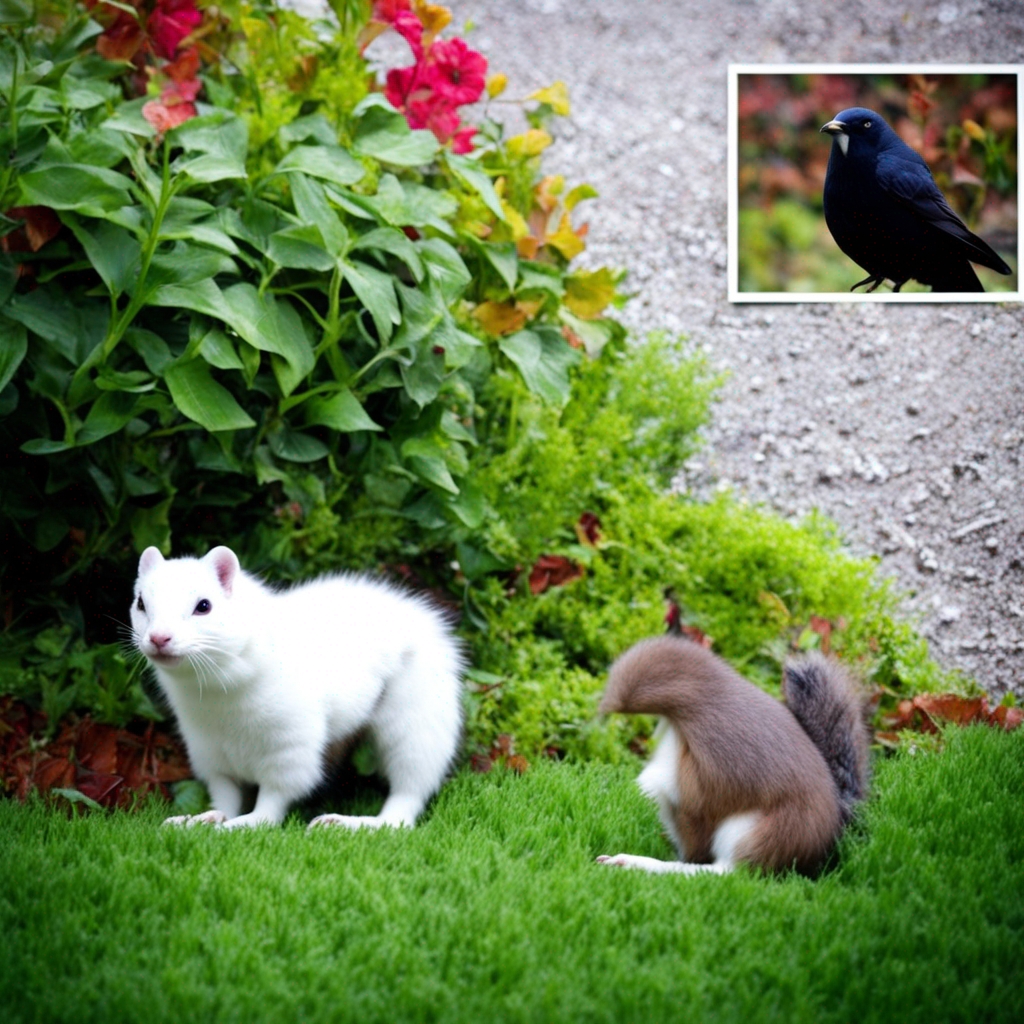} &
        \includegraphics[width=0.14\textwidth]{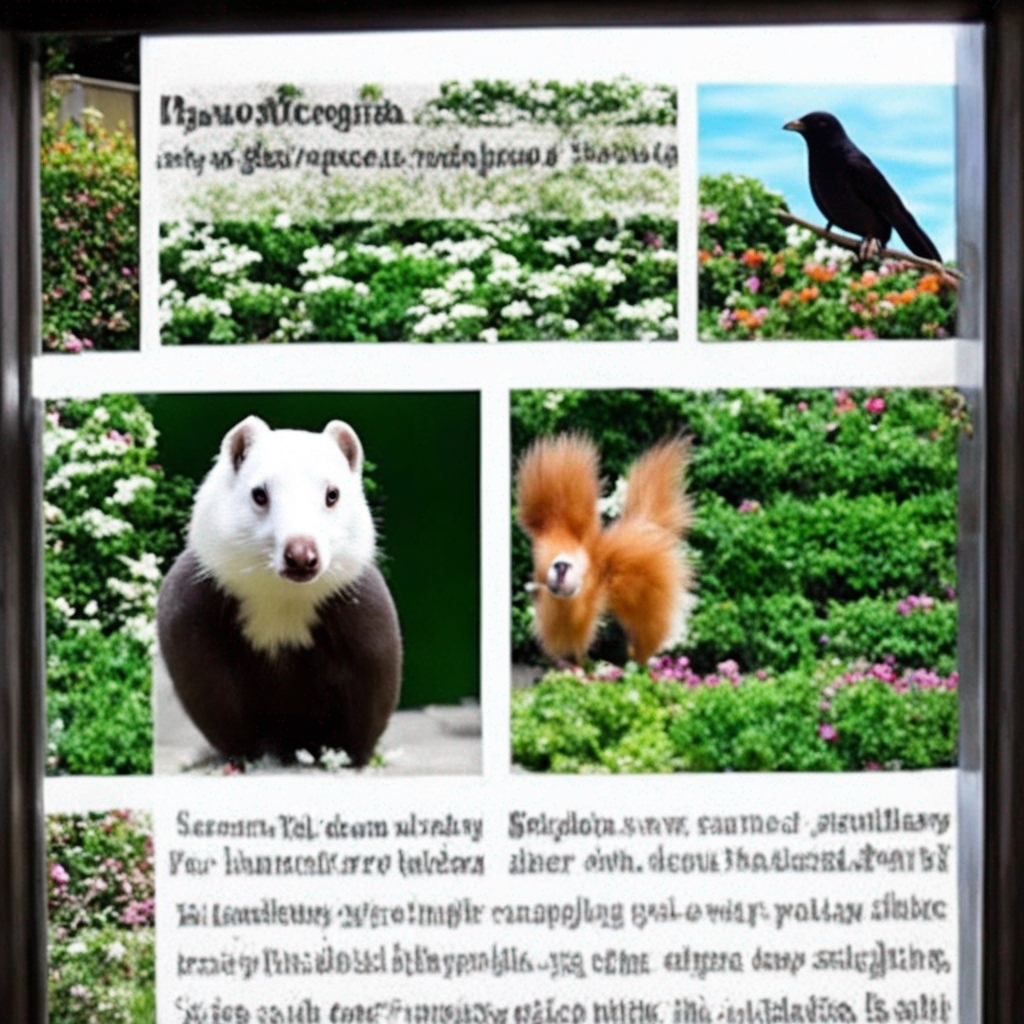} &
        \includegraphics[width=0.14\textwidth]{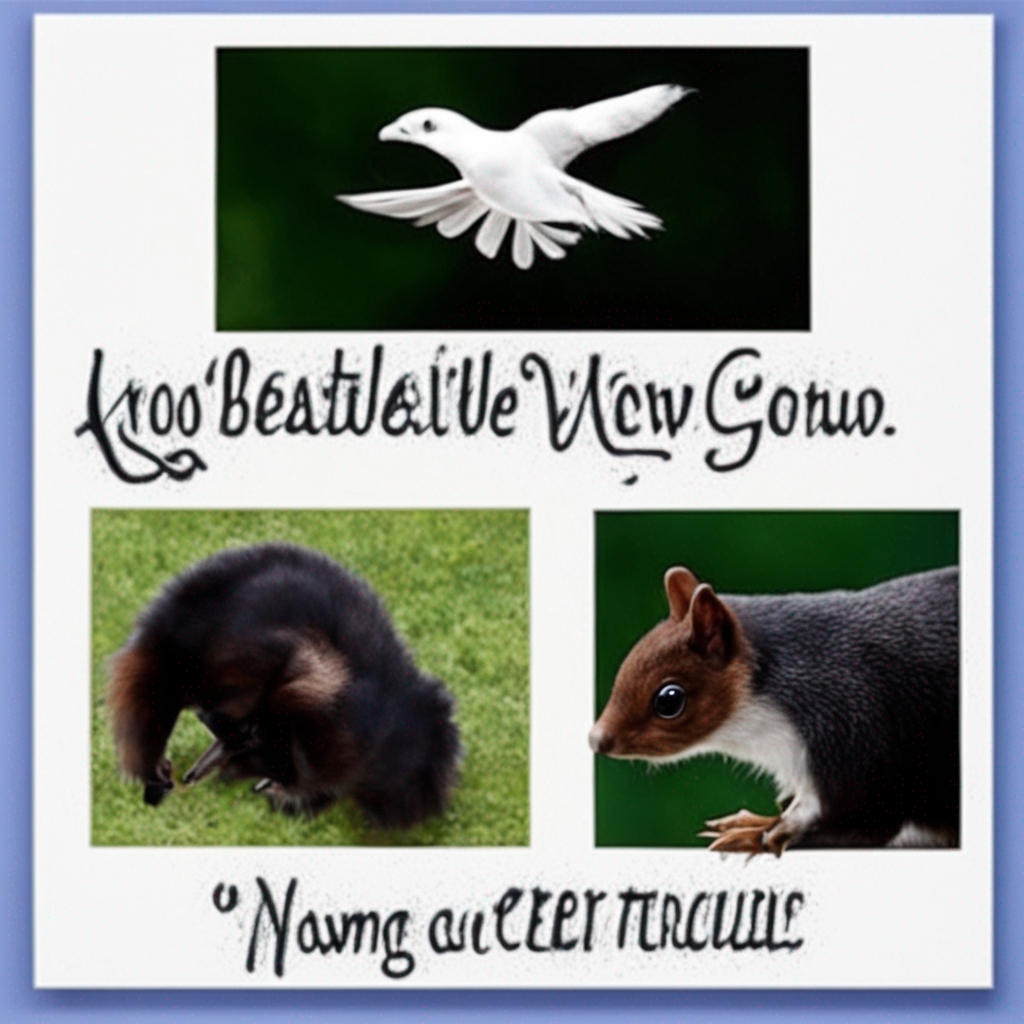} &
        \includegraphics[width=0.14\textwidth]{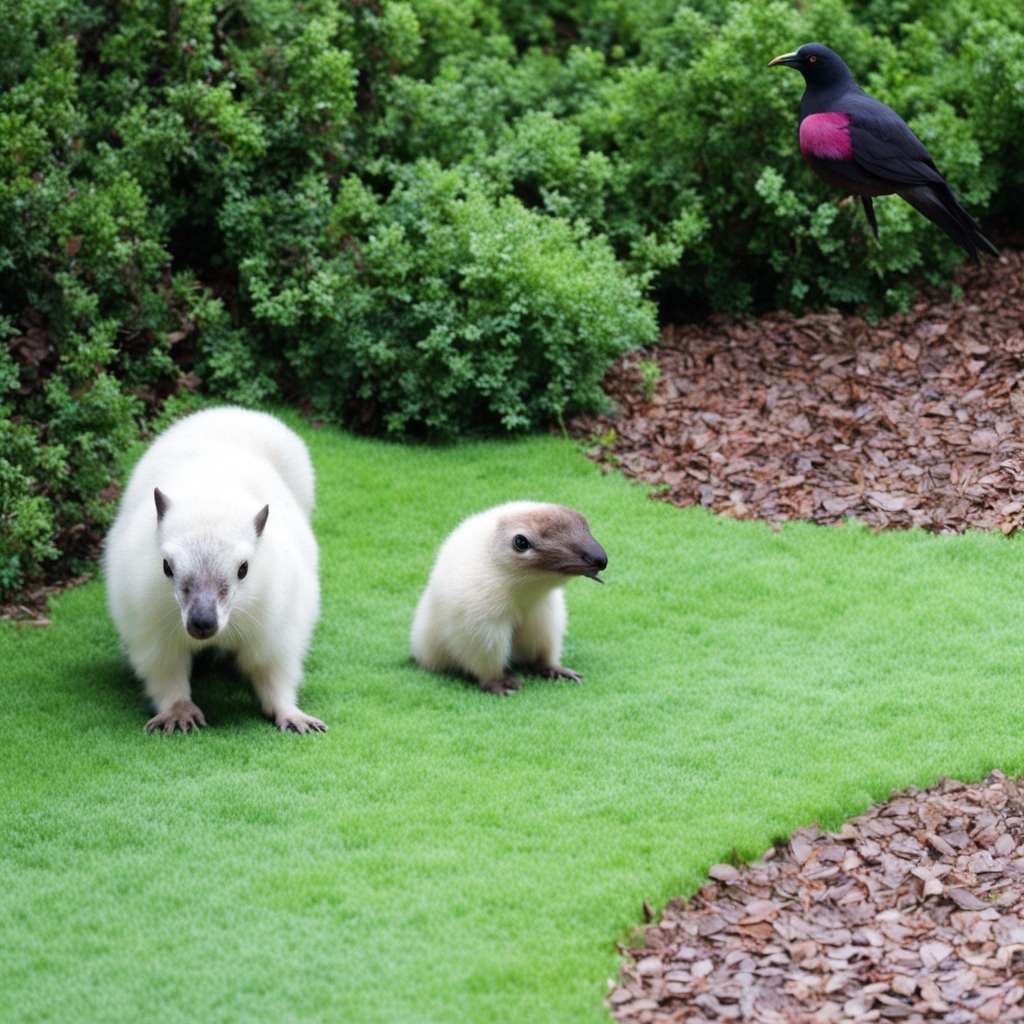} &
        \includegraphics[width=0.14\textwidth]{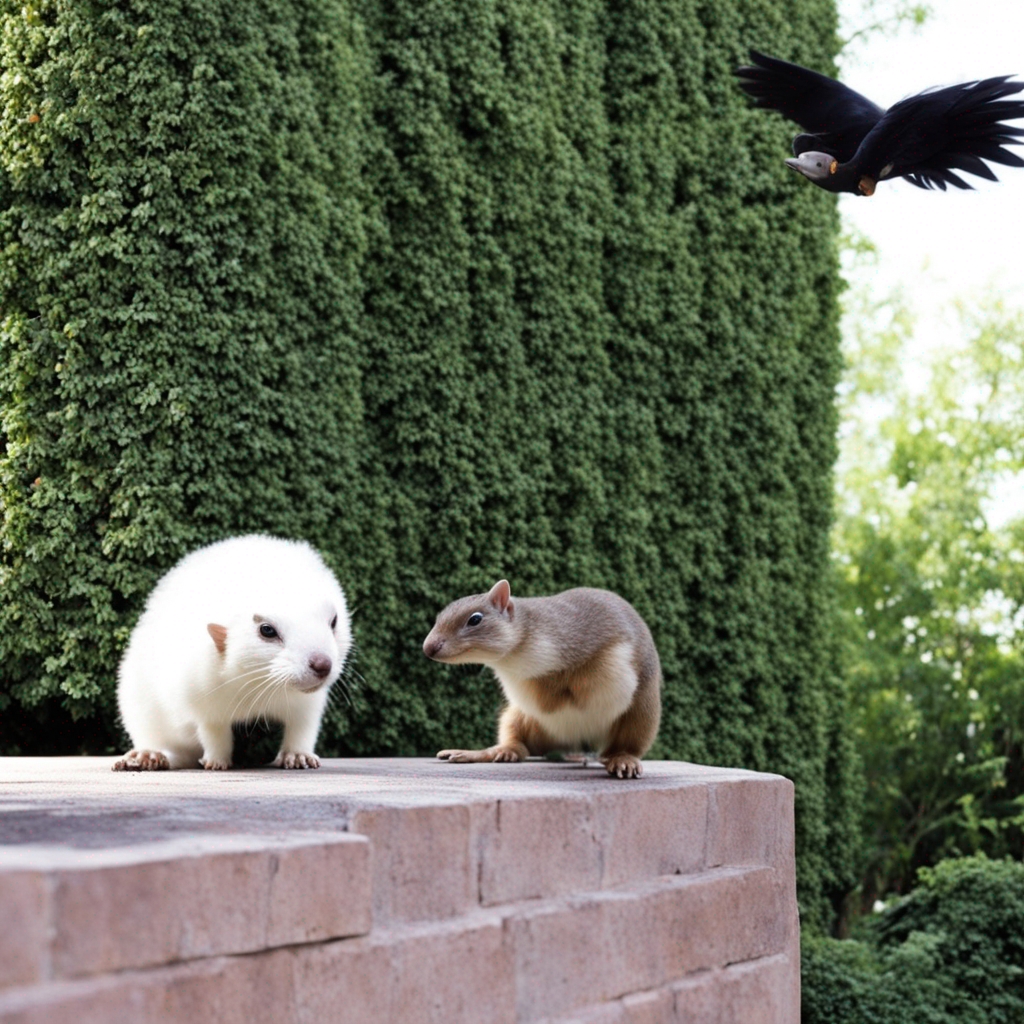} &
        \includegraphics[width=0.14\textwidth]{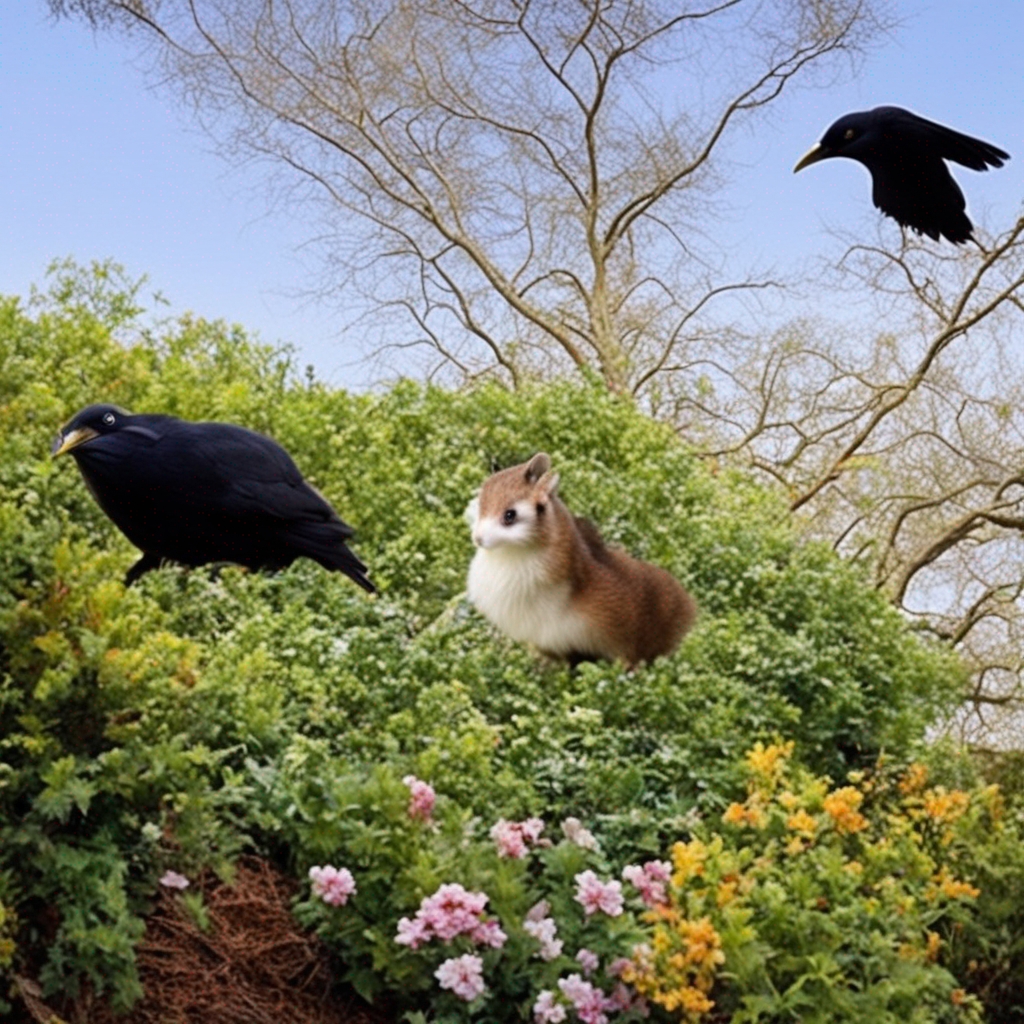} &
        \includegraphics[width=0.14\textwidth]{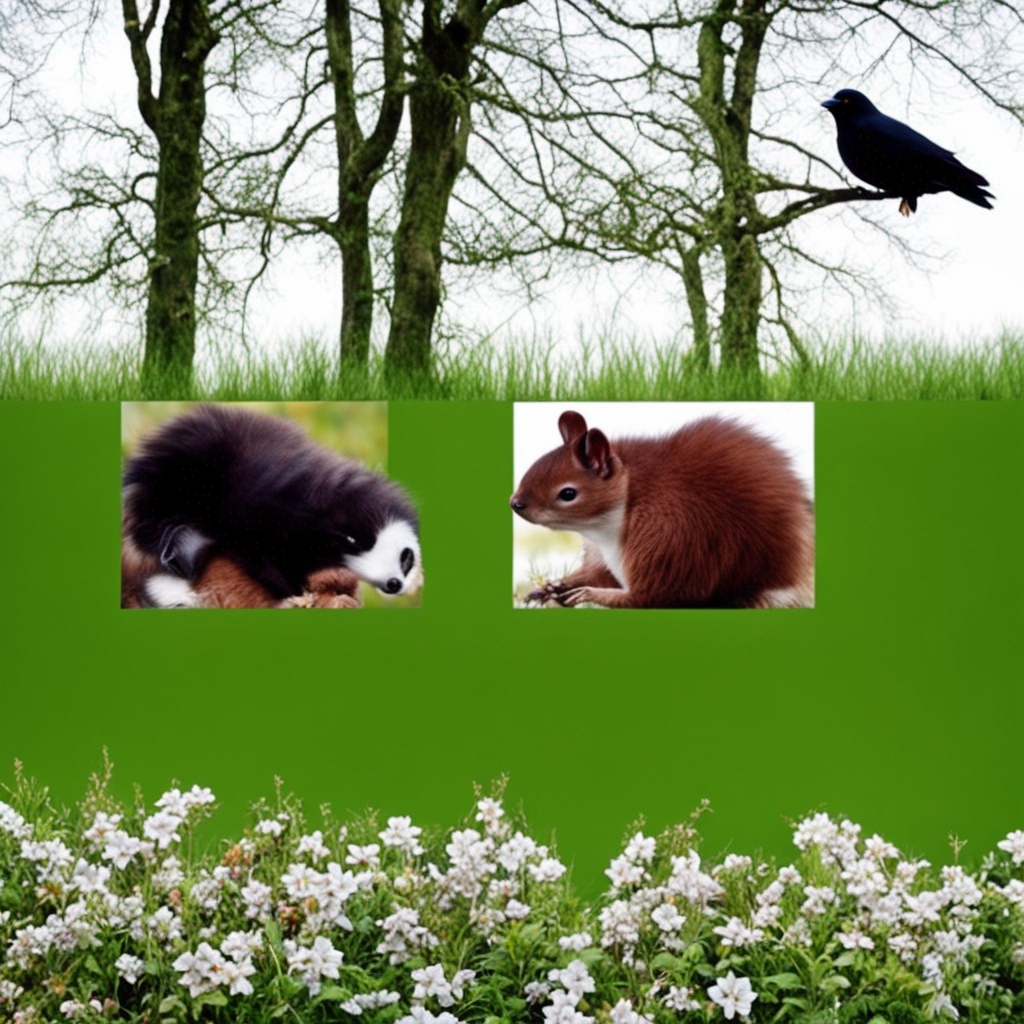} \\

    \end{tabular}
    }
    \captionof{figure}{Comparison of non-curated images generated from seeds 0 to 6.}
    \label{fig:non_curated}
\end{figure*}

\begin{acks}
We thank Elad Richardson and Narek Tumanyan for their early feedback and helpful suggestions.
We also thank the anonymous reviewers for their
meticulous comments which have helped improve our work.
This work was partially supported by ISF (grant 3441/21).
\end{acks}

\bibliographystyle{ACM-Reference-Format}
\bibliography{main}

\clearpage

\begin{appendices}
\section*{\LARGE Appendix}
\section{Technical Details}

\subsection{Architecture and Hyperparameters}

In our experiments, we used SDXL~\cite{podell2023sdxl} as the backbone model.

Our soft-layout network follows the Readout Guidance architecture with spatially aligned heads~\cite{luo2024readout}, with minor modifications: \begin{itemize} \item We use the self-attention keys and cross-attention queries from the decoder layers as inputs to the network, adjusting the input channels accordingly. \item We modify the final convolutional layer to output 10 channels instead of 3. \end{itemize}

The network was trained for 5{,}000 steps using a learning rate of $10^{-4}$. To compute the triplet loss, we sample 50 pixel triplets per image, selecting subject pixels with a probability of 0.75 and background pixels with a probability of 0.25. A similarity margin of $\alpha = 0.5$ is applied.

During image generation, we use 50 denoising steps, a DDPM scheduler, and a guidance scale of 7.5. Guidance is applied during the first 15 denoising steps, with 5 gradient descent iterations per step. The decisive loss is computed using the weights $\alpha_{\textit{cross}} = 0.3$, $\alpha_{\textit{var}} = 0.21$, and $\alpha_{\textit{dice}} = 0.49$. For $\mathcal{L}_{\textit{dice}}$, a temperature value of $\tau = 15$ is used. For hard-clustering, we set a sliding window size of $w=30$ and a variance threshold of $\sigma_{\textit{cluster}}^2 = 0.025$.

\subsection{Dataset Generation}

Our dataset consists of approximately 1,500 generated images and their corresponding segmentation maps.

To construct training prompts, we use the same 20 MSCOCO classes as in MIC~\cite{binyamin2024make}. Each prompt randomly includes 1–3 classes. For each selected class, we assign a quantity between 1 and 10, with a probability of 0.9. We optionally prepend a prefix (with probability 0.8) and append a postfix (with probability 0.6), both sampled from fixed lists: \begin{itemize} \item Prefixes: ``a photo of'', ``an image of'', ``a picture of'', ``a painting of''. \item Postfixes: ``on the grass'', ``on the road'', ``on the ground'', ``in a yard''. \end{itemize}

We observe that our soft-layout network generalizes well to unseen object classes, backgrounds, and prompt structures, owing to its lightweight design and the use of expressive attention features from the pre-trained diffusion model.

\subsection{Computational Resource Usage}

All experiments were conducted on an NVIDIA A100 GPU, with all computations — including clustering — performed on the GPU. Similar to Readout Guidance, our sampling process takes approximately 77 seconds and utilizes 36 GB of VRAM, compared to 7 seconds and 8 GB for vanilla SDXL. Our implementation builds upon the Readout Guidance and Bounded Attention codebases, which were not optimized for resource efficiency. As such, further code optimization is likely to reduce both runtime and memory usage.

\section{Additional Results}

\begin{figure*}[th!]
    \centering
    \setlength{\tabcolsep}{0.001\textwidth}
    {\small
    \begin{tabular}{c c c c c c c}

        \multicolumn{7}{c}{"\textbf{Three apples}, \textbf{two oranges}, and an \textbf{avocado} on the kitchen counter"} \\
        \includegraphics[width=0.14\textwidth]{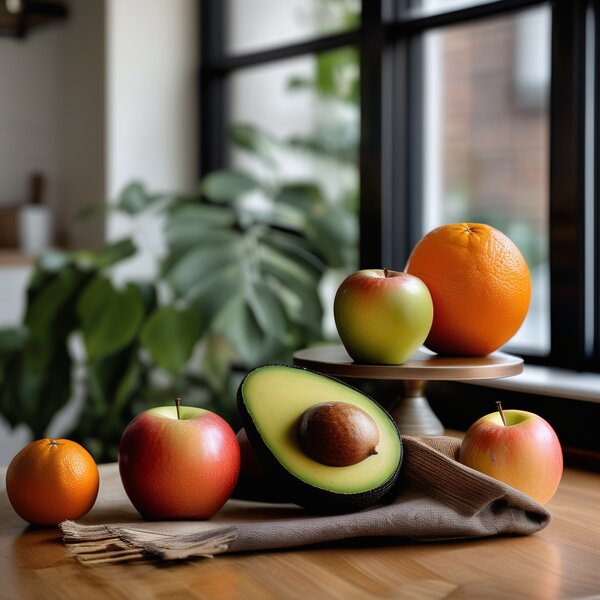} &
        \includegraphics[width=0.14\textwidth]{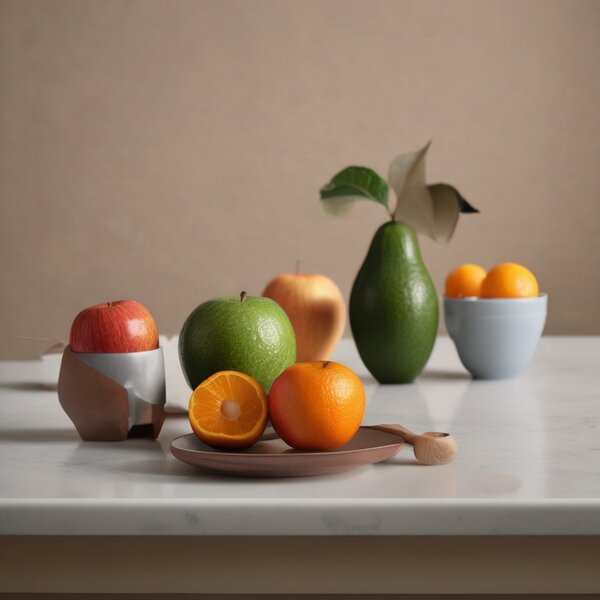} &
        \includegraphics[width=0.14\textwidth]{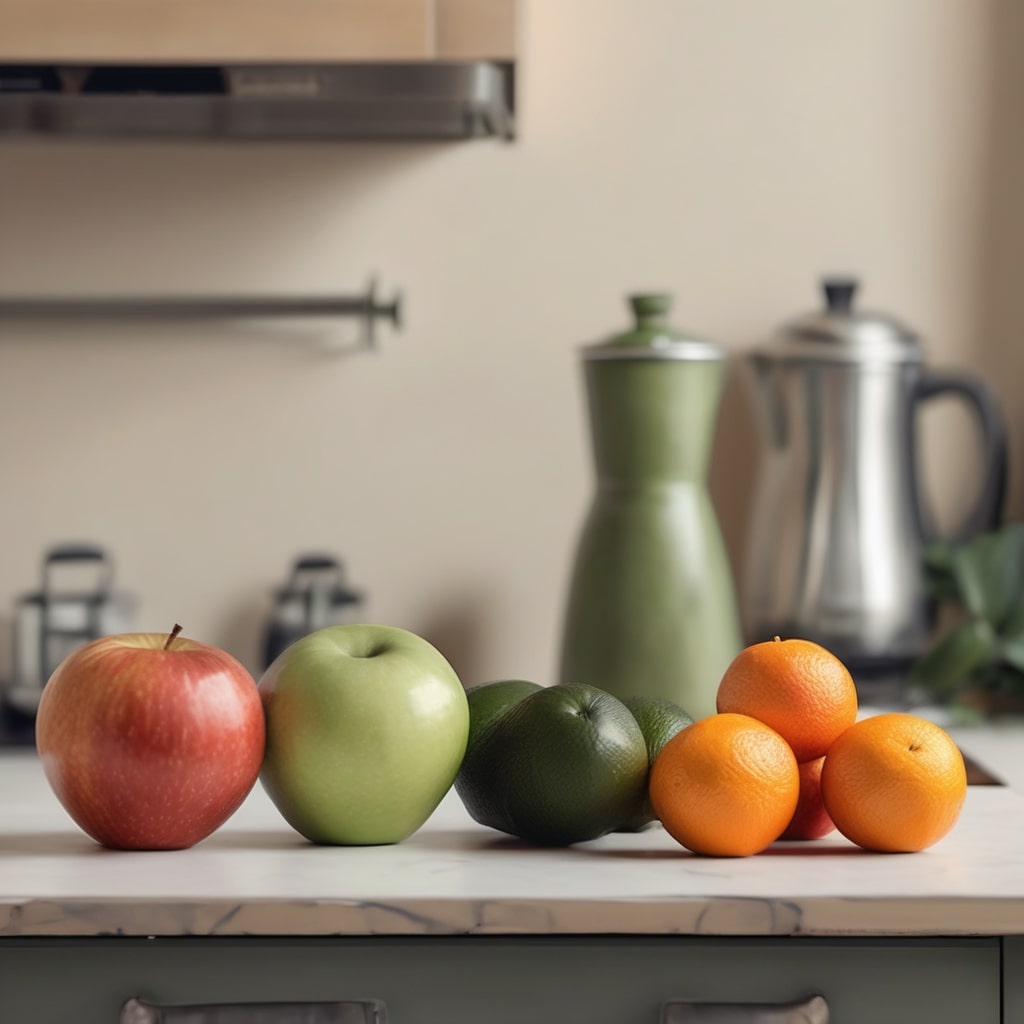} &
        \includegraphics[width=0.14\textwidth]{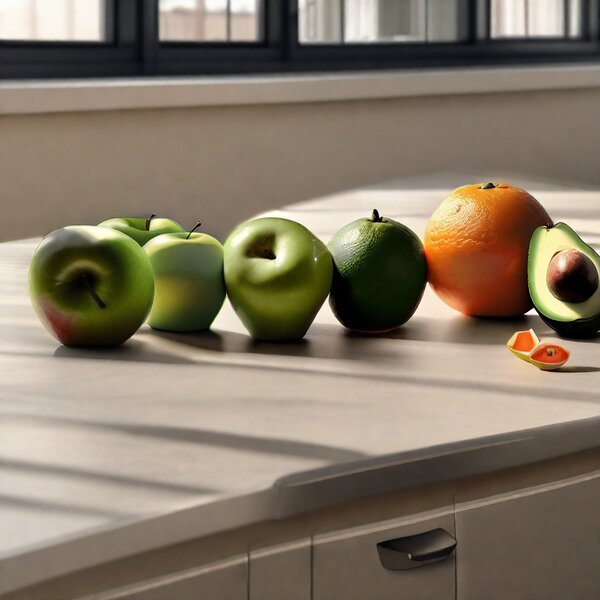} &
        \includegraphics[width=0.14\textwidth]{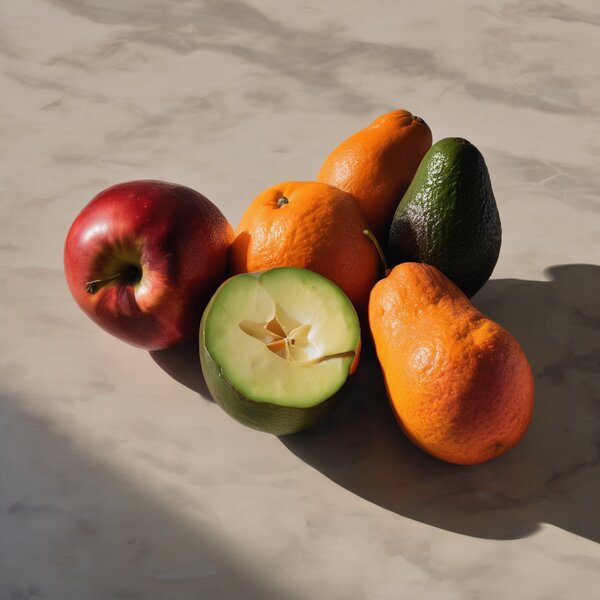} &
        \includegraphics[width=0.14\textwidth]{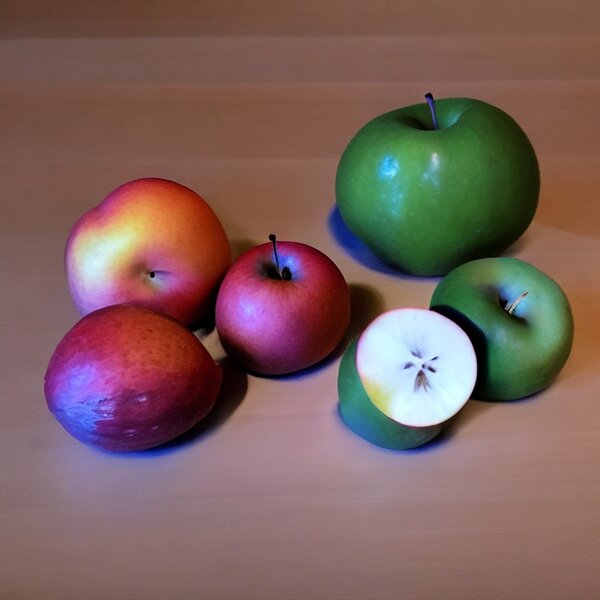} &
        \includegraphics[width=0.14\textwidth]{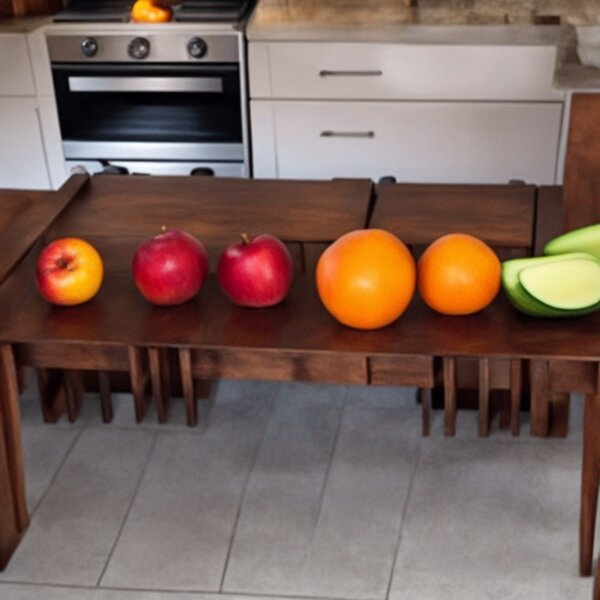} \\

        \multicolumn{7}{c}{"A \textbf{monkey} and a \textbf{sloth} and a \textbf{frog} in the rainforest"} \\
        \includegraphics[width=0.14\textwidth]{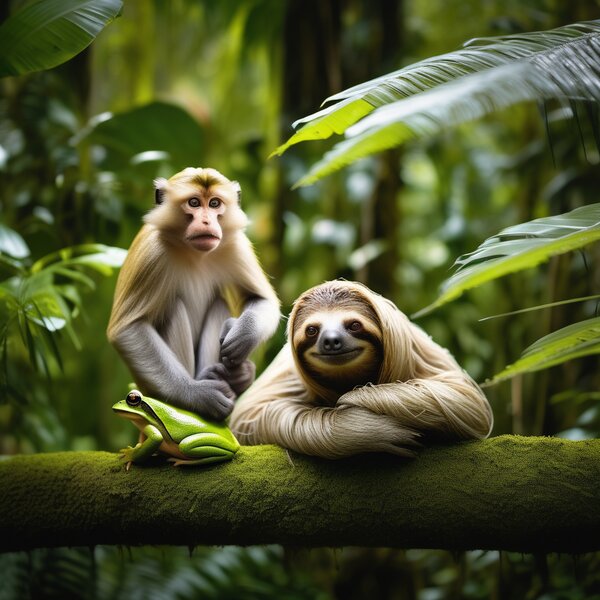} &
        \includegraphics[width=0.14\textwidth]{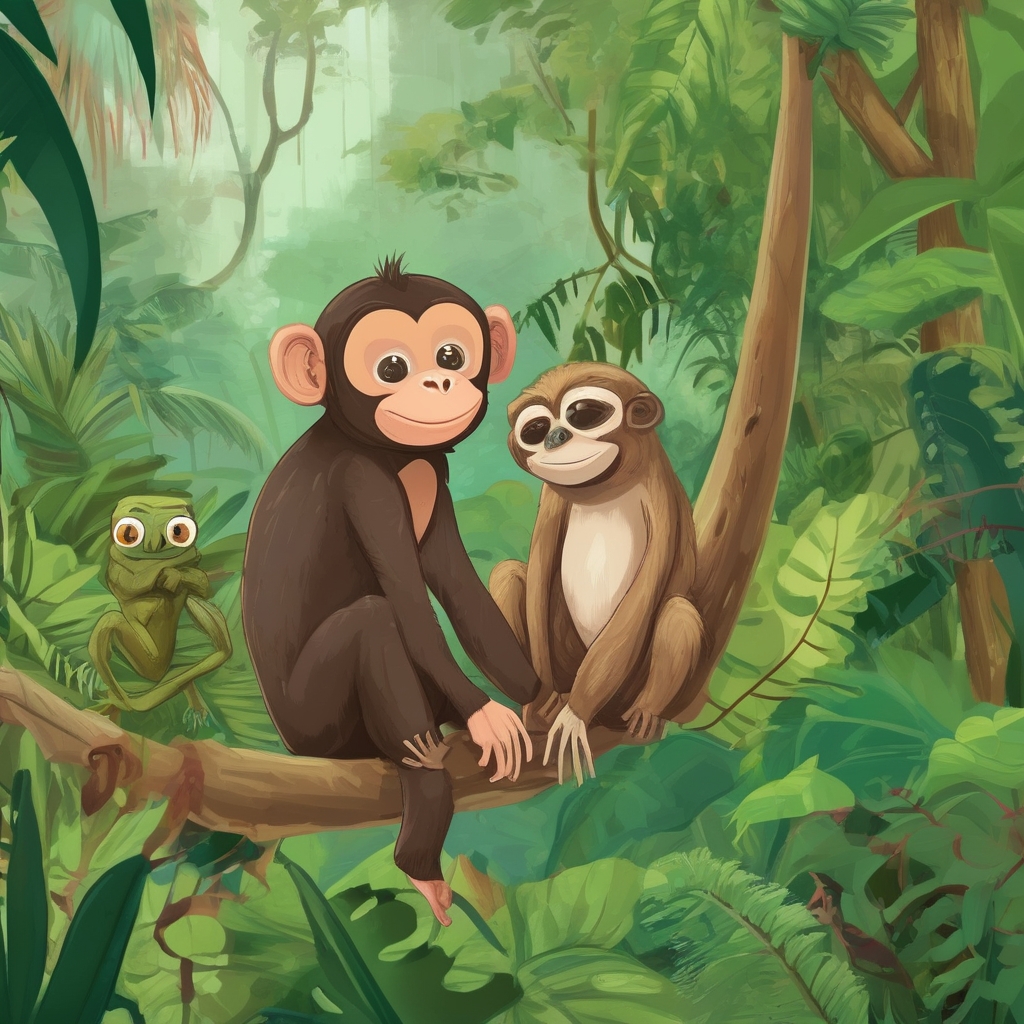} &
        \includegraphics[width=0.14\textwidth]{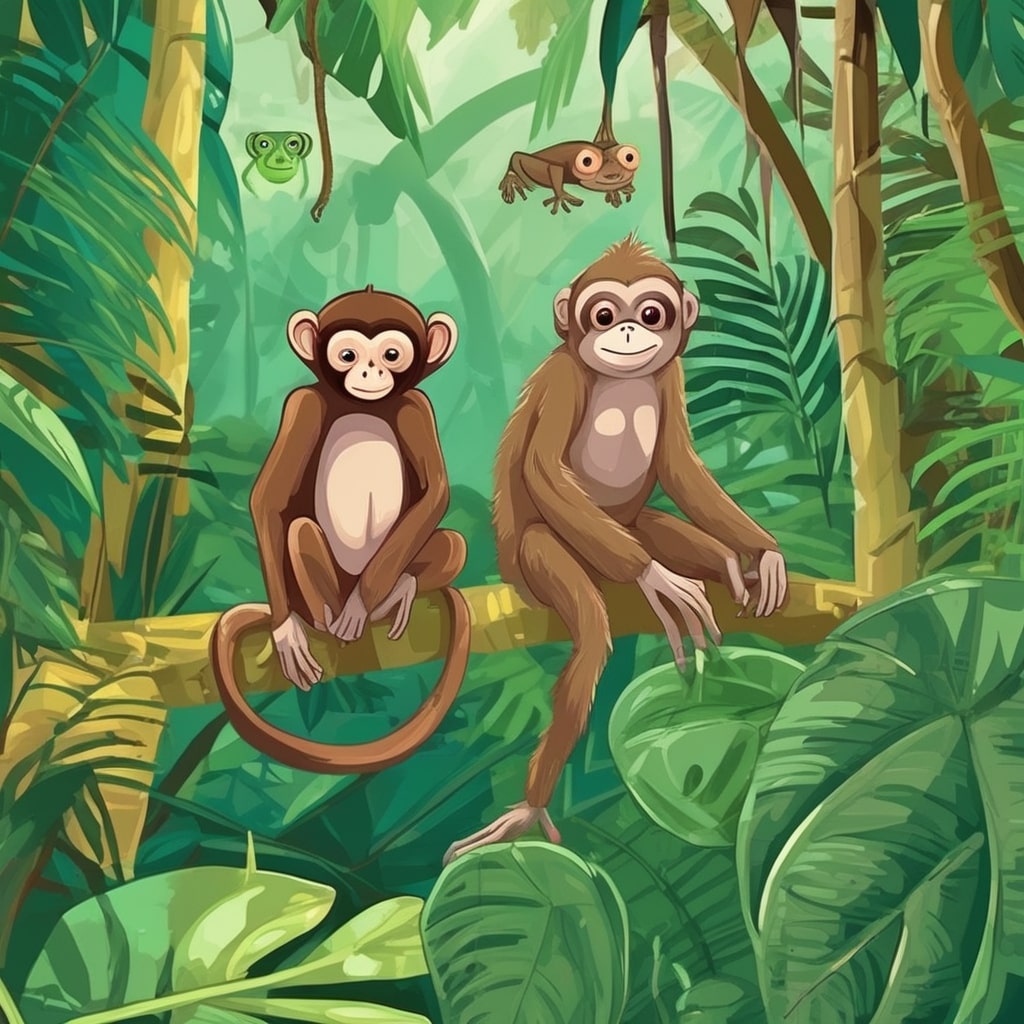} &
        \includegraphics[width=0.14\textwidth]{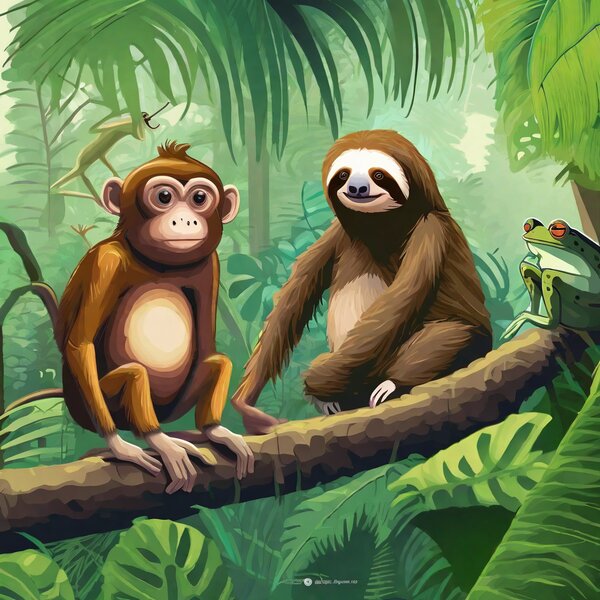} &
        \includegraphics[width=0.14\textwidth]{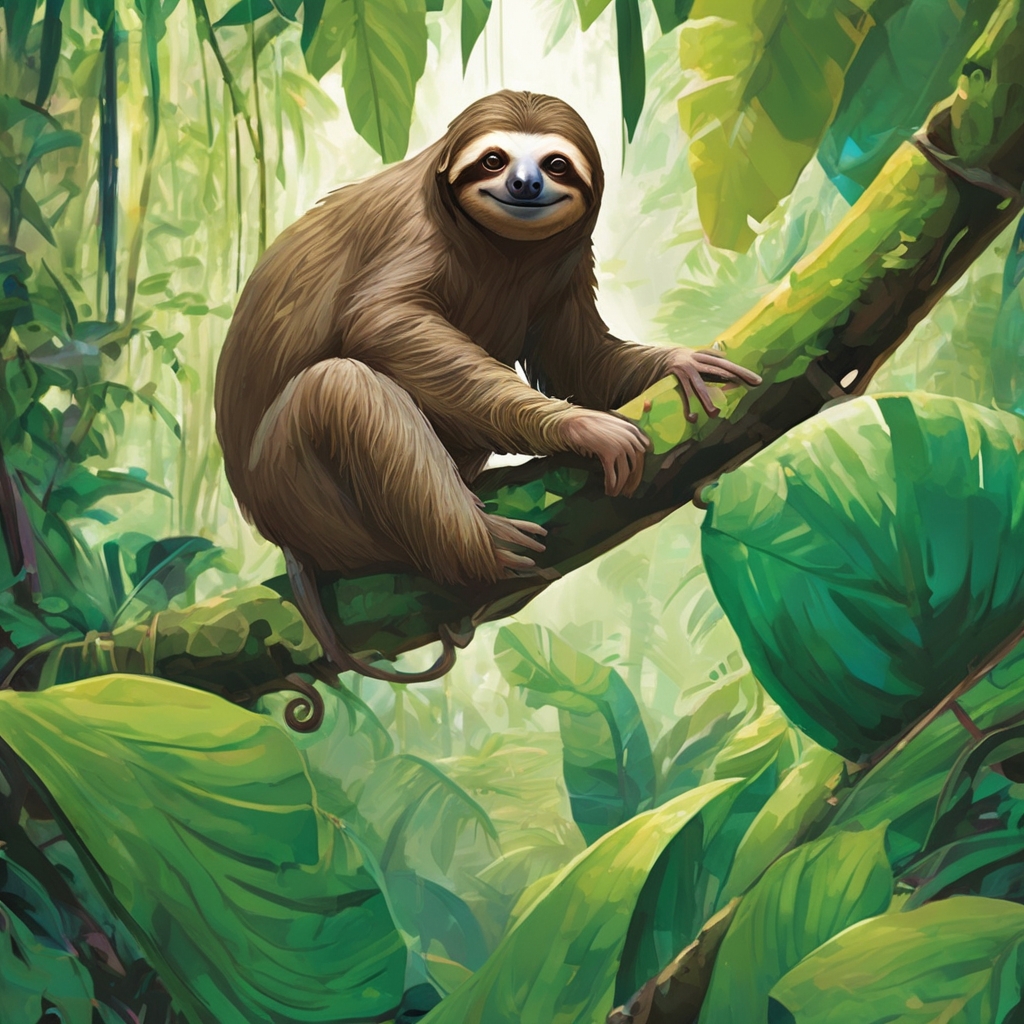} &
        \includegraphics[width=0.14\textwidth]{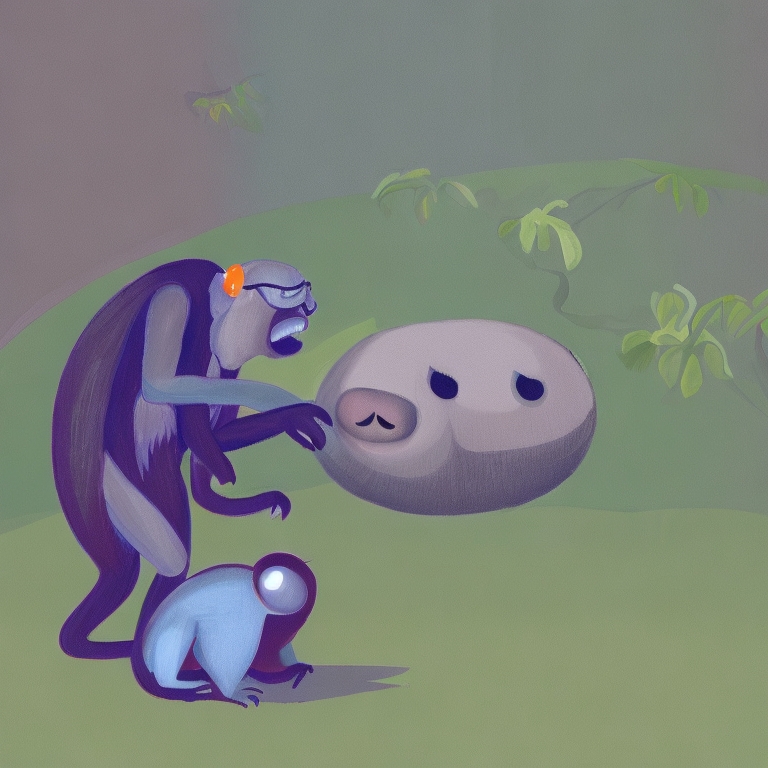} &
        \includegraphics[width=0.14\textwidth]{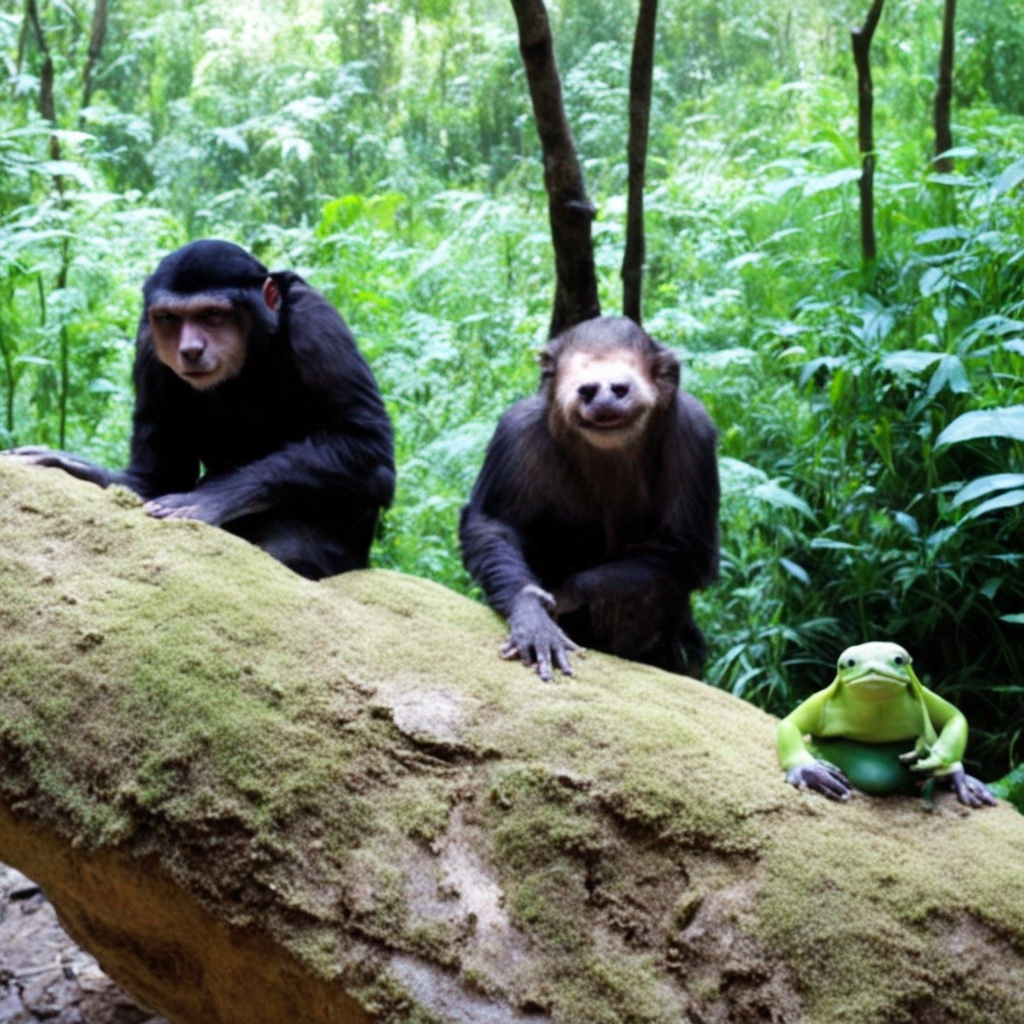} \\

        \multicolumn{7}{c}{"\textbf{Two electric guitars} and \textbf{two violins} on the floor"} \\
        \includegraphics[width=0.14\textwidth]{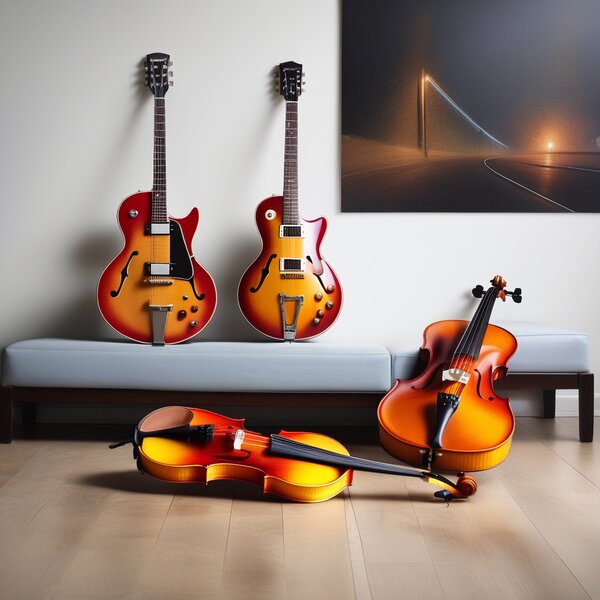} &
        \includegraphics[width=0.14\textwidth]{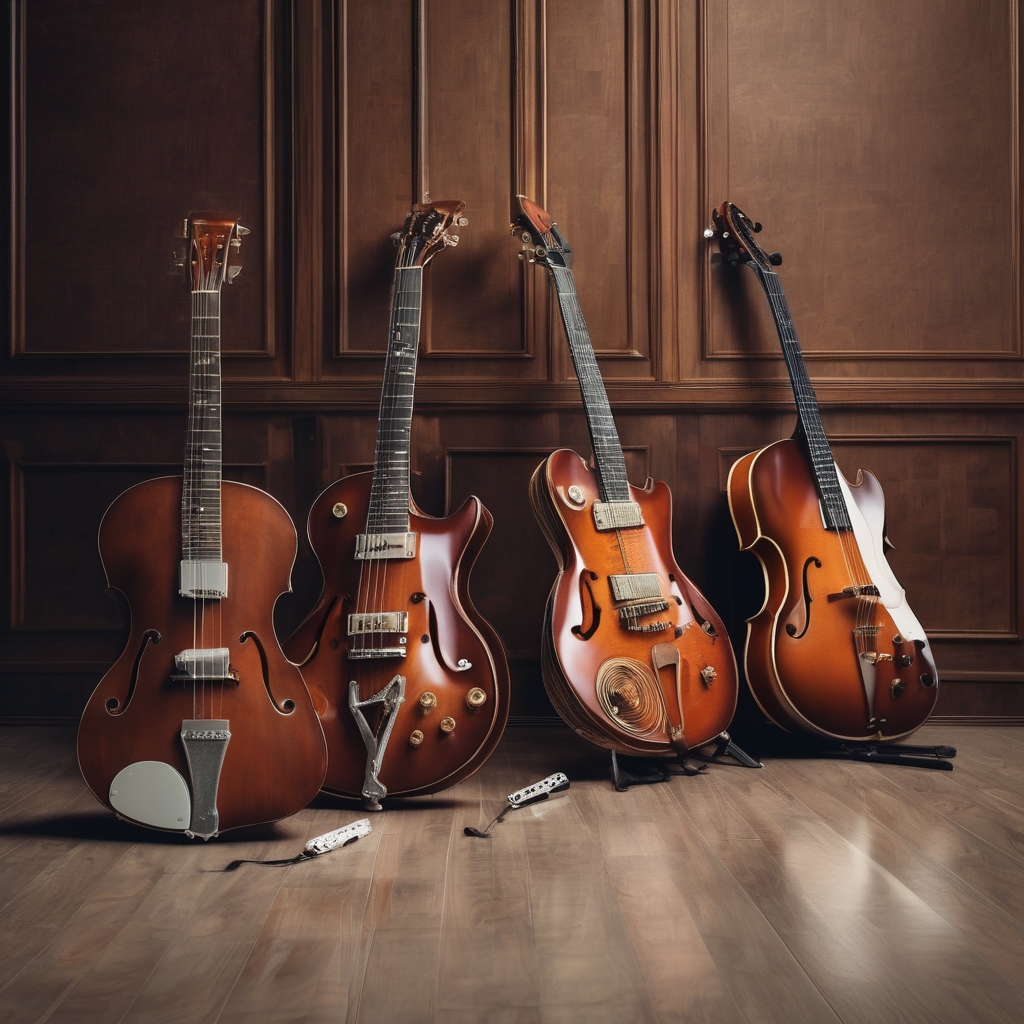} &
        \includegraphics[width=0.14\textwidth]{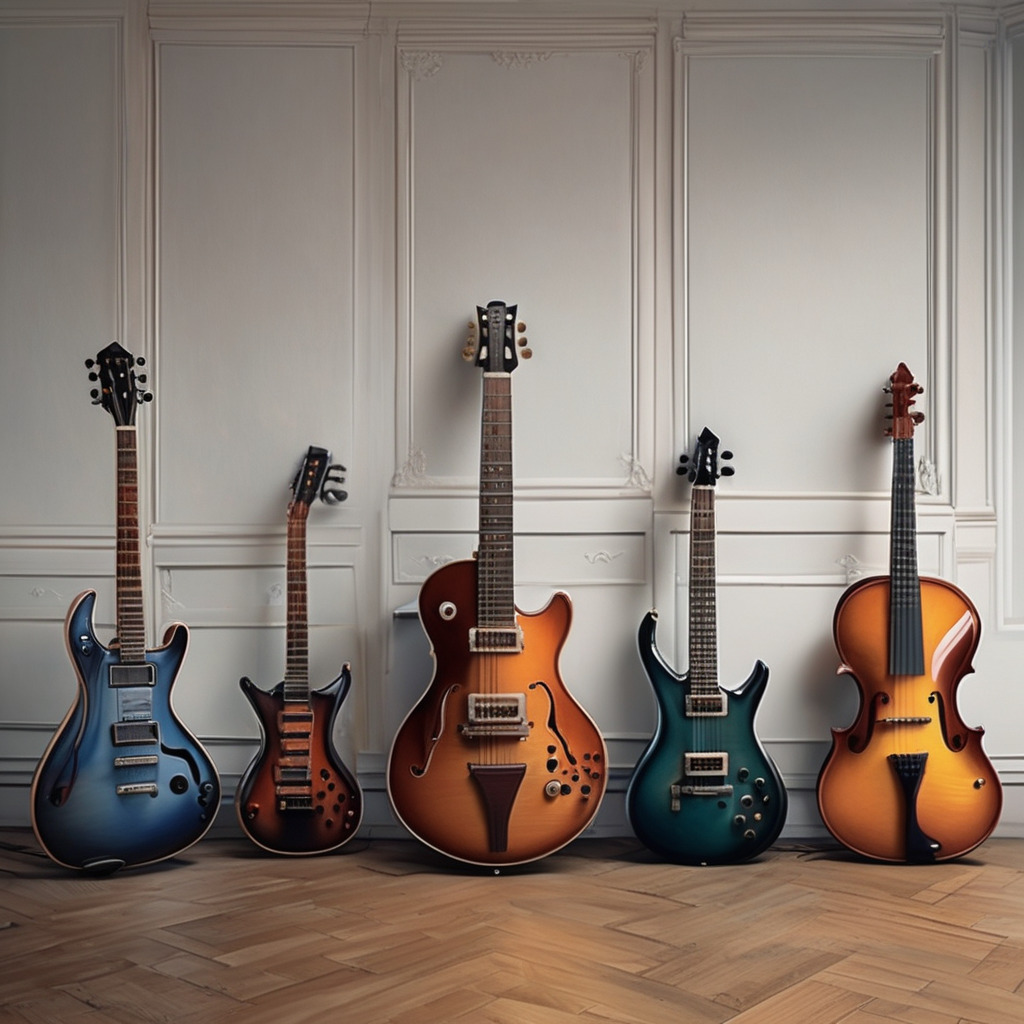} &
        \includegraphics[width=0.14\textwidth]{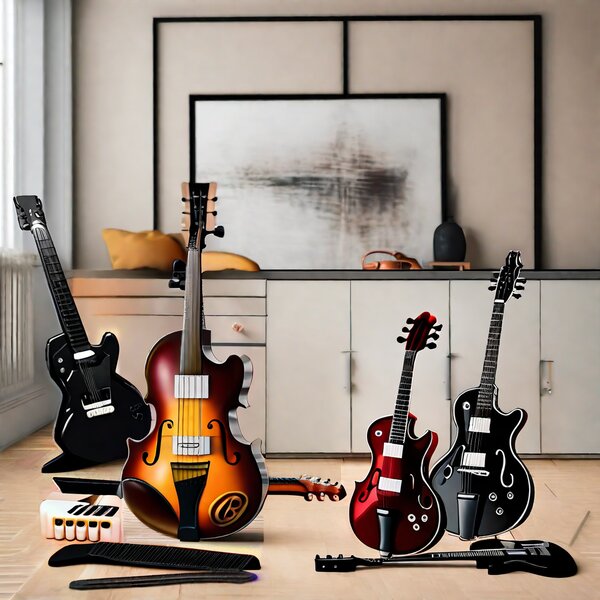} &
        \includegraphics[width=0.14\textwidth]{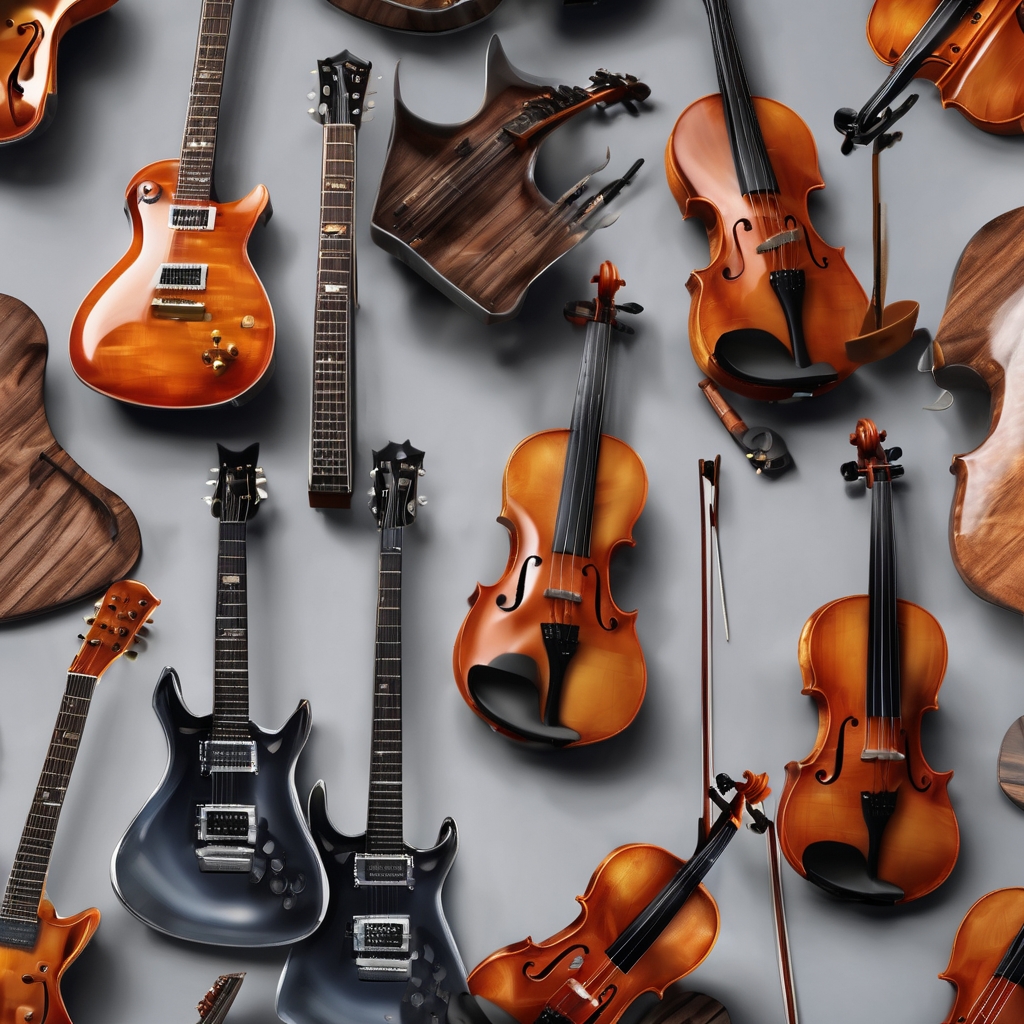} &
        \includegraphics[width=0.14\textwidth]{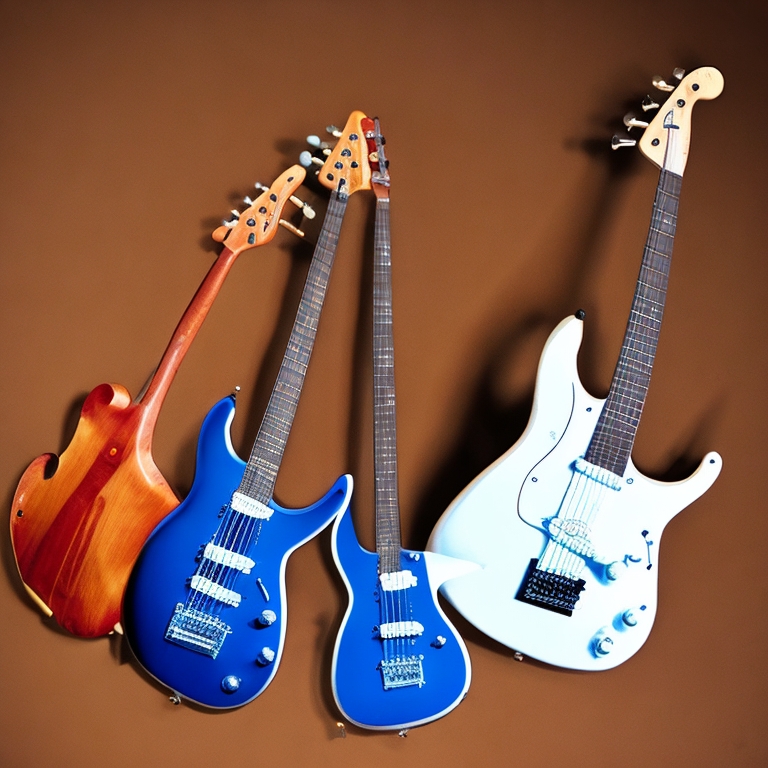} &
        \includegraphics[width=0.14\textwidth]{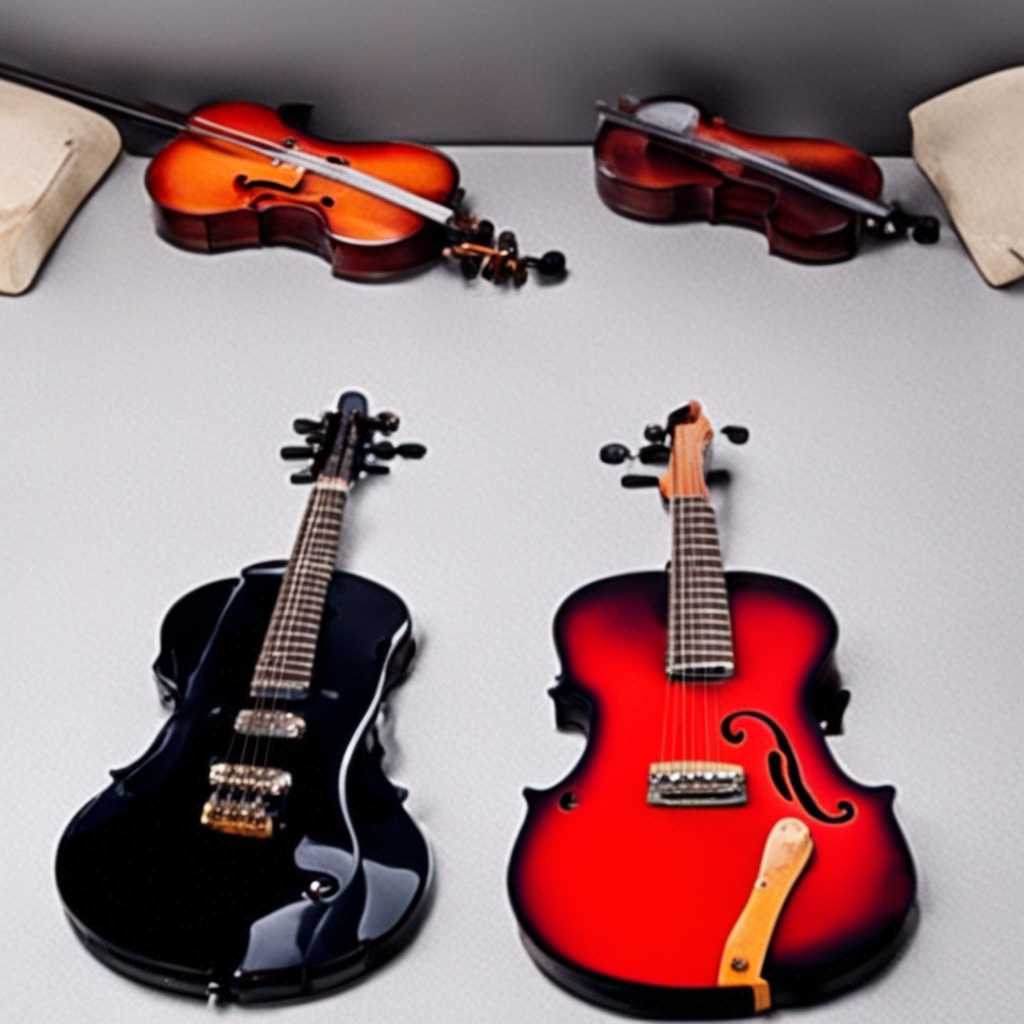} \\

        \multicolumn{7}{c}{"\textbf{Two stone towers} and a \textbf{carriage} in medieval times"} \\
        \includegraphics[width=0.14\textwidth]{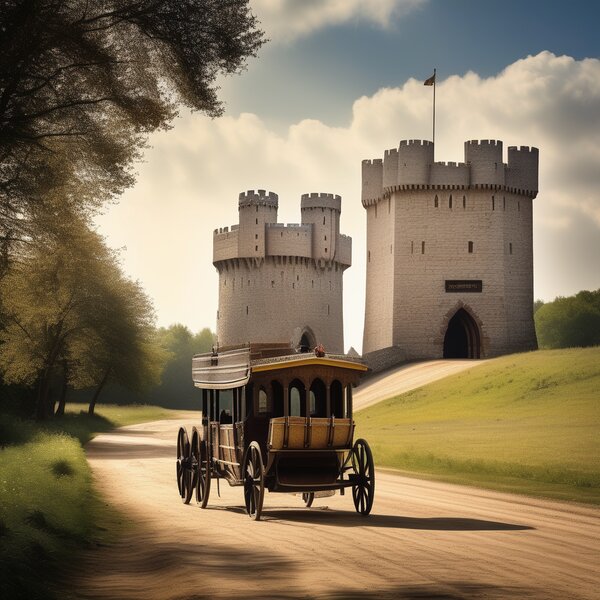} &
        \includegraphics[width=0.14\textwidth]{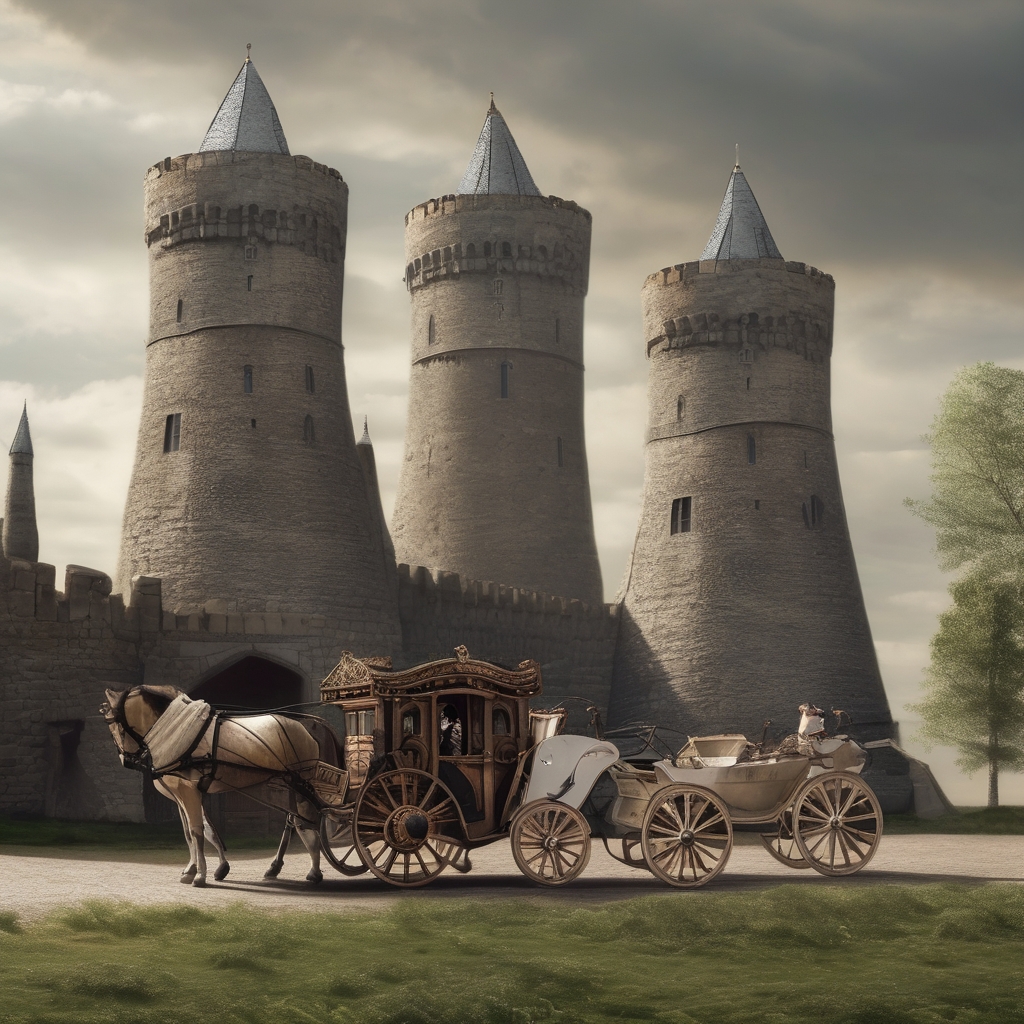} &
        \includegraphics[width=0.14\textwidth]{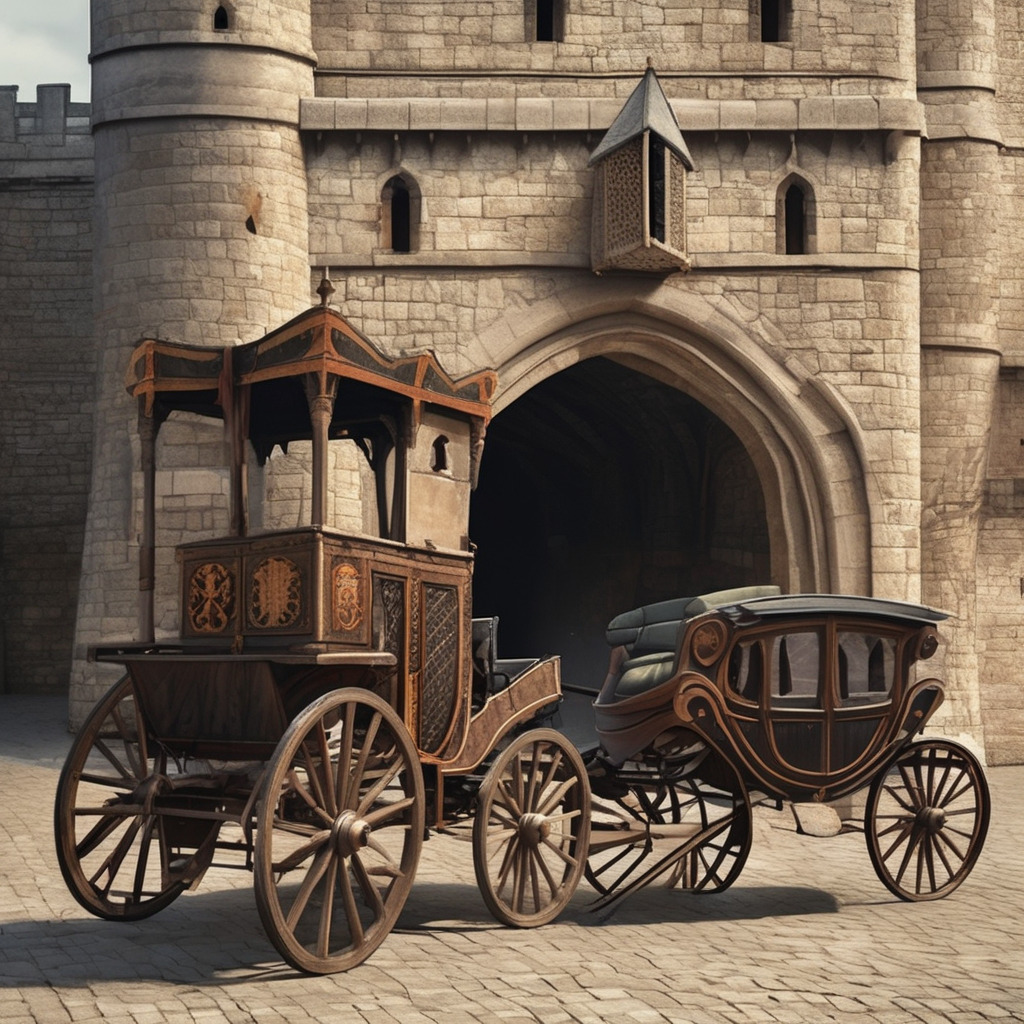} &
        \includegraphics[width=0.14\textwidth]{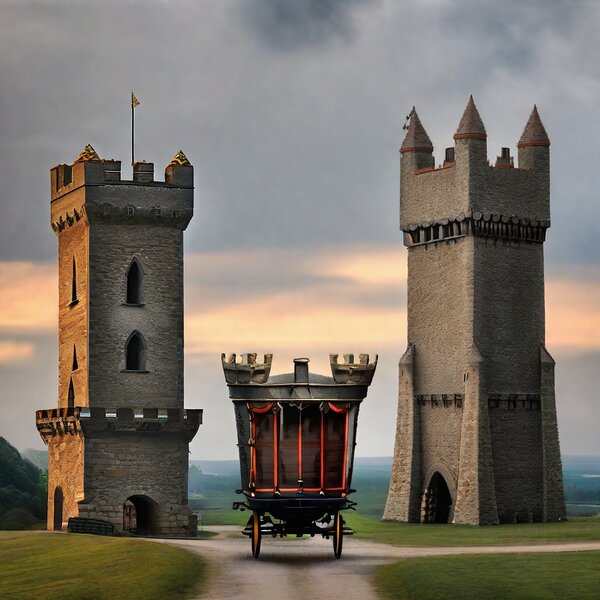} &
        \includegraphics[width=0.14\textwidth]{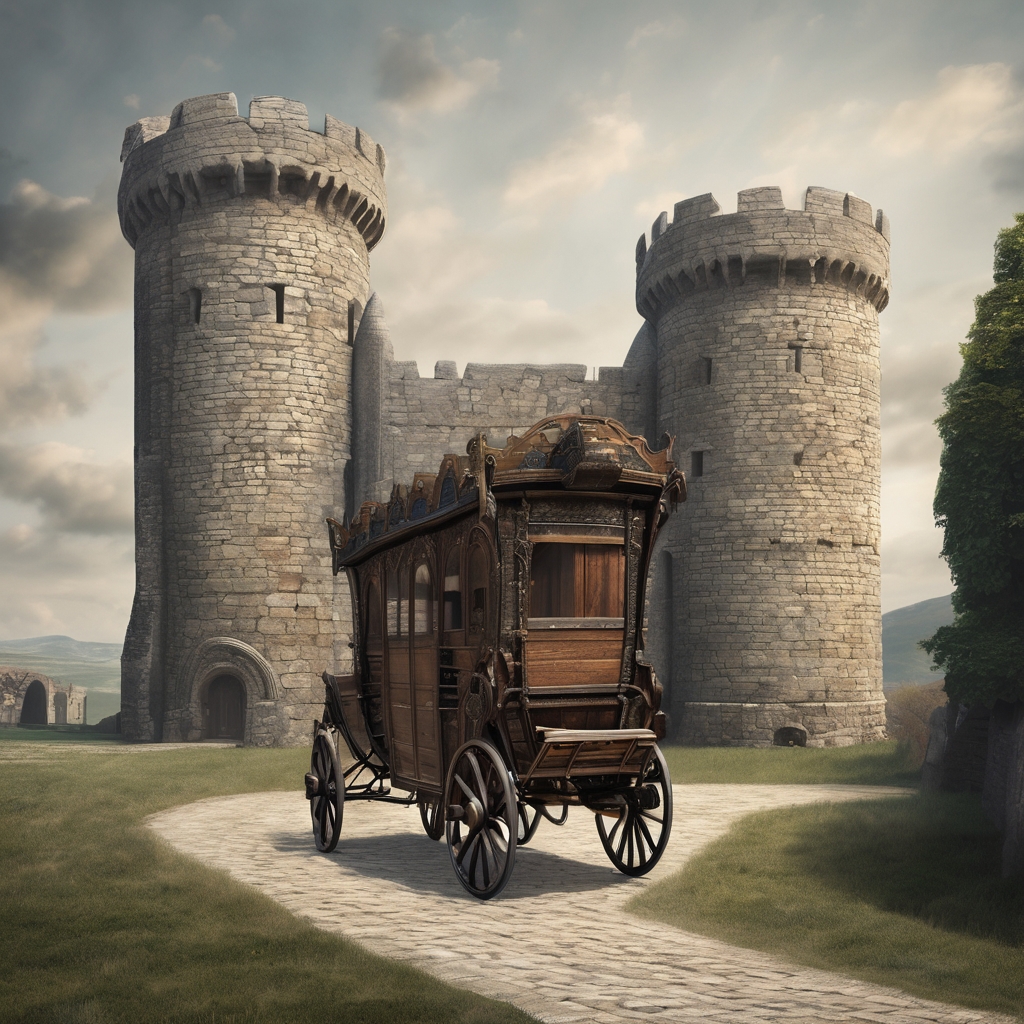} &
        \includegraphics[width=0.14\textwidth]{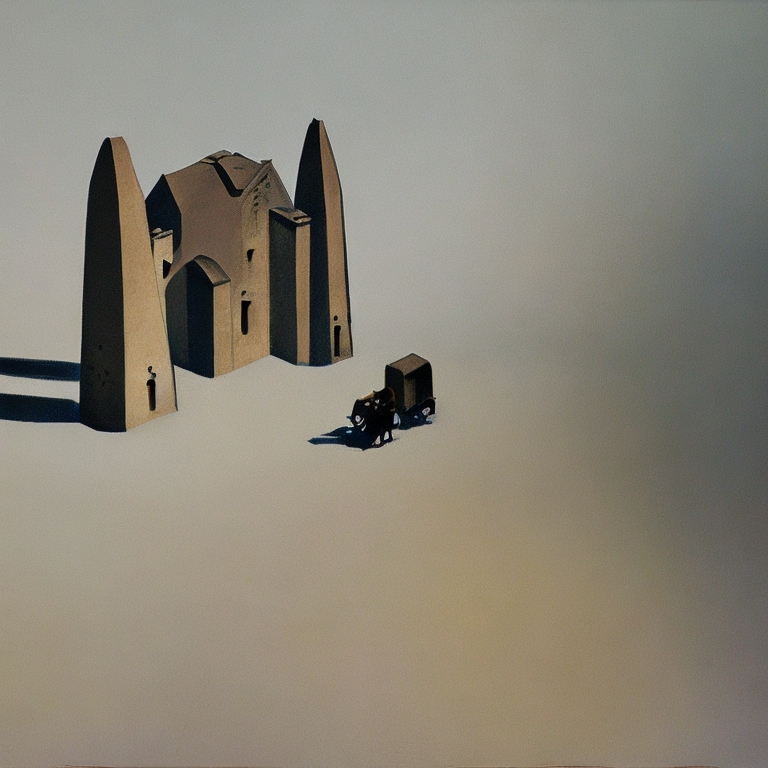} &
        \includegraphics[width=0.14\textwidth]{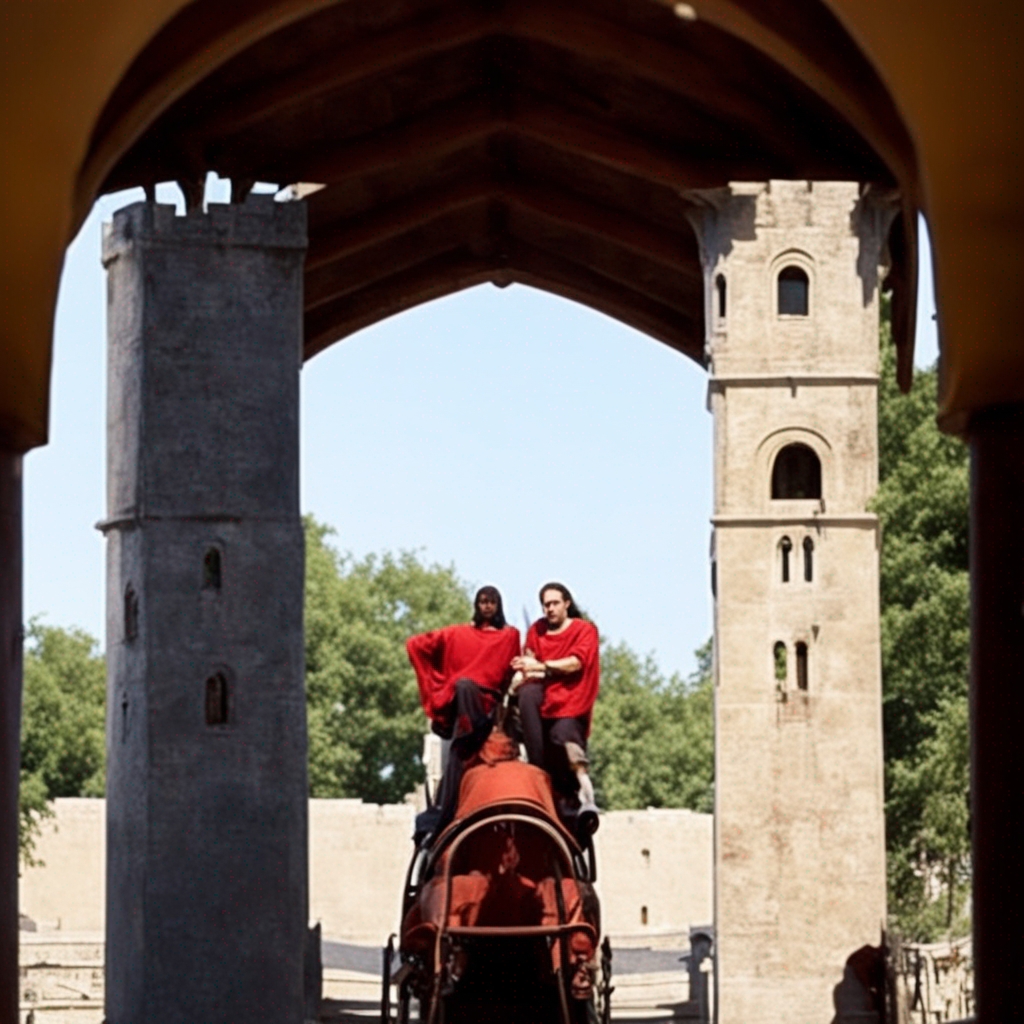} \\

        \multicolumn{7}{c}{"An \textbf{iguana}, a \textbf{red lizard}, a \textbf{turtle} and a \textbf{porcupine} in the desert"} \\
        \includegraphics[width=0.14\textwidth]{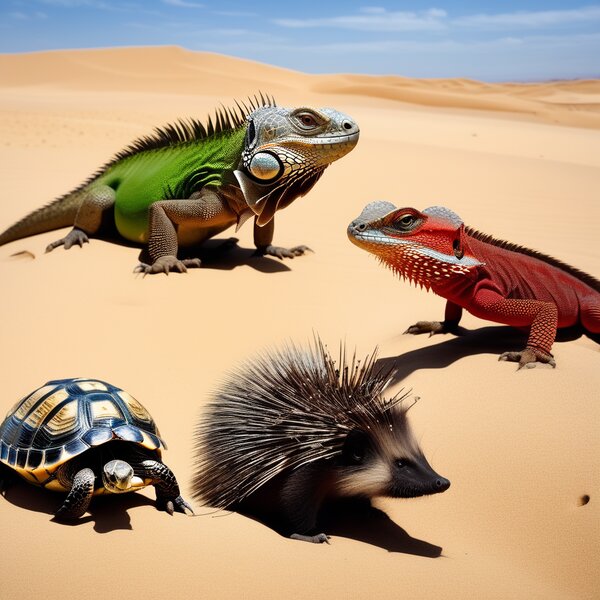} &
        \includegraphics[width=0.14\textwidth]{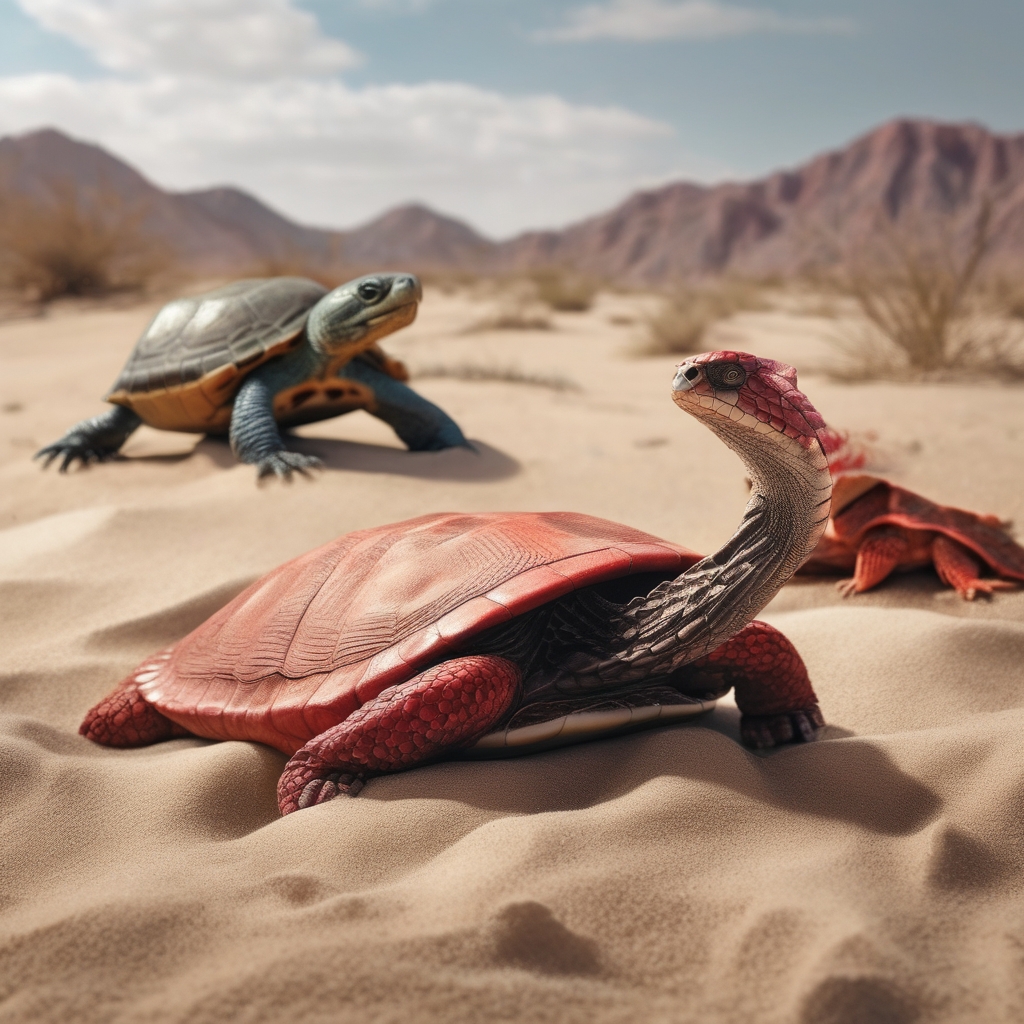} &
        \includegraphics[width=0.14\textwidth]{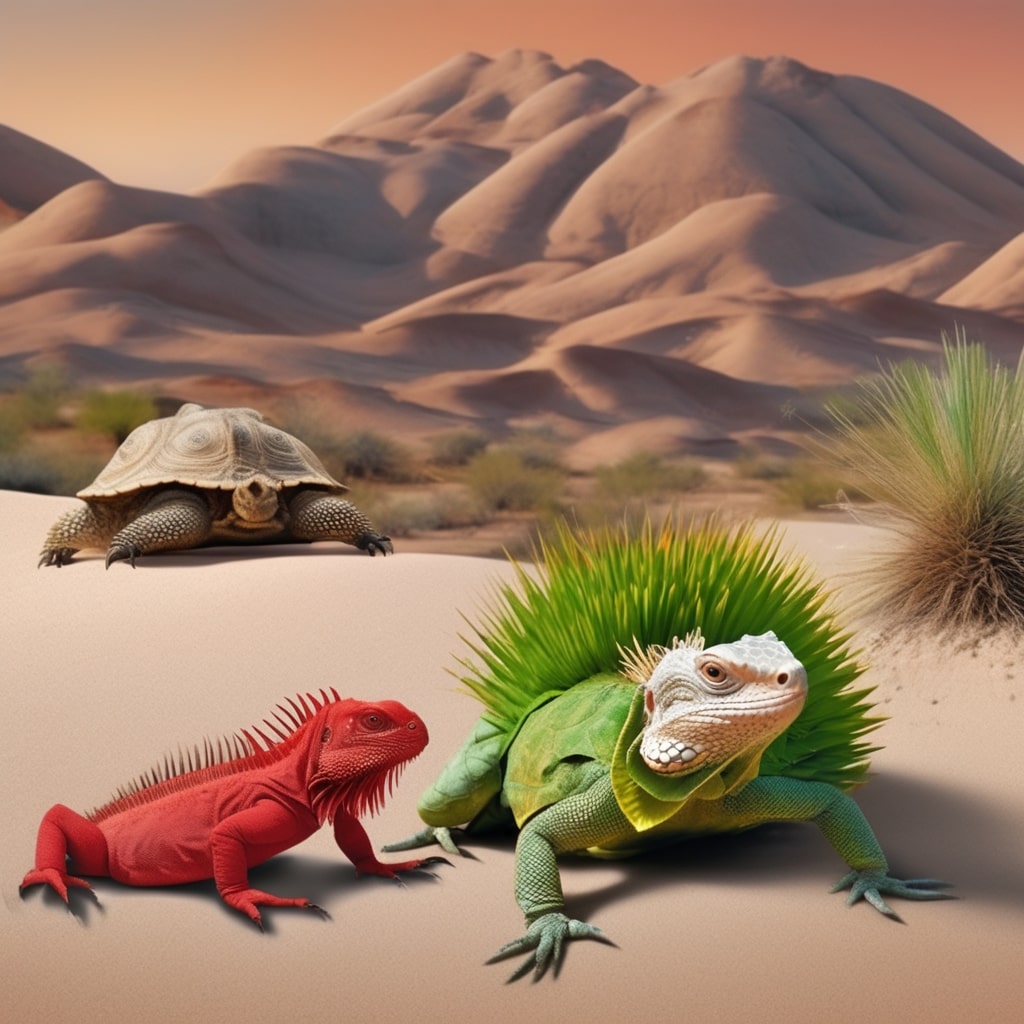} &
        \includegraphics[width=0.14\textwidth]{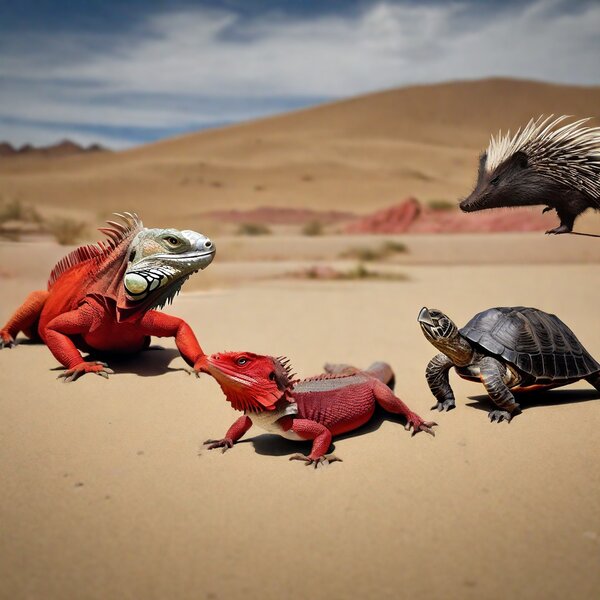} &
        \includegraphics[width=0.14\textwidth]{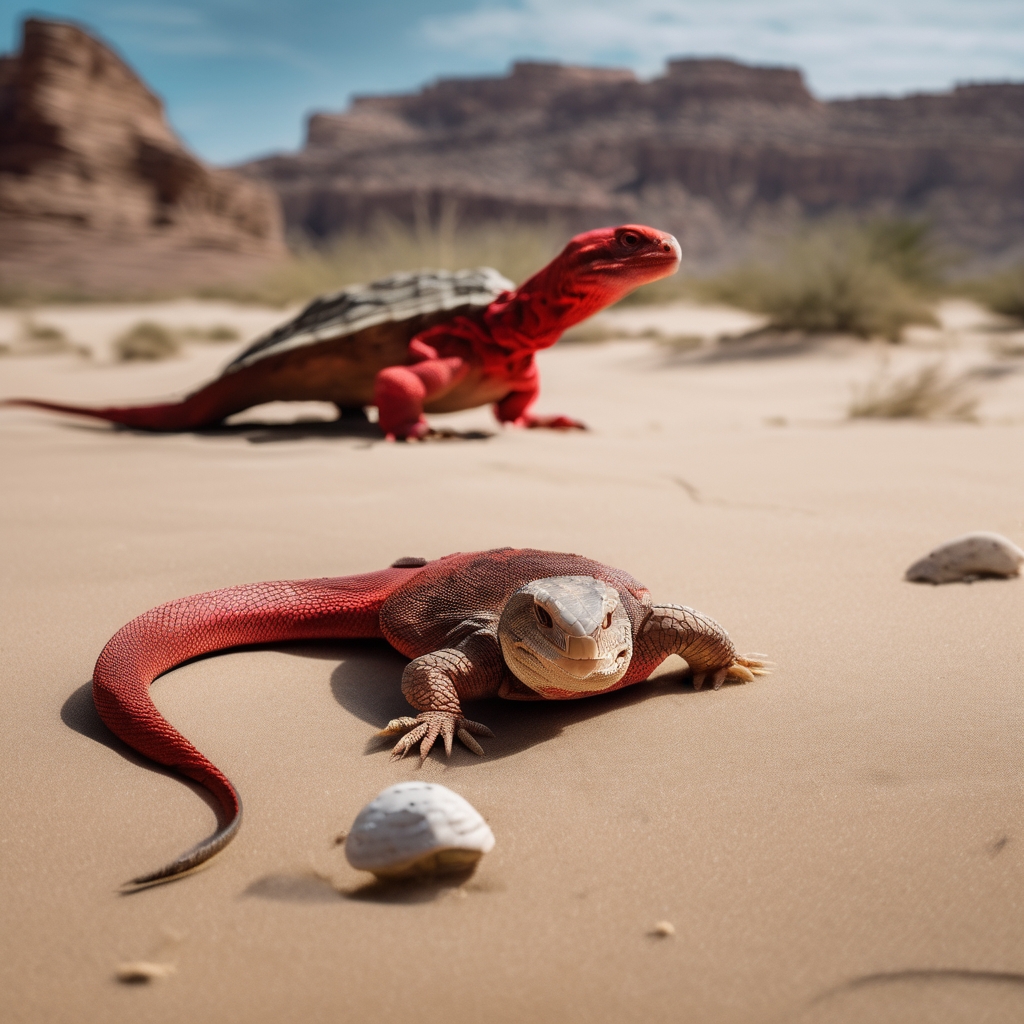} &
        \includegraphics[width=0.14\textwidth]{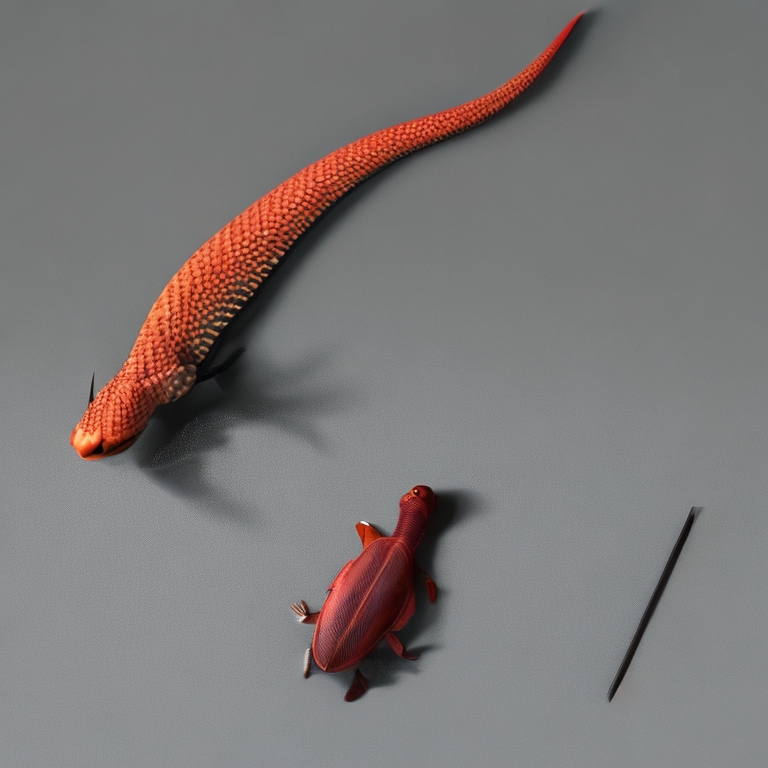} &
        \includegraphics[width=0.14\textwidth]{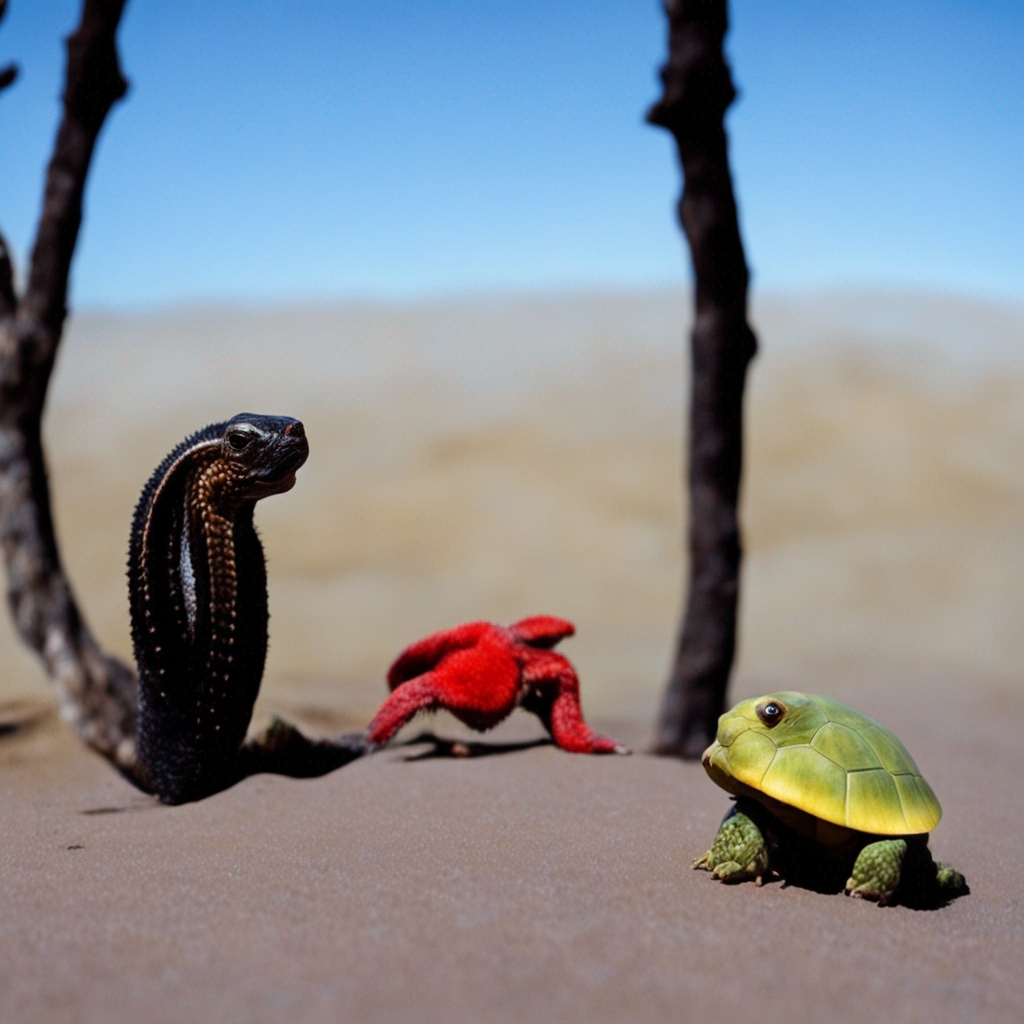} \\
        
        Ours & SDXL & A\&E & LLM+BA & RPG & Ranni & LMD+ \\
    \end{tabular}
    }
    \captionof{figure}{\textbf{Qualitative comparison} of our method with baseline methods.
    }
    \label{fig:comparisons_sup}
\end{figure*}

\begin{figure*}[h!]
    \setlength{\tabcolsep}{0.002\textwidth}
    \scriptsize
    \centering
    {
    \begin{tabular}{c c c c c c}
        \multicolumn{6}{c}{"... a \textbf{parrot}, and \textbf{two doves} sitting on a branch in a lush forest at daylight"} \\
        \includegraphics[width=0.163\textwidth]{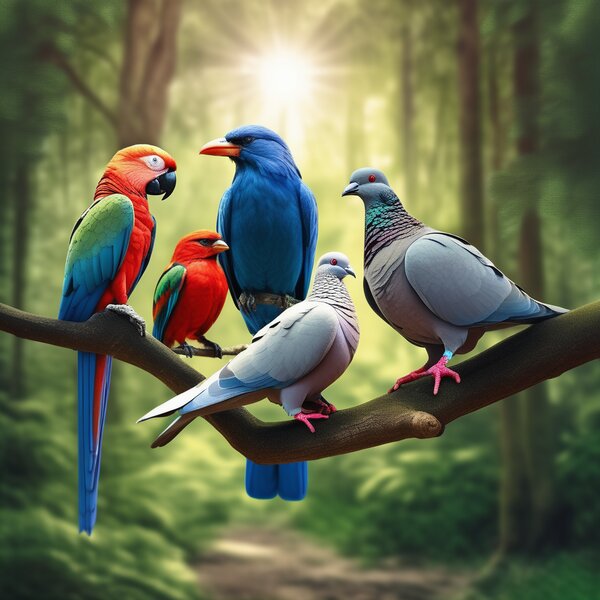} &
        \includegraphics[width=0.163\textwidth]{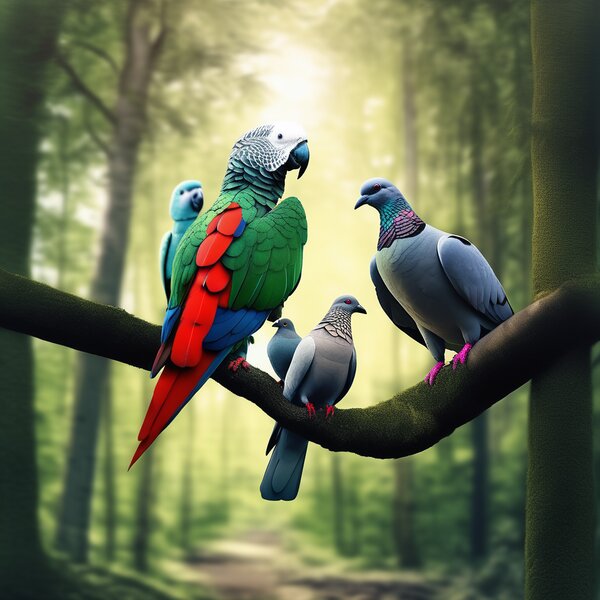} &
        \includegraphics[width=0.163\textwidth]{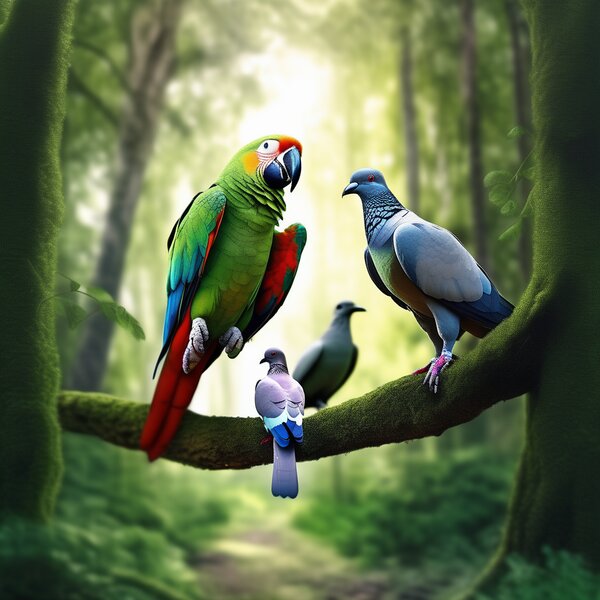} &
        \includegraphics[width=0.163\textwidth]{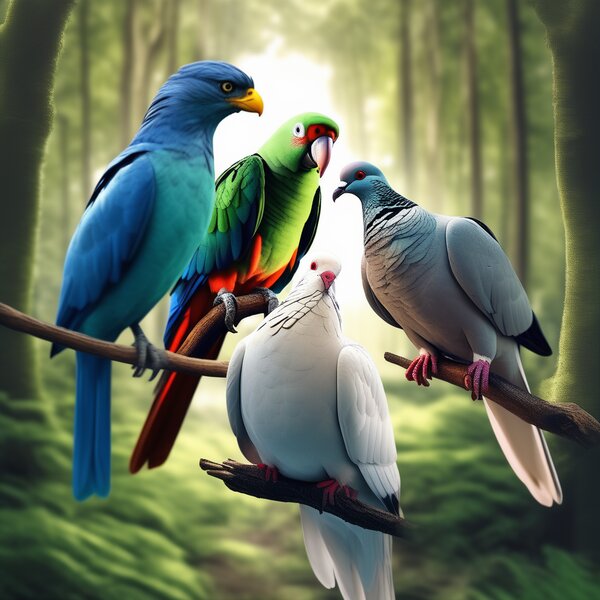} &
        \includegraphics[width=0.163\textwidth]{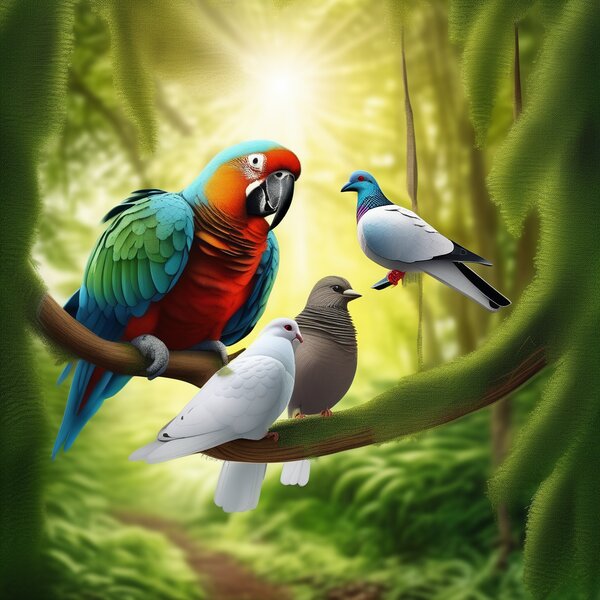} &
        \includegraphics[width=0.163\textwidth]{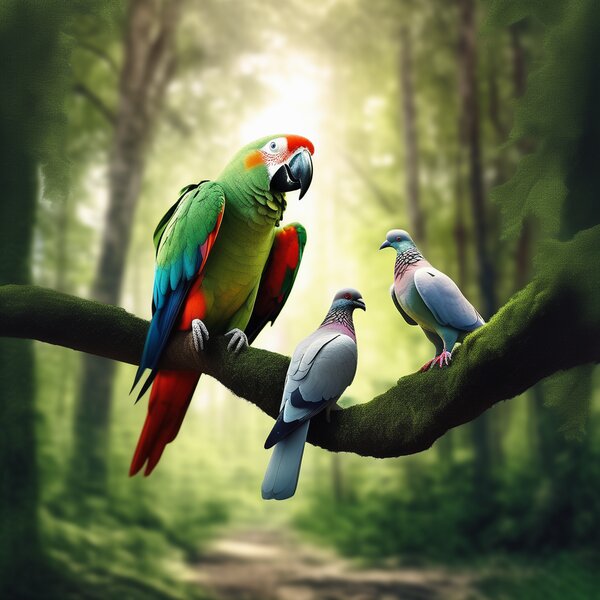} \\

        \multicolumn{6}{c}{"... \textbf{two cows} and a \textbf{donkey} in a farm"} \\
        \includegraphics[width=0.163\textwidth]{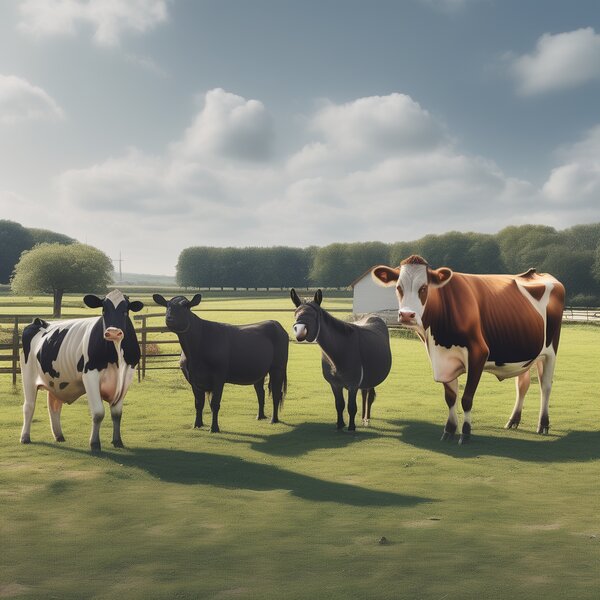} &
        \includegraphics[width=0.163\textwidth]{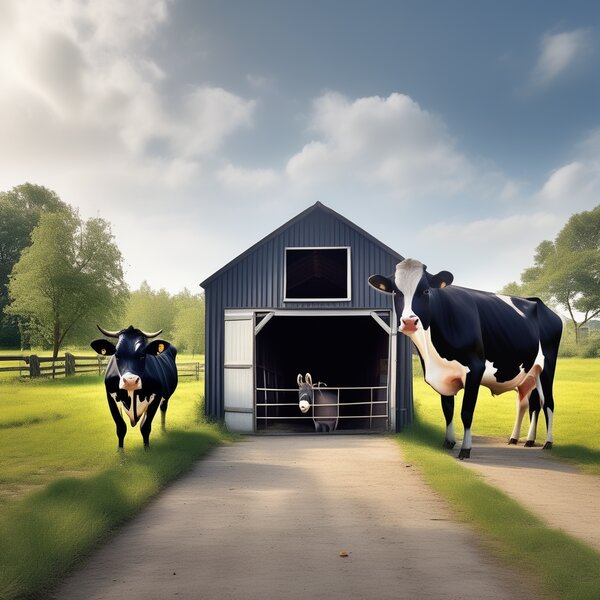} &
        \includegraphics[width=0.163\textwidth]{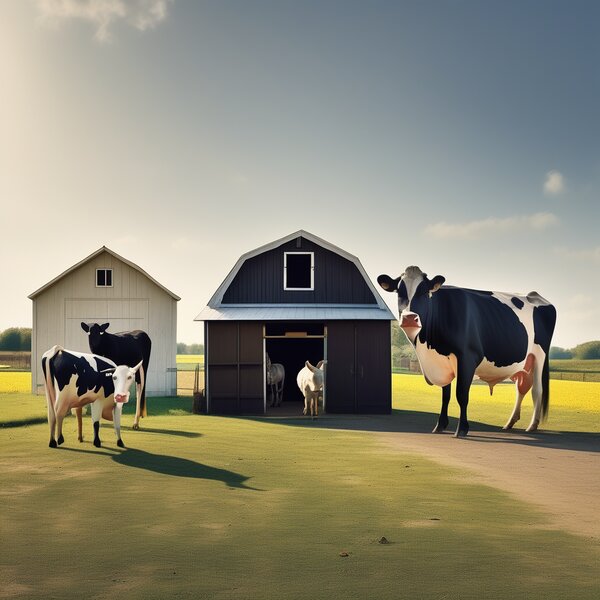} &
        \includegraphics[width=0.163\textwidth]{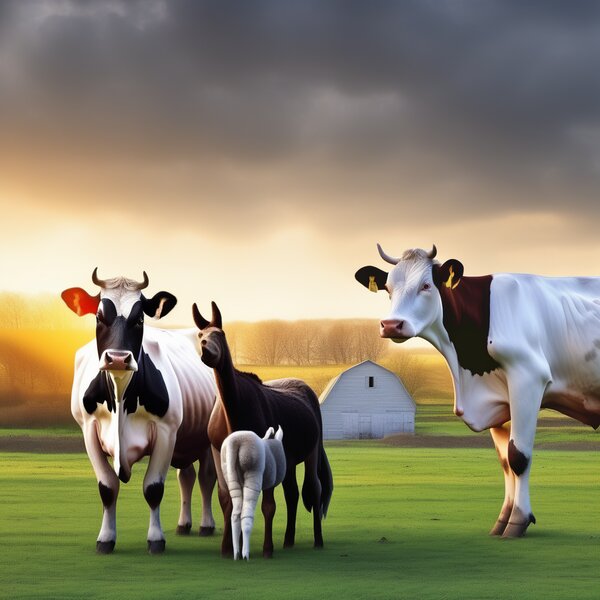} &
        \includegraphics[width=0.163\textwidth]{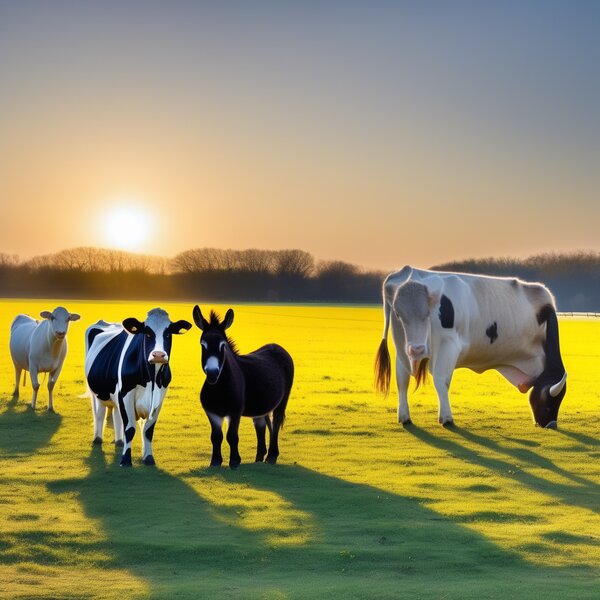} &
        \includegraphics[width=0.163\textwidth]{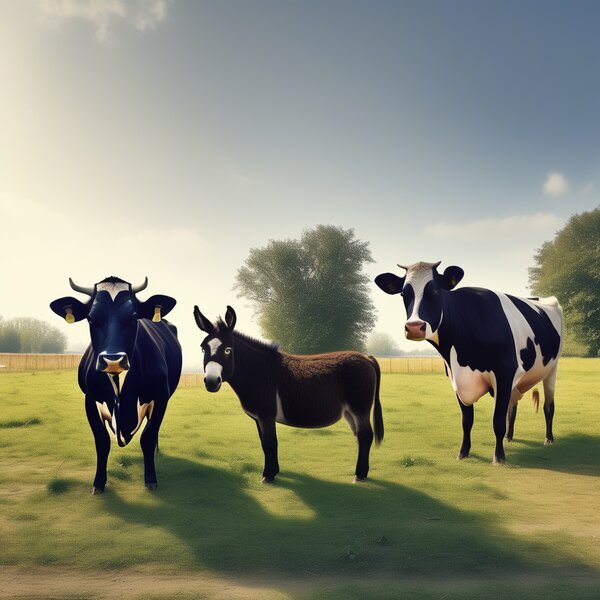} \\
        
        w/o $\mathcal{L}_{\text{decisive}}$ & w/o $\mathcal{L}_{\text{cross}}$ & w/o $\mathcal{L}_{\text{var}}$ & w/o $\mathcal{L}_{\text{dice}}$ & $\mathcal{L}_{\text{decisive}}\left(S^{t-1},M^{t-1}\right)$ & Full method \\
    \end{tabular}
    }
    \captionof{figure}{\textbf{Qualitative ablation.} We ablate our method by skipping the guidance steps (w/o $\mathcal{L}_{\text{decisive}}$), dropping a loss term when optimizing (w/o $\mathcal{L}_{\text{cross}}$, w/o $\mathcal{L}_{\text{var}}$, w/o $\mathcal{L}_{\text{dice}})$, and performing an alternative guidance step ($\mathcal{L}_{\text{decisive}}\left(S^{t-1},M^{t-1}\right)$), where the loss is computed between the soft- and hard-layouts of the same timestep (instead of $\mathcal{L}_{\text{decisive}}\left(S^{t-1},M^t\right)$). All images in each row are generated using the same seed.}
    \label{fig:ablation}
\end{figure*}

We use the same baseline as in the main paper: LLM+BA~\cite{dahary2025yourself}, RPG~\cite{yang2024mastering}, Ranni~\cite{feng2024ranni}, LMD+\cite{lian2023llm}, and A\&E~\cite{chefer2023attend}.

\subsection{Quantitative Results}

\begin{table}
    \centering
    \setlength{\tabcolsep}{0.005\textwidth}
    \captionof{table}{
    Quantitative evaluation.
    \vspace{-10pt}
    }
    \begin{tabular}{l c c c c c c c}
        \toprule
        Method & Color & Texture & Single-Class & Multi-Class & Layout Diversity \\
        \midrule
        Ours & 0.704 & 0.686 & 0.837 & 0.723 & 0.718 \\
        SDXL & 0.568 & 0.660 & 0.746 & 0.676 &  -  \\
        A\&E & 0.537 & 0.659 & 0.742 & 0.682 &  -  \\
        LLM+BA & 0.685 & 0.665 & 0.659 & 0.603 & 0.408 \\
        RPG & 0.604 & 0.643 & 0.609 & 0.635 & 0.155 \\
        Ranni & 0.259 & 0.445 & 0.729 & 0.579 & 0.679 \\
        LMD+ & 0.457 & 0.614 & 0.885 & 0.898 & 0.408 \\
        \bottomrule
    \end{tabular}
    \label{table:eval}
\end{table}

\begin{table}
    \captionof{table}{
    Ablation user study results.
    \vspace{-10pt}
    }
    \centering
    \setlength{\tabcolsep}{0.005\textwidth}
    \begin{tabular}{l c c c c c c c}
        \toprule
        Method & Prompt-Alignment Accuracy \\
        \midrule
        w/o $\mathcal{L}_{\textit{decisive}}$ & 0.016 \\
        w/o $\mathcal{L}_{\textit{cross}}$ & 0.442 \\
        w/o $\mathcal{L}_{\textit{var}}$ & 0.447 \\
        w/o $\mathcal{L}_{\textit{dice}}$ & 0.105 \\
        $\mathcal{L}_{\text{decisive}}\left(S^{t-1},M^{t-1}\right)$ & 0.289 \\
        Full method & 0.832 \\
        \bottomrule
    \end{tabular}
    \label{table:ablation}
\end{table}

Table~\ref{table:eval} presents the quantitative results comparing our method with the baseline. Unlike other approaches that balance trade-offs between metrics, our method consistently delivers high performance across all metrics.

\subsection{Qualitative Results}

Figure~\ref{fig:comparisons_sup} showcases additional qualitative comparison results. Unlike competing methods, which fail to accurately generate all subjects from the prompt, our method consistently preserves the intended semantics of each subject. For instance, in the first row, none of the methods successfully generate all the fruits specified in the prompt. Similarly, in the last row, none of the methods accurately capture all the animals in the prompt, with most suffering from attribute leakage.

\subsection{Ablation Studies}

\paragraph{Quantitative Evaluation}

To quantitatively assess the contribution of each component, we conducted a user study, following the same format as our benchmark user study, and composed of a subset of 10 random prompts from the full user study. We report the percentage of user-selected images for each methods, i.e. prompt-alignment accuracy, as recorded by 19 participants, in Table~\ref{table:ablation}.

\paragraph{Qualitative Evaluation}

We display qualitative ablation studies in Figure~\ref{fig:ablation}, where we systematically vary our method's configuration to assess each component's importance.

As can be seen in the leftmost four columns, neglecting our decisive guidance, or any of its terms, promotes subject over-generation due to the instability of the hard-layouts during generation.
Omitting $\mathcal{L}_{\text{cross}}$ or $\mathcal{L}_{\text{var}}$ causes clusters to fragment internally, leading them to span multiple, disconnected subject instances. Additionally, parts of a subject may be absorbed into the background, resulting in shrunken or incomplete subject regions.
Omitting $\mathcal{L}_{\text{dice}}$ also promotes over-generation, as oscillating cluster boundaries across timesteps lead to the emergence of redundant subjects with mixed appearances at cluster edges.

Finally, we ablate the choice of computing our $\mathcal{L}_{\text{decisive}}$ loss between the intermediate soft-layout $S^{t-1}$ and the previous hard-layout $M^{t}$. Instead, we apply an additional denoising step on $z_{t-1}$ before optimizing it using guidance. That denoising step is employed to extract the denoising model's features and compute an updated hard-layout $M^{t-1}$. Then, during guidance, we compute $\mathcal{L}_{\text{decisive}}$ between $S^{t-1}$ and $M^{t-1}$. As evident in the second-to-right column, this approach also compromises accurate subject generation, yielding redundant subject instance due to clustering inconsistencies between timesteps.

\subsection{Limitations}

\begin{figure}
    \setlength{\tabcolsep}{0.002\textwidth}
    \scriptsize
    \centering
    
    \begin{tabular}{c c}
        & ``... a \textbf{cactus} in a \textbf{clay pot} and \\
        ``... \textbf{ten apples} ...'' & a \textbf{fern} in a \textbf{porcelain pot} ...'' \\
        \includegraphics[width=0.4\linewidth]{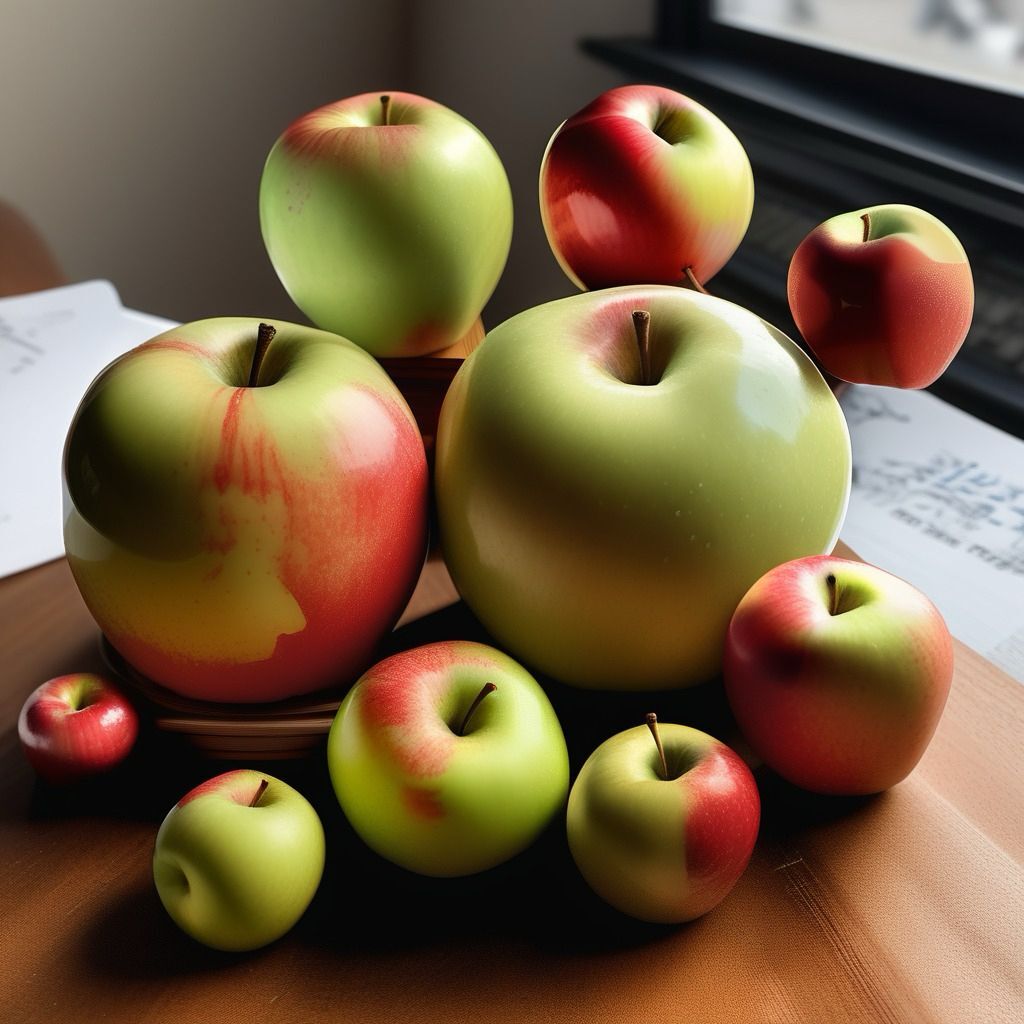} & 
        \includegraphics[width=0.4\linewidth]{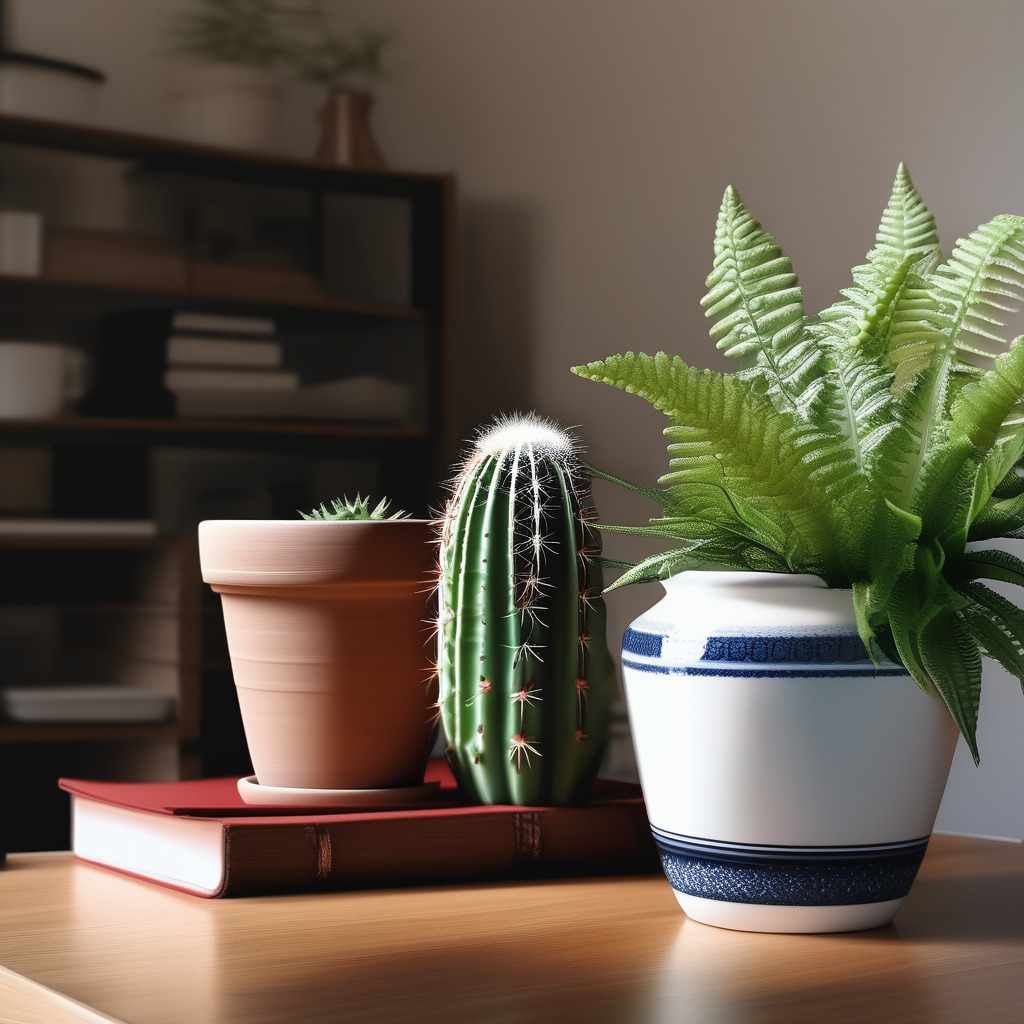}
    \end{tabular}
    \caption{Limitations.}
    \label{fig:limitations}
\end{figure}

In Figure~\ref{fig:limitations}, we present two limitations of our method. First, in cluttered scenes, subjects may appear with irregular sizes or exhibit poor interaction with the background (left image). We observe that this issue also occurs with vanilla SDXL and can often be mitigated by increasing the number of denoising steps.

Second, since the layouts are derived from the model’s prior — which lacks a robust understanding of spatial relationships~\cite{chatterjee2024getting} — subjects may sometimes fail to respect spatial constraints specified in the prompt (right image).

\end{appendices}

\end{document}